%% file: main_princeton.tex
\pdfoutput=1

\documentclass[phd,black,twoside,10pt]{PrincetonThesis}
\usepackage{epsfig}
	\usepackage{setspace}

	\title{Message Passing and Combinatorial Optimization}
	\author{Siamak Ravanbakhsh}
	\department{Department of Computing Science}
	\advisor{Russell Greiner}
	\degreemonth{January}
	\degreeyear{2015}

	\input{header_princeton}

    \glsaddall
    \newglossary[slg]{notation}{not}{ntn}{Notation}
    \makeglossaries
    \loadglsentries{gloss}

    \glsaddall
    \makeindex

\begin{document}
        \doublespacing	
	\begin{frontmatter}

	  \begin{thesisabstract}
	  \input{abstract}
	  \end{thesisabstract}
          
\clearpage
\begin{center}
To the memory of my grandparents.
\end{center}
\clearpage

	  \begin{acknowledgements}
	  \input{acknowledgements}

	  \end{acknowledgements}
    %

	\end{frontmatter}

	\cleardoublepage
\pagestyle{fancyplain}
\onehalfspacing
\input{chap1}
\input{chap2_new}
\input{chap3}

\input{chap4}

\input{conclusion.tex}

	\appendix


\cleardoublepage
	
\bibliographystyle{abbrvnat-without-url}
{\small
\bibliography{thesis} 
}%

\cleardoublepage

\part*{Appendix}
\addcontentsline{toc}{part}{Appendix}

\input{proofs}


\end{document}

%% file: header_princeton.tex
\usepackage{etex}
\usepackage{url}
\usepackage[]{algorithm2e}
\usepackage{tikz}
\usepackage{verbatim}
\usepackage{amsmath}
\usepackage{amsthm}

\usepackage{amsfonts}
\usepackage{amssymb}    
\usepackage{mathrsfs}
\usepackage[ampersand]{easylist}
\ListProperties(Hide=100, Hang=true, Progressive=3ex, Style*=-- ,
Style2*=$\bullet$ ,Style3*=$\circ$ ,Style4*=\tiny$\blacksquare$ )
\usepackage{graphicx}
\usepackage{tocbibind}
\usepackage{subcaption}
\usepackage[font={small,it},labelfont=bf]{caption}%

\usepackage{multicol}
\usepackage{multirow}
\usepackage{tabu}
\usepackage{rotating}

\usepackage[numbers,square,sort&compress]{natbib}
\usepackage[pdftex,plainpages=false,breaklinks=true,colorlinks=true,urlcolor=cyan,citecolor=cyan,linkcolor=cyan,bookmarks=true,bookmarksopen=cyan,bookmarksopenlevel=0,pagebackref,linktocpage=true,bookmarksnumbered=true]{hyperref}
\usepackage{hypernat}
\usepackage[nosuper,xindy,nonumberlist]{glossaries}
\usepackage{cleveref}

    \DeclareMathOperator{\prob}{Pr}
\renewcommand{\Re}[0]{\mathcal{R}} 
\newcommand{\spsign}[0]{\ensuremath{\frown}}

\newcommand{\ex}[0]{\mathbb{E}}

\newcommand{\expo}[1]{\exp\left(#1\right)}

\usepackage{marginnote}
\usepackage{framed}

\newcommand{\marnote}[1]{}
\newcommand{\marnoteloc}[2]{}

\newcommand{\magn}[1]{{\textbf{#1}}}

\usepackage{enumerate}
\newcommand{\refChapter}[1]{\cref{#1}}
\newcommand{\RefChapter}[1]{\Cref{#1}}
\newcommand{\refSection}[1]{\cref{#1}}
\newcommand{\RefSection}[1]{\Cref{#1}}
\newcommand{\refEq}[1]{\cref{#1}}
\newcommand{\refEqs}[2]{\cref{#1}}
\newcommand{\RefEq}[1]{\Cref{#1}}
\newcommand{\RefEqs}[2]{\Cref{#1} and \ref{#1}}
\newcommand{\refFigure}[1]{\cref{#1}}
\newcommand{\RefFigure}[1]{\Cref{#1}}
\newcommand{\refTheorem}[1]{\cref{#1}}
\newcommand{\refExample}[1]{\cref{#1}}
\newcommand{\RefProposition}[1]{\Cref{#1}}
\newcommand{\refProposition}[1]{\cref{#1}}
\newcommand{\refTable}[1]{\cref{#1}}
\newcommand{\RefTable}[1]{\Cref{#1}}
\newcommand{\RefAlgorithm}[1]{\Cref{#1}}
\newcommand{\refDefinition}[1]{\cref{#1}}
\newcommand{\bullitem}[0]{\ensuremath{\bullet~}}
\newcommand{\pcref}[1]{\cref{#1}}

\newcommand{\ie}[0]{\emph{i.e.},~}
\newcommand{\eg}[0]{\emph{e.g.},~}
\newcommand{\aka}[0]{a.k.a.~}
\newcommand{\wrt}[0]{w.r.t.~}
\newcommand{\etal}{\emph{et al.~}}
\newcommand{\etc}{\emph{etc.~}}

\renewcommand{\Re}[0]{\mathbb{R}} 
\newcommand{\Qe}[0]{\mathbb{Q}} 
\newcommand{\Ce}[0]{\mathbb{C}} 
\newcommand{\mb}{\ensuremath{\Delta}}
\newcommand{\pancake}{\nabla}

\newcommand{\xx}[0]{\ensuremath{x}}
\newcommand{\xxh}[0]{\ensuremath{\hat{x}}}
\newcommand{\xxx}[2]{\ensuremath{\xx_{#1:#2}}}
\newcommand{\xxs}[2]{\ensuremath{\xs_{#1:#2}}}
\newcommand{\ww}[0]{\ensuremath{w}}
\newcommand{\yy}[0]{\ensuremath{y}}
\newcommand{\zz}[0]{\ensuremath{z}}
\newcommand{\zzz}[2]{\ensuremath{\zz_{#1:#2}}}
\DeclareRobustCommand{\xs}{\ensuremath{\underline{\xx}}}
\newcommand{\xsh}[0]{\ensuremath{\hat{\underline{\xx}}}}

\newcommand{\zs}[0]{\ensuremath{\underline{\zz}}}
\newcommand{\zzs}[2]{\ensuremath{\zs_{#1:#2}}}

\newcommand{\opp}[1]{{\overset{#1}{\oplus}}}
\newcommand{\bigopp}[1]{\overset{#1}{\bigoplus}}

\newcommand{\falsemath}{\ensuremath{\text{\textsc{false}}}}
\newcommand{\truemath}{\ensuremath{\text{\textsc{true}}}}
\newcommand{\nulll}{\ensuremath{\text{\textsc{null}}}}

\newcommand{\sn}{\ensuremath{\settype{S}_{N}}}

\newcommand{\func}[1]{\mathsf{#1}}
\newcommand{\jfun}{\ensuremath{\jmath}}
\newcommand{\ff}{\ensuremath{\func{f}}}
\newcommand{\fg}{\ensuremath{\func{g}}}
\newcommand{\pp}{\ensuremath{\func{p}}}
\newcommand{\mm}{\ensuremath{\func{m}}}
\newcommand{\ph}{\ensuremath{\widehat{\pp}}}
\newcommand{\PP}{\ensuremath{\func{P}}}

\newcommand{\hp}{\ensuremath{\func{h}}}
\newcommand{\hh}{\ensuremath{\widehat{\hp}}}

\newcommand{\qq}{\ensuremath{\func{q}}}
\newcommand{\vs}{\underbar{\ensuremath{\func{v}}}}
\newcommand{\strat}{\ensuremath{\func{s}}}
\newcommand{\QQ}{\ensuremath{\func{Q}}}

\newcommand{\settype}[1]{\ensuremath{\mathcal{#1}}}
\newcommand{\II}{\ensuremath{\mathrm{I}}}
\newcommand{\AAA}{\ensuremath{\mathrm{A}}}
\newcommand{\BBB}{\ensuremath{\mathrm{B}}}
\newcommand{\JJ}{\ensuremath{\mathrm{J}}}
\newcommand{\KK}{\ensuremath{\mathrm{K}}}
\newcommand{\LL}{\ensuremath{\mathrm{L}}}
\newcommand{\HH}{\ensuremath{\mathrm{H}}}
\newcommand{\NN}{\ensuremath{\settype{N}}}
\newcommand{\interaction}{\ensuremath{{\mathit{J}}}}
\newcommand{\localfield}{\ensuremath{{\mathit{h}}}}
\newcommand{\xor}{\ensuremath{{\mathrm{xor}}}}

\newcommand{\kernel}{\ensuremath{\func{k}}}
\newcommand{\cn}{\ensuremath{\func{c}}}
\newcommand{\rank}{\ensuremath{\func{r}}}
\newcommand{\modul}{\ensuremath{\func{m}}}

\newcommand{\OO}{\mathbf{\mathcal{O}}}
\newcommand{\GG}{\settype{G}}

\newcommand{\HG}{\settype{H}}
\newcommand{\KG}{\settype{K}}
\newcommand{\CG}{\settype{C}}
\newcommand{\VV}{\settype{V}}

\newcommand{\EE}{\settype{E}}

\newcommand{\CC}{\settype{C}}
\renewcommand{\SS}{\settype{S}}
\newcommand{\DD}{\settype{D}}
\newcommand{\TT}{\settype{T}}
\newcommand{\FF}{\settype{F}}
\newcommand{\YY}{\settype{Y}}
\newcommand{\XX}{\settype{X}}
\newcommand{\RR}{\settype{Y}^*}

\newcommand{\WW}{\settype{P}}

\newcommand{\solutions}{\settype{S}}
\newcommand{\region}{\rho}
\newcommand{\msgrg}{\gamma}
\newcommand{\rg}[2]{\region_{#1:#2}}
\newcommand{\sumset}[0]{{ \settype{S}}}
\newcommand{\pset}[0]{{ \settype{D}}}
\newcommand{\semig}{\mathscr{G}}
\newcommand{\semiring}{\mathscr{S}}
\newcommand{\auto}{\func{Aut}}
\newcommand{\homo}{\func{Hom}}
\newcommand{\orbit}{\func{orbit}}

\newcommand{\nn}[1]{[\mathrm{#1}]}
\newcommand{\tst}[1]{\ensuremath{^{(\mathrm{#1})}}}
\newcommand{\Dn}{\ensuremath{\mathbf{A}}}

\newcommand{\Dt}{\ensuremath{\mathbf{B}}}
\newcommand{\nb}{\partial}

\newcommand{\functional}[1]{\mathsf{#1}}
\newcommand{\entropy}{\functional{H}}
\newcommand{\partition}{\functional{Z}}

\newcommand{\energy}{\functional{e}}
\newcommand{\expenergy}{\functional{U}}
\newcommand{\divergence}{\functional{D}}

\newcommand{\compclass}{\mho}

\newcommand{\msg}[2]{{\ph}_{#1 \to #2}}
\newcommand{\msgover}[2]{\tilde{\func{p}}_{#1 \to #2}}
\newcommand{\msga}[0]{\func{m}}
\newcommand{\msgaover}[0]{\tilde{\func{m}}}
\newcommand{\msgs}{{\underline{\ph}}}
\newcommand{\msgss}[2]{{\underline{\ph}}_{#1 \to #2}}

\newcommand{\msgq}[2]{\func{q}_{#1 \to #2}}
\newcommand{\msgssq}[2]{\underline{\func{q}}_{#1 \to #2}}

\newcommand{\msgt}[2]{{\widetilde{\func{p}}}_{#1 \to #2}}
\newcommand{\pht}{\ensuremath{\widetilde{\pp}}}

\newcommand{\MSGT}[2]{{\widetilde{\func{P}}}_{#1 \to #2}}
\newcommand{\MSGSST}[2]{{\widetilde{\underline{\func{P}}}}_{#1 \to #2}}

\newcommand{\PHT}{\ensuremath{\widetilde{\PP}}}
\newcommand{\MSGBIT}[2]{{\widetilde{{\func{P}}}}_{#1 \leftrightarrow #2}}
\newcommand{\msgbit}[2]{{{\underline{\widetilde{\pp}} }}_{#1 \leftrightarrow #2}}
\newcommand{\MSG}[2]{{\func{P}}_{#1 \to #2}}

\newcommand{\MSP}[2]{\func{S}_{#1 \to #2}} 
 
\newcommand{\PSP}[0]{\func{S}} 
\newcommand{\msp}[2]{\func{S}_{#1 \to #2}}
\newcommand{\back}[1]{\ensuremath{\setminus #1}}

\newcommand{\ident}{\mathbf{\func{1}}}

\newcommand{\identt}[1]{\overset{#1}{\ident}}
\newcommand{\NP}{\ensuremath{\mathbb{NP}}}
\newcommand{\coNP}{\ensuremath{\mathbf{co}\mathbb{NP}}}
\newcommand{\Aclass}{\ensuremath{\mathbb{A}}}
\newcommand{\PPclass}{\ensuremath{\mathbb{PP}}}
\newcommand{\Poly}{\ensuremath{\mathbb{P}}}
\newcommand{\RP}{\ensuremath{\mathbb{RP}}}
\newcommand{\poly}{\Poly}
\newcommand{\pspace}{\ensuremath{\mathbb{PSPACE}}}
\newcommand{\sharpP}{\ensuremath{\mathbb{\#P}}}
\newcommand{\defeq}{\ensuremath{\overset{\text{def}}{=}}}
\newcommand{\bptimes}{\ensuremath{\overset{\text{}}{\otimes}}}
\newcommand{\bpplus}{\ensuremath{\overset{\text{}}{\oplus}}}
\newcommand{\bigbpplus}{\ensuremath{\overset{\text{}}{\bigoplus}}}
\newcommand{\bigbptimes}{\ensuremath{\overset{\text{}}{\bigotimes}}}
\newcommand{\sptimes}{\ensuremath{\overset{\spsign}{\otimes}}}
\newcommand{\spplus}{\ensuremath{\overset{\spsign}{\oplus}}}
\newcommand{\bigspplus}{\ensuremath{\overset{\spsign}{\bigoplus}}}

\newcommand{\sumop}{\ensuremath{\mathrm{{sum}}}}
\newcommand{\prodop}{\ensuremath{\mathrm{{prod}}}}
\newcommand{\sumprod}{\ensuremath{(+, \times)}}

\newcommand{\minsum}{\ensuremath{(\min, +)}}
\newcommand{\maxsum}{\ensuremath{(\max, +)}}
\newcommand{\minmax}{\ensuremath{(\min, \max)}}
\newcommand{\average}{\ensuremath{\mathsf{avg}}}
\newcommand{\powerop}{\ensuremath{{\odot}}}

\newcommand{\pbpiI}[2]{\ensuremath{{\func{X}}_{#1 \to #2}}}
\newcommand{\pbp}{\ensuremath{{\func{X}}}}

\newcommand{\gsiI}[2]{\ensuremath{{\func{G}_{#1 \to #2}}}}

\theoremstyle{plain}
\newtheorem{theorem}{Theorem}[section]
\newtheorem{claim}[theorem]{Claim}
\newtheorem{lemma}[theorem]{Lemma}
\newtheorem{conjecture}[theorem]{Conjecture}
\newtheorem{proposition}[theorem]{Proposition}
\newtheorem{corollary}[theorem]{Corollary}
\theoremstyle{definition}
\newtheorem{definition}{Definition}[section]
\newtheorem{example}{Example}[section]
\theoremstyle{remark}
\newtheorem{remark}{Remark}
 \AtBeginDocument{%
    \crefname{equation}{equation}{equations}%
    \crefname{chapter}{chapter}{chapters}%
    \crefname{section}{section}{sections}%
    \crefname{appendix}{appendix}{appendices}%
    \crefname{enumi}{item}{items}%
    \crefname{footnote}{footnote}{footnotes}%
    \crefname{figure}{figure}{figures}%
    \crefname{table}{table}{tables}%
    \crefname{theorem}{theorem}{theorems}%
    \crefname{lemma}{lemma}{lemmas}%
    \crefname{conjecture}{conjecture}{conjectures}%
    \crefname{corollary}{corollary}{corollaries}%
    \crefname{proposition}{proposition}{propositions}%
    \crefname{definition}{definition}{definitions}%
    \crefname{result}{result}{results}%
    \crefname{example}{example}{examples}%
    \crefname{remark}{remark}{remarks}%
    \crefname{note}{note}{notes}%
}

\usepackage{fancyhdr}
\renewcommand{\chaptermark}[1]{\markboth{\textit{\chaptername}\ \thechapter.\ #1}{}}

\renewcommand{\chaptermark}[1]{\markboth{#1}{}}



%% file: gloss.tex

\newglossaryentry{variables}{type=notation,
name=\ensuremath{$\xx, \yy, \zz$} (lower case roman letters),
description={single variables},
sort=1
}

\newglossaryentry{sets}{type=notation,
name=\ensuremath{$\XX, \YY,\CC$} (upper case caligraphic letters),
description={sets},
sort=2
}

\newglossaryentry{indsets}{type=notation,
name=\ensuremath{$\II, \JJ, \KK$} (upper case roman letters),
description={sets of variable indices},
sort=3
}

\newglossaryentry{func}{type=notation,
name=\ensuremath{$\ff, \qq, \pp$} (lower case sans serif letters),
description={functions over discrete domains (equivalent to arrays/tensors)},
sort=4
}

\newglossaryentry{functional}{type=notation,
name=\ensuremath{$\PP, \QQ, \pbp$} (upper case sans serif letters),
description={functionals},
sort=5
}

\newglossaryentry{tuples}{type=notation,
name=\ensuremath{\underline{x}, \underline{\mathsf{q}}, \underline{\mathsf{P}}} (underline),
description={``tuples'' of variables, functions or functionals},
sort=6
}

\newglossaryentry{matrix}{type=notation,
name=\ensuremath{\Dn, \Dt} (bold upper case roman letters),
description={Matrices},
sort=601
}

\newglossaryentry{messages}{type=notation,
name=\ensuremath{\msgq{i}{\mathrm{I}}, \MSGSST{\partial i}{\mathrm{I}},\msp{\mathrm{I}}{i}} (arrow in the subscript), 
description={(tuple of) ``messages'' as functions or functionals},
sort=602
}

\newglossaryentry{probs}{type=notation,
name=\ensuremath{\pp, \msg{\mathrm{I}}{i}, \MSG{I}{i}} (variants of letter $p$),
description={``normalized'' marginals or messages as functions or functionals},
sort=61
}


\newglossaryentry{complexityclasses}{type=notation,
name=\ensuremath{\Poly, \NP, \PPclass, \sharpP, \pspace} (blackboard bold letters),
description={complexity classes},
sort=7
}

\newglossaryentry{inffamily}{type=notation,
name=\ensuremath{\Sigma, \Phi, \Psi} (capital greek letters),
description={inference families},
sort=72
}











%% file: abstract.tex
%
%
%
\vspace{-.3in}
Graphical models use the intuitive and well-studied methods of graph theory to implicitly represent dependencies between variables in large systems. 
They can model the global behaviour of a complex system by specifying only local factors.This thesis studies inference in discrete graphical models from an ``algebraic perspective'' and the ways inference can be used to express and approximate \NP-hard combinatorial problems.

We investigate the complexity and reducibility of various inference problems, in part by organizing them in an inference hierarchy. We then investigate tractable approximations for a subset of these problems using distributive law in the form of message passing. The quality of the resulting message passing procedure, called Belief Propagation (BP), depends on the influence of loops in the graphical model. We contribute to three classes of approximations that improve BP for loopy graphs (I) loop correction techniques; (II) survey propagation, another message passing technique that surpasses BP in some settings; and (III) hybrid methods that interpolate between deterministic message passing and Markov Chain Monte Carlo inference.

We then review the existing message passing solutions and provide novel graphical models and inference techniques for combinatorial problems under three broad classes: (I) constraint satisfaction problems (CSPs) such as satisfiability, coloring, packing, set / clique-cover and dominating / independent set and their optimization counterparts; (II) clustering problems such as hierarchical clustering, K-median, K-clustering, K-center and modularity optimization; (III) problems over permutations including (bottleneck) assignment,  graph ``morphisms'' and alignment, finding symmetries and (bottleneck) traveling salesman problem. In many cases we show that message passing is able to find solutions that are either near optimal or favourably compare with today's state-of-the-art approaches.

%% file: acknowledgements.tex

Many have helped me survive and grow during my graduate studies. First, I would like to thank my supervisor Russell Greiner, for all I have learned from him, most of all for teaching me the value of common sense in academia and also for granting me a rare freedom in research. During these years, I had the chance to learn from many great minds in Alberta and elsewhere. It has been a pleasure working with David Wishart, Brendan Frey, Jack Tuszynski and Barnab\'{a}s P\'{o}czos. I am also grateful to my committee members, Dale Schuurmans, Csaba Szepesv\'{a}ri, Mohammad Salavatipour and my external examiner Cristopher Moore for their valuable feedback. 
 
I have enjoyed many friendly conversations with colleagues: Babak Alipanahi, Nasimeh Asgarian, Trent Bjorndahl, Kit Chen, Andrew Delong, Jason Grant, Bret Hoehn, Sheehan Khan, Philip Liu, Alireza Makhzani, Rupasri Mandal, James Neufeld, Christopher Srinivasa, Mike Wilson, Chun-nam Yu and others.

I am thankful to my friends: Amir, Amir-Mashoud, Amir-Ali, Amin, Arash, Arezoo, Azad, Babak(s), Fariba, Farzaneh, Hootan, Hossein, Kent, Kiana, Maria, Mariana, Meys, Meysam, Mohsen, Mohammad, Neda, Niousha, Pirooz, Sadaf, Saeed, Saman, Shaham, Sharron, Stuart, Yasin, Yavar and other for all the good time in Canada. Finally I like to thank my family and specially Reihaneh and Tare for their kindness and constant support. 

Here, I acknowledge the help and mental stimulation from the playful minds on the Internet, from Reddit to stackexchange and the funding and computational/technical support from computing science help desk, Alberta Innovates Technology Futures, Alberta Innovates Center for Machine Learning and Compute Canada.

%% file: chap1.tex



\chapter*{Introduction}
\addcontentsline{toc}{chapter}{Introduction}
{\Huge \bf M}any complicated systems can be modeled as a graphical structure with interacting local functions. Many fields have (almost independently) discovered this: graphical models have been used in
bioinformatics (protein folding, medical imaging and spectroscopy, pedagogy
trees, regulatory networks~\cite{yanover2002approximate,ravanbakhsh2014accurate,krogh2001predicting,beal2005bayesian,murphy1999modelling}),
neuroscience (formation of associative memory and neuroplasticity~\cite{churchland1992computational,amit1992modeling}), 
communication theory (low density parity check codes~\cite{tanner1981recursive,gallager1962low}), 
statistical physics (physics of dense matter and spin-glass theory~\cite{Mezard1987}), 
image processing (inpainting, stereo/texture reconstruction, denoising and super-resolution~\cite{felzenszwalb2006efficient,freeman2000learning}),
compressed sensing~\cite{donoho2009message}, robotics~\cite{thrun2002particle} (particle filters), sensor
networks~\cite{ihler2005nonparametric,crick2002loopy}, social networks~\cite{wasserman1994social,mccallum2005topic}, 
natural language processing~\cite{manning1999foundations}, 
speech recognition~\cite{cooke2010monaural,hershey2010super},
artificial intelligence (artificial neural networks, Bayesian networks~\cite{welling2004exponential,Pearl88})
and \magn{combinatorial optimization}. 
This thesis is concerned with the application of graphical models in 
solving \marnote{combinatorial problem}
combinatorial optimization problems~\cite{moore2011nature,grotschel1995combinatorial,steiglitz1982combinatorial}, which broadly put,
seeks \textit{an ``optimal'' assignment to a discrete set of variables, where a brute force 
approach is infeasible.}

To see how the decomposition offered by a graphical model can model a complex system, consider
 a joint distribution over $200$ binary variables. A naive way to represent this would require a table with $2^{200}$ entries. However if variables are
conditionally independent such that their dependence structure forms a tree, we
can exactly represent the joint distribution using only $200 \times 2^2$ values.
Operations such as marginalization, which require computation time linear in the
original size, are now reduced to local computation in the form of message
passing on this structure (\ie tree), which in this case, reduces the cost to linear
in the new exponentially smaller size. It turns out even if the dependency structure has loops, we can use message passing to perform ``approximate'' inference. 

\index{perspectives on inference}
Moreover, we approach the problem of inference from an algebraic point of view~\cite{Aji2000}.
This is in contrast to the variational perspective on local computation~\cite{Wainwright2007}.
These two perspectives are to some extent ``residuals'' from the different origins of research in AI and statistical physics.

\index{Boltzmann}
In the statistical study of physical systems, the Boltzmann distribution   
 relates the 
probability of each state of a physical system to its energy, 
which is often decomposed due to local interactions \cite{Mezard1987,Mezard09}.
These studies have been often interested in 
 modeling systems at the thermodynamic limit of infinite variables and the 
average behaviour through the study of random ensembles.  
Inference techniques with this origin (\eg mean-field and cavity methods) are often asymptotically exact under these assumptions. 
Most importantly these studies have reduced inference to optimization through the notion of free energy --\aka variational approach. 

In contrast, graphical models in the AI community have emerged in the study of knowledge representation
and reasoning under uncertainty \cite{pearl_probabilistic_1988}. These advances are characterized by  their attention to the 
theory of computation and logic \cite{bacchus1991representing}, where  interest in computational (as opposed to analytical) 
solutions has motivated the study of approximability, computational complexity \cite{Cooper90:Computational,roth1993hardness} and 
 invention of inference techniques such as belief propagation that are efficient and exact on tree structures.  
Also, these studies have lead to algebraic abstractions 
in modeling systems that allow local computation \cite{shenoy1990axioms,lauritzen1997local}.

The common foundation underlying these two approaches is information theory, 
where derivation of probabilistic principles from logical axioms \cite{jaynes2003probability} leads to
notions such as entropy and divergences that are closely linked to their physical counter-parts
\ie entropy and free energies in  physical systems. At a less abstract level,  it was shown that
inference techniques in AI and communication are attempting to minimize  (approximations to) free energy \cite{Yedidia2001a,Aji01}.

\index{critical phenomenon}
Another exchange of ideas between the two fields was in the study of critical phenomenon 
in random constraint satisfaction problems by both computer scientists and physicists \cite{fu_application_1986,mitchell1992hard,monasson1999determining};
satisfiability is at the heart of theory of computation and an important topic to investigate reasoning in AI. 
On the other hand, the study of critical phenomena and phase transitions 
is central in statistical physics of disordered systems. This was culminated when
a variational analysis lead to discovery of survey propagation~\cite{mezard_analytic_2002} for constraint satisfaction, which
significantly advanced the state-of-the-art in solving random satisfiability problems.

Despite this convergence, variational and algebraic perspectives are to some extent complementary
-- \eg the variational approach does not extend beyond (log) probabilities, while the algebraic approach cannot justify application of message passing to graphs with loops.
Although we briefly review the variational perspective, this thesis is mostly concerned with the algebraic perspective. In particular, rather than the study of phase transitions 
and the behaviour of the set of solutions for combinatorial problems, we are concerned with 
finding solutions to individual instances.

 \Cref{part1} starts by expressing the general form of inference, proposes a novel inference hierarchy and studies its complexity in \cref{chapter:background}.
\marnote{\cref{chapter:background}}
Here, we also show how some of these problems are reducible to others and introduce the algebraic structures that make efficient inference possible.
The general form of notation and the reductions that are proposed in this chapter are  used in later chapters.

\Cref{chapter:inference} studies some forms of approximate inference, by first introducing belief propagation.
\marnote{\cref{chapter:inference}}
It then considers the problems with intractably large number of factors and factors with large cardinality, then proposes/reviews solutions to both problems.
We then study different modes of inference as optimization and review alternatives such as convergent procedures and convex and linear programming relaxations for
some inference classes in the inference hierarchy.
Standard message passing using belief propagation is only guaranteed to be  exact if the graphical structure has no loops.
This optimization perspective (\aka\ variational perspective) has also led to design of approximate inference techniques that account for short loops in the graph.
A different family of loop correction techniques can account for long loops by taking message dependencies into account. 
This chapter reviews these methods and introduces a novel loop correction scheme that can account for both short and long loops, resulting in more accurate inference over difficult instances.

Message passing over loopy graphs can be seen as a fixed point iteration procedure, 
and the existence of loops means there may be more than one fixed point.
Therefore an alternative to loop correction is to in some way incorporate all fixed points. 
This can be performed also by a message passing procedure, known as  
survey propagation. The next section of this chapter
introduces survey propagation from a novel algebraic perspective that enables performing inference on the set of fixed points.
Another major approach to inference is offered by Markov Chain Monte Carlo (MCMC) techniques. After a minimal review of MCMC, the final section of this chapter
introduces a hybrid inference procedure, called perturbed belief propagation that interpolates between belief propagation and Gibbs sampling. We show that this technique can outperform both belief propagation and Gibbs sampling in particular settings.

\Cref{part2} of this thesis uses the inference techniques derived in the first part to solve a wide range of combinatorial problems.
We review the existing message passing solutions and provide novel formulations  for 
three broad classes of problems: 1) constraint satisfaction problems (CSPs), 2) clustering problems and 3) combinatorial problems over permutations.

In particular, in \cref{chapter:csp} we use perturbed belief propagation and perturbed survey propagation 
\marnote{\cref{chapter:csp}}
to obtain state-of-the-art performance in random satisfiability and coloring problems.
We also introduce novel message passing solutions and review the existing methods for sphere packing, set-cover, clique-cover, dominating-set and independent-set and several of their optimization counterparts.
By applying perturbed belief propagation to graphical representation of packing problem, we are able to compute long ``optimal'' nonlinear binary codes with large number of digits.

\Cref{chapter:clustering} proposes message passing solutions to several clustering problems such as K-clustering, K-center and Modularity optimization and shows that message passing is able
to find near-optimal solutions on moderate instances of these problems.
Here, we also review the previous approaches 
\marnote{\cref{chapter:clustering}}
to K-median and hierarchical clustering and also the related graphical models for minimum spanning tree and prize-collecting Steiner tree. 

\Cref{chapter:permutations} deals with combinatorial problems over permutations, by first reviewing the existing 
\marnote{\cref{chapter:permutations}}
graphical models for matching, approximation of permanent, and graph alignment and
introducing two novel message passing solutions for min-sum and min-max versions of traveling salesman problem (\aka bottleneck TSP).
We then study  graph matching problems, including (sub-)graph isomorphism, monomorphism, homomorphism, graph alignment and ``approximate'' symmetries.
In particular, in the study of graph homomorphism we show that its graphical model generalizes that of  of several other problems, including Hamiltonian cycle, clique problem and
coloring. We further show how graph homomorphism can be used as a
surrogate for isomorphism to find symmetries.

\section*{Contributions and acknowledgment}
\addcontentsline{toc}{section}{Contributions}
The results in this thesis are a joint work with my supervisor Dr.~Greiner and other researchers.
In detail, 
the algebraic approach to inference is presented in~\cite{ravanbakhsh_algebra};
the loop correction ideas are published in \cite{ravanbakhsh_loop}; perturbation schemes for CSP are presented in \cite{ravanbakhsh_csp};
performing min-max inference was first suggested by Dr.~Brendan Frey and Christopher Srinivasa, and many of the related ideas including min-max reductions 
are presented in our joint paper \cite{ravanbakhsh_minmax}.
Finally, the augmentation scheme for TSP and Modularity maximization is discussed in \cite{ravanbakhsh_augmentation}.

The contribution of this thesis, including all the published work is as follows:
\begin{easylist}
& Generalization of inference problems in graphical models including: 
&& The inference hierarchy. 
&& The limit of distributive law on tree structures.
&& All the theorems, propositions and claims on complexity of inference, including
&&& \NP-hardness of inference in general commutative semirings.
& A unified treatment of different modes of inference over factor-graphs and identification of their key properties (\eg significance of inverse operator) in several settings including: 
&& Loop correction schemes.
&& Survey propagation equations.
& Reduction of min-max inference to min-sum and sum-product inference.
& Simplified form of loop correction in Markov networks and their generalization to incorporate short loops over regions.
& A novel algebraic perspective on survey propagation.
& Perturbed BP and perturbed SP and their application to constraint satisfaction problems.
& Factor-graph augmentation for inference over intractably large number of constraints.
& Factor-graph formulation for several combinatorial problems including 
&& Clique-cover.
&& Independent-set, set-cover and vertex cover.
&& Dominating-set and packing (the binary-variable model) 
&& Packing with hamming distances 
&& K-center problem, K-clustering and clique model for modularity optimization.
&& TSP and bottleneck TSP. 
& The general framework for study of graph matching, including
&& Subgraph isomorphism.\footnote{Although some previous work~\cite{bradde2010aligning} claim to address the same problem, we note that their formulation is for sub-graph monomorphism rather than isomorphism.} 
&& Study of message passing for Homomorphism and finding approximate symmetries. 
&& Graph alignment with a diverse set of penalties.
\end{easylist}

%% file: chap2_new.tex
\part{Inference by message passing}\label{part1}
This part of the thesis first studies the representation formalism, hierarchy of inference problems, reducibilities and the underlying algebraic structure
that allows efficient inference in the form of message passing in graphical models, in \cref{chapter:background}.
By viewing inference under different lights, we then review/introduce procedures that allow better approximations in \cref{chapter:inference}.

\chapter{Representation, complexity and reducibility}\label{chapter:background}
In this chapter, we use a simple algebraic structure  -- \ie commutative semigroup -- 
to express a general form for inference in graphical models.
To this end, we first introduce the factor-graph representation and formalize inference in \refSection{sec:inference}.
\RefSection{sec:hierarchy} focuses on four operations defined by summation, multiplication, minimization 
and maximization,  to construct a hierarchy of inference problems within \pspace, such that the problems in the 
same class of the hierarchy belong to the same complexity class.
Here, we encounter some new inference problems and establish the completeness of problems at lower
levels of hierarchy \wrt their complexity classes.
In \refSection{sec:gdl} we augment our simple structures with  two properties to obtain message passing on commutative semirings. Here, we also observe that replacing a semigroup with an Abelian group, gives us normalized marginalization as a form of inference inquiry. 
Here, we show that inference in any
commutative semiring is \NP-hard and postpone further investigation of message passing 
to the next chapter.
\RefSection{sec:reductions} shows how some of the inference problems introduced so far are reducible to others.

\section{The problem of inference}\label{sec:inference}
We use commutative semigroups to both define what a graphical 
model represents and also to define inference over this graphical model. 
The idea of using structures such as semigroups, monoids and semirings in expressing inference has a long history\cite{lauritzen1997local,schiex1995valued,bistarelli1999semiring}. 
Our approach, based on factor-graphs~\cite{kschischang_factor_2001} 
and commutative semigroups, generalizes
a variety of previous frameworks, including
Markov networks~\cite{clifford1990markov}, 
Bayesian networks~\cite{pearl1985bayesian}, 
Forney graphs~\cite{forney2001codes}, hybrid models~\cite{dechter2001hybrid}, 
influence diagrams \cite{howard2005influence} and valuation
networks \cite{shenoy1992valuation}. 
\index{graphical models}

In particular, the combination of factor-graphs and
semigroups that we consider here  
generalizes the plausibility, feasibility and utility framework of \citet{pralet2007algebraic},
which is explicitly reduced to  the graphical models mentioned above and many more.
The main difference in our approach is in keeping the framework free of semantics (\eg 
decision and chance variables, utilities, constraints),
that are often associated with variables, factors and operations, without changing the
expressive power.
These notions can later be associated with individual inference problems
 to help with interpretation.

\index{semigroup} \index{semigroup!commutative semiring}
\index{monoid} \index{semigroup!commutative monoid}
\index{Abelian group}
\begin{definition}\label{def:semigroup}
A \magn{commutative semigroup} is a pair $\semig = (\RR, \otimes)$, where $\RR$ is a set and $\otimes: \RR \times \RR \to \RR$ is a binary operation that is 
(I) associative: $ a \otimes (b \otimes c) = (a \otimes b) \otimes c$ and (II) commutative: $a \otimes b = b \otimes a$ for all $a, b, c \in \RR$.
A \magn{commutative monoid} is a commutative semigroup plus an identity element $\identt{\otimes}$
such that  $a \otimes \identt{\otimes} = a$. 
If every element $a \in \RR$ has an inverse $a^{-1}$ (often written $\frac{1}{a}$), 
 such that $a \otimes a^{-1} = \identt{\otimes}$, 
and $a \otimes \identt{\otimes} = a$,
the commutative monoid is an \magn{Abelian group}.
\end{definition}
Here, the associativity and commutativity properties of a commutative semigroup make the operations invariant to the order of elements.  In general, these properties are not 
``vital'' and one may define inference starting from a \textit{magma}.\footnote{A \index{magma} magma \cite{pinter2012book} generalizes a semigroup, as it does not require associativity property nor an identity element. Inference in graphical models can be also extended to use magma (in \cref{def:fg}). For this, 
the elements of $\RR$ and/or $\XX$ should be ordered and/or parenthesized so as to avoid ambiguity in the order of pairwise operations over the set. Here, to avoid unnecessary complications, we confine our treatment to commutative semigroups.}

\begin{example} Some examples of semigroups are:
\begin{easylist}
& The set of strings with the concatenation operation forms a semigroup with the 
empty string as the identity element. However this semigroup is not commutative.
& The set of natural numbers $\settype{N}$ with summation defines a commutative semigroup. 
& Integers modulo $n$ with addition defines an Abelian group.
& The power-set $2^{\SS}$ of any set $\SS$, with intersection operation defines a commutative semigroup with $\SS$ as its identity element.
& The set of natural numbers with greatest common divisor defines a commutative monoid with $0$ as its identity. In fact any semilattice is a commutative semigroup~\cite{davey2002introduction}.
& Given two commutative semigroups on two sets $\RR$ and $\settype{Z}^*$, their Cartesian product is also a commutative semigroup.
\end{easylist}
\end{example}

\index{variable}
\index{factor}
Let $\xs = (\xx_1,\ldots,\xx_N)$ be a tuple of $N$ discrete
variables $\xx_i\in \XX_i$, where $\XX_i$ is the domain of
$\xx_i$  and $\xs \in \XX =  \XX_1 \times \ldots \times \XX_N$.  
Let $\II \subseteq \NN = \{1,2,\ldots,N\}$ denote a subset
of variable indices and $\xs_\II \!=\! \{ \xx_i\! \mid\! i\in \II\} \in \XX_\II$ be the
tuple of variables in $\xs$ indexed by the subset $\II$.
A factor $\ff_{\II}: \XX_{\II} \to \YY_\II$ is a function over a subset of variables and 
$\YY_\II = \{ \ff_\II(\xs_\II) \mid \xs_\II \in \XX_\II\}$ is the range of this factor. 
\index{factor!range}

\begin{definition}\label{def:fg}
\index{factor-graph}
 A \magn{factor-graph}  is a pair $(\FF, \semig)$  such that
\begin{itemize}
\item $\FF = \{ \ff_{\II}\}$ is a collection of factors with collective range  $\YY = \bigcup_\II \YY_\II$.
\item  $|\FF| = \mathrm{Poly}(N)$.  
\item $\ff_\II$ has a polynomial representation in $N$ and it is possible to evaluate $\ff_\II(\xs_\II)\; \forall \II, \xs_\II$ in polynomial time.
\item $\semig = (\RR, \otimes)$ is a commutative semigroup, where \index{closure} $\RR$ is the closure of $\YY$ w.r.t. $\otimes$.
\end{itemize}
\index{expanded form}
\index{joint form}
The factor-graph compactly represents the expanded (joint) form
\begin{align}\label{eq:expanded}
  \qq(\xs) = \bigotimes_\II \ff_\II(\xs_\II)
\end{align}
\end{definition}


Note that the connection between the set of factors $\FF$ and the commutative semigroup is through the ``range'' \index{factor!range} of factors.
The conditions of this definition are necessary and sufficient to 1) compactly represent a factor-graph and 2) evaluate the expanded form, $\qq(\xs)$, in polynomial time.
A stronger condition to ensure that a factor has a compact representation is $|\XX_\II| = \mathrm{Poly}(N)$, which means  $\ff_\II(\xs_\II)$ can be explicitly expressed for each $\xs_\II \in \XX_\II$ as 
an $|\II|$-dimensional array.



\begin{figure}
\centering
\includegraphics[width=.5\textwidth]{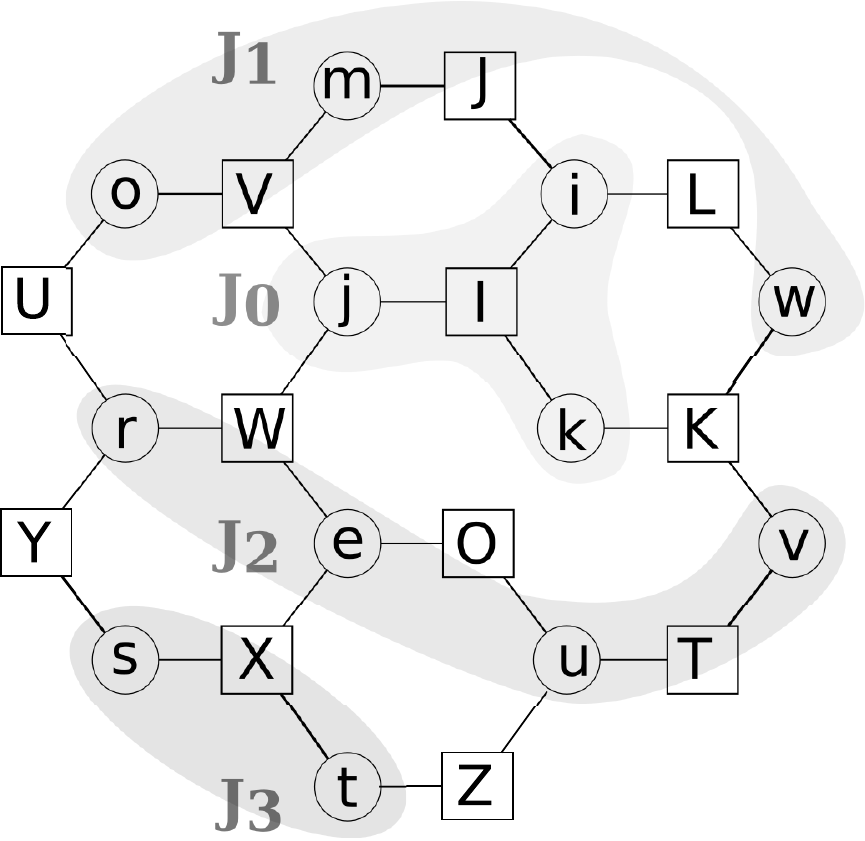}
\caption{A factor-graph with variables as circles and factors as squares.}
\label{fig:fg}
\end{figure}

$\FF$ can be conveniently represented as a bipartite graph 
\index{bipartite graphical model}
that includes two sets of nodes: variable nodes $\xx_i$, and factor 
 nodes ${\II}$. A variable node $i$ (note that we will often
identify a variable $\xx_i$ with its index ``$i$'') is connected to a
factor node $\II$ if and only if $i \in \II$ --\ie $\II$ is a set that is also an index. 
We will use $\nb$ to denote the neighbours of a variable or factor node in the factor
\index{variable!neighbours}
graph -- that is $\nb \II = \{ i \; \mid \; i \in \II \}$ (which is
the set $\II$) and $\nb i = \{ \II \; \mid \; i \in \II \}$.
\index{Markov blanket}
Also, we use $\mb i$ to denote the \magn{Markov blanket} of node $\xx_i$ -- \ie $\mb i 
= \{ j \in \nb \II\; \mid \; \II \in \nb i,\; j \neq i\}$.
\begin{example}\label{example:fig1}
\Cref{fig:fg} shows a factor-graph with 12 variables and 12 factors.
Here $\xs = (\xx_i, \xx_j,\xx_k, \xx_e, \xx_m, \xx_o, \xx_r, \xx_s, \xx_t, \xx_u, \xx_v, \xx_w)$, $\II = \nb \II =  \{i,j,k\}$,  $\xs_\mathrm{K} = \xs_{\{k,w,v\}}$
and $\nb j = \{\II, \mathrm{V}, \mathrm{W}\}$. 
Assuming $\semig_e = (\Re, \min)$, the expanded
form represents  $$\qq(\xs) = \min \{ \ff_{\II}(\xs_\II), \ff_{\JJ}(\xs_\JJ),\ldots,\ff_{\mathrm{Z}}(\xs_{\mathrm{Z}}) \}.$$

Now, assume that all variables are binary -- \ie $\XX = \{0,1\}^{12}$ and $\qq(\xs)$ is 12-dimensional hypercube, with one assignment at each corner. Also assume
all the factors count the number of non-zero variables -- \eg for $\zs_{\mathrm{W}} = (1,0,1) \in \XX_{\mathrm{W}}$ we have $\ff_{\mathrm{W}}(\zs_{\mathrm{W}}) = 2$. Then, for the complete assignment 
$\zs = (0,1,0,1,0,1,0,1,0,1,0,1) \in \XX$, it is easy to check that the expanded form is 
$\qq(\zs) = \min \{2, 0, 1,\ldots, 1\} = 0$. 
\end{example}

\index{marginalization}
A marginalization operation shrinks the expanded form $\qq(\xs)$ using another commutative semigroup with binary operation $\oplus$.
Inference is a combination of an expansion and one or more marginalization operations,
which can be computationally intractable due to the exponential size of the expanded form.
\begin{definition} 
Given a function $\qq: \XX_{\JJ} \to \YY$, and a commutative semigroup $\semig = (\RR, \oplus)$, where $\RR$ is the closure of $\YY$ w.r.t. $\oplus$,
 the marginal of $\qq$ for $\II \subset \JJ$ is 
\begin{align}\label{eq:marginalization}
 \qq(\xs_{\JJ \back \II}) \quad \defeq \quad \bigoplus_{\xs_{\II}} \qq(\xs_{\JJ}) 
\end{align}
where $\bigoplus_{\xs_{\II}} \qq(\xs_{\JJ})$ is short for $\bigoplus_{\xs_{\II} \in \XX_{\II}} \qq(\xs_{\JJ \back \II}, \xs_{\II})$, 
and it means to compute $\qq(\xs_{\JJ \back \II})$ for each $\xs_{\JJ \back \II}$, one should perform the operation $\oplus$
over the set of all the assignments to the tuple $\xs_{\II} \in \XX_{\II}$.
\end{definition}
We can think of $\qq(\xs_\JJ)$ as a $\vert \JJ \vert$-dimensional tensor and marginalization as performing $\oplus$ operation over the axes
in the set $\II$. The result is another $\vert \JJ \back \II \vert$-dimensional tensor (or function) that we call the \magn{marginal}. 
Here if the marginalization is over all the dimensions in $\JJ$, we denote the marginal by $\qq(\emptyset)$ instead of $\qq(\xs_\emptyset)$ and call it the \magn{integral} of $\qq$.
\index{integration}

Now we define an inference problem as a sequence of marginalizations over the expanded form of a factor-graph.
\index{inference!problem}
\index{marginal}
\index{integral}
\begin{definition}\label{def:inference}
An \magn{inference problem} seeks
  \begin{align}
    \qq(\xs_{\JJ_0}) = \bigopp{M}_{\xs_{\JJ_M}} \bigopp{M-1}_{\xs_{\JJ_{M-1}}} \ldots \bigopp{1}_{\xs_{\JJ_1}} \bigotimes_{\II} \ff_\II(\xs_\II) 
  \end{align}
where
  \begin{itemize}
  \item \index{closure} $\RR$ is the closure of $\YY$ (the \index{collective range|see {factor!range}} collective range of factors), w.r.t. $\opp{1},\ldots,\opp{M}$ and $\otimes$. 
  \item $\semig_m = (\RR, \opp{m})\; \forall 1 \leq m \leq M$ and $\semig_e=(\RR, \otimes)$ are all  commutative semigroups.
\item $\JJ_0,\ldots,\JJ_L$ partition the set of variable indices $\NN = \{1,\ldots,N\}$. 
\item $\qq(\xs_{\JJ_0})$ has a polynomial representation in $N$ -- \ie $|\XX_{\JJ_0}| = \mathrm{Poly}(N)$ 
\end{itemize}
\end{definition}

Note that $\opp{1},\ldots,\opp{M}$  refer to potentially different operations as each belongs to a different semigroup.
When $\JJ_0 = \emptyset$, we call the inference problem \magn{integration} (denoting the inquiry by $\qq(\emptyset)$) and otherwise we call it \magn{marginalization}.
Here, having a constant sized $\JJ_0$ is not always enough to ensure that $\qq(\xs_{\JJ_0})$  has a polynomial representation in $N$. This is because 
the size of $\qq(\xs_{\JJ_0})$ for any individual $\xs_{\JJ_0} \in \XX_{\JJ_0}$ may grow exponentially with $N$ (\eg see \cref{th:product_pspace}).
In the following we call $\semig_e = (\RR, \otimes)$ the expansion semigroup and $\semig_m = (\RR, \opp{m}) \; \forall 1 \leq m \leq M$
the marginalization semigroup.\index{semigroup!expansion} \index{semigroup!marginalization}
\index{marginalization!semigroup} \index{expansion semigroup}


\begin{example}\label{example:fig2}
Going back to \refExample{example:fig2}, the shaded region in \refFigure{fig:fg}
shows a partitioning of the variables that we use to define the following inference problem:
$$
\qq(\xs_{\JJ_0}) = \max_{\xs_{\JJ_3}} \sum_{\xs_{\JJ_2}} \min_{\xs_{\JJ_1}} \min_\II \ff_\II(\xs_\II) 
$$
We can associate this problem with the following semantics: we may 
think of each factor as an agent, where
$\ff_{\II}(\xs_\II)$ is the payoff for agent $\II$, which only depends
on a subset of variables $\xs_\II$. 
We have adversarial variables ($\xs_{\JJ_1}$), environmental or chance variables ($\xs_{\JJ_2}$), controlled variables ($\xs_{\JJ_3}$) and query variables ($\xs_{\JJ_0}$).  
The inference problem above  for 
each query $\xs_{\JJ_0}$ seeks to maximize 
the expected minimum payoff of all agents,
without observing the adversarial or chance variables, and 
assuming the adversary makes its decision after observing control and chance variables.
\end{example}

\begin{example}\label{example:ising}
\index{probabilistic graphical model}
\index{graphical models!probabilistic}

A ``probabilistic'' graphical model is defined using a expansion semigroup $\semig_e = (\Re^{\geq 0}, \times)$ and often a marginalization semigroup $\semig_m = (\Re^{\geq 0}, +)$. The expanded form represents the unnormalized joint probability
$\qq(\xs) = \prod_{\II}\ff_{\II}(\xs_{\II})$, whose marginal probabilities are simply called marginals.
Replacing the summation with marginalization semigroup $\semig_m = (\Re^{\geq 0}, \max)$, seeks the
 maximum probability state and the resulting integration problem 
$\qq(\emptyset) =  \max_{\xs} \prod_{\II} \ff_\II(\xs_\II)$
is known as \magn{maximum a posteriori (MAP)} inference. 
\index{MAP}
\index{maximum a posteriori|see{MAP}}
\index{max-product}
 Alternatively by adding a second marginalization operation to the summation, we get the \magn{marginal MAP} inference
\index{marginal MAP}
\index{max-sum-product}
\begin{align}\label{eq:marginalmap}
    \qq(\xs_{\JJ_0}) \quad = \quad \max_{\xs_{\JJ_2}} \sum_{\xs_{\JJ_1}} \prod_{\II} \ff_{\II}(\xs_\II).
\end{align}
where here $\bigotimes = \prod$, $\bigopp{1} = \sum$ and $\bigopp{2} = \max$.

If the object of interest is the negative 
\index{energy}
log-probability (\aka energy), the product expansion semigroup is
replaced by $\semig_e=(\Re, +)$. Instead of sum marginalization semigroup, we can
\index{log-sum-exp semigroup}
\index{semigroup!log-sum-exp}
use the \magn{log-sum-exp} semigroup, $\semig_m = (\Re, +)$ where 
$a \oplus b \defeq \log(e^{-a} + e^{-b})$. The integral in this case is the log-partition function.
If we change the marginalization semigroup to $\semig_m = (\Re, \min)$, the integral is the minimum energy (corresponding to MAP).

A well-known example of a probabilistic graphical model is the 
\index{Ising}
 \magn{Ising model} of ferromagnetism. This model is an
extensively studied in physics, mainly to model the phase transition
in magnets.
The model consists of binary variables ($\xx_i \in \{-1,1\}$) -- denoting magnet spins --
arranged on the nodes of a graph $\GG = (\VV, \EE)$ (usually a grid or Cayley tree). 
\index{Hamiltonian}
\index{Boltzmann}
\index{energy function}
The energy function (\ie Hamiltonian) associated with a 
\index{configuration} configuration $\xs$ is the joint form 
\marnote{energy function}
\begin{align}\label{eq:isinghamiltonian}
\energy(\xs) \quad = \quad \qq(\xs) \quad = \quad - \sum_{(i,j) \in \EE}  \xx_i \; \interaction_{ij}\; \xx_j  - \sum_{i
\in \VV} \localfield_{i}
\end{align}
\index{variable interaction}
Variable interactions are denoted by $\interaction$ and $\localfield$ is called the
\index{local field} local field. Here each $\interaction_{i,j}$ defines a factor over $\xx_i, \xx_j$: $\ff_{\{i,j\}}(\xs_{\{i,j\}}) = -\xx_i \; \interaction_{ij}\; \xx_j$ 
and local fields define \index{local factors} local factors $\ff_{\{i\}}(\xx_i) = - \localfield_i \xx_i$.

Depending on the type of interactions, we call the resulting Ising model:\\
\noindent \index{ferromagnetic}
\bullitem\ \magn{ferromagnetic}, if all $\interaction_{ij} > 0$. In this setting, neighbouring
  variables are likely to take similar values.
\\
\noindent\index{anti-ferromagnetic} \bullitem \magn{anti-ferromagnetic}, if all $\interaction_{ij} < 0$.
\\ \noindent
\bullitem non-ferromagnetic, if both kind of interactions are allowed. In
  particular, if the ferromagnetic and anti-ferromagnetic interactions have
  comparable frequency, the model is called \magn{spin-glass}. \index{spin-glass}
\marnote{spin-glass}
  This class of problem shows most interesting behaviours, which is not completely
  understood~\cite{mezardspin}. 
As we will see, the studied phenomena in these materials have
  important connections to difficult inference
  problems including combinatorial optimization
  problems.
  Two well studied models of spin glass are 
  Edward-Anderson (EA \cite{Edwards1975}) and Sherrington-Kirkpatrick (SK \cite{sherrington1975solvable}) models.
  While the EA model is defined on a grid (\ie spin-glass interactions over a
  grid), the SK model is a complete graph.

\end{example}

\section{The inference hierarchy}\label{sec:hierarchy}
\index{inference!decision}
Often, the complexity class is concerned with the \magn{decision version} of the 
\index{decision problem}
inference problem in \refDefinition{def:inference}. The decision version of an inference problem asks a yes/no question about the integral: $\qq(\emptyset) \overset{?}{\geq} q$ for a given $q$. 



\index{inference!hierarchy}
\index{inference hierarchy}

Here, we produce a hierarchy of inference problems  
in analogy to polynomial~\cite{stockmeyer1976polynomial}, the counting~\cite{wagner1986complexity} and  arithmetic~\cite{rogers1987theory} hierarchies. 

To define the hierarchy, we 
assume the following in \refDefinition{def:inference}:
\begin{easylist}
& Any two consecutive marginalization operations are distinct ($\opp{l} \neq \opp{{l+1}} \; \forall 1 \leq l < M$).
& The marginalization index sets $\JJ_l \; \forall 1 \leq l \leq M$ are non-empty. Moreover if 
\index{polynomial marginalization}
\index{marginalization!polynomial}
$| \JJ_l| = \OO(\log(N))$ we call this marginalization operation a \magn{polynomial marginalization} as here $| \XX_{\JJ_{l}} | = \mathrm{Poly}(N)$. 
& In defining the factor-graph, we required each factor to be polynomially computable.
In building the hierarchy, we require the operations over each semigroup to be polynomially computable as well. To this end we consider 
the set of rational numbers $\RR \subseteq \Qe^{\geq 0} \cup \{\pm \infty\}$. Note that this automatically eliminates semigroups that involve operations such as exponentiation and logarithm (because $\Qe$ is not closed under these operations) and only consider summation, product, minimization
and maximization. 
\end{easylist}
We can always re-express any inference problem to enforce the first two conditions and therefore they do not impose any restriction. In the following we will use the a \index{inference!language} \magn{language} to identify inference problems for an arbitrary set of factors $\FF = \{\ff_\II\}$. For example, sum-product refers to the inference problem $\sum_{\xs} \prod_\II \ff_\II(\xs_\II) \overset{?}{\geq} q$.
In this sense the rightmost ``token'' in the language (here \textit{product}) identifies the expansion semigroup $\semig_e = (\mathbb{Q}, \prod)$ and the rest of tokens identify the marginalization semigroups over $\mathbb{Q}$ in the given order. Therefore, this minimal language exactly identifies the inference problem. The only information that affects the computational complexity of an inference
problem but is not specified in this language is whether each of the marginalization operations are polynomial or exponential.

\index{inference!hierarchy!families}
We define five \magn{inference families}: $\Sigma, \Pi, \Phi, \Psi, \Delta$. The families
are associated with that ``outermost'' marginalization operation  -- \ie $\opp{M}$ in \refDefinition{def:inference}). 
$\Sigma$ is the family of
inference problems where $\opp{M} = \sumop$. Similarly,
$\Pi$ is associated with product, $\Phi$ with minimization and $\Psi$ with maximization.
$\Delta$ is the family of inference problems where the last marginalization is polynomial (\ie $|\JJ_M| = \OO(\log(N))$ regardless of $\opp{M}$).

\index{inference!hierarchy!classes}
Now we define \magn{inference classes} in each family, such that all the problems in the same class
have the same computational complexity. Here, the hierarchy is exhaustive -- \ie
 it includes all inference problems with four operations sum, min, max and product
 whenever the integral $\qq(\emptyset)$ has a polynomial representation (see \cref{th:product_pspace}). 
 Moreover the inference classes are disjoint.
 For this, each family
is parameterized by a subscript $M$ and two sets $\sumset$ and $\pset$ (\eg $\Phi_M(\sumset, \pset)$ is an inference ``class'' in family $\Phi$). As before, $M$
is the number of marginalization operations, $\sumset$ is the set of indices
of the (exponential) $\sumop$-marginalization and $\pset$ is the set of indices of
polynomial marginalizations.
\begin{example}
Sum-min-sum-product identifies the decision problem
\begin{align*}
\sum_{\xs_{\JJ_3}} \min_{\xs_{\JJ_2}} \sum_{\xs_{\JJ_1}} \prod_{\II} \ff_\II(\xs_\II) \overset{?}{\geq} q
\end{align*}
where $\JJ_1$, $\JJ_2$ and $\JJ_3$ partition $\settype{N}$. Assume
$\JJ_1 = \{2,\ldots,\frac{N}{2}\}$, $\JJ_2 = \{\frac{N}{2} + 1,\ldots, N\}$ and $\JJ_3 = \{1\}$. Since we have three marginalization operations $M = 3$.
Here the first and second marginalizations are exponential and the third one is polynomial (since $|\JJ_3|$ is constant). Therefore $\pset = \{3\}$.
Since the only exponential summation is $\bigopp{1}_{\xs_{\JJ_1}} = \sum_{\xs_{\JJ_1}}$,
$\sumset = \{1\}$.
In our inference hierarchy, this problem belongs to the class $\Delta_{3}( \{1\}, \{3\})$. 

Alternatively, if we use different values for $\JJ_1$, $\JJ_2$ and $\JJ_3$ that all linearly grow with $N$,
the corresponding inference problem becomes a member of  $\Sigma_3(\{1,3\}, \emptyset)$.
\end{example}
\begin{remark}
\index{inference!hierarchy!valid class}
Note that arbitrary assignments to $M$, $\sumset$ and $\pset$ do not necessarily define a valid inference class.
For example we require that $\sumset \cap \pset = \emptyset$ and no index in $\pset$ and $\sumset$
are is larger than $M$. Moreover, the values in $\sumset$ and $\pset$ should be compatible with the inference class. For example, for inference class $\Sigma_M(\sumset,\pset)$, $M$ is a member of $\sumset$.
For notational convenience, if an inference class notation is invalid we equate it with an empty set -- \eg $\Psi_1(\{1\}, \emptyset) = \emptyset$, because $\sumset = \{1\}$ and $M=1$ means the inference class is $\Sigma$ rather than $\Psi$. 
\end{remark}

In the definition below, we ignore the inference problems in which product
appears in any of the marginalization semigroups (\eg product-sum).
\index{marginalization!product}
The following claim, explains this choice. 
\begin{claim}\label{th:product_pspace}
For $\oplus_M = \prodop$, the inference query $\qq(\xs_{\JJ_0})$ can have an exponential representation in $N$. 
\end{claim}
\begin{proof}
The claim states that when the product appears in the marginalization operations, the marginal (and integral) can become very large, such that we can no longer represent them in polynomial space in $N$.
 We show this for an integration problem. The same idea can show the exponential representation of a marginal query. 

To see why this integral has an exponential representation in $N$, consider its simplified form
\begin{align*}
  \qq(\emptyset) = \prod_{\xs_{\II}} \qq(\xs_{\II})
\end{align*}
where $\qq(\xs)$ here is the result of inference up to the last marginalization step $\opp{M}$, which is product, where $\XX_\II$ grows exponentially with $N$. 
Recall that the hierarchy is defined for operations on $\mathbb{Q}^{\geq 0}$. Since $\qq(\xs_\II)$ for each $\xs_\II \in \XX_\II$ 
has a constant size, say  $c$, the size of representation of $\qq(\emptyset)$ using a binary scheme is 
\begin{align*}
  \left \lceil \log_2(\qq(\emptyset)) \right \rceil = \left \lceil \log_{2} \big ( \prod_{\xs_{\II}} \qq(\xs_{\II}) \big ) \right \rceil = \left \lceil \sum_{\xs_{\II}} c \right \rceil = \left \lceil c \vert \XX_\II \vert \right \rceil
\end{align*}
which is exponential in $N$.
\end{proof}

\index{inference!hierarchy!base members}
Define the \magn{base members} of families as
\begin{align} \label{eq:basemembers}
\Sigma_0(\emptyset, \emptyset) \defeq \{\sumop\} \quad &
\Phi_0(\emptyset, \emptyset) \defeq \{ \min \} \\
\Psi_0(\emptyset, \emptyset) \defeq \{ \max \} \quad &
\Pi_0(\emptyset, \emptyset) \defeq \{ \prodop \} \notag \\
\Delta_0(\emptyset, \emptyset) = \emptyset \quad &\Delta_{1}(\emptyset, \{1\}) \defeq \{\sumop-\sumop, \min-\min, \max-\max \} \notag
\end{align}
where the initial members of each family only identify the expansion semigroup -- \eg $\sumop$ in $\Sigma_0(\emptyset, \emptyset)$ identifies $\qq(\xs) = \sum_{\II} \ff_{\II}(\xs_\II)$.  
Here, the exception is $\Delta_1(\emptyset, \{1\})$, which contains three \textit{inference problems}.\footnote{We treat $M=1$ for $\Delta$ specially as in this case the marginalization operation can not be polynomial. This is because if $|\JJ_1| = \OO(\log(N))$, then $|\JJ_0| = \Omega(N)$ which violates the conditions in the definition of the inference problem. 
}

Let $\Xi_M(\sumset, \pset)$ denote the union of corresponding classes within all families:
\begin{align*}
  \Xi_M(\sumset, \pset) = \Sigma_M(\sumset, \pset) \cup \Pi_M(\sumset, \pset) \cup \Phi_M(\sumset, \pset) \cup \Psi_M(\sumset, \pset) \cup \Delta_M(\sumset, \pset)
  \end{align*}
Now define the \magn{inference family members} recursively, by adding a marginalization operation to all the problems in each inference class. If this marginalization is polynomial then the new class belongs to the $\Delta$ family and the set $\pset$ is updated accordingly. Alternatively, if this outermost marginalization is exponential, depending on the new marginal operation (\ie $\min, \max, \sumop$) the new class is defined to be a member of $\Phi, \Psi$ or $\Sigma$. For the case that the last marginalization is summation set $\sumset$ is updated.

\index{inference!hierarchy!recursion}
\noindent {\bullitem\ \textbf{Adding an exponential marginalization}} $ \forall \; |\XX_{\JJ_M}| = \mathrm{Poly}(N), M > 0$
\begin{align}
\Sigma_{M+1}(\sumset \cup \{M+1\}, \pset) &\defeq \big \{ \sumop - \xi \mid \xi \in  \Xi_M(\sumset, \pset) \back \Sigma_{M}(\sumset, \pset)  \} \label{eq:hierarchymembers1}\\
\Phi_{M+1}(\sumset, \pset) &\defeq \big \{ \min - \xi \mid \xi \in \Xi_M(\sumset, \pset) \back \Phi_{M}(\sumset, \pset) \big \} \notag \\
\Psi_{M+1}(\sumset, \pset) &\defeq \big \{ \max - \xi \mid \xi \in \Xi_M(\sumset, \pset) \back \Psi_{M}(\sumset, \pset) \big \} \notag \\
\Pi_{M+1}(\sumset, \pset) &\defeq  \emptyset \notag 
\end{align}

\noindent \bullitem\ \textbf{Adding a polynomial marginalization} $\quad \forall \; |\XX_{\JJ_M}| =  \mathrm{Poly}(N), M > 1$  
  \begin{align}\label{eq:hierarchymembers2}
\Delta_{M+1}(\sumset, \pset \cup \{M+1\}) &\defeq \big \{ \oplus-\xi \mid 
 \xi \in  \Xi_M(\sumset, \pset) \; , \oplus \in \{\min,\max,\sumop \} \big \}
\end{align}


\subsection{Single marginalization}
The inference classes in the hierarchy with one marginalization are 
\begin{align}
\Delta_1(\emptyset, \{1\}) &= \{\min-\min,\; \max-\max,\; \sumop-\sumop\}\\
\Psi_1(\emptyset, \emptyset)  &= \{\max-\min,\; \max-\sumop,\; \max-\prodop\} \\
\Phi_1(\emptyset, \emptyset)  &= \{\min-\max,\; \min-\sumop,\; \min-\prodop\} \\
\Sigma_1(\{1\}, \emptyset)  &=  \{ \sumop-\prodop,\; \sumop-\min,\; \sumop-\max \}
\end{align}

Now we review all the problems above and prove that $\Delta_1, \Psi_1, \Phi_1$ and $\Sigma_1$ are complete w.r.t. \poly, \NP, \coNP\ and \PPclass\ respectively.  
Starting from $\Delta_1$:
\index{polynomial inference}
\index{inference!polynomial}
\index{complexity!sum-sum}
\index{complexity!min-min}
\begin{proposition}\label{th:sumsum_poly}
sum-sum, min-min and max-max inference are in \poly.
\end{proposition}
\begin{proof}
To show that these inference problems are in \poly, we 
provide polynomial-time algorithms for them:

\noindent \bullitem\ $\sumop-\sumop$ is short for
\begin{align*}
  \qq(\emptyset) = \sum_{\xs} \sum_{\II} \ff_\II(\xs_\II)
\end{align*}
which asks for the sum over all assignments of $\xs \in \XX$, of the sum
of all the factors. It is easy to see that each factor value $\ff_\II(\xs_\II)\; \forall \II,\ \XX_\II$
is counted $\vert \XX_{\back \II} \vert$ times in the summation above.
Therefore we can rewrite the integral above as
\begin{align*}
  \qq(\emptyset) = \sum_{\II} \vert \XX_{\back \II} \vert \big ( \sum_{\xs_{\II}} \ff_{\II}(\xs_\II) \big )
\end{align*}
where the new form involves polynomial number of terms and therefore is easy to calculate.

\noindent \bullitem\ $\min-\min$ (similar for $\max-\max$) is short for
\begin{align*}
  \qq(\emptyset) = \min_{\xs} \min_{\II} \ff_\II(\xs_\II)
\end{align*}
where the query seeks the minimum achievable value of any factor.
We can easily obtain this by seeking the range of all factors and reporting the minimum value in polynomial time. 
\end{proof}

Max-sum and max-prod are widely studied and it is known that their decision version are \NP-complete~\cite{shimony1994finding}.
By reduction from satisfiability we can show that max-min inference~\cite{ravanbakhsh_minmax} is also \NP-hard.
\index{min-max!complexity}
\index{complexity!min-max}
\begin{proposition}\label{th:minmaxnp}
The decision version of max-min inference that asks $\max_{\xs} \min_{\II} \ff_{\II}(\xs_\II) \overset{?}{\geq} q$ is \NP-complete.
\end{proposition}
\begin{proof}
Given $\xs$ it is easy to verify the decision problem, so max-min decision belongs to \NP. 
To show \NP-completeness, we reduce the 3-SAT to a max-min inference problem, such that
3-SAT is satisfiable \textit{iff} the max-min value is $\qq(\emptyset) \geq 1$ and unsatisfiable otherwise. 

Simply define one factor per clause of 3-SAT, such that $\ff_{\II}(\xs_{\II}) = 1$
if $\xs_{\II}$ satisfies the clause and any number less than one otherwise.
With this construction, the max-min value $\max_{\xs} \min_{\II \in \FF} \ff_{\II}(\xs_\II)$ is one \emph{iff} the original SAT problem was satisfiable, otherwise it is less than one. This reduces 3-SAT to Max-Min-decision.
\end{proof}

This means all the problems in $\Psi_1(\emptyset, \emptyset)$ are in \NP\ (and in fact are complete w.r.t. this complexity class).
In contrast, problems in $\Phi_1(\emptyset, \emptyset)$ are in \coNP, which is the class of decision problems in which the ``NO instances'' result has a polynomial time verifiable witness or proof.
Note that by changing the decision problem from $\qq(\emptyset) \overset{?}{\geq} q$ to $\qq(\emptyset) \overset{?}{\leq} q$, the complexity classes
of problems in $\Phi$ and $\Psi$ family are reversed (\ie problems in $\Phi_1(\emptyset, \emptyset)$ become $\NP$-complete and the problems in $\Psi_1(\emptyset, \emptyset)$ become \coNP-complete).

\index{complexity!sum-product}
\index{complexity!PP-complete}
Among the members of $\Sigma_1(\{1\}, \emptyset)$, sum-product is known to be \PPclass-complete~\cite{littman2001stochastic,roth1993hardness}.
It is easy to show the same result for sum-min (sum-max) inference.
\index{complexity!sum-min}
\begin{proposition}\label{th:summinsharpp}
The sum-min decision problem $\sum_{\xs} \min_{\II} \ff_\II(\xs_\II) \overset{?}{\geq} q$ is \PPclass-complete for $\YY = \{0,1\}$.
\end{proposition}
 \PPclass\ is the class of problems that are polynomially solvable using a non-deterministic Turing machine, where the acceptance condition is that the majority of computation paths accept.
\begin{proof}
To see that $\sum_{\xs} \min_\II \ff_{\II}(\xs_\II) \overset{?}{\geq} q$ is in \PPclass, 
 enumerate all $\xs \in \XX$ non-deterministically and for each assignment
calculate $\min_\II \ff_{\II}(\xs_\II)$ in polynomial time 
(where each path accepts iff $\min_\II \ff_{\II}(\xs_\II) = 1$) and accept
iff at least $q$ of the paths accept.

Given a matrix $\Dn \in \{0,1\}^{N \times N}$ the problem of calculating its permanent
\begin{align*}
  \mathsf{perm}(\Dn) = \sum_{\zs \in \sn} \prod_{i = 1}^{N} \Dn_{i, \zz_i}
\end{align*}
where $\sn$ is the set of permutations of $1,\ldots,N$ 
is \sharpP-complete and the corresponding decision problem is \PPclass-complete~\cite{valiant1979complexity}.
To show completeness w.r.t. \PPclass\ it is enough to reduce the problem of computing the matrix permanent to sum-min inference in a graphical model.
\index{complexity!permanent}
\index{permanent}
The problem of computing the permanent has been reduced to sum-product inference in graphical models  \cite{huang2009approximating}. However, when $\ff_{\II}(\xs_\II) \in \{0,1\}\;\forall \II$,
sum-product is isomorphic to sum-min. This is because $y_1 \times y_2 = \min(y_1,y_2) \forall y_i \in \{0,1\}$. Therefore, the problem of computing the permanent for such matrices reduces to sum-min
inference in the factor-graph of \cite{huang2009approximating}.
\end{proof}



\subsection{Complexity of general inference classes}
Let $\compclass(.)$ denote the complexity class of an inference class in the hierarchy.
In obtaining the complexity class of problems with $M > 1$, we use the following fact, which is also used in the polynomial hierarchy: $\Poly^{\NP} = \Poly^{\coNP}$~\cite{arora2009computational}. 
In fact $\Poly^{\NP^{\Aclass}} = \Poly^{\coNP^{\Aclass}}$, for any oracle $\Aclass$. 
This means that by adding a polynomial marginalization to the problems in $\Phi_M(\sumset, \pset)$ and $\Psi_M(\sumset, \pset)$,
we get the same complexity class $\compclass(\Delta_{M+1}(\sumset, \pset \cup \{M+1\}))$.
The following gives a recursive definition of complexity class for problems in the inference hierarchy.\footnote{
We do not prove the completeness w.r.t. complexity classes beyond the first level of the hierarchy and only assert the membership.
} Note that the definition of the complexity for each class is very similar to the recursive definition
of members of each class in \cref{eq:hierarchymembers1,eq:hierarchymembers2}
\index{complexity!recursive}
\index{inference!hierarchy!complexity}
\begin{theorem} The complexity of inference classes in the hierarchy is given by the recursion
\begin{align}
&\compclass(\Phi_{M+1}(\sumset, \pset)) =  \coNP^{\compclass(\Xi_M(\sumset, \pset) \back \Phi_M(\sumset, \pset))}\label{eq:compmin}\\
&\compclass(\Psi_{M+1}(\sumset, \pset)) =  \NP^{\compclass(\Xi_M(\sumset, \pset) \back \Psi_M(\sumset, \pset))} \label{eq:compmax}\\
&\compclass(\Sigma_{M+1}(\sumset\cup \{M+1\}, \pset)) = \PPclass^{\compclass(\Xi_M(\sumset, \pset) \back \Sigma_M(\sumset, \pset))} \label{eq:compsum}\\
&\compclass(\Delta_{M+1}(\sumset, \pset\cup \{M+1\})) =  \Poly^{\compclass(\Xi_M(\sumset, \pset))} \label{eq:comppoly}
\end{align}
where the base members are defined in \cref{eq:basemembers} and belong to $\Poly$.
\end{theorem}
\begin{proof}
  Recall that our definition of factor graph ensures that $\qq(\xs)$ can be evaluated in polynomial time and therefore the base members are in $\Poly$ (for complexity of base members of $\Delta$ see \cref{th:sumsum_poly}). We use these classes as the base of our induction and assuming the complexity classes above are correct for $M$ we show that are correct for $M+1$. We consider all the above statements one by one:
  
\noindent \bullitem\ \textit{Complexity for members of} $\Phi_{M+1}(\sumset, \pset)$:\\
 Adding an exponential-sized \emph{min}-marginalization to an inference problem with known complexity $\Aclass$,
 requires a Turing machine to non-deterministically enumerate 
$\zs_{\JJ_M} \in \XX_{\JJ_M}$ possibilities, then call the $\Aclass$
oracle with the ``reduced factor-graph'' -- in which $\xs_{\JJ_M}$ is clamped to $\zs_{\JJ_M}$ -- 
and reject iff any of the calls to oracle rejects. This means 
$\compclass(\Phi_{M+1}(\sumset, \pset)) = \coNP^{\Aclass}$.

Here, \refEq{eq:compmin} is also making another assumption expressed in the following claim.
\begin{claim}\label{claim:inproof}
All inference classes in 
$\Xi_M(\sumset, \pset) \back \Phi_M(\sumset, \pset)$ have the same complexity $\Aclass$.
\end{claim}
\begin{itemize}
  \item $M = 0$: the fact that $\qq(\xs)$ can be evaluated in polynomial time means that $\Aclass = \Poly$.
\item  $M > 0$:  $\Xi_M(\sumset, \pset) \back \Phi_M(\sumset, \pset)$  only contains one inference class -- that is exactly only one of the following cases
is correct:
\begin{itemize}
\item $M \in \sumset \; \Rightarrow \; \Xi_M(\sumset, \pset) \back \Phi_M(\sumset, \pset) = \Sigma_M(\sumset, \pset)$
\item $M \in \pset \; \Rightarrow \; \Xi_M(\sumset, \pset) \back \Phi_M(\sumset, \pset) =  \Delta_M(\sumset, \pset)$
\item $M \notin \sumset \cup \pset \; \Rightarrow \; \Xi_M(\sumset, \pset) \back \Phi_M(\sumset, \pset) =  \Psi_M(\sumset, \pset)$. \\
(in constructing the hierarchy we assume two consecutive marginalizations are distinct and the current marginalization is a minimization.)
\end{itemize}
But if $\Xi_M(\sumset, \pset) \back \Phi_M(\sumset, \pset)$ contains a single class, the inductive hypothesis ensures that all problems in $\Xi_M(\sumset, \pset) \back \Phi_M(\sumset, \pset)$ have the same complexity class $\Aclass$.
\end{itemize}
This completes the proof of our claim.

\noindent \bullitem\ \textit{Complexity for members of} $\Psi_{M+1}(\sumset, \pset)$:\\
Adding an exponential-sized \emph{max}-marginalization to an inference problem with known complexity $\Aclass$,
 requires a Turing machine to non-deterministically enumerate 
$\zs_{\JJ_M} \in \XX_{\JJ_M}$ possibilities, then call the $\Aclass$
oracle with the reduced factor-graph  
and accept iff any of the calls to oracle accepts. 
This means 
$\compclass(\Psi_{M+1}(\sumset, \pset)) = \NP^{\Aclass}$.
Here, an argument similar to that of \cref{claim:inproof} ensures that
$\Xi_M(\sumset, \pset) \back \Psi_M(\sumset, \pset)$
in \refEq{eq:compmax} contains a single inference class.

\noindent \bullitem\ \textit{Complexity for members of} $\Sigma_{M+1}(\sumset\cup \{M+1\}, \pset)$:\\
Adding an exponential-sized \emph{sum}-marginalization to an
inference problem with known complexity $\Aclass$,
 requires a Turing machine to non-deterministically enumerate 
$\zs_{\JJ_M} \in \XX_{\JJ_M}$ possibilities, then call the $\Aclass$
oracle with the reduced factor-graph  
and accept iff majority of the calls to oracle accepts. 
This means $\compclass(\Psi_{M+1}(\sumset, \pset)) = \PPclass^{\Aclass}$.
\begin{itemize}
\item $M = 0$: the fact that $\qq(\xs)$ can be evaluated in polynomial time means that $\Aclass = \Poly$.
\item  $M > 0$: 
  \begin{itemize}
\item $M \in \pset \; \Rightarrow \; \Xi_M(\sumset, \pset) \back \Sigma_M(\sumset, \pset) = \Delta_M(\sumset, \pset)$.
  \item $M \notin \pset \cup \sumset \; \Rightarrow \; \Xi_M(\sumset, \pset) \back \Sigma_M(\sumset, \pset) = \Psi_M(\sumset, \pset) \cup \Phi_M(\sumset, \pset)$: 
 despite the fact that $\Aclass = \compclass(\Psi_{M}(\sumset, \pset))$ is different from
$\Aclass' = \compclass(\Phi_{M}(\sumset, \pset))$, since \textit{$\PPclass$ is closed under complement},
which means $\PPclass^\Aclass = \PPclass^{\Aclass}$ and the recursive definition of complexity  \refEq{eq:compsum} remains correct.
  \end{itemize}
\end{itemize}

\noindent \bullitem\ \textit{Complexity for members of} $\Delta_{M+1}(\sumset, \pset\cup \{M+1\})$:\\
Adding a polynomial-sized marginalization to an
inference problem with known complexity $\Aclass$,
 requires a Turing machine to deterministically enumerate 
 $\zs_{\JJ_M} \in \XX_{\JJ_M}$ possibilities in polynomial time,
 and each time call the $\Aclass$ oracle with the reduced factor-graph  
and accept after some polynomial-time calculation. This means 
$\compclass(\Psi_{M+1}(\sumset, \pset)) = \Poly^{\Aclass}$. Here, there are
three possibilities:
\begin{itemize}
\item $M = 0$: here again  $\Aclass = \Poly$.
\item $M \in \sumset \; \Rightarrow \; \Xi_M(\sumset, \pset) = \Sigma_M(\sumset, \pset)$.
  \item $M \in \pset \; \Rightarrow \; \Xi_M(\sumset, \pset) = \Delta_M(\sumset, \pset)$.
   \item $M \notin \pset \cup \sumset \; \Rightarrow \; \Xi_M(\sumset, \pset) = \Psi_M(\sumset, \pset) \cup \Phi_M(\sumset, \pset)$, in which case since $\PPclass^{\NP^{\mathbb{B}}} = \PPclass^{\coNP^{\mathbb{B}}}$, the recursive definition of complexity in \refEq{eq:comppoly} remains correct.
\end{itemize}
\end{proof}

\begin{example}\label{example:marginalmap}
\index{marginal MAP}
\index{complexity!marginal MAP}
  Consider the marginal-MAP inference of \refEq{eq:marginalmap}.
The decision version of this problem, $\qq(\emptyset) \overset{?}{\geq} q$,
 is a member of $\Psi^{2}(\{1\}, \emptyset)$ which also includes
 $\max-\sumop-\min$ and $\max-\sumop-\max$.
The complexity of this class according to \refEq{eq:compmax} is $\compclass(\Psi^{2}(\{1\}, \emptyset)) = \NP^{\PPclass}$.
However, marginal-MAP is also known to be ``complete'' w.r.t. $\NP^{\PPclass}$~\cite{park2004complexity}.
Now suppose that the max-marginalization over $\xs_{\JJ_2}$ is polynomial (\eg $|\JJ_2|$ is constant). 
Then marginal-MAP belongs to $\Delta_{2}(\{1\}, \{2\})$ with complexity $\poly^{\PPclass}$.
This is because a Turing machine can enumerate all $\zs_{\JJ_2} \in \XX_{\JJ_2}$ in polynomial time and
call its $\PPclass$ oracle to see if 
  \begin{align*}
    &\qq(\xs_{\JJ_0} \mid \zs_{\JJ_2}) \overset{?}{\geq} q\\
     \text{where} \quad &\qq(\xs_{\JJ_0} \mid \zs_{\JJ_2}) = \sum_{\xs_{\JJ_2}} \prod_{\II} \ff_{\II}(\xs_{\II \back \JJ_2}, \zs_{\II \cap \JJ_2}) 
  \end{align*}
and \emph{accept} if any of its calls to oracle accepts, and rejects otherwise.
Here, $\ff_{\II}(\xs_{\II \back \JJ_2}, \zs_{\II \cap \JJ_2})$ is the reduced factor, in which all the variables in $\xs_{\JJ_2}$ are fixed to $\zs_{\JJ_2\cap \II}$.  
\end{example}

The example above also hints at the rationale behind the recursive definition of 
complexity class for each inference class in the hierarchy.
Consider the inference family $\Phi$:

\index{Toda theorem}
Here, \magn{Toda's theorem} \cite{toda1991pp} has an interesting implication w.r.t. the hierarchy. 
This theorem states that  $\PPclass$ is as hard as the polynomial hierarchy, which means
$\min-\max-\min-\ldots-\max$ inference for an arbitrary, but constant, number of min and max operations appears below the
sum-product inference in the inference hierarchy.

\subsection{Complexity of the hierarchy}\label{sec:logical}
By restricting the domain $\RR$ to $\{0,1\}$, min and max become isomorphic to logical AND ($\wedge$) and OR ($\vee$) respectively,
where $1 \cong \truemath, 0 \cong \falsemath$. By considering the restriction of the inference hierarchy to these
\index{quantified satisfiability}
\index{satisfiability!quantified}
two operations we can express quantified satisfiability (QSAT) as inference in a graphical model, where $\wedge \cong \forall$ and $\vee \cong \exists$. Let each factor $\ff_\II(\xs_\II)$ be a disjunction --\eg $\ff(\xs_{i,j,k}) = \xx_i \vee \neg \xx_j \vee \neg \xx_k$. Then we have
\begin{align*}
\forall_{\xs_{\JJ_M}} \exists_{\xs_{\JJ_{M-1}}} \ldots \exists_{\xs_{\JJ_2}} \forall_{\xs_{\JJ_1}}  \bigwedge_{\II} \ff_\II(\xs_\II)\; \cong  
\min_{\xs_{\JJ_M}} \max_{\xs_{\JJ_{M-1}}} \ldots \max_{\xs_{\JJ_2}} \min_{\xs_{\JJ_1}} \min_{\II} \ff_\II(\xs_\II)
\end{align*}
\index{stochastic satisfiability}
\index{satisfiability!stochastic}
By adding the summation operation, we can express the stochastic satisfiability~\cite{littman2001stochastic} and by generalizing the constraints from disjunctions 
we can represent any quantified constraint problem (QCP)~\cite{bordeaux2002beyond}.
\index{quantified constraint problem}
QSAT, stochastic SAT and QCPs are all \pspace-complete,
\index{PSPACE-complete}
\index{complexity!PSPACE-complete}
where \pspace\ is the class of problems that can be solved by a (non-deterministic) Turing machine in polynomial space.
Therefore if we can show that inference in the inference hierarchy is in \pspace, 
it follows that inference hierarchy is in \pspace-complete as well.
\begin{theorem}\label{th:hierarchy_complexity}
The inference hierarchy is \pspace-complete.
\end{theorem}
\index{inference!hierarchy!complexity}
\begin{proof} (\pcref{th:hierarchy_complexity})
\begin{algorithm}[h]
\SetKwInOut{Input}{input}\SetKwInOut{Output}{output}
\DontPrintSemicolon
\Input{$\bigopp{M}_{\xs_{\JJ_M}} \bigopp{{M-1}}_{\xs_{\JJ_{M-1}}} \ldots \bigopp{1}_{\xs_{\JJ_1}} \bigotimes_{\II} \ff_\II(\xs_\II)$}
\Output{$\qq(\xs_{\JJ_0})$}
\DontPrintSemicolon
 \For(\tcp{loop over the query domain}){ \textbf{each} $\zs_{\JJ_0} \in \XX_{\JJ_0}$}{
 \For(\tcp{loop over $\XX_{i_N}$}){ \textbf{each} $\zz_{i_N} \in \XX_{i_N}$}{
.\;
.\;
.\;
\For(\tcp{loop over $\XX_{i_1}$}){ \textbf{each} $\zz_{i_{1}} \in \XX_{i_1}$}{
$\qq_{1}(\zz_{i_1}) := \bigotimes_{\II} \ff_\II(\zs_\II)$;
}
$\qq_{i_2}(\zz_{i_2}) := \bigopp{\jfun(i_1)}_{\xx_{i_1}} \qq_1(\xx_{i_1})$\;
.\;
.\;
.\;
$\qq_{N}(\zz_{i_N}) := \bigopp{\jfun(i_{{N-1}})}_{\xx_{i_{{N-1}}}} \qq_{{N-1}}(\xx_{i_{{N-1}}})$\;
}
$\qq(\zs_{\JJ_{0}}) := \bigopp{\jfun(i_{{N}})}_{\xx_{i_N}} \qq_N(\xx_{i_N})$\;
}
\caption{inference in \pspace}\label{alg:pspace_inference}
\end{algorithm}
To prove that a problem is \pspace-complete, we have to show that 1) it is in \pspace\ and 2) a \pspace-complete problem reduces to it.
We already saw that QSAT, which is \pspace-complete, reduces to the inference hierarchy.
But it is not difficult to show that inference hierarchy is contained in \pspace.
Let 
  \begin{align*}
    \qq(\xs_{\JJ_0}) = \bigopp{M}_{\xs_{\JJ_M}} \bigopp{M-1}_{\xs_{\JJ_{M-1}}} \ldots \bigopp{1}_{\xs_{\JJ_1}} \bigotimes_{\II} \ff_\II(\xs_\II)
  \end{align*}
be any inference problem in the hierarchy.
We can simply iterate over all values of $\zs \in \XX$ in nested loops or using a recursion.
Let $\jfun(i): \{1,\ldots,N\} \to \{1,\ldots,M\}$ be the index of the marginalization that involves $\xx_i$ -- that is $i \in \JJ_{\jfun(i)}$. Moreover let $i_1,\ldots,i_N$ be an ordering of variable indices
such that $\jfun(i_k) \leq \jfun(i_{k+1})$.
\RefAlgorithm{alg:pspace_inference} uses this notation to demonstrate this procedure using nested loops.
Note that here we loop over individual domains $\XX_{i_k}$ rather than $\XX_{\JJ_m}$ and track
only temporary tuples $\qq_{i_k}$, so that the space complexity remains polynomial in $N$. 

\end{proof}


\section{Polynomial-time inference}\label{sec:gdl}
\index{inference!polynomial}
\index{polynomial inference}
Our definition of inference 
was based on an expansion operation $\otimes$ and one or more marginalization operations $\opp{1}, \ldots, \opp{M}$.
If we assume only a single marginalization operation, polynomial time inference
is still not generally possible.
However, if we further assume that the expansion operation is distributive
over marginalization and the factor-graph has no loops,
exact polynomial time inference is possible. 
\begin{definition}
\index{commutative semiring}
\index{semiring!commutative}
A \magn{commutative semiring} $\semiring=(\RR, \oplus, \otimes)$ is the combination of two 
commutative semigroups $\semig_e = (\RR, \otimes)$ and $\semig_m = (\RR, \oplus)$ with two additional properties
\begin{itemize} 
\index{identity}
\index{annihilator}
\item identity elements $\identt{\oplus}$ and $\identt{\otimes}$ such that $\identt{\oplus} \oplus a = a$ and $\identt{\otimes} \otimes a = a$. Moreover  $\identt{\oplus}$ is an \magn{annihilator} for $\semig_e = (\otimes, \RR)$: $a \otimes \identt{\oplus} = \identt{\oplus}\quad \forall a \in \RR$.\footnote{
That is when dealing with reals, this is $\identt{\oplus} = 0$; this means $a \times 0 = 0$.}
\index{distributive law}
\item distributive property: $$a \otimes (b \oplus c) = (a \otimes b) \oplus (a \otimes b)\quad \forall a,b,c \in \RR$$
\end{itemize}
\end{definition}

The mechanism of efficient inference using distributive law can be seen in a simple example: instead of  calculating $ \min(a +  b , a + c)$,
using the fact that summation distributes over minimization, we may instead obtain the same result using $a + \min(b , c)$, which requires fewer operations.

\begin{example}\label{example:semirings}
The following are some examples of commutative semirings:
\index{sum-product}
\index{max-product}
\index{min-max}
\index{min-sum}
\index{or-and}
\index{union-intersection}
\begin{easylist}
& Sum-product $(\Re^{\geq 0}, + , \times)$. 
& Max-product $(\Re^{\geq 0} \cup  \{-\infty\}, \max, \times)$ and $(\{0,1\}, \max, \times)$. 
& Min-max $(\settype{S}, \min, \max)$ on any ordered set $\settype{S}$. 
& Min-sum $(\Re \cup \{ \infty \}, \min, +)$ and $(\{0,1\}, \min, +)$. 
& Or-and $(\{\truemath,\falsemath\}, \vee, \wedge)$.
& Union-intersection $(2^{\SS}, \cup, \cap)$ for 
any power-set $2^{\SS}$.
& The semiring of natural numbers with greatest common divisor and least common multiple $(\settype{N},\mathsf{lcm}, \mathsf{gcd})$. 
& Symmetric difference-intersection semiring for any power-set $(2^\SS,\nabla, \cap)$.
\end{easylist}
Many of the semirings above are isomorphic --\eg $y' \cong -\log(y)$ defines an isomorphism between min-sum and max-product. 
It is also easy to show that the or-and semiring is 
isomorphic to min-sum/max-product semiring on $\RR = \{0,1\}$.
\end{example}

The inference problems in the example above have different properties indirectly inherited from their commutative semirings:
\index{choice function}
for example, the operation $\min$ (also $\max$) is a \magn{choice function},
 which means $\min_{a \in \settype{A}} a \quad \in \settype{A}$. 
The implication is that if $\sumop$ of the semiring is  
$\min$ (or $\max$), we can replace it with $\arg_{\xs_{\JJ_M}} \max$
and (if required) recover $\qq(\emptyset)$
using $\qq(\emptyset) = \bigotimes_{\II} \ff_\II(\xs^*)$ in polynomial time.


As another example, since both operations have inverses, sum-product is a \magn{field} \cite{pinter2012book}. 
\index{field}
The availability of inverse for $\otimes$ operation -- \ie when $\semig_e$ is an Abelian group -- 
\index{Abelian group}
has an important implication for inference:
the expanded form of \refEq{eq:expanded} can be normalized, and we may inquire about \textbf{normalized marginals}
\index{normalized marginal}
\index{marginalization!normalized}
\begin{align}
\quad \pp(\xs_{\JJ}) \quad = \quad &  \bigoplus_{ \xs_{\back \JJ}}  \pp(\xs)&\label{eq:semiring_marginalization}\\
\text{where}\quad \pp(\xs) \quad \defeq \quad & \frac{1}{\qq(\emptyset)} \otimes \big (\bigotimes_{\II}\ff_{\II}(\xs_{\II}) \big ) &\quad \text{if} \quad \qq(\emptyset) \neq \identt{\oplus}\label{eq:p}\\
\pp(\xs) \quad \defeq \quad \identt{\oplus} &\quad \text{if} \quad \qq(\emptyset) = \identt{\oplus} \label{eq:specialp}
\end{align}
\index{joint form!normalized}
where $\pp(\xs)$ is the normalized joint form. We deal with the case where the integral evaluates to the annihilator as a special case because
division by annihilator may not be well-defined.
This also means, when working with normalized expanded form and normalized marginals, we always have
$\bigoplus_{\xs_{\JJ}} \pp(\xs_{\JJ}) = \identt{\otimes}$ 
\begin{example}
Since $\semig_{e} = (\Re^{>0}, \times)$ and $\semig_{e} = (\Re, +)$
are both Abelian groups, min-sum and sum-product inference have normalized marginals. 
For min-sum inference this means $\min_{\xs_{\JJ}}\pp(\xs_{\JJ}) = \identt{\sumop} = 0$. However, for min-max inference, since $(\SS, \max)$ is not Abelian, normalized marginals are not defined.
\end{example}

We can apply the identity and annihilator of a commutative semiring to define constraints.
\index{constraint!factor}
\begin{definition}\label{def:constraint}
A \magn{constraint} is a factor $\ff_\II: \XX_{\II} \to \{\identt{\otimes}, \identt{\oplus}\}$
whose range is limited to identity and annihilator of the expansion monoid.\footnote{Recall that a monoid is a semigroup with an identity. 
The existence of identity here is a property of the semiring.}
\end{definition}
Here, $\ff_{\II}(\xs) = \identt{\oplus}$ iff $\xs$ is forbidden and 
$\ff_{\II}(\xs) = \identt{\otimes}$ iff it is permissible. 
A \magn{constraint satisfaction problem} (CSP) 
is any inference problem on a semiring
in which all factors are constraints. Note that this allows definition 
of the ``same'' CSP on any commutative semiring. 
\index{constraint satisfaction problem}
The idea of using different semirings to define CSPs has been studied in the 
past~\cite{bistarelli1999semiring}, however its implication about inference on commutative
semirings has been ignored.
\index{commutative semiring!complexity}
\index{inference!complexity}
\begin{theorem}\label{th:semiring_inference}
Inference in any commutative semiring is \NP-hard under randomized polynomial-time reduction.
\end{theorem}
\begin{proof}
To prove that inference in any semiring $\semiring = (\RR, \identt{\oplus}, \identt{\otimes})$ is \NP-hard under randomized polynomial
\index{unique satisfiability}
\index{satisfiability!unique}
 reduction, we deterministically reduce \textit{unique satisfiability} (USAT) to an inference problems on any semiring.
\index{promise problem}
USAT is a so-called ``promise problem'', that asks whether a satisfiability problem that is promised to have either zero or one satisfying assignment is satisfiable.
\citet{valiant1986np} prove that a polynomial time randomized algorithm (\RP) for USAT
implies a \RP=\NP.

For this reduction consider a set of binary variables $\xs \in \{0,1\}^{N}$, 
one per each variable in the given instance of USAT.
For each clause, define a constraint factor $\ff_{\II}$ such that $\ff_{\II}(\xs_{\II}) = \identt{\otimes}$ if $\xs_{\II}$ satisfies that clause and  $\ff_{\II}(\xs_{\II}) = \identt{\oplus}$ otherwise. This means, $\xs$ is a satisfying assignment for USAT iff $\qq(\xs) = \bigotimes_{\II} \ff_\II(\xs_{\II}) = \identt{\otimes}$. If the instance is unsatisfiable, the integral $\qq(\emptyset) = \bigoplus_{\xs} \identt{\oplus} = \identt{\oplus}$ (by definition of $\identt{\oplus}$).
If the instance is satisfiable there is only a single instance $\xs^*$ for which $\qq(\xs^*) = \identt{\otimes}$, and therefore the integral evaluates to $\identt{\otimes}$.
Therefore we can decide the satisfiability of USAT by performing inference on any semiring, by only relying on the properties of identities.  
The satisfying assignment can be recovered using a decimation procedure, assuming access to an oracle for inference on the semiring.

\end{proof}
\index{xor-and}
\index{inference!xor-and}
\begin{example}
Inference on xor-and semiring $(\{\truemath,\falsemath\}, \xor, \wedge)$, where each factor has
a disjunction form, is called parity-SAT, 
which asks whether the number of SAT solutions is even or odd. A corollary to theorem~\ref{th:semiring_inference} is that parity-SAT is \NP-hard under randomized reduction,
which is indeed the case \cite{valiant1986np}.
\end{example}

We find it useful to use the same notation for the \marnote{identity function}
\index{identity function}
\index{factor!identity}
 \magn{identity function} $\ident(\mathrm{condition})$:
\begin{align}\label{eq:identity}
  \ident(\mathrm{cond.}) \; \defeq \;  \left \{ 
\begin{array}{r c c c}
 &(+,\times) & (\min,+) & (\min,\max)\\
\mathrm{cond.} = \truemath & 1 & 0 & -\infty\\
\mathrm{cond.} = \falsemath & 0 & +\infty & +\infty\\
\end{array}   
\right .
\end{align}
where the intended semiring for $\ident(.)$ function will be clear from the context. 

\input{reductions.tex}


%% file: reductions.tex

\section{Reductions}\label{sec:reductions}
Several of the inference problems over commutative semirings are reducible to each other. \RefSection{sec:marg2int} reviews the well-known reduction of marginalization to integration for general commutative semirings. 
We use this reduction to obtain approximate message 
dependencies in performing loop corrections in \refSection{sec:loops}.
\index{reduction}

In \refSection{sec:decimation}, we introduce a procedure to reduce
integration to that of finding normalized marginals.
The same procedure, called \magn{decimation}, reduces sampling to marginalization.
The problem of \magn{sampling} from a distribution is known to be almost as difficult \marnote{sampling complexity}
\index{complexity!sampling}
as sum-product integration~\cite{jerrum1986random}. 
As we will see in \cref{sec:csp}, constraint satisfaction can be reduced to sampling and therefore marginalization. In \refSection{sec:pbp} we introduce a perturbed message passing scheme to perform approximate sampling and use it to solve CSPs.
Some recent work perform 
approximate sampling by finding the MAP solution in the perturbed factor-graph, 
in which a particular type of noise is added to the factors~\cite{papandreou2011perturb,hazan2012partition}.
Approximate sum-product integration has also 
been recently reduced to MAP inference~\cite{ermon2013optimization,ermon2014low}.
In \refSection{sec:variational}, 
we see that min-max and min-sum inference can be obtained as limiting cases of min-sum and sum-product inference respectively.

\RefSection{sec:minmaxnphard} reduces the min-max inference to min-sum also to a sequence of CSPs (and therefore sum-product inference) over factor-graphs.
This reduction gives us a powerful procedure to solve min-max problems, which we use in \refChapter{chapter:combinatorial} to solve bottleneck combinatorial problems.

In contrast to this type of reduction between various modes of inference, many have studied reductions of different types of factor-graphs \cite{eaton2013model}.
Some examples of these special forms are factor-graphs with: binary variables, pairwise interactions, constant degree nodes, and planar form. 
For example~\citet{sanghavi2007message} show that min-sum integration is reducible to \textit{maximum independent-set} problem. However since a pairwise binary 
factor-graph can represent a maximum independent-set problem (see \refSection{sec:csp_opt}), this means that min-sum integration in any factor-graph can be reduced to the 
same problem on a pairwise binary model.

\index{independent-set!maximum}
These reductions are in part motivated by the fact that under some further restrictions the restricted factor-graph allows more efficient inference.
\index{planar}
\index{constant degree}
For example, (I) it is possible to calculate the sum-product integral of the \textit{planar} spin-glass Ising model (see \refExample{example:ising}) in polynomial time, in the absence of local fields \cite{fisher1966dimer}; (II) the complexity of the loop correction method that we study in \refSection{sec:message_dependency} grows exponentially with the degree of each node and therefore it may be beneficial to consider reduced factor-graph where $|\nb i| = 3$; and (III) if the factors in a factor-graphs with pairwise factors satisfy certain metric property, polynomial algorithms can obtain the exact min-sum integral using graph-cuts~\cite{boros2002pseudo}.

\subsection{Marginalization and integration}
This section shows how for arbitrary commutative semirings 
there is a reduction from marginalization to integration and vice versa.

\subsubsection{Marginalization reduces to integration}\label{sec:marg2int}
For any fixed assignment to a subset of variables $\xs_\AAA = \zs_\AAA$ (\aka \magn{evidence}), \marnote{evidence}
\index{evidence}
we can reduce all the factors $\ff_{\II}(\xs_{\II})$ that have non-empty
intersection with $\AAA$ (\ie $\II \cap \AAA \neq \emptyset$) accordingly:\marnote{clamping}
\index{clamping}
\begin{align}\label{eq:reduce_factor}
\ff_{\II \back \AAA}(\xs_{\II \back \AAA} \mid \zs_\AAA) \quad \defeq \quad \bigoplus_{\xs_{\II \cap \AAA}}\ff_{\II}(\xs_{\II}) \otimes \ident(\xs_{\II \cap \AAA} = \zs_{\II \cap \AAA}) \quad \forall \II \; s.t.\; 
\AAA \cap \II \neq \emptyset 
\end{align}
\index{factor graph!reduced}
where the identity function $\ident(.)$ is defined by \cref{eq:identity}.
The new factor graph produced by \magn{clamping} all factors in this manner, 
has effectively accounted for the evidence.
Marginalization or integration, can be performed on this reduced factor-graph. 
We use similar notation for the \textit{integral and marginal in the new factor graph} -- \ie $\qq(\emptyset \mid \xs_\AAA)$ and $\qq(\xs_\BBB \mid \xs_\AAA)$.
Recall that the problem of integration is that of calculating $\qq(\emptyset)$.
We can obtain the marginals $\qq(\zs_\AAA)$ by integration on reduced factor-graphs for all $\zs_\AAA \in \XX_\AAA$ reductions. 
\index{reduce!marginal to integral}
\begin{claim}\label{th:marg2int}
\begin{align}
\qq(\zs_{\AAA}) \quad = \quad \qq(\emptyset \mid \zs_\AAA)
\end{align}
\end{claim}
\begin{proof}  \begin{align*}
\qq(\zs_{\AAA}) & = \bigoplus_{\xs_{ \back \AAA}} \bigotimes_{\II} \ff_{\II} (\xs_{\II \back \AAA}, \zs_{\AAA \cap \II}) \\
& = \bigoplus_{\xs} \bigg ( \ident(\xs_\AAA = \zs_\AAA) \otimes \bigotimes_{\II} \ff_{\II}(\xs_{\II \back \AAA}, \zs_{\AAA \cap \II}) \bigg ) \\
& = \bigoplus_{\xs} \bigotimes_{\II} \bigg ( \ff_{\AAA}(\xs_{\AAA}) \otimes \ident(\xs_{\II \cap \AAA} = \zs_{\II \cap \AAA}) \bigg ) \\
& = \bigoplus_{\xs} \bigotimes_{\II}  \ff_{\II \back \AAA} (\xs_{\II \back \AAA} \mid \zs_\AAA) \quad = \quad  \qq(\emptyset \mid \zs_\AAA)
  \end{align*}
\end{proof}
where we can then normalize $\qq(\xs_{\II})$
values to get $\pp(\xs_{\II})$ (as defined in \cref{eq:semiring_marginalization}).

\subsubsection{Integration reduces to marginalization}\label{sec:decimation}
Assume we have access to an oracle that can produce the 
normalized marginals of \refEq{eq:semiring_marginalization}.
We show how to calculate $\qq(\emptyset)$ by making $N$ calls to the oracle. 
Note that if the marginals are not normalized, the integral is trivially given by $\qq(\emptyset) = \bigoplus_{\xs_{\JJ}} \qq(\xs_{\JJ})$

Start with $t = 1$, $\BBB(t=0) = \emptyset$ and given the normalized marginal over a variable $\pp(\xx_{i(t)})$, 
fix the $\xx_{i(t)}$ to an arbitrary value $\zz_{i(t)} \in \XX_{i(t)}$.
Then reduce all factors according to \cref{eq:reduce_factor}. 
Repeat this process of marginalization and clamping 
$N$ times until all the variables are fixed. 
At each point, $\BBB(t)$ denotes the subset of variables fixed up to step $t$ (including $i(t)$) and 
$\pp(\xx_{i(t)} \mid \zs_{\BBB(t-1)}) = \frac{\qq(\xx_{i(t)} \mid \zs_{\BBB(t-1)})}{\qq(\emptyset \mid \zs_{\BBB(t-1)})}$ refers to the new marginal. 
Note that we require $i(t) \notin \BBB(t-1)$ -- that is 
at each step we fix a different variable.

We call an assignment to $\xx_{i(t)} = \zz_{i(t)}$ \textit{invalid}, 
if  $\pp(\zz_{i(t)} \mid \zs_{\BBB(t)}) = \identt{\oplus}$. 
This is because $\identt{\oplus}$ is the annihilator
of the semiring and we want to avoid division by the annihilator. 
Using \cref{eq:semiring_marginalization,eq:p,eq:specialp}, it is easy to show that if 
$\qq(\emptyset) \neq \identt{\oplus}$, a valid assignment always exists (this is because $\bigoplus_{\xx_{i(t)}}\pp(\xx_{i(t)} \mid \zs_{\BBB(t-1)}) = \identt{\otimes}$). Therefore if we are unable to find a valid assignment, it means $\qq(\emptyset) = \identt{\oplus}$. 

Let $\zs = \zs_{\BBB(N+1)}$ denote the final 
joint assignment produced using the procedure above.
\index{reduce!integral to marginal}
\begin{proposition}\label{th:int2marg} The integral in the original factor-graph is given by
  \begin{align}
\qq(\emptyset) \quad = \quad \left (\bigotimes_{\II} \ff_{\II}(\zs_\II)\right) \otimes \left( \bigotimes_{1 \leq t \leq N} \pp(\zz_{i(t)} \mid \zs_{\BBB(t-1)})\right)^{-1}
  \end{align}
where the inverse is defined according to $\otimes$-operation.
\end{proposition}
\begin{proof}
First, we derive the an equation for ``conditional normalized marginals'' for semirings where $\otimes$ defines an inverse. 
\begin{claim} For any semiring with normalized joint form we have
  \begin{align*}
  \pp(\xs) = \pp(\xx_i) \otimes \pp(\xs_{\back i} \mid \xx_i)  
  \end{align*}
where $\pp(\xs_{\back i} \mid \xx_i) = \frac{\qq(\xs_{\back i} \mid \xx_i)}{\qq(\emptyset \mid \xx_i)}$
\end{claim}
To arrive at this equality first note that since $\xs = \xs_{\back i}, \xx_i$, $\qq(\xs_{\back i} \mid \xx_i) = \qq(\xs)$.
Then multiply both sides by $\qq(\xx_i) = \qq(\emptyset \mid \xx_i)$ (see \cref{th:marg2int}) to get
\begin{align*}
  \qq(\xs) \otimes \qq(\emptyset \mid \xx_i) &= \qq(\xx_i) \otimes \qq(\xs_{\back i} \mid \xx_i) &\Rightarrow \\
  \frac{\qq(\xs)}{\qq(\emptyset)} &=  \frac{\qq(\xx_i)}{\qq(\emptyset)} \otimes \frac{\qq(\xs_{\back i} \mid \xx_i)}{\qq(\emptyset \mid \xx_i)} &\Rightarrow \\
  \pp(\xs) &= \pp(\xx_i) \otimes \pp(\xs_{\back i} \mid \xx_i) &
\end{align*}
where we divided both sides by $\qq(\emptyset)$ and moved a term from left to right in the second step.

\index{semiring!chain rule}
Now we can apply this repeatedly to get a chain rule for the semiring:
\begin{align*}
\pp(\xs) \quad = \quad \pp(\xx_{i_1}) \otimes \pp(\xx_{i_2} \mid \xx_{i_1}) \otimes \pp(\xx_{i_3} \mid \xs_{\{i_1, i_2 \}}) \otimes \ldots \otimes \pp(\xx_{i_N} \mid \xs_{\{i_1,\ldots,i_{N-1} \}})
\end{align*}
which is equivalent to
\begin{align*}
\pp(\xs) \quad = \quad \pp(\xx_{i(1)}) \otimes \pp(\xx_{i(2)} \mid \xs_{\BBB(1)}) \otimes \ldots \otimes \pp(\xx_{i(N)} \mid \xs_{\BBB(N-1)})
\quad =\quad \bigotimes_{1 \leq t \leq N} \pp(\xs_{i(t)} \mid \xs_{\BBB(t-1)})
\end{align*}

Simply substituting this into definition of $\pp(\xs)$ (\refEq{eq:p}) and re-arranging we get 
\begin{align*}
  &\bigotimes_{1 \leq t \leq N} \pp(\xs_{i(t)} \mid \xs_{\BBB(t-1)}) \quad = \frac{1}{\qq(\emptyset)} \bigotimes_{\II \in \FF} \ff_{\II}(\xs_\II) \quad \Rightarrow\\
  &\qq(\emptyset) \quad  = \quad \left (\bigotimes_{\II} \ff_{\II}(\zs_\II)\right) \otimes \left( \bigotimes_{1 \leq t \leq N} \pp(\zz_{i(t)} \mid \zs_{\BBB(t-1)})\right)^{-1} 
\end{align*}
\end{proof}


\index{decimation}
The procedure of incremental clamping is known as \magn{decimation}, and its variations are typically used for two objectives: \marnote{decimation}
(I) recovering the MAP assignment from (max) marginals (assuming a max-product semiring). Here instead of an arbitrary $\zs_{\JJ} \in \XX_{\JJ}$, one picks 
$\zs_{\JJ} = \arg_{\xs_{\JJ}}\max \; \pp(\xs_{\JJ})$. 
(II)~producing an unbiased sample from a distribution $\pp(.)$ (\ie assuming sum-product semiring). 
For this we sample from $\pp(\xs_\II)$: 
$\zs_{\JJ} \sim \pp(\xs_{\JJ})$. 

\subsection{Min-max reductions}\label{sec:minmaxnphard}
The min-max objective appears in various fields, particularly in building \label{minmax objectives}
robust models under uncertain and adversarial settings.  In the
\index{adversarial}
\index{min-max}
context of probabilistic graphical models, several  min-max
objectives different from inference in min-max semiring have been previously studied 
\cite{kearns2001graphical,ibrahimi2011robust} (also see \cref{sec:limits}).  
In combinatorial optimization, min-max may refer to the relation between
maximization and minimization in dual combinatorial objectives and
their corresponding linear programs \cite{schrijver1983min},
or it may refer to min-max settings due to uncertainty in the problem
specification
\cite{averbakh2001complexity,aissi2009min}.

\index{bottleneck problems}
In \refChapter{chapter:combinatorial} we will see that several problems that \marnote{bottleneck problems}
are studied under the class of \magn{bottleneck problems} can be formulated
using the min-max semiring.
Instances of these problems include bottleneck traveling
salesman problem \cite{parker1984guaranteed}, K-clustering
\cite{gonzalez1985clustering}, K-center problem
\cite{dyer1985simple,khuller2000capacitated} and
bottleneck assignment problem \cite{gross1959bottleneck}.

\citet{edmonds1970bottleneck} introduce a bottleneck framework with
a duality theorem that relates the min-max objective in one problem
instance to a max-min objective in a dual problem. An intuitive example
is the duality between the min-max cut separating nodes $a$ and $b$ --
the cut with the minimum of the maximum weight -- and min-max path between $a$
and $b$, which is the path with the minimum of the maximum weight
\cite{fulkerson1966flow}. \citet{hochbaum1986unified}
leverages the triangle inequality in metric spaces to find constant factor
approximations to several \NP-hard min-max problems under a unified
framework.

The common theme in a majority of heuristics for min-max or bottleneck
problems is the relation of the min-max objective to a CSP 
\cite{hochbaum1986unified,panigrahy1998log}. 
We establish a similar relation within the context of
factor-graphs, by reducing the min-max inference problem on the
original factor-graph to inference over a CSP factor-graph (see \refSection{sec:gdl}) on the
reduced factor-graph in \refSection{sec:minmax2sumprod}. In particular, since 
we use sum-product inference to solve the resulting CSP, we call this 
reduction, sum-product reduction of min-max inference.

\subsubsection{Min-max reduces to min-sum}\label{sec:minmax2minsum}
\index{reduce!min-max to min-sum}
Here, we show that min-max inference reduces to min-sum, although
 in contrast to the sum-product reduction of the next subsection, this is not a polynomial time reduction.
First, we make a simple observation about min-max inference.
Let \index{factor!range} $\YY = \bigcup_{\II \in \FF} \YY_\II$ denotes the union over the range
of all factors.  The min-max value belongs to this set $\max_{\II \in
  \FF} \ff_\II(\xx^*_\II) \in \YY$.  In fact for any assignment
$\xs$, $\max_{\II \in \FF} \ff_\II(\xs_\II) \in \YY$.

Now we show how to manipulate the factors in the original factor-graph to produce new factors over the same domain
such that the min-max inference on the former corresponds to the min-sum inference on the later.

\begin{lemma}\label{th:lemma}
  Any two sets of factors, $\{\ff_\II\}$ and
  $\{\fg_\II\}$, over the identical domains $\{\XX_\II\}$ have identical min-max solutions
  \begin{align*}
    \arg_{\xs}\min \max_\II \ff_\II(\xs_\II) = \arg_{\xs}\min \max_\II \fg_\II(\xs_\II)
  \end{align*}
  if $\;\;\forall \II,\JJ \in \FF, \xs_\II \in \XX_\II,\xs_\JJ\in
  \XX_\JJ$
  \begin{align*}
    \quad \ff_\II(\xs_\II) < \ff_\JJ(\xs_\JJ) \quad \Leftrightarrow  \quad \fg_\II(\xs_\II) < \fg_\JJ(\xs_\JJ)
  \end{align*}
\end{lemma}
\begin{proof}
  Assume they have different min-max assignments\footnote{For
    simplicity, we are assuming each instance has a single min-max
    assignment.  In case of multiple assignments there is a one-to-one
    correspondence between them. Here the proof instead starts with the assumption
    that there is an assignment $\xs^*$ for the first factor-graph that
    is different from all min-max assignments in the second
    factor-graph.}  --\ie $\xs^* = \arg_{\xs}\min \max_\II \ff_\II(\xs_\II)$, $\xs'^*
  = \arg_{\xs} \min \max_\II \fg_\II(\xs_I)$ and $x^* \neq x'^*$.  Let $y^*$
  and $\yy'^*$ denote the corresponding min-max values.
  \begin{claim}
    \begin{align*}
      \yy^* > \max_\II \ff_\II(\xs'^*_I) \quad \Leftrightarrow \quad \yy'^* < \max_\II \fg_\II(\xs^*_\II)\\
      \yy^* < \max_\II \ff_\II(\xs'^*_\II) \quad \Leftrightarrow \quad \yy'^* > \max_\II \ff'_\II(\xs^*_\II)
    \end{align*}
  \end{claim}
  This simply follows from the condition of the Lemma.  But
  in each case above, one of the assignments $\yy^*$ or $\yy'^*$ is not
 an  optimal min-max assignment as there is an alternative assignment that has a lower
  maximum over all factors.
\end{proof}

This lemma simply states that what matters in the min-max solution is
the \emph{relative ordering} in the factor-values.

Let $\yy\nn{1} \leq \ldots \leq \yy\nn{\vert \YY \vert}$ be an ordering of elements in
$\YY$, and let $\rank(f_I(x_I))$ denote the rank in $\{1,\ldots,|\YY|\}$ of $\yy_\II = \ff_\II(\xs_\II)$
in this ordering.  Define the min-sum reduction of $\{\ff_\II\}_{\II \in
  \FF}$ as
\begin{align*}
  \fg_\II(\xs_\II) = 2^{\rank(\ff_\II(\xs_\II))} \quad  \forall \II \in \FF
\end{align*}

\begin{theorem}\label{th:minsum}
  \begin{align*}
    \arg_{\xs} \min  \sum_\II \fg_\II(\xs_\II) \quad = \quad \arg_{\xs} \min \max_\II \ff_\II(\xs_\II)
  \end{align*}
  where $\{\fg_\II\}_\II$ is the min-sum reduction of $\{\ff_\II\}_\II$.
\end{theorem}
\begin{proof}
  First note that since $g(z) = 2^z$ is a monotonically increasing
  function, the rank of elements in the range of $\{\fg_\II \}_\II$ is the
  same as their rank in the range of $\{\ff_\II \}_\II$.  Using
  Lemma~\ref{th:lemma}, this means 
  \begin{align}\label{eq:minsumproof}
    \arg_{\xs}\min \max_\II \fg_\II(\xs_\II) \quad = \quad \arg_{\xs} \min \max_\II \ff_\II(\xs_\II).
  \end{align}

  Since $2^z > \sum_{l = 0}^{z-1} 2^l$, by definition of $\{\fg_\II\}$ we
  have
  \begin{align*}
    \max_{\II \in \FF} \fg_\II(\xs_\II)  > \sum_{\II \in \FF \back \II^*} \fg_\II(\xs_\II) \quad \text{where}\;\; \II^* = \arg_\II\max \fg_\II(\xs_\II) 
  \end{align*}
  It follows that for $\xs^1, \xs^2 \in \XX$,
  \begin{align*}
    \max_\II \fg_\II(\xs^1_\II) < \max_\II \fg_\II(\xs^2_\II) \quad \Leftrightarrow \quad \sum_\II \fg_\II(\xs^1_\II) < \sum_\II \fg_\II(\xs^2_\II)
  \end{align*}
  Therefore
  \begin{align*}
    \arg_{\xs} \min \max_\II \fg_\II(\xs_\II) \quad = \quad \arg_{\xs} \min \sum_\II \fg_\II(\xs_\II).
  \end{align*}
  This equality, combined with \cref{eq:minsumproof}, prove the
  statement of the theorem.
\end{proof}

An alternative approach is to use an inverse temperature parameter $\beta$ and
re-state the min-max objective as the min-sum objective at the low temperature limit
\begin{align}\label{eq:tmp_limit}
  \lim_{\beta \to +\infty} \arg_{\xs} \min  \sum_\II \ff^{\beta}_\II(\xs_\II) \quad = \quad \arg_{\xs} \min \max_\II \ff_\II(\xs_\II)
\end{align}

\subsubsection{Min-max reduces to sum-product}\label{sec:minmax2sumprod}
\index{reduce!min-max to sum-product}
\index{reduce!min-max to CSP}
Recall that
$\YY = \bigcup_{\II \in \FF} \YY_\II$ denote the union over the range
of all factors.
For any $\yy \in \YY$, we
\emph{reduce} the original min-max problem to a CSP using the
following reduction.  \marnote{$\pp_{\yy}$-reduction}
\index{Py-reduction}
\begin{definition}
For any $\yy \in \YY$,
\magn{$\pp_{\yy}$-reduction} of the min-max problem:
\begin{align}\label{eq:min-max}
  \xs^* \quad = \quad \arg_{\xs} \min \; \max_{\II \in \FF} \; \ff_{\II}(\xs_\II)
\end{align}
is given by
\begin{align}\label{eq:minmax_reduction}
  \pp_{\yy}(\xs) \quad \defeq \quad \frac{1}{\qq_{\yy}(\emptyset)} \prod_{\II \in \FF} \ident(\ff_{\II}(\xs_{\II}) \leq \yy) 
\end{align}
where $\qq_{\yy}(\emptyset)$ is the normalizing constant.\footnote{ To always have a well-defined probability, we
  define $\frac{0}{0} \defeq 0$.}  
\end{definition}
This distribution defines a CSP
over $\XX$, where $\pp_{\yy}(\xs) > 0$ \textit{iff} $\xs$ is a
satisfying assignment.  Moreover, $\qq_\yy(\emptyset)$ gives the number of satisfying
assignments.
The following theorem is the basis of our reduction.

\begin{theorem}\label{th:minmax}
  Let $\xs^*$ denote the min-max solution and $\yy^*$ be its
  corresponding value --\ie $\yy^* = \max_{\II} \; \ff_{\II}(\xs^*_{\II})$.  Then
  $\pp_\yy(\xs)$ is satisfiable for all $\yy \geq \yy^*$ (in particular
  $\pp_\yy(\xs^*) > 0$) and unsatisfiable for all $\yy < \yy^*$.
\end{theorem}
\begin{proof}
  \emph{(A) $\pp_\yy$ for $\yy \geq \yy^*$ is satisfiable:} It is enough
  to show that for any $\yy \geq \yy^*$, $\pp_\yy(\xs^*) > 0$.  But since
  \begin{align*}
    \pp_{\yy}(\xs^*) \quad = \quad \frac{1}{\qq_\yy(\emptyset)} \prod_{\II} \ident(\ff_\II(\xs^*_\II) \leq \yy) 
  \end{align*}
  and $\ff_\II(\xs^*_\II) \leq \yy^* \leq \yy$, all the indicator functions on
  the rhs evaluate to $1$, showing that $\pp_{\yy}(\xs^*) > 0$.

  \emph{(B) $\pp_\yy$ for $\yy < \yy^*$ is not satisfiable:} Towards a contradiction assume
  that for some $\underline{y} < \yy^*$, $\pp_{\underline{y}}$ is
  satisfiable.  Let $\underline{\xs}$ denote a satisfying assignment
  --\ie $\pp_{\underline{y}}(\underline{\xs}) > 0$. Using the definition
  of $\pp_\yy$-reduction, this implies that $\ident(\ff_\II(\underline{\xs}_\II)
  \leq \underline{y}) > 0$ for all $\II \in \FF$.  However this
  means that $\max_\II \ff_\II(\underline{\xs}_\II) \leq \underline{y} < \yy^*$,
  which means $\yy^*$ is not the min-max value. 
\end{proof}

\index{binary search}
This theorem enables us to find a min-max assignment by solving a \marnote{binary search}
sequence of CSPs.  Let $\yy\nn{1} \leq \ldots \leq \yy\nn{\vert \YY \vert}$ be an
ordering of $\yy \in \YY$.  Starting from $\yy = \yy\nn{\lceil N/2
  \rceil}$, if $\pp_\yy$ is satisfiable then $\yy^* \leq \yy$. On the
other hand, if $\pp_\yy$ is not satisfiable, $\yy^* > \yy$.  Using \textbf{binary
search}, we need to solve $\log(\vert \YY \vert)$ CSPs to find the min-max
solution.  Moreover at any time-step during the search, we have both
upper and lower bounds on the optimal solution. That is \marnote{lower \& upper bounds}
$\underline{\yy} < \yy^* \leq \overline{\yy}$, where
$\pp_{\underline{\yy}}$ is the latest unsatisfiable and
$\pp_{\overline{\yy}}$ is the latest satisfiable reduction.

However, finding an assignment $\xs^*$ such that $\pp_\yy(\xs^*) > 0$ or
otherwise showing that no such assignment exists, is in general, \NP-hard.
\index{incomplete solver}
Instead, we can use an incomplete solver \cite{kautz2009incomplete}, which may find a solution \marnote{incomplete solver}
if the CSP is satisfiable, but its failure to find a solution does not guarantee unsatisfiability.
By using an incomplete
solver, we lose the lower bound
$\underline{y}$ on the optimal min-max solution.\footnote{To maintain the lower bound one should be able to correctly assert unsatisfiability.}
However the following theorem states that, as we increase
$\yy$ from the min-max value $\yy^*$, the number of satisfying assignments
to $\pp_\yy$-reduction increases, making it potentially easier to solve.

\begin{proposition}\label{th:z}
  \begin{align*}
    \yy_1 < \yy_2 \quad \Rightarrow \quad \qq_{\yy_1}(\emptyset) \leq \qq_{\yy_2}(\emptyset) \quad \quad \forall \yy_1, \yy_2 \in \YY
  \end{align*}
  where $\qq_{\yy}(\emptyset)$ (\ie partition function) is the number of solutions of $\pp_\yy$-reduction.
\end{proposition}
\begin{proof}
Recall the definition $\qq_{\yy}(\emptyset)  =
  \sum_{\xs} \prod_{\II} \ident(\ff_\II(\xs_\II) \leq \yy)$.  For $\yy_1 < \yy_2$
  we have:
  \begin{align*}
    &\ff_\II(\xs_\II) \leq \yy_1 \quad \rightarrow \quad \ff_\II(\xs_\II) \leq \yy_2 \;& \Rightarrow\\
    &\ident(\ff_\II(\xs_\II) \leq \yy_1) \leq \ident(\ff_\II(\xs_\II) \leq \yy_2) \; &\Rightarrow\\
    &\sum_{\xs} \prod_{\II} \ident(\ff_\II(\xs_\II) \leq \yy_1) \leq  \sum_{\xs} \prod_{\II} \ident(\ff_\II(\xs_\II) \leq \yy_2)\;& \Rightarrow \\
    &\qq_{\yy_1}(\emptyset) \leq \qq_{\yy_2}(\emptyset)&
  \end{align*} 
\end{proof}

This means that the sub-optimality of our solution is related to our
ability to solve CSP-reductions -- that is, as the gap $\yy - \yy^*$
increases, the $\pp_\yy$-reduction potentially becomes easier to solve.

%% file: chap3.tex
\chapter{Approximate inference}\label{chapter:inference}

\section{Belief Propagation}\label{sec:bp}
\input{bp.tex}

\section{Tractable factors}\label{sec:tractable}
\input{tractible_factors.tex}

\section{Inference as optimization}\label{sec:variational}
\input{variational_methods.tex}

\section{Loop corrections}\label{sec:loops}
\input{loops.tex}


\section{Survey Propagation: semirings on semirings}\label{sec:meta}
\input{meta_constructions.tex}

\section{Messages and particles}\label{sec:hybrid}
\input{hybrid_methods.tex}

%% file: bp.tex
A naive approach to inference over commutative semirings  
\begin{align}\label{eq:semiringinference}
\qq(\xs_{\JJ}) \quad = \quad \bigoplus_{\xs_{\back \JJ}} \bigotimes_{\II} \ff_{\II}(\xs_\II) 
\end{align}
or its normalized version (\refEq{eq:semiring_marginalization}), is to construct a complete $N$-dimensional array of $\qq(\xs)$
using the tensor product $\qq(\xs) \; = \; \bigotimes_\II \ff_\II(\xs_\II)$ and then perform $\oplus$-marginalization.
However, the number of elements in $\qq(\xs)$ is $\vert \XX \vert$, which is exponential in $N$, the number of variables.

If the factor-graph is loop free, 
we can use distributive law to make inference tractable. 
Assuming $\qq(\xs_\KK)$ (or $\qq(\xx_k)$) is the marginal of interest, form a tree with $\KK$ (or $k$) as its root.
Then starting from the leaves, using the distributive law, we can move the $\oplus$ inside the $\otimes$ 
\index{distributive law}
 and define ``messages'' from leaves towards the root as follows:
\index{message passing!BP}
\index{BP}
\index{belief propagation|see {BP}}
\index{semiring!message passing}
\index{inference!BP|see {BP}}
\begin{align}
\msgq{i}{{{\II}}}(\xx_i) \quad & = \quad  \bigotimes_{\JJ \in \nb{i} \back \II} \msgq{{\JJ}}{i}(\xx_i) \label{eq:miI_semiring}\\
\msgq{{{\II}}}{i}(\xx_i) \quad & = \quad  \bigoplus_{ \xs_{\back i}} \ff_{{\II}}(\xs_{\II})
\bigotimes_{j \in \nb \II \back i} \msgq{j}{{\II}}(\xx_{j})  \label{eq:mIi_semiring}
\end{align}
where \refEq{eq:miI_semiring} defines the message from a variable to a factor, 
closer to the root and similarly \refEq{eq:mIi_semiring} defines the message from factor ${\II}$ to a variable $i$ closer to the root.
Here, the distributive law allows moving the $\bpplus$ over the domain $\XX_{\II \back i}$ from outside to inside of \refEq{eq:mIi_semiring} --
the same way $\oplus$ moves its place in $(a \otimes b) \oplus (a \otimes c)$ to give $a \otimes (b \oplus c)$, where 
$a$ is analogous to a message.

\begin{figure}
\centering
\includegraphics[width=.5\textwidth]{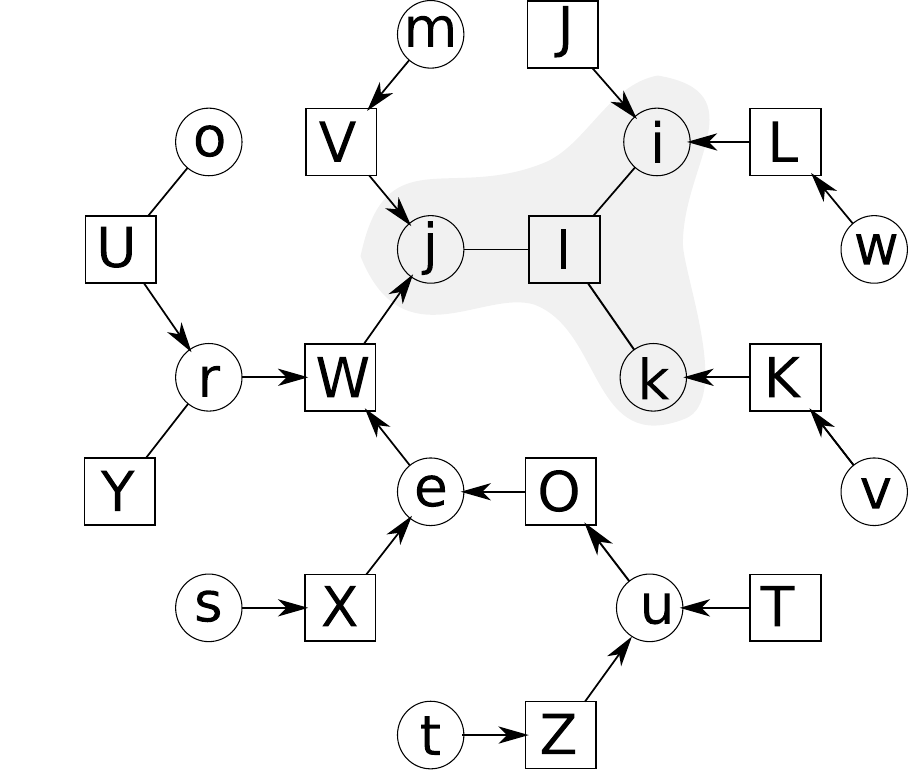}
\caption[Belief Propagation on a loop-free factor-graph]{The figure shows a loop-free factor-graph and the direction of 
messages sent between variable and factor nodes in order to calculate the marginal over the grey region.}
\label{fig:bp-tree}
\end{figure}

By starting from the leaves, and calculating the messages towards the root, we obtain 
the marginal over the root node as the product of incoming messages
\begin{align}
\qq(\xx_{k}) \quad &= \quad  \bigotimes_{\II \in \nb k} \msgq{\II}{k}(\xx_k)
\end{align}
In fact, we can assume any
subset of variables $\xs_{\AAA}$ (and factors within those variables) to be the root. Then, the set of all incoming messages to $\AAA$, 
produces the marginal 
\index{marginalization!BP}
\begin{align}\label{eq:bpmarg_region}
\qq(\xs_{\AAA}) \quad = \quad \left ( \bigotimes_{\II \subseteq \AAA} \ff_{\II}(\xs_{\II}) \right ) \left (\bigotimes_{i \in \AAA, \JJ \in \nb i, \JJ \not \subseteq \AAA} \msgq{\JJ}{i}(\xs_i) \right )
\end{align}

\begin{example}
Consider the joint form represented by the factor-graph of  \refFigure{fig:bp-tree} 
\begin{align*}
  \qq(\xs) \quad = \quad \bigotimes_{\AAA \in \{\II,\JJ,\KK, \mathrm{L},\mathrm{O},\mathrm{T},\mathrm{U},\mathrm{V},\mathrm{W},\mathrm{X},\mathrm{Y},\mathrm{Z}\}}  \ff_{\AAA}(\xs_{\AAA}) 
\end{align*}
and the problem of calculating the marginal over $\xs_{\{i,j,k\}}$ (\ie the shaded region).
\begin{align*}
  \qq(\xs_{\{i,j,k\}}) \quad = \quad \bigoplus_{\xs_{\back \{i,j,k\}}} \bigotimes_{\AAA \in \{\II,\JJ,\KK, \mathrm{L},\mathrm{O},\mathrm{T},\mathrm{U},\mathrm{V},\mathrm{W},\mathrm{X},\mathrm{Y},\mathrm{Z}\}}  \ff_{\AAA}(\xs_{\AAA}) 
\end{align*}

We can move the $\oplus$ inside the $\otimes$ to obtain
\begin{align*}
 \qq(\xs_{\{i,j,k\}}) \quad = \quad  \ff_{\II}(\xs_{\II}) \bptimes \msgq{\mathrm{L}}{i}(\xx_i) \bptimes \msgq{\mathrm{K}}{i}(\xx_i) \bptimes \msgq{\mathrm{V}}{j}(\xx_j) \bptimes \msgq{\mathrm{W}}{j}(\xx_j) \bptimes \msgq{\mathrm{K}}{k}(\xx_k)
\end{align*}
where each term $\msgq{\AAA}{i}$ factors the summation on the corresponding sub-tree.
For example 
\begin{align*}
  \msgq{\mathrm{L}}{i} \quad = \quad \bigoplus_{\xx_w} \ff_{\mathrm{L}}(\xs_{\mathrm{L}})
\end{align*}

Here the message $\msgq{\mathrm{W}}{j}$ is itself a computational challenge
\begin{align*}
  \msgq{\mathrm{W}}{j} \quad = \quad \bigoplus_{\xs_{\back j}} \bigotimes_{\AAA \in \{\mathrm{W}, \mathrm{U}, \mathrm{Y}, \mathrm{X}, \mathrm{O}, \mathrm{T}, \mathrm{Z} \} }\ff_{\mathrm{\AAA}}(\xs_{\AAA})
\end{align*}

However we can also decompose this message over  sub-trees
\begin{align*}
  \msgq{\mathrm{W}}{j} \quad = \quad \bigoplus_{\xs_{\back j}} \ff_{\mathrm{\AAA}}(\xs_{\AAA})
\bptimes \msgq{e}{\mathrm{W}}(\xx_{e}) \bptimes \msgq{r}{\mathrm{W}}(\xx_{r})
\end{align*}
where again using distributive law $\msgq{e}{\mathrm{W}}$ and  $\msgq{r}{\mathrm{W}}$ 
further simplify based on the incoming messages to the variable nodes $\xx_r$ and $\xx_e$.
\end{example}

This procedure is known as Belief Propagation (BP), which is sometimes prefixed with the corresponding semiring \eg sum-product BP.
Even though BP is only guaranteed to produce correct answers when the factor-graph is a
tree (and few other cases \cite{Aji1998,Weiss2001a,bayati2005maximum,weller2013map}), 
it performs surprisingly well when applied as a fixed point iteration to graphs with loops \cite{Murphy1999,gallager1962low}. In the case of loopy graphs the message updates are repeatedly applied in the hope of convergence. This is in contrast with BP on trees, where
the messages -- from leaves to the root -- are calculated only once. 
\index{factor graph!loopy}
The message update can be applied
to update the messages either synchronously or asynchronously and the update
\index{convergence!BP}
\index{update schedule}
schedule can play an important role in convergence (\eg \cite{Elidan2006,kolmogorov2006convergent}).
Here, for numerical stability, when the $\otimes$ operator has an inverse, the messages are normalized. We use $\propto$ to indicate this normalization according to the mode of inference  
\index{BP!normalized}
\index{fixed point}
\begin{align}
\msg{{{\II}}}{i}(\xx_i) \quad & \propto \quad   \bigoplus_{ \xs_{\back i}} \ff_{{\II}}(\xs_{\II})
\bigotimes_{j \in \nb \II \back i} \msg{j}{{\II}}(\xx_{j})  & \;\propto\; \MSG{\II}{i}(\msgss{\nb \II \back i}{\II})(\xx_i)\label{eq:mIi_semiring_norm}\\
\msg{i}{{{\II}}}(\xx_i) \quad & \propto \quad   \bigotimes_{\JJ \in \nb{i} \back \II} \msg{{\JJ}}{i}(\xx_i)  &\;\propto\; \MSG{i}{\II}(\msgss{\nb i \back \II}{i})(\xx_i)\label{eq:miI_semiring_norm}\\
\ph(\xs_{\II}) \quad &\propto \quad   \ff_{\II}(\xs_{\II}) \bigotimes_{i \in \nb \II} \msgs{i}{\II}(\xx_i)  &\quad \label{eq:marg_semiring_factor_norm}\\
\ph(\xx_{i}) \quad & \propto \quad    \bigotimes_{\II \in \nb i} \msg{\II}{i}(\xx_i)  &\quad \label{eq:marg_semiring_norm}
\end{align}
Here, for general graphs, $\ph(\xx_i)$ and $\ph(\xs_{\II})$ are approximations to $\pp(\xx_i)$
and $\pp(\xs_\II)$ of \refEq{eq:semiring_marginalization}.
\index{BP!functional form}
The \textbf{functionals} $\MSG{i}{\II}(\msg{\nb i \back \II}{i})(.)$ and 
$\MSG{\II}{i}(\msgss{\nb \II \back i}{\II})(.)$ cast the BP message updates as an operator on a subset of incoming messages -- \ie $\msgss{\nb i \back \II}{i} = \{ \msg{\JJ}{i} \mid \JJ \in \nb i \back \II \}$. We use these functional notation in presenting the algebraic form of survey propagation in \refSection{sec:meta}.

Another heuristic that is often employed with sum-product and min-sum BP is the \textbf{Damping} of messages.
This often improves the convergence when BP is applied to loopy graphs. Here a damping parameter $\lambda \in (0,1]$
is used to partially update the new message based on the old one -- \eg for sum-produt BP we have
\index{damping}
\begin{align}
\msg{{{\II}}}{i}(\xx_i) \quad & \propto \quad \lambda \msg{{{\II}}}{i}(\xx_i) + (1 - \lambda) \bigg (\sum_{ \xs_{\back i}} \ff_{{\II}}(\xs_{\II})
\prod_{j \in \nb \II \back i} \msg{j}{{\II}}(\xx_{j}) \bigg ) \label{eq:mIi_sumprod}\\
\msg{i}{{{\II}}}(\xx_i) \quad & \propto \quad \lambda \msg{i}{{{\II}}}(\xx_i) + (1 - \lambda) \bigg (\prod_{\JJ \in \nb{i} \back \II} \msg{{\JJ}}{i}(\xx_i)\bigg )  \label{eq:miI_sumprod}\\
\end{align}
where as an alternative one may use the more expensive form of geometric damping (where $\lambda$ appears in the power) or apply damping to either variable-to-factor or factor-to-variable messages but not both. Currently -- similar to several other ideas that we explore in this thesis -- damping is a ``heuristic'', which has proved its utility in applications but lacks theoretical justification.

\subsection{Computational Complexity}\label{sec:bpcomplexity}
\index{complexity!BP}
The time complexity of a single variable-to-factor message update (\cref{eq:miI_semiring}) 
is $\OO(| \nb i | \; |
\XX_i|)$.
To save on computation, when a variable has a large number of neighbouring factors,
and if none of the message values is equal to the annihilator $\identt{\bpplus}$ (\eg zero for the sum-product), 
and the inverse of $\otimes$ is defined, we can derive the marginals once, and 
produce variable-to-factor messages as
\begin{align}\label{eq:belief_update}
  \msg{i}{\II}(\xx_i) \quad = \quad \ph(\xx_i) \otimes \left ( \msg{\II}{i}(\xx_i) \right )^{-1} \quad \forall \II \in \nb i  
\end{align}

This reduces the cost of calculating all variable-to-factor messages
\index{v-sync update}
\index{variable-synchronous update}
leaving a variable from  $\OO(\vert \XX_i \vert\; \vert \nb i \vert^2 )$ to $\OO(\vert \XX_i \vert\; \vert \nb i \vert )$.\marnote{v-sync update}
We call this type of  BP update, \magn{variable-synchronized (v-sync)} update. Note that since $\max$ is not Abelian on any non-trivial
ordered set, min-max BP does not allow this type of variable-synchronous update. This further motivates using the sum-product reduction
of min-max inference.
\index{BP!update schedule}
The time complexity of a single factor-to-variable message update (\refEq{eq:mIi_semiring}) 
is $\OO(\vert \XX_{\II}\vert)$. However as we see in \refSection{sec:tractable},
sparse factors allow much faster updates. Moreover 
in some cases, we can reduce the time-complexity by calculating all the 
factor-to-variable messages that leave a particular factor at the same time 
\index{factor-synchronous update}
\index{f-sync update}
(\eg \refSection{sec:tsp}). We call this type of synchronized update, \marnote{$\ff$-sync update} \textbf{factor-synchronized ($\ff$-sync)} update.

\subsection{The limits of message passing}\label{sec:limits}
By observing the application of distributive law in semirings, a natural question to ask is: can we use distributive law for polynomial time inference on loop-free graphical models over any of the inference problems at higher levels of inference hierarchy or in general any inference problem with more than one marginalization operation?
The answer to this question is further motivated by the fact that, when loops exists, the same scheme may become a powerful approximation technique. 
When we have more than one marginalization operations, a natural assumption in using distributive law 
is that the expansion operation distributes over all the marginalization operations -- \eg  as in min-max-sum (where sum distributes over both min and max), min-max-min, xor-or-and.
\index{marginalization!two operations}
\index{inference!efficient}
Consider the simplest case with three operators $\opp{1}$, $\opp{2}$ and $\otimes$, where $\otimes$ distributes over both $\opp{1}$ and $\opp{2}$.
Here the integration problem is
$$\qq(\emptyset) \quad = \quad \bigopp{2}_{\xs_{\JJ_2}} \bigopp{1}_{\xs_{\JJ_1}} \bigotimes_{\II} \ff_{\II}(\xs_\II)$$
where $\JJ_1$ and $\JJ_2$ partition $\{1,\ldots,N\}$.

In order to apply distributive law for each pair $(\opp{1}, \otimes)$ and $(\opp{2}, \otimes)$, we need to be able to commute $\opp{1}$ and $\opp{2}$ operations.
That is, we require 
\begin{align}\label{eq:op_commute}
\bigopp{1}_{\xs_{\AAA}} \bigopp{2}_{\xs_{\BBB}} \fg(\xs_{\AAA \cup \BBB}) = \bigopp{2}_{\xs_\BBB} \bigopp{1}_{\xs_{\AAA}} \fg(\xs_{\AAA \cup \BBB}).
\end{align}
for the specified $\AAA \subseteq \JJ_1$ and $\BBB \subseteq \JJ_2$.

Now, consider a simple case involving two binary variables $\xx_i$ and $\xx_j$, where $\fg(\xs_{\{i,j\}})$ is 
\vspace{.05in}
\begin{center}
\scalebox{.8}{
\begin{tabu}{r r |[2pt] c | c }
\multicolumn{2}{c}{}&\multicolumn{2}{c }{$\xx_j$}\\
\multicolumn{2}{c}{}&0&1\\\tabucline[2pt]{3-4}
\multirow{2}{*}{$\xx_i$}&0&a&\multicolumn{1}{c |[2pt]}{b}\\\cline{2-4}
&1&c&\multicolumn{1}{c |[2pt]}{d}\\\tabucline[2pt]{3-4}
\end{tabu}
}
\end{center}
\vspace{.05in}
Applying \refEq{eq:op_commute} to this simple case (\ie $\AAA = \{i\}, \BBB=\{j\}$), we require
$$
  (a \opp{1} b) \opp{2} (c \opp{1} d) \quad = \quad (a \opp{2} b) \opp{1} (c \opp{2} d). 
$$

\index{Eckmann-Hilton Lemma}
The following theorem leads immediately to a  negative result:
\begin{theorem}\cite{eckmann1962group}:
\begin{align*}
  (a \opp{1} b) \opp{2} (c \opp{1} d) \; = \; (a \opp{2} b) \opp{1} (c \opp{2} d)
\quad \Leftrightarrow \quad \opp{1} = \opp{2} \quad \forall a,b,c
\end{align*}
\end{theorem}
\noindent which implies that \emph{direct application of distributive law to tractably and exactly solve any  inference problem with more than one marginalization operation is unfeasible, even for tree structures.} This limitation was previously known for marginal MAP inference~\cite{park2004complexity}.
\index{marginal MAP}

Min and max operations have an interesting property in this regard.
Similar to any other operations for min and max we have
\begin{align*}
  \min_{\xs_{\JJ}} \max_{\xs_{\II}} \fg(\xs_{\II \cup \JJ}) \neq \max_{\xs_{\II}} \min_{\xs_{\JJ}} \fg(\xs_{\II \cup \JJ})
\end{align*}

However,
\index{minimax theorem}
\index{game theory}
\index{graphical games}
\index{graphical models!game theory}
 if we slightly change the inference problem (from pure assignments $\xs_{\JJ_l}  \in \XX_{\JJ_l}$ to a distribution over assignments; \aka mixed strategies), as a result of the celebrated \emph{minimax theorem} \cite{von2007theory}, the min and max operations commute -- \ie
\begin{align*}
   \min_{\strat(\xs_{\JJ})} \max_{\strat(\xs_{\II})} \sum_{\xs_{\II \cup \JJ}} \strat(\xs_{\JJ}) \fg(\xs_{\II \cup \JJ}) \strat(\xs_{\II}) \quad = \quad \max_{\strat(\xs_{\II})} \min_{\strat(\xs_{\JJ})} \sum_{\xs_{\II \cup \JJ}}  \strat(\xs_{\II}) \fg(\xs_{\II \cup \JJ}) \strat(\xs_{\JJ_1})
\end{align*}
where $\strat(\xs_{\JJ_1})$ and $\strat(\xs_{\JJ_2})$ are mixed strategies.
This property has enabled addressing problems with min and max marginalization operations using message-passing-like procedures. For example, \citet{ibrahimi2011robust} solve this (mixed-strategy) variation of min-max-product inference. Message passing procedures that operate on graphical models for game theory  (\aka ``graphical games'')  also rely on this property~\cite{ortiz2002nash,kearns2007graphical}.

%% file: tractible_factors.tex
The applicability of graphical models to discrete optimization problems is limited by the size and number of factors in the factor-graph. In \refSection{sec:hop} we review some of the large order factors that allow efficient message passing, focusing on the sparse factors used in \refChapter{chapter:combinatorial} to solve combinatorial problems. 
In \refSection{sec:augmentation} we introduce an augmentation procedure
similar to cutting plane method to deal with large
number of ``constraint'' factors.

\subsection{Sparse factors}\label{sec:hop}
\index{factor!sparse}
\index{sparse factors}
The factor-graph formulation of many interesting combinatorial problems
involves sparse (high-order) factors. 
\index{high-order potentials}
Here, either the factor involves a large number of variables, or the variable domains, $\XX_i$, have 
large cardinality.
In all such factors, we are able
to significantly reduce the $\OO(\vert\XX_{\II} \vert)$ time complexity of
calculating factor-to-variable messages.  Efficient message passing
over such factors is studied by several works in the context of
sum-product and min-sum inference classes 
\cite{potetz2008efficient,gupta2007efficient,tarlow2010hop,tarlow2012fast,Rother2009}.
Here we confine our discussion to some of the factors used in \refChapter{chapter:combinatorial}.

The application of such sparse factors are common in vision.
Many image labelling solutions
to problems such as
image segmentation and stereo reconstruction, operate using priors that
enforce similarity of neighbouring pixels. The image processing task is then usually
reduced to finding the MAP solution.
 However pairwise potentials are insufficient for capturing the statistics of natural
images and therefore higher-order-factors have been employed \cite{Roth2009,Paget1998,Kohli2007,Kohli2008,Kohli2009,
Komodakis2009,Lan2006}. 

\index{Potts factor}
\index{factor!Potts}
The simplest form of sparse factor in combinatorial applications is the \marnote{Potts factors}
\magn{Potts} factor, $\ff_{\{i,j\}}(\xx_i, \xx_j) = \ident(\xx_i = \xx_j)$. 
This factor assumes the same domain for all the variables
($\XX_i = \XX_j \; \forall i,j$) and its tabular form is
non-zero only across the diagonal.  It is easy to see that this allows
the marginalization of \cref{eq:mIi_semiring} to be performed in
$\OO(\vert\XX_i\vert)$ rather than $\OO(\vert\XX_i\vert\;
\vert\XX_j\vert)$.  Another factor of similar form is the inverse Potts
factor, $\ff_{\{i,j\}}(\xx_i, \xx_j) = \ident(\xx_i \neq \xx_j)$, which
ensures $\xx_i \neq \xx_j$.  In fact any pair-wise factor that is a
constant plus a \magn{band-limited} matrix allows $\OO(\vert\XX_i\vert)$ \marnote{band-limited factors}
inference (\eg factors used for bottleneck TSP in \refSection{sec:btsp}).

\index{cardinality factors}
Another class of sparse factors is the class of \magn{cardinality factors}, where
$\XX_i = \{0,1\}$ and the factor is defined based on only the number of \marnote{cardinality factors}
non-zero values --\ie $\ff_{\II}(\xs_{\II}) = \fg(\sum_{i \in \nb \II} \xx_i)$. 
\citet{gail1981likelihood} proposes a simple $\OO(\vert\nb \II \vert \;
K)$ method for $\ff(\xs_{\II}) = \ident((\sum_{i \in \nb \II} \xx_i) =
K)$.  We refer to this factor as K-of-N factor and use similar
algorithms for at-least-K-of-N $\ff_{\II}(\xs_{\II}) = \ident((\sum_{i
  \in  \nb \II} \xx_i) \geq K)$ and at-most-K-of-N $\ff_{\II}(\xs_{\II}) =
\ident((\sum_{i \in \nb \II} \xx_i) \leq K)$ factors.
\index{factor!sparse!K-of-N}

\index{linear clique potentials}
An alternative is  the \magn{linear clique potentials} of \citet{potetz2008efficient}.  \marnote{linear clique potentials}
The authors propose a $\OO(\vert\nb \II \vert\; 
\vert\XX_i\vert^2)$ (assuming all variables have the same domain 
$\XX_i$) marginalization scheme for a general family of factors,
called linear clique potentials, where $\ff_{\II}(x_{\II}) =
\fg(\sum_{i \in \nb \II}\xx_i \ww_i)$ for a nonlinear $\fg_{\II}(.)$. 
For sparse factors with larger non-zero values (\ie larger $k$),
more efficient methods evaluate the sum of pairs of variables using
auxiliary variables forming a binary tree and use the Fast Fourier
Transform to reduce the complexity of K-of-N factors to $\OO( \vert \nb \II \vert 
\log(\vert \nb \II \vert)^2)$ (see \cite{tarlow2012fast} and references in there).

Here for completeness we provide a brief description of efficient message passing 
through at-least-K-of-N factors for sum-product 
and min-sum inference. 

\subsubsection{K of N factors for sum-product}
Since variables are binary, it is convenient to assume all
variable-to-factor messages are normalized such that $\msg{j}{\II}(0) =
1$. Now we calculate $\msg{\II}{i}(0)$ and
$\msg{\II}{i}(1)$ for at-least-K-of-N factors, and then normalize them 
such that $\msg{\II}{i}(0) = 1$.

In deriving $\msg{\II}{i}(0)$, we should assume that at least $K$ other variables that are adjacent 
to the factor $\ff_{\II}$ are nonzero and extensively use the assumption that $\msg{j}{I}(0) = 1$. 
The factor-to-variable message of \refEq{eq:mIi_semiring_norm} becomes
\begin{align}
  \msg{\II}{i}(0) \quad = &\quad \sum_{\xs_{\back i}} \ident \bigg( (\sum_{j \in \nb \II \back i} \xx_j) \geq K \bigg) \prod_{j \in \nb \II \back i } \msg{j}{\II}(\xx_j) \notag \\ 
   &\quad = \sum_{\AAA \subseteq \nb \II \back i,\; \vert\AAA\vert \geq K}\;\; \prod_{j \in \AAA} \msg{j}{\II}(1)\label{eq:kofn}
\end{align}
where the summation is over all subsets $\AAA$ of $\nb \II \back i$ that have at least $K$ members. 

Then, to calculate $\msg{\II}{i}(1)$ we follow the same
procedure, except that here the factor is replaced by $\ident\bigg ( (\sum_{j
  \in \nb \II \back i} \xx_j) \geq K-1 \bigg )$. This is because here we
assume $\xx_i = 1$ and therefore it is sufficient for $K-1$ other
variables to be nonzero.

Note that in \refEq{eq:kofn}, the sum iterates over ``all'' $\AAA \subseteq \nb \II \back i$ of size at least $K$. For high-order factors $\ff_\II$ (where $|\II|$ is large), this summation contains 
\index{dynamic programming}
an exponential number of terms. Fortunately, we can use dynamic programming to perform
this update in $\OO(\vert \nb \II \vert\; K)$.
The basis for the recursion of dynamic programming is that,
starting from $\BBB = \II \back i$, a variable $\xx_k \in \xs_{\KK}$ can be either zero or one
\begin{align*}
&  \sum_{\AAA \in \{ \KK \subseteq \BBB, \vert \KK \vert \geq k \}}  \prod_{j \in \AAA} \msg{j}{\II}(1)  = \\   
 & \sum_{\AAA \in \{ \KK \subseteq  \BBB \back k,\; \vert \KK \vert \geq K \}} \prod_{j \in \AAA} \msg{j}{\II}(1)
  \;+ \; \msg{k}{\II}(1) \left (\sum_{\AAA \in \{\KK \subseteq  \BBB \back k,\; \vert \AAA \vert \geq K - 1 \}} \prod_{j \in \AAA} \msg{j}{\II}(1) \right )
\end{align*}
where each summation on the r.h.s.~can be further decomposed using similar recursion. Here, dynamic 
program reuses these terms so that each is calculated only once.


\subsubsection{K of N factors for min-sum}
Here again, it is more convenient to work with normalized variable-to-factor messages such that $\msg{j}{\II}(0) = \identt{\otimes} = 0$.
Moreover in computing the factor-to-variable message $\msg{\II}{i}(\xx_i)$ we also normalize it such that $\msg{\II}{i}(0) = 0$.
Recall the objective is to calculate
\begin{align*}
  \msg{\II}{i}(\xx_i) \quad = \quad \min_{\xs_{\back i}} \ident\bigg( (\sum_{j \in \nb \II} \xx_j) = K \bigg) \; \sum_{j \in \nb \II \back i} \msg{j}{\II}(\xx_j)
\end{align*}
for $\xx_i = 0$ and $\xx_i = 1$.

We can assume the constraint factor is satisfied, since if it is violated, 
the identity function evaluates to $+\infty$ (see \cref{eq:identity}).
For the first case, where $\xx_i = 0$, $K$ out of $\vert \nb \II \back i \vert$ neighbouring variables to factor $\II$ should be non-zero (because $\ident((\sum_{j \in \nb \II} \xx_j) = K)$ and $\xx_i = 0$).
The minimum is obtained if we assume the neighbouring variables with smallest $\msg{j}{\II}(1)$ are non-zero and the rest are zero. 
For $\xx_i = 1$, only $K-1$ of the remaining neighbouring variables need to be non-zero and therefore we need to find $K -1$ smallest of incoming messages ($\msg{j}{\II}(1)\; \forall j \in \nb \II \back i$)
as the rest of messages are zero due to normalization.

By setting the $\msg{\II}{i}(0) = 0$, and letting $\AAA(K) \subset \nb \II \back i$ identify the set of $K$ smallest incoming messages to factor $\II$, 
the  $\msg{\II}{i}(1)$ is given by
\begin{align*}
\msg{\II}{i}(1) \quad = \quad \bigg (\sum_{j \in \AAA(K)}  \msg{j}{\II}(1) \bigg ) - \bigg ( \sum_{j \in \AAA(K-1)}  \msg{j}{\II}(1) \bigg ) \quad = \quad \msg{j^{K}}{\II}(1)
\end{align*}
where $j^K$ is the index of $K^{th}$ smallest incoming message to $\II$, excluding $\msg{i}{\II}(1)$.
A similar procedure can give us the at-least-K-of-N and at-most-K-of-N factor-to-variable updates.

If $K$ is small (\ie a constant) we can obtain the $K^{th}$ smallest incoming message in $\OO(K\; \vert \nb \II \vert)$ time, and if $K$ is in the order of  $\vert \nb \II \vert$ this requires $\OO(\vert \nb \II \vert \log(\vert \nb \II \vert ))$ computations.
For both min-sum and sum-product, we incur negligible additional cost by calculating ``all'' the outgoing messages from factor $\II$ simultaneously (\ie $\ff$-sync update).

\subsection{Large number of constraint factors}\label{sec:augmentation}
We consider a scenario where an (exponentially) large number of factors represent hard constraints (see \refSection{def:constraint})
and ask whether it is possible to find a feasible solution by considering
only a small fraction of these constraints.
\index{augmentation}
The idea is to start from a graphical model corresponding to a computationally tractable subset of constraints, and after obtaining a solution for a sub-set of constraints (\eg using min-sum BP),
augment the model with the set of constraints that are violated in the current solution. 
This process is repeated in the hope that we might arrive at a solution that does not violate any of the
constraints, before augmenting the model with ``all'' the constraints.
Although this is not theoretically guaranteed to work, experimental results suggest this can be very 
efficient in practice.

\index{cutting plane}
This general idea has been extensively \marnote{cutting plane}
studied under the term \magn{cutting plane methods} in different settings. 
\citet{dantzig1954solution} first investigated this idea in the context 
\index{integer program}
of TSP and \citet{gomory1958outline} provided a elegant method to identify violated \marnote{integer program}
constraints in the context of finding integral solutions to linear programs (LP).
It has since been used to also solve a variety of nonlinear optimization problems.
In the context of graphical models, \citet{sontag2007new} (also \cite{sontag2007cutting}) use cutting plane method to iteratively tighten the
marginal polytope -- that enforces the local consistency of marginals; see \refSection{sec:variational} -- in order to 
improve the variational approximation. Here, we are interested in the augmentation process that
changes the factor-graph (\ie the inference problem) rather than improving the 
approximation of inference. 

The requirements of the cutting plane method are availability
of an optimal solver -- often an LP solver -- and a procedure to identify the violated
constraints. Moreover, they operate in real domain $\Re^d$; hence the term ``plane''. 
However, message passing  
can be much faster than LP in finding approximate MAP assignments for structured 
\index{linear program}
optimization problems~\cite{yanover2006linear}.
This further motivates using augmentation in the context of message passing.

In \cref{sec:tsp,sec:modularity}, 
we use this procedure to approximately solve
 TSP and graph-partitioning respectively.
Despite losing the guarantees that make cutting plane method very powerful, 
augmentative message passing has several 
advantages:
First, message passing is highly parallelizable.
Moreover by directly obtaining integral solutions, 
it is much easier to find violated constraints.
(Note the cutting plane method for combinatorial problems operates on 
\index{fractional solution}
{\em fractional} solutions, whose rounding may eliminate its guarantees.)
However, due to non-integral assignments, cutting plane methods require sophisticated tricks to find violations. For example, see \cite{applegate2006traveling} 
for application of cutting plane to TSP.

%% file: variational_methods.tex


\index{variational}
The variational approach is concerned with probabilities, and therefore this
section is limited to operations on real domain.
In the variational approach, \magn{sum-product} inference is expressed as \marnote{variational sum-product}
\begin{align}\label{eq:inf}
\ph \quad = \quad \arg_{\ph}\min \quad \divergence(\ph \mid \pp^{\beta})
\end{align}
\index{KL-divergence}
where $\divergence(\ph \mid \pp^{\beta})$ is the KL-divergence between our
approximation $\ph$ and the true distribution $\pp$ at inverse temperature ${\beta}$ (see  \refExample{example:ising}). 
Here $\ph$ is formulated in terms of desired marginals.

Expanding the definition of KL-divergence and substituting $\pp$ from \refEq{eq:p},
\cref{eq:inf} becomes \marnote{variational free energy}
\index{variational free energy}
 \begin{align}
\ph \quad = \quad  \arg_{\ph}\min & \quad  \sum_{\xs} \ph(\xs) \log(\ph(\xs)) 
\; - \; \beta \sum_{\xs} \ph(\xs) \log(\pp(\xs)) 
\equiv \label{eq:freeenergyinf}\\ 
  \arg_{\ph}\min & \quad \sum_{\xs} \ph(\xs) \log(\ph(\xs)) 
 \; - \;
\beta \sum_{\xs} \ph(\xs) \left (\sum_{\II} \log(\ff_{\II}(\xs_{\II})) \right ) 
\label{eq:helmholtz}
 \end{align}
where we have removed the log partition function $ \log(\qq(\emptyset,\beta)) =  
\log \left (\sum_{\xs} \prod_{\II} \ff_{\II}(\xs_{\II})^{\beta} \right )$
from \cref{eq:freeenergyinf} because it does not depend on $\ph$.
This means that the minimum of \refEq{eq:helmholtz} is  $ \log(\qq(\emptyset,\beta))$,
which appears when $\divergence(\ph\mid\pp^{\beta}) = 0$ -- \ie $\ph = \pp^\beta$.

The quantity being minimized in \cref{eq:helmholtz},
known as \magn{variational  free energy}, has two terms:
\index{expected energy}
 the \magn{(expected) energy} term 
$$\expenergy(\ph, \pp) \; \defeq \; -\sum_{\xs} \ph(\xs) \left (\sum_{\II} \log (\ff_{\II}(\xs_{\II})) \right ) $$ 
and the \magn{entropy} term 
\index{entropy}
$$\entropy(\ph) = -\sum_{\xs} \ph(\xs) \log(\ph(\xs)) \quad .$$
Different families of representations for $\ph(.)$ (in terms of its marginals)
produces different inference procedures such as BP, Generalized BP and
\index{mean-field}
Mean-field method \cite{Wainwright2007}.

\index{temperature limit}
Max-product (or \magn{min-sum}) inference is retrieved as zero-temperature limit of sum-product inference: \marnote{variational min-sum}
\begin{align}
\ph(.) \; &=\; \lim_{\beta \to +\infty} \; \arg_{\ph} \min \beta \expenergy(\ph, \pp) - \entropy(\ph) \notag \\ 
&\equiv \; \arg_{\ph} \min  - \sum_{\xs}  \ph(\xs) \big ( \sum_{\II} \log(\ff_\II(\xs_\II)) \big )\label{eq:minsum_variational}
\end{align}
where the energy term is linear in $\ph$
and therefore the optima will be at a corner of probability simplex, reproducing
the MAP solution.

Here by defining $\ff'(\xs) \leftarrow \frac{1}{\ff(\xs)}$, we get the min-sum form
\begin{align*}
\ph(.) \; = \; \arg_{\ph}\min \sum_{\xs} \ph(x_\II) \big ( \sum_{\II}\log(\ff'_\II(\xs_{\II}) \big )
\end{align*}

We observe that using a second \marnote{variational min-max}
parameter $\alpha$, \magn{min-max} inference is also retrievable\footnote{Here we assume there are no ties at the min-max solution \ie $\ff_{\II}(\xs^*_\II) > \ff_{\JJ}(\xs^*_\JJ) \;\; \forall \JJ \neq \II$.}
\begin{align}
\ph(.) \; &=\; \lim_{\alpha \to +\infty} \; \arg_{\ph} \min \sum_{\xs}\; \ph(\xs) \big (\sum_{\II}\log(\ff_\II(\xs_\II))^{\alpha}\big ) \;\notag \\
&\equiv\; \arg_{\ph} \min \; \sum_{\xs} \ph(\xs_\II) \big (\max_{\II} \log(\ff_\II(\xs_\II))\big )\label{eq:minmax_variational}
\end{align}
where again due to the linearity of the objective in $\ph$, the optima are at the extreme points of the probability simplex.

To retrieve sum-product BP update equations 
from divergence minimization of \cref{eq:inf},
\index{reparametrization}
\index{semiring!reparametrizations}
we will \magn{reparameterize} $\ph$ using its marginals $\ph(\xs_{\II})$ and $\ph(\xx_i)$. \marnote{reparametrization}
Here, we present this reparametrization in a more general form, as it holds for a any commutative semiring
where $(\RR, \otimes)$ is an abelian group.

\begin{proposition} \label{th:reparam}If the $\otimes$ operator of the semiring has an inverse and the factor-graph is loop-free, we can write $\pp(\xs)$ as
\begin{align}\label{eq:semiring_reparam}
  \ph(\xs) \; = \; \frac{\bigbptimes_\II \ph(\xs_\II)}{\bigbptimes_i \big ( \ph(\xx_i) \powerop ({ \vert \nb i \vert - 1 }) \big )}
\end{align}
where the inverse is w.r.t $\otimes$ and the exponentiation operator is defined as $a \powerop b \defeq \underbrace{a \otimes \ldots \otimes a}_{b\; \text{times}}$.
\end{proposition}
\begin{proof}
For this proof we use the exactness of BP on trees and substitute BP marginals \refEqs{eq:marg_semiring_norm}{eq:marg_semiring_factor_norm}
into \refEq{eq:semiring_reparam}:
\begin{align*}
  &\frac{\bigbptimes_\II \ph(\xs_\II)}{\bigbptimes_i \big ( \ph(\xx_i) \powerop ({ \vert \nb i \vert - 1 }) \big )} & \quad = \quad 
  &\frac{\bigbptimes_\II  \ff_{\II}(\xs_{\II}) \bigotimes_{i \in \nb \II} \msg{i}{\II}(\xx_i)}{\bigbptimes_i 
\big ( \bigotimes_{\II \in \nb i} \msg{\II}{i}(\xx_i) \powerop ({ \vert \nb i \vert - 1 }) \big )} & = \\
&\frac{\bigbptimes_\II  \ff_{\II}(\xs_{\II}) \bigotimes_{i \in \nb \II} \msg{i}{\II}(\xx_i)}{\bigbptimes_i 
\big ( \bigotimes_{\II \in \nb i} \msg{i}{\II}(\xx_i) \big )} & \quad = \quad
&\bigbptimes_\II  \ff_{\II}(\xs_{\II}) \quad = \quad  \pp(\xs)
\end{align*}
where we substituted the variable-to-factor messages in the denominator with factor-to-variable messages 
according to \refEq{eq:miI_semiring} and used the definition of inverse (\ie $a \bptimes a^{-1} = \identt{\bptimes}$)
to cancel out the denominator. 
\end{proof}

Intuitively, the denominator is simply cancelling the double counts --
that is since 
$\ph(\xx_i)$ is counted once for any $\II \in \nb i$ in the nominator,
the denominator removes all but one of them.

\subsection{Sum-product BP and friends}\label{sec:sumprod_n_friends}
Rewriting \cref{eq:semiring_reparam} for sum-product ring
$\ph(\xs)  = \frac{\prod_{\II} \ph(\xs_{\II})}{\prod_{i} \ph(\xx_i)^{\mid \nb i \mid -1}}$
and replacing $\ph$ in the variational energy minimization, we get \marnote{Bethe approximation}
\begin{align}
  \ph \quad &= \quad \arg_{\ph} \min \quad  \beta \sum_{\II} \ph(\xs_{\II}) \ff_{\II}(\xs_{\II}) \label{eq:bpenergy} \\
- &\left (\sum_{\II} \sum_{\xs_{\II}}\ph(\xs_{\II}) \log(\ph(\xs_{\II})) \right ) 
- \left ( \sum_{i} (1 - \mid \nb i \mid) \sum_{\xx_i} \ph(\xx_{i}) \log(\ph(\xx_{i})) \right ) \label{eq:bethe}\\
\text{such that}\quad &  \quad \sum_{\xx_{\back i}} \ph(\xs_{\II}) = \ph(\xx_i) \quad \forall i, \II \in \nb i \label{eq:marginalpoly} \\
 & \quad \sum_{\xx_i} \ph(\xx_i) = 1\label{eq:sumtoone}
\end{align}
where the energy term \cref{eq:bpenergy} is exact and the quantity that is minimized  is known as
\index{Bethe free energy}
 \magn{Bethe free energy}~\cite{bethe_statistical_1935,Yedidia2001}. 
The constraints \cref{eq:marginalpoly,eq:sumtoone} ensure that marginals are consistent and sum to one. 
Following the lead of \citet{Yedidia2001}, \citet{Heskes2003} showed that stable fixed points of sum-product BP are local optima of Bethe free energy.

The optimization above approximates the KL-divergence minimization of 
\index{marginal polytope}
\refEq{eq:helmholtz} in two ways: (I) While the marginal constraint  ensure local consistency, \marnote{marginal polytope}
for general factor-graphs there is no guarantee that even a joint probability $\ph$ with such 
marginals exists (\ie local consistency conditions outer-bound \magn{marginal polytope}; the polytope of marginals realizable by a join probability $\pp(\xs)$).
(II) Bethe entropy is not exact for loopy factor-graphs.
Using the method of Lagrange multipliers to enforce the local consistency 
constraints and setting the derivatives of \cref{eq:bethe} \wrt $\ph$ to zero,
recovers sum-product BP updates~\cite{Yedidia2001,Yedidia2005}. 
This optimization view of inference has inspired
\index{convex approximations}
\index{message passing!convergent}
many sum-product inference techniques with convex entropy approximations and 
convergence guarantees~\cite{Wainwright2005,wiegerinck2003fractional,Heskes2006,Teh2002,Yuille2002,Meshi2009,globerson2007approximate,Hazan2008}. 

\subsection{Min-sum message passing and LP relaxation}
LP relaxation of min-sum problem seeks marginals $\ph(\xs_\II) \forall \II$ \marnote{min-sum LP}
\begin{align}
  \ph \quad &= \quad \arg_{\ph} \min \quad  \sum_{\II} \ph(\xs_{\II}) \ff_{\II}(\xs_{\II}) \label{eq:lp} \\
\text{such that}\quad &  \quad \sum_{\xx_{ \back i}} \ph(\xs_{\II}) = \ph(\xx_i) \quad \forall i, \II \in \nb i \label{eq:lpmarginalpoly} \\
 & \quad \sum_{\xx_i} \ph(\xx_i) = 1\notag
\end{align}

If integral (\ie $\ph(\xx_i) = \ident(\xx_i = \xx^*_i)$
for some $\xs^* = \{\xx^*_1,\ldots,\xx^*_N\}$), this LP solution
is guaranteed to be optimal (\ie identical to \cref{eq:minsum_variational}).
Taking the zero temperature limit ($\lim \beta \to \infty$) of the Bethe free energy of \cref{eq:bpenergy,eq:bethe}, 
for any convex entropy approximation \cite{Wainwright2005,Wainwright2005a,Heskes06,Hazan2008}, \marnote{convex approximations}
\index{linear program}
ensures that sum-product message passing solution recovers the Linear Programming (LP) solution \cite{Weiss2007}.
Moreover, replacing the summation with maximization (which again corresponds to temperature limit) in the resulting convex message passing, 
produces the convex min-sum message passing, which agrees with LP relaxations, under some conditions (\eg when there are no ties in beliefs).
The general interest in recovering LP solutions by message passing is to retain its optimality guarantees while 
benefiting from the speed and scalability of message passing that stems from exploitation of graphical structure~\cite{Yanover2006}.
\index{replicates and splitting}
One may also interpret some of these convex variations as replicating variables and factors while keeping the corresponding messages identical over the \magn{replicates}~\cite{ruozzi2013message}. \marnote{replicates \& splitting}
After obtaining message updates, the number of replicates are allowed to take rational values (Parisi introduced a similar trick  for estimation of 
the partition function using replica trick \cite{Mezard1987,castellani_spin-glass_2005}). 

\index{message passing!MPLP}
Another notable variation for approximate MAP inference is max-product-linear-program \marnote{MPLP},
which performs block coordinate descend in the space of duals for \refEq{eq:lp}. MPLP is guaranteed  to converge and is often able to recover LP solution~\cite{globerson2008fixing,sontag2012tightening}.
\index{dual decomposition}
Finally \magn{dual (and primal) decomposition} methods minimize factors separately and combine their estimates in a way that agrees with sub-gradient in each \marnote{dual decomposition} iteration~\cite{Boyd2010,komodakis2007mrf,Komodakis2010a,jojic2010accelerated}.

\subsection{Min-max and other families}\label{sec:minmax_variational}
By rephrasing the variational min-max inference of \refEq{eq:minmax_variational} $$\ph = \arg_{\ph} \min \sum_{\xs} \ph(\xs_\II) \big (\max_{\II} \log(\ff_\II(\xs_\II))\big )$$
in terms of marginals $\ph(\xs_\II)$ and enforcing marginal consistency constraints, we obtain the following
LP relaxation \marnote{min-max LP}
\index{min-max}
\index{linear program! LP}
\begin{align}
  \ph \quad &= \quad \arg_{\ph} \min \quad y \label{eq:minmax_lp}\\
\text{such that}\quad  & \quad \sum_{\xs_\II} \ph(\xs_{\II}) \ff_{\II}(\xs_{\II}) \leq y \quad \forall \II\notag \\
\quad &  \quad \sum_{\xx_{\back i}} \ph(\xs_{\II}) = \ph(\xx_i) \quad \forall i, \II \in \nb i \notag \\
 & \quad \sum_{\xx_i} \ph(\xx_i) = 1\notag
\end{align}
which surprisingly resembles our sum-product reduction of min-max inference in \refSection{sec:minmax2sumprod}.
Here $\sum_{\xs_\II} \ph(\xs_{\II}) \ff_{\II}(\xs_{\II}) \leq y$ is a relaxation of our sum-product factor $\ident(\ff_\II(\xs_\II) \leq \yy)$ in \refEq{eq:minmax_reduction}.
\begin{claim}\label{th:minmax_lp}
$\yy$ in  \cref{eq:minmax_lp} lower bounds the min-max objective $\yy^*$. Moreover, if $\ph$  is integral, then $\yy = \yy^*$ and $\xs^*$
is the optimal min-max assignment.
\end{claim}
\begin{proof}
The integral solution $\ph$, corresponds to the following optimization problem
\begin{align}
  \xs^* \quad &= \quad \arg_{\xs} \min \quad y \label{eq:minmax_ip}\\
\text{such that}\quad  & \quad  \ff_{\II}(\xs_{\II}) \leq y \notag \\
&\equiv \quad \arg_{\xs} \min \; \max_{\II}\ff_{\II}(\xs_{\II}) \notag
\end{align}
which is the exact min-max inference objective. Therefore, for integral $\ph$, we obtain optimal min-max solution.
On the other hand by relaxing the integrality constraint, because of the optimality guarantee of LP, the LP solution $\yy$ can not be worse than
the integral solution and its corresponding value $\yy^*$.
\end{proof}

This lower bound complements \marnote{min-max lower bound}
 the upper bound that we obtain using a combination of sum-product reduction and an  incomplete solver (such as perturbed BP of \refSection{sec:pbp}) 
and can be used to assess the optimality of a min-max solution.

The only other extensively studied inference problem in the inference hierarchy of \refSection{sec:hierarchy} is
\magn{max-sum-product} (\aka marginal MAP) inference \cite{de2011new,jiang2011message,maua2012anytime}. 
In particular \marnote{variational marginal MAP}
variational formulation of max-sum-product inference~\cite{liu2013variational}, substitutes the entropy term in \cref{eq:freeenergyinf} with conditional entropy.

%% file: loops.tex
In this section we first review the region-based methods that account for short loops in \refSection{sec:regions} and 
then show how to perform loop correction by taking into account the message dependencies in \refSection{sec:message_dependency}. In \refSection{sec:generalizedloop}
we introduce a loop correction method that can benefit from both types of loop
corrections, producing more accurate marginals. While the region-based techniques can be used to directly estimate the integral,
the approximation techniques that take message dependencies into account are only applied for estimation of marginals.

\subsection{Short loops}\label{sec:regions}
We consider a general class of methods that improve inference in a
loopy graphical model by performing exact inference over 
regions that contain small
loops.

\index{junction-tree}
\index{junction-graph}
The earliest of such methods is \magn{junction-tree} \marnote{junction-tree}
\cite{Lauritzen1988,Jensen1990}, which performs exact inference  
with computation cost that grows exponentially in the size of largest
region --\ie tree width \cite{Chandrasekaran2008}. Here, regions form a
tree and the  messages are passed over regional intersections. While this
algorithm is still popular in applications that involve certain class of graphs \cite{Bodlaender1986} or
when the exact result is required, most graphs do not have a low tree width
\cite{Kloks1992,Kaminski2009}. 

An extension to junction tree is the \marnote{junction/cluster graph}
\magn{junction graph} method \cite{Aji01} that removes the requirement for the
regions to form a tree.
For this, the proxy between two regions is a subset of their intersection
(rather than the whole intersection) and one still requires the regions that
contain a particular variable to form a tree. Similar ideas are discussed under
the name of cluster graphs in \cite{koller_probabilistic_2009}.

\index{generalized BP}
Inspired by the connection between Bethe free energy and belief 
propagation (see \refSection{sec:variational}),
\citet{Yedidia2001} proposed \magn{Generalized BP} 
 that minimizes Kikuchi approximation to free energy
(\aka \magn{Cluster Variational Method} \cite{Kikuchi1951,Pelizzola2005a}). 
Here the entropy approximation is obtained from a region-graph.
\index{cluster variational method}

\index{region}
A \magn{region} $\region$ is a collection of connected variables $\VV(\region)$ \marnote{region} 
and a set of factors $\FF(\region)$ such that each
participating factor depends only on the variables included in the region.
To build the CVM region-graph\footnote{Here we are making a distinction between
a general region graph and a CVM region-graph.}, one starts with predefined top\marnote{region-graph}
\index{region-graph}
(or outer) regions such that each factor is included in at least one region.
Then, we add the intersection of two regions (including variables and factors) 
recursively until no more sub(inner)-region can be added. Each region is then
connected to its immediate parent.

A region-graph, reparameterizes $\ph(.)$ in terms of its marginals over \marnote{region-based reparametrization}
the regions
\index{reparametrization}
\begin{align}\label{eq:reparam_region}
  \ph(\xs) \quad = \quad \prod_{\region} \ph(\xs_{\VV(\region)})^{\cn(\region)}
\end{align}
where $\cn(\region)$ is the \magn{counting number} for region $\region$ and 
\index{counting number}
ensures that each variable and factor is counted only once.
This number is recursively defined by M\"{o}bius formula for inner regions: \marnote{M\"{o}bius formula}
\index{Mobius formula}
\begin{align}\label{eq:mobius}
\cn(\region) \quad = \quad 1 \; - \;\sum_{\region' \supset \region} \cn(\region) 
\end{align}
where $\region'$ is an ancestors of $\region$ in the region graph.\footnote{
  More accurately $\VV(\region) \subseteq
\VV(\region')$.
}

Similar to BP, by substituting the reparametrization of \refEq{eq:reparam_region} into \marnote{Kikuchi approximation}
the variational free energy minimization of \refEq{eq:helmholtz}
\index{Kikuchi free energy}
we get 
\begin{align}
  \ph \; = &\; \arg_{\ph} \min \;\sum_{\region} \cn(\region) \left ( \sum_{\xs_{\VV(\region)}}\ph(\xs_{\VV(\region)}) \left ( \big (\sum_{\II \in \FF(\region)} \ff_{\II}(\xs_{\II}) \big ) - \ph(\xs_{\VV(\region)})\log(\ph(\xs_{\VV(\region)}))\right) \right )\notag
\\
\text{s.t.}\;&\; \sum_{ \xs_{\back \VV(\region)}} \ph(\xs_{\VV(\region')}) = \ph(\xs_{\VV(\region)}) \quad \forall \region \subset \region' \label{eq:marginalpoly_region}
\end{align}
which is known as \magn{Kikuchi approximation} to free energy~\cite{Kikuchi1951}. 
The constraints of \cref{eq:marginalpoly_region} ensure that marginals are consistent 
across overlapping regions. Solving this constraint optimization using the method of Lagrange multipliers, \marnote{generalized BP}
 yields a set of recursive equations that are known as Generalized BP equations \cite{Yedidia2001}.
Again a region-based approximation is exact only if the region-graph has no loops.
 
A region-graph without restrictions on the choice of regions generalizes
junction-graph method as well. 
The general construction of the region
 graph only requires that the counting numbers of all the regions to which 
 a variable (or a factor) belong, sum to $1$ \cite{Yedidia2005}.
 For different criteria on the choice of regions see also \cite{Pakzad2005,Welling2004,Welling2005}.

\subsection{Long loops}\label{sec:message_dependency}
In graphical models with long-range correlation
between variables, region-based methods are 
insufficient. This is because their complexity grows exponentially with the number 
of variables in each region and therefore they are necessarily inefficient account for long loops in the graph.

\index{cut-set conditioning}
A class of methods for reducing long-range correlations are methods based on \magn{cut-set conditioning} 
\cite{pearl_probabilistic_1988}, where a subset of variables are clamped, so as to 
remove the long-range correlations that are formed through the paths that include the cut-set.
For example, consider a Markov network in the form of a cycle. By fixing any single variable, the reduced factor graph
becomes a tree and therefore allows exact inference. Several works investigate more sophisticated ideas in 
performing better inference by clamping a subset and the resulting theoretical guarantees \cite{darwiche2001recursive,eaton2009choosing,weller2014clamping}.
A closely related idea is \textbf{Rao-Blackwellization} \index{Rao-Blackwellization} (\aka collapsed MCMC; see \cref{sec:mcmc}), 
a hybrid approach to inference
\cite{Gelfand1990} where particles represent a partial assignment
of variables and inference over the rest of variables is performed using a
deterministic method. The deterministic inference method such as BP is used to calculate
the value of the partition function, for each possible joint assignment of the
variables that are not collapsed. Then collapsed particles are sampled
\index{collapsed particles}
accordingly. 
This process in its general form is very expensive, but one could
reduce the cost, depending on the structure of the network
\cite{Bidyuk2007}.

\index{loop calculus}
The \magn{loop calculus} of \citet{Chertkov2006a}\cite{Chertkov2006b} expands the free energy \marnote{loop calculus}
around the Bethe approximation, 
with one term per each so-called generalized loop in the graph.
Since the number of loops grows exponentially in the number of variables, this
expansion does not provide a practical solution. Some attempts have been made to
to make this method more practical by truncating the loop series \cite{Gomez2006}.
While the original loop series was proposed for binary valued and pairwise
factors, it has been generalized to arbitrary factor-graphs
\cite{Xiao2011,Welling2012} and even region-graphs \cite{Zhou2011}.

\index{loop-corrected BP}
\index{BP!loop-correction}
Another class of approximate inference methods perform loop correction by \marnote{message dependencies}
estimating the message dependencies in a graphical model
\cite{Montanari2005,Rizzo2006,Mooij2008}.
These methods are particularly interesting as they directly compensate
for the violated assumption of BP -- \ie corresponding to independent set of incoming messages.

For the benefit of clarity, we confine the loop correction equations 
in this section and its generalization in the next section to 
\index{Markov network}
Markov networks (\ie $\vert \nb \II \vert = 2$); see \cite{ravanbakhsh_loop}
for our factor-graph versions. Although the previous works on loop corrections
have been only concerned with sum-product inference, here we present
loop corrections for a general commutative semiring $(\RR, \oplus, \otimes)$ in which the 
operation $\otimes$ has an inverse (\ie $(\RR, \otimes)$ is a group). 
In particular this means these loop corrections may be used for min-sum class of inference.

Here we rewrite BP update
 \cref{eq:mIi_semiring,eq:miI_semiring} for Markov networks\footnote{
Note that $\msg{i}{j}(\xx_{i})$ is over $\XX_i$ rather the conventional way of defining it on $\XX_j$.
This formulation is the same as original BP equation for Markov network if the
graph does not have a loop of size two.
}
\begin{align}\label{eq:bprewrite}
\msg{i}{j}(\xx_{i}) \quad& \propto \quad \bigoplus_{\back \xx_i} \bigotimes_{k\in
\mb i \back j} \ff_{\{k,i\}}(\xs_{\{k,i\}}) \otimes \msg{k}{i}(\xx_k)\\
\ph(\xx_i) \quad & \propto \quad \bigoplus_{\back \xx_i}\bigotimes_{k\in \mb i}
\ff_{\{k,i\}}(\xs_{k,i})\;
\msg{k}{i}(\xx_k)
\end{align}

\begin{figure}[htp!]
\centering
\includegraphics[width=.45\textwidth]{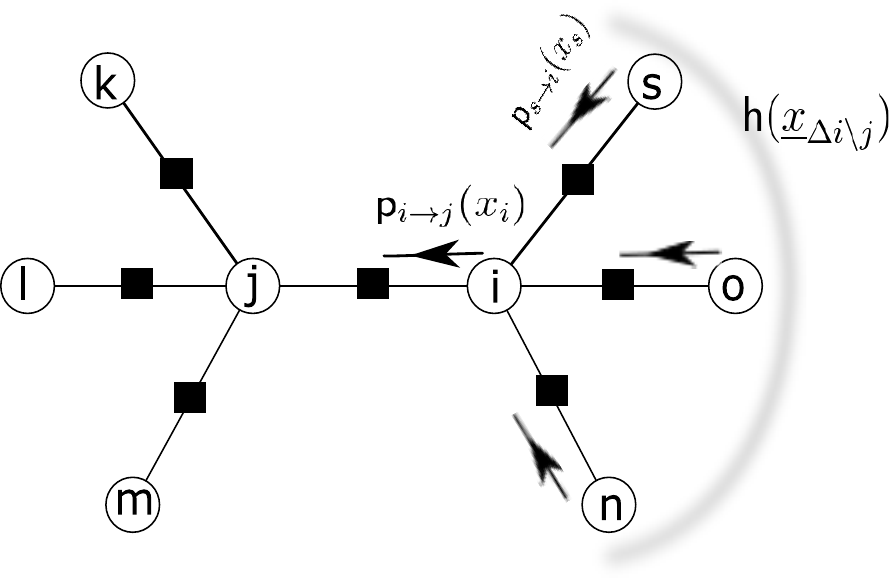}
\includegraphics[width=.45\textwidth]{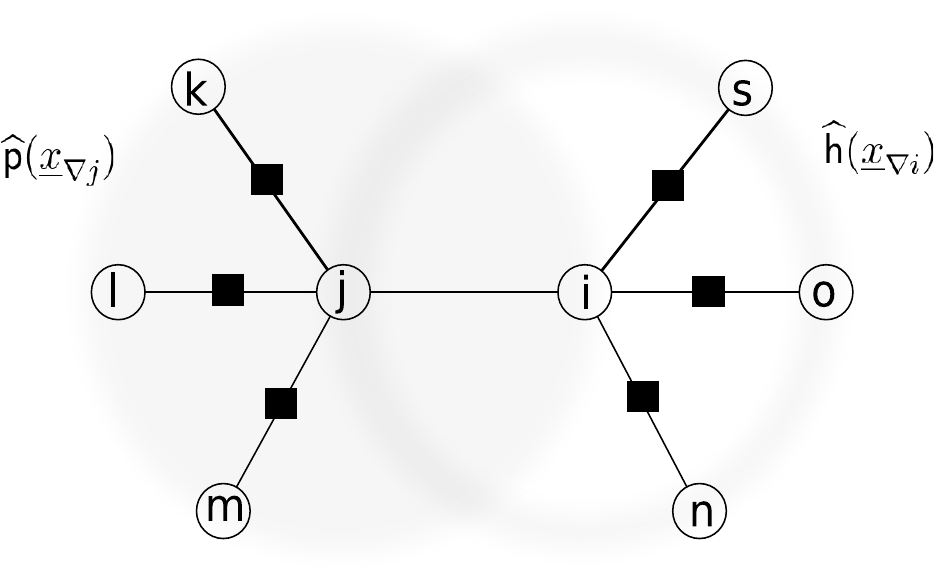}
\caption[Illustration of distributions used for loop correction.]{(left) BP messages on a Markov network and the ideal way dependencies should be taken into account.
 (right) BP marginal over extended Markov blanket $\pancake j$ for node $j$ and the message dependencies over the Markov blanket $\mb i$ for node $i$.
}\label{fig:lcbp}
\end{figure}

\index{message dependencies}
\Cref{fig:lcbp}(left) shows the BP messages on a part of Markov network. Here if the Markov network is a tree,
BP's assumption that $\msg{s}{i}$, $\msg{o}{i}$ and $\msg{n}{i}$ are independent is valid, 
because these messages summarize the effect of separate sub-trees on the node $i$. However if the graph has loops, then  
we use $\hh(\xs_{\mb i \back j})$ to denote \magn{message dependencies}.
If we had access to this function,
we could easily change the BP message update of \cref{eq:bprewrite} to
\begin{align*}
\msg{i}{j}(\xx_{i}) \quad \propto \quad \hh(\xs_{\mb i \back j}) \otimes \bigoplus_{\back \xx_i} \bigotimes_{k\in
\mb i \back j} \ff_{\{k,i\}}(\xs_{\{k,i\}}) \otimes \msg{k}{i}(\xx_k)
\end{align*}

Since it is not clear how to estimate $\mathsf{h}(\xs_{\mb i \back j})$, we follow a different path, and instead estimate
the so-called \magn{cavity distribution}, which we denote by $\hh(\xs_{\mb i})$, which is simply the joint marginal over the Markov blanket after making a cavity -- \ie removing a variable $i$ and its neighbouring factors $\ff_\II\ \forall \II \in \nb i$. 
However, since the missing information is the ``dependence'' between the messages, 
$\hh(\xs_{\mb(i)})$ has a degree of freedom w.r.t. individual marginals -- \ie it can be inaccurate by a factor of $\bigotimes_{j \in \mb(i)}\fg_j(\xx_j)$ for some $\fg_j \forall j$, without affecting the loop-corrected message passing procedure. This degree of freedom is essentially equivalent to the freedom in initialization of BP messages. 
In the following, we show the resulting loop-corrected message passing.
\index{cavity distribution}
But first  we write BP updates in a different form.

\index{extended Markov blanket}
\index{Markov blanket! extended}
Define the \magn{extended Markov blanket} \marnote{extended Markov blanket}
$\pancake i = \mb i \cup \{ i \}$ be the Markov blanket of $i$ plus
$i$ itself, see \refFigure{fig:lcbp} (right). 
We can write BP marginals over $\pancake i$ 
\begin{align}\label{eq:pancake}
\ph(\xs_{\pancake i}) \; \propto \; \bigotimes_{k \in \mb i} \ff_{\{i,k\}}(\xs_{\{i,k\}})
\otimes \msg{k}{i}(\xx_k)  
\end{align}
Using this, \cref{eq:bprewrite} simplifies to:
\begin{align}
\msg{i}{j}(\xx_{i}) \quad& \propto \quad \bigoplus_{\back \xx_i} \ph(\xs_{\pancake i}) /
\ff_{\{i,j\}}(\xs_{\{i,j\}}) \\
\ph(\xx_k) \quad & \propto \quad \bigoplus_{\back \xx_i} \ph(\xs_{\pancake i})
\end{align}

Now assume we are given the dependency between the messages $\msg{k}{i}(\xx_k) \;\forall k \in \mb i$
in the form of $\hh(\xs_{\mb i})$. This means we can re-express \cref{eq:pancake} as
\begin{align}
\ph(\xs_{\pancake i}) \quad \propto \quad \hh(\xs_{\mb i}) \bigotimes_{k \in \mb i}
\ff_{\{i,k\}}(\xs_{\{i,k\}}) \otimes \msg{k}{i}(\xx_k) 
\end{align}

By enforcing the marginal consistency of $\ph(\xs_{\pancake i})$ and $\ph(\xs_{\pancake j})$ over $\xx_i$ (and $\xx_j$)
\begin{align*}
  \bigoplus_{\xs_{\back i,j}} \ph(\xs_{\pancake i})/\ff_{\{i,j\}}(\xx_i, \xx_j) \quad = \quad \bigoplus_{\xs_{\back i,j}} \ph(\xs_{\pancake j})/\ff_{\{i,j\}}(\xx_i, \xx_j)
\end{align*}
we retrieve a message update similar to that of BP (in \cref{eq:bprewrite}) that incorporates 
the dependency between BP messages \marnote{loop-corrected BP update}
\begin{align}\label{eq:lcbp}
\msg{i}{j}\tst{t+1}(\xx_i) \propto \frac 
{\bigoplus_{\back \xx_i} \ph(\xs_{\pancake_i}) / \ff_{\{i,j\}}(\xs_{\{i,j\}}) }
{\bigoplus_{\back \xx_i} \ph(\xs_{\pancake_j}) / \ff_{\{i,j\}}(\xs_{\{i,j\}})} \otimes
\msg{i}{j}\tst{t}(\xx_i)
\end{align}

\index{message passing! loop-corrected BP}
It is easy to verify that this update reduces to BP updates (\refEq{eq:bprewrite}) when $\hh(\xs_{\partial
i})$ is uniform -- that is we do not have any dependency between messages. 
The loop-correction method of Mooij \etal\ \cite{Mooij2008} is similar, however
this interpretation does not apply to their updates for factor graphs. 
We extend the same idea to perform loop-correction for overlapping regions of connected
variables in \refSection{sec:generalizedloop} where we pass the messages from one region to the outer boundary of
another region.

The main computational cost in these loop correction methods is
estimating  $\hh(\xs_{\mb i})$, the message dependencies. 
We use clamping to perform this task. 
For
this we remove $\xx_i$ and all the immediately depending factors from the graph. 
Then we approximate the marginal $\hh(\xs_{\mb i})$ by reduction to integration; see \refSection{sec:marg2int}.
Note that the $\hh(\xs_{\mb i})$ obtained this way contains not only dependencies but also the individual marginals in the absence of node $i$ ($\hh(\xx_{j}) \; \forall j \in \mb i$).
However since the messages updates for $\msg{j}{i}(\xx_j) \; \forall j \in \mb i$, perform ``corrections''
to this joint probability, we do not need to divide $\hh(\xs_{\mb i})$ by the individual marginals.  


\subsection{Both message dependencies and short loops}\label{sec:generalizedloop}
\RefSection{sec:regions} presented loop correction methods that improve loopy BP by considering
interactions within small clusters of variables, 
thus taking small loops within these clusters into account. The previous section showed how to 
account for dependency between BP messages -- thus taking long-range correlations into account.
In this section we introduce a generalization that performs both types of loop correction.

The basic idea is to form regions, and perform exact inference over regions, to take short loops into account. However in performing message passing between these regions, we introduce a method
to perform loop correction over these messages. 

\begin{figure}[ht!]
\centering
\includegraphics[width=.5\textwidth]{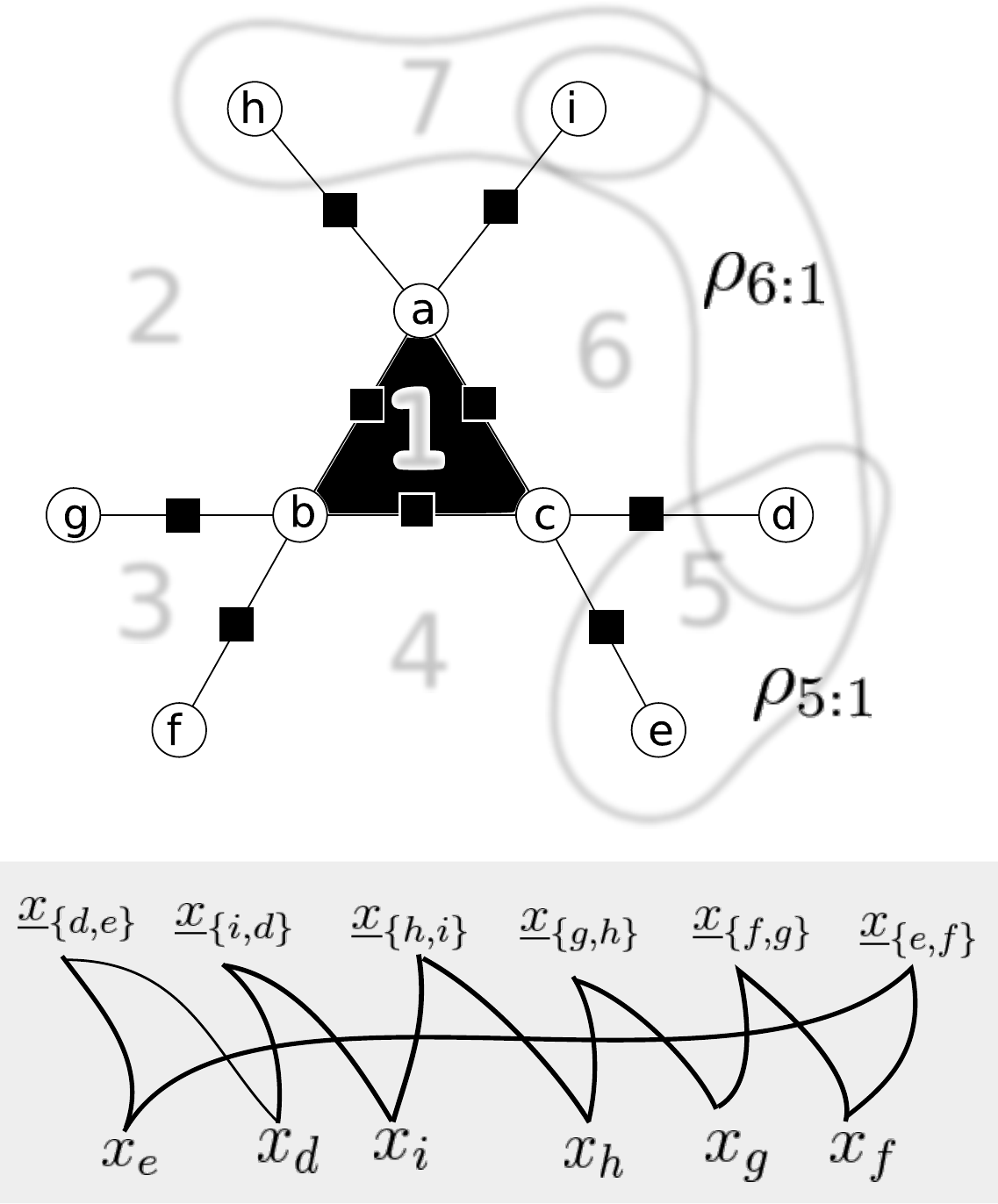}
\caption[An example for generalized loop correction method.]{(top) seven regions $\rho_1,\ldots,\rho_7$ and the domain of messages 
sent from each region to $\rho_1$. Here $\rho_{5:1}$ is the domain of message from region 5 to region 1.
(bottom) the message region-graph shows how these overlapping messages are combined to prevent double-counts.}
\label{fig:glc}
\end{figure}

We start by defining a \magn{region} $\region = \{i,\ldots,l\}$ \marnote{region}
\index{region}
as a set of connected variables. 
Note that this definition is different from definition of region for region-based methods 
as it only specifies the set of variables, and not factors.
Let $\mb \region = \{i \in \mb j, i \not \in \region \mid j \in \region \}$ be the Markov blanket of region $\region$, 
 and as before let $\pancake \region = \region \cup \mb \region$.

Region $\region_1$ is a neighbour of $\region_2$ with \magn{neighbourhood} \marnote{regional neighbourhood}
$\rg{1}{2}$ \textit{iff} 
$\rg{1}{2} \defeq (\mb \region_1) \cap \region_2 \neq \emptyset$ -- \ie the Markov blanket of $\region_1$ intersects with $\region_2$ (note that $\rg{1}{2}$ is different 
from $\rg{2}{1}$).
The messages are exchanged on these neighbourhoods and $\msg{1}{2}(\xs_{\rg{1}{2}})$ is a message from region $\region_1$ to $\region_2$.

\begin{example}
\Cref{fig:glc} shows a set of neighbouring regions (indexed by $1, 2, 3, 4, 5, 6$ and $7$). 
Here the region $\region_1$ receives ``overlapping'' messages from four other regions. 
For example the message $\msg{6}{1}(\xs_{d,i})$ overlaps with the message
$\msg{5}{1}(\xs_{d,e})$ as well as $\msg{7}{1}(\xs_{h,i})$. Therefore simply 
writing $\pp(\xs_{\pancake \region})$ in terms of the factors inside 
$\pancake \region$ and the incoming messages (as in \cref{eq:bpmarg_region}) 
will double-count some variables.
\end{example}

\subsubsection{Message region-graphs}
\index{region-graph}
Here, similar to \refSection{sec:regions}, we construct a region-graph to track the 
double-counts. However, here we have one message-region-graph per region $\region$.
The construction is similar to that of cluster variational methods; 
we start with the original message-domains (\eg $\rg{2}{1}$) and recursively 
add the intersections, until no more message-region $\msgrg$ can be added. 
Each message-region $\msgrg$ is connected to its immediate parent.
\refFigure{fig:glc} shows the two-layered message-region-graph for region $\region_1$.
Here, for discussions around a particular message-region-graph we will drop the region-index $1$.

Let $\msga(\xs_\msgrg) \defeq \msg{\pi}{1}(\xs_{\rg{\pi}{1}}) \; \forall \pi$ be the top regions in the region-graph, consisting of all the incoming messages to region of interest $\region_1$. 
The M\"{o}bius formula (\cref{eq:mobius}) gives counting number for message-region (here again the top regions' counting number is one).
A \magn{downward pass}, starting from top regions, calculates the belief $\msga(\xs_\msgrg)$ over each message-region, as the average of beliefs over its parents.

\begin{example}
In \refFigure{fig:glc} the belief $\msga(\xx_{e})$ is the average of beliefs over
$\msga(\xs_{\{d,e\}}) = \msg{5}{1}(\xs_{\rg{5}{1}})$ and $\msga(\xs_{\{e,h\}}) = \msg{4}{1}(\xs_{\rg{4}{1}})$
when marginalized over $\xx_e$.
Here the counting number of $\xx_e$ is $\cn(\{e\}) = 1 - (1 + 1) = -1$. 
\end{example}

We require the $\otimes$ operator of the semiring to have an inverse. 
Recall the power operator $\powerop$ 
$\yy \powerop k \defeq \underbrace{\yy \otimes \ldots \otimes \yy}_{\text{k times}}$. 
For sum-product and min-sum semirings, this operator corresponds to exponentiation and product respectively and it is well-defined also for rational numbers $k$. 
Now define the average as 
\begin{align}
  \average(\{\yy_1,\ldots, \yy_{k}\}) \defeq \big ( \bigotimes_{i} \yy_i \big ) \powerop {\frac{1}{k}}   
\end{align}

Using $\mathsf{Pa}(\msgrg)$ to denote the parents of region $\msgrg$, \marnote{downward pass}
in the downward pass 
 \begin{align}\label{eq:downward}
 \msga(\xs_{\msgrg}) \propto  \bigotimes_{\msgrg' \in \mathsf{Pa}(\msgrg)}  \average ( \bigoplus_{\xs_{\back \msgrg}}\msga(\msgrg'))
 \end{align}
where $\msga(\xs_{\msgrg})$ for top regions are just the incoming messages.

Let $\ff_{\AAA}(\xs_{\AAA}) = \bigotimes_{\II \subseteq \AAA} \ff_{\II}(\xs_{\II})$ be the semiring-product
of all the factors defined over a subset of $\AAA$.
For example in \refFigure{fig:glc}, $\ff_{\pancake \region_a}(\xs_{\pancake \region_a})$ is the product
of $9$ pairwise factors (\ie all the edges in the figure).

After a downward pass, the belief over $\pancake \region$ (analogous to \refEqs{eq:pancake}{eq:bpmarg_region}):
\begin{align}
  \ph(\xs_{\pancake \region}) \quad \propto \quad  \ff_{\pancake \region}(\xs_{\pancake \region}) 
\otimes 
\bigg ( \bigotimes_{\region} \msga(\xs_{\msgrg}) \powerop \cn(\msgrg)
\bigg )
\end{align}
that is $\ph(\xs_{\pancake \region})$ is the semiring product of all the factors inside this region
and all the beliefs over message-regions inside its message-region-graph, 
where double counts are taken into account.
For our example, assuming sum-product ring, 
$$\ph(\xs_{\pancake \region_a}) = \ff_{\pancake \region_a}(\xs_{\pancake \region_a})  
 \msga(\xs_{4,5})\ldots \msga(\xs_{9,4})  \bigg( \msga(\xx_{4})^{-1}\ldots\msga(\xx_9)^{-1} \bigg )$$ where
the semiring-product and inverse are product and division on real domain.

At this point we can also introduce an  
estimate for message dependencies $\hh(\xs_{\nb \region})$
into the equation above, and generalize the update of \refEq{eq:lcbp} to 
\begin{align}
\msg{b}{a}\tst{t+1}(\xs_{\rg{b}{a}}) \propto 
\frac{\bigoplus_{\back \xs_{\rg{b}{a}}} \ph(\xs_{\pancake \region_b}) / \ff_{\pancake \rho_a \cap \pancake \rho_b}(\xs_{\pancake \rho_a \cap \pancake \rho_b}) }
{\bigoplus_{\back xs_{\rg{b}{a}}} \ph(\xs_{\pancake \region_a}) / \ff_{\pancake \rho_a \cap \pancake \rho_b}(\xs_{\pancake \rho_a \cap \pancake \rho_b}) } \otimes
\msgover{b}{a}\tst{t}(\xs_{\rg{b}{a}})
\end{align}

One last issue to resolve is to define the ``effective'' message $\msgover{b}{a}\tst{t}(\xs_{\rg{b}{a}})$, which is different from 
$\msg{b}{a}\tst{t}(\xs_{\rg{b}{a}})$.
Since we did not directly use $\msg{b}{a}\tst{t}(\xs_{\rg{b}{a}})$ in the previous iteration, we should not include it directly in this update.
Instead we use the message region-graph for region $a$ to calculate the effective message:
\begin{align}\label{eq:effective_msg}
  \msgover{b}{a}(\xs_{\rg{b}{a}}) = \bigotimes_{\msgrg \subseteq \rg{b}{a}} \big ( \msga(\xs_{\msgrg}) \powerop \cn(\msgrg) \big )
\end{align}

\index{upward pass}
The effective message, as defined above (\cref{eq:effective_msg}), can be efficiently calculated in \marnote{upward pass} an \textbf{upward pass} in the message region-graph. Starting from the parents of the lowest regions,
update the belief $\msga(\xs_{\msgrg})$ obtained in downward pass \cref{eq:downward}
using the new beliefs over its children:
\begin{align}\label{eq:upward}
  \msgaover(\xs_{\msgrg}) \quad = \quad \msga(\xs_{\msgrg}) \bigotimes_{\msgrg' \in \mathsf{Ch}(\msgrg)}\frac{\msgaover(\xs_{\msgrg'})}{\bigoplus_{\xs_{\back \msgrg'}} \msga(\xs_{\msgrg})}
\end{align}
where $\mathsf{Ch}(\msgrg)$ is the set of children of $\msgrg$ in the message region-graph.
After the upward pass, the new beliefs over top regions gives us the effective messages
\begin{align}
\msgover{a}{b}(\xs_{\rg{a}{b}}) = \msgaover(\xs_{\rg{a}{b}})
\end{align}

\begin{example}
In our example of \cref{fig:glc}(bottom), assuming a sum-product ring, since the message-region-graph only has two layers,
 we can write the effective message 
$\msgover{h}{a}(\xs_{\{4,9\}})$ as
\begin{align}
 \msgover{h}{a}(\xs_{\{4,9\}}) = \msg{h}{a}(\xs_{\{4,9\}})\frac{\msga(\xx_{4}) \msga(\xx_{9})}{(\sum_{\xx_4} \msg{h}{a}(\xs_{\{4,9\}}))(\sum_{\xx_9} \msg{h}{a}(\xs_{\{4,9\}}))}
\end{align}
\end{example}

\index{generalized loop correction}
\index{message passing! GLC}
This form of loop correction, which we call Generalized Loop Correction (GLC), generalizes both correction schemes of \cref{sec:regions,sec:message_dependency}.
The following theorem makes this relation to generalized BP more explicit.
\begin{theorem}\footnote{See \cite{ravanbakhsh_loop} for the proof.}
For Markov networks, if the regions $\{\region\}$ partition the variables,
then any Generalized BP fixed
point of a particular region-graph construction is also a
fixed point for GLC, when using uniform message dependencies (\ie $\hh(\xs_{\mb i}) \propto 1\; \forall i,\xs_{\mb i}$).
\end{theorem}

\subsection{Experiments} 
\label{sec:experiments_lcbp}
This section compares different variations of our generalized loop correction method (GLC)
for sum-product ring against
BP as well as Cluster Variational Method (CVM; \refSection{sec:regions}), loop correction of
\cite{Mooij2008} (LCBP), which does not exactly account for short loops and the Tree Expectation Propagation (TreeEP)~\cite{Minka03} method,
which also performs some kind of loop correction.
For CVM, we use the double-loop algorithm \cite{Heskes06},
which is slower than Generalized BP but has better convergence 
properties.\footnote{All methods are applied without any damping.
We stop each method
after a maximum of 1E4 iterations or 
if the change in the probability distribution (or messages) is less than 1E-9.
}
We report the time in seconds and the error for each method
as the average of absolute error in single variable marginals -- 
\ie $\frac{1}{N} \sum_{i} \sum_{\back \xx_i} \mid  \ph(\xx_i) - \pp(\xx_i) \mid$.
For each setting, we report the average results over $10$ random instances of
the problem.
We experimented with grids and 3-regular random graphs.\footnote{The evaluations are based on
  implementation in \emph{libdai} inference toolbox~\cite{Mooij10}.}

\index{Ising}
\begin{figure} 
\begin{subfigure}[t]{.48\columnwidth}
\includegraphics[width=.98\textwidth]{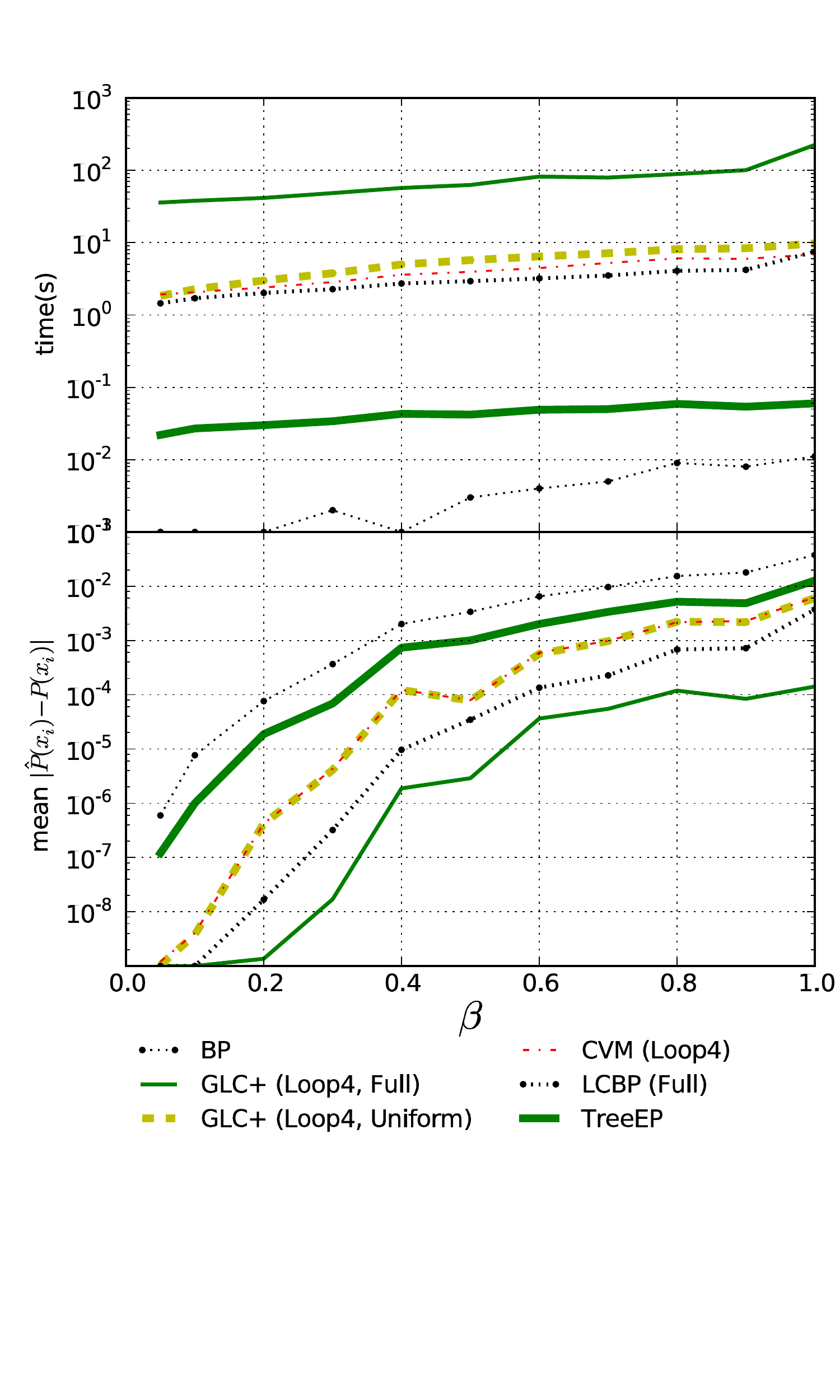}
\caption{spin-glass Ising grid}
\label{fig:gridTempTime}
\end{subfigure}
~
\begin{subfigure}[t]{.48\columnwidth}
\includegraphics[width=.98\textwidth]{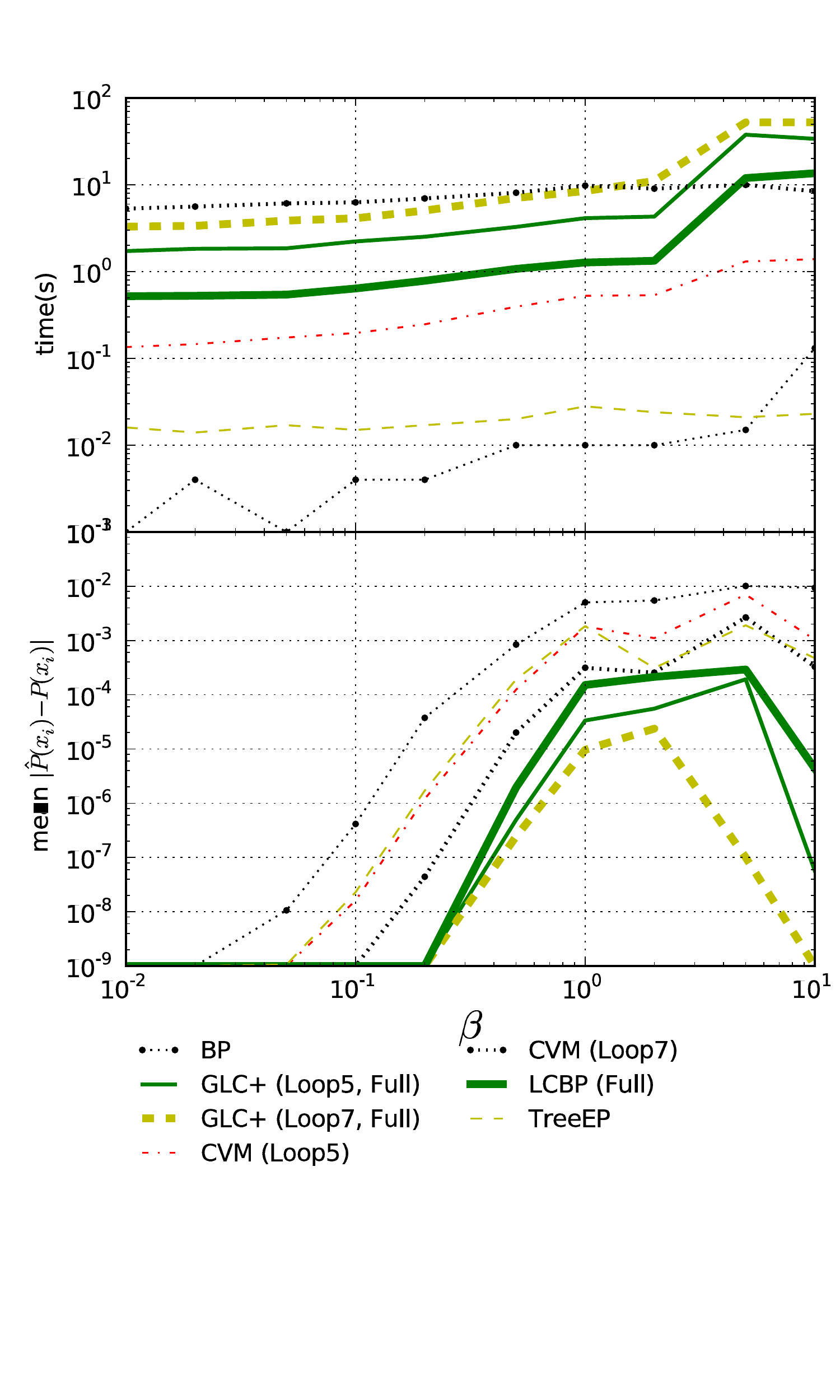}
\caption{spin-glass Ising model on a 3-regular graph}
\label{fig:3regIsingTemp}
\end{subfigure}
\caption[Run-time and accuracy at different levels of difficult for various loop correction methods.]{
Average Run-time and accuracy for 
6x6 spinglass Ising grids and 3-regular Ising model for different values of 
$\beta$. Variable interactions
are sampled from $\mathcal{N}(0, \beta^2)$ and local fields are sampled
from $\mathcal{N}(0,1)$.}
\end{figure}

Both LCBP and GLC can be used without any information on message dependencies.
with an initial cavity distribution estimated via clamping cavity variables. 
In the experiments, {\em full} means message dependencies $\hh$ was estimated 
while {\em uniform} means $\hh$ was set to uniform distribution 
(\ie loop-correction was over the regions only).
We use GLC to denote the case where the regions were selected such that they have 
no overlap (\ie $\region_a \cap \region_b = \emptyset \; \forall a, b$)
and GLC+ when overlapping clusters of some form
are used.
For example, {\em GLC+(Loop4, full)} refers to a setting
with message dependencies that contains all overlapping loop clusters of length
up to 4.
If a factor does not appear in any loops, it forms its own cluster. 
The same form of clusters are used for CVM.   

\begin{figure}
\centering 
\begin{subfigure}[t]{.48\columnwidth}
\includegraphics[width=.95\textwidth]{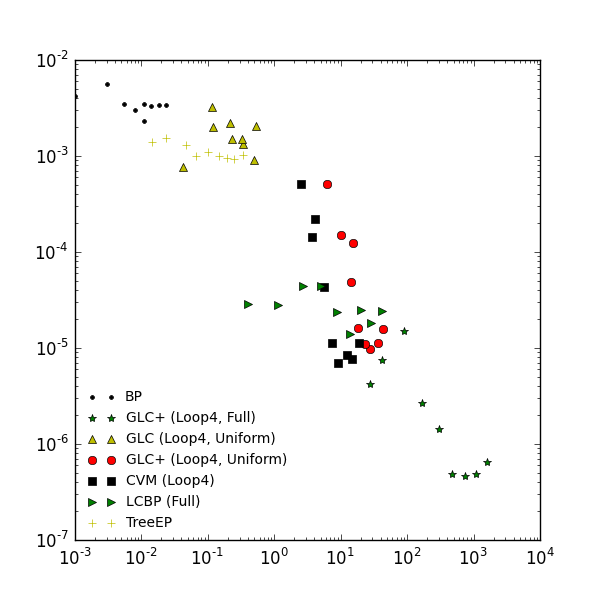}
\caption{spin-glass Ising grid}
\label{fig:gridSize}
\end{subfigure}
\begin{subfigure}[t]{.48\columnwidth}
\includegraphics[width=.95\textwidth]{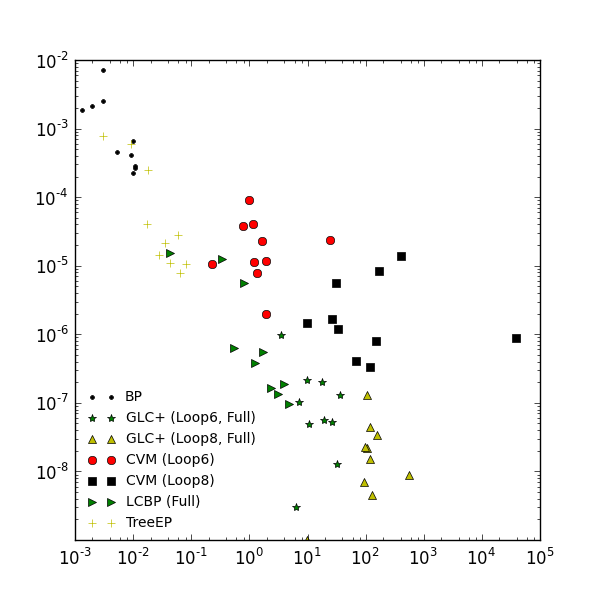}
\caption{spin-glass Ising model on 3-regular graph}
\label{fig:3regIsingSize}
\end{subfigure}
\caption[Run-time versus accuracy over different problem sizes for various loop correction methods.]{Time vs error for Ising grids  and 3-regular Ising models  with local field and interactions sampled from a standard normal. 
Each method in the graph has 10 points, each representing an Ising model of different size ($10$ to $100$ variables). }
\label{fig:IsingSize}
\end{figure}

\subsubsection{Grids} 
We experimented with periodic spin-glass Ising grids
of \pcref{example:ising}. 
In general,
smaller local fields and larger variable interactions result
in more difficult problems. 
We sampled local fields independently from $\mathcal{N}(0, 1)$ and
interactions from $\mathcal{N}(0, \beta^2)$. 
\RefFigure{fig:gridTempTime}
summarizes the results for 6x6 grids for different values of $\beta$.

We also experimented with periodic grids of different sizes,
generated by sampling all factor entries independently from $\mathcal{N}(0,1)$. 
\RefFigure{fig:gridSize} compares the
computation time and error of different methods for grids of sizes
that range from 4x4 to 10x10.

\subsubsection{Regular Graphs} 
We generated two sets of experiments with random 3-regular graphs
 (all nodes have degree 3) over 40 variables. 
Here we used Ising model when both
 local fields and couplings are independently sampled from
$\mathcal{N}(0, \beta^2)$.
\RefFigure{fig:3regIsingTemp}
shows the time and error for different values of $\beta$. 
\RefFigure{fig:3regIsingSize} shows time versus error for graph
size between $10$ to $100$ nodes for $\beta = 1$.

Our results suggest that by taking both long and short loops into account we can significantly improve the accuracy of 
inference at the cost of more computation time. In fact both plots in \refFigure{fig:IsingSize} show a log-log trend
in time versus accuracy which suggests that taking short and long loops into account has almost independently improved the
quality of inference.

%% file: meta_constructions.tex
\index{message passing!survey propagation| see {SP}}
\index{SP}
\index{survey propagation|see {SP}}
Survey propagation (SP) was first introduced as a message passing solution to satisfiability~\cite{braunstein_survey_2002} and was later generalized to general CSP
\cite{braunstein_constraint_2002} and arbitrary inference problems over factor-graphs~\cite{Mezard09}.
Several works offer different interpretations and generalizations of survey propagation \cite{Kroc2002,braunstein_survey_2003,maneva_new_2004}.
Here, we propose a generalization based the same notions that extends the application of BP to arbitrary commutative semirings.
\index{commutative semiring}
Our derivation closely follows and generalizes the variational approach of  
\citet{Mezard09}, in the same way that the algebraic approach to BP (using commutative semirings) generalizes the variational derivation of sum-product and min-sum BP.

\index{fixed point}
As a fixed point iteration procedure,
if BP has more than one fixed points,
it may not converge at all. Alternatively, if the messages are initialized properly BP may converge to one of its fixed points.
SP equations, take ``all'' BP fixed points into account.
In our algebraic perspective, this accounting of all fixed points is using a third operation
$\spplus$. In particular, we require that $\otimes$ also distribute over $\spplus$, forming 
a second commutative semiring. We refer to the this new semiring as \magn{SP semiring}.
\index{commutative semiring! SP}

Let $\msgss{\cdot}{\cdot}$ be a BP fixed point -- that is
$$
\msgss{\cdot}{\cdot} = \{ \forall i, \II \in \nb i \quad \msg{i}{\II} = \MSG{i}{\II}(\msgss{\nb i\back \II}{i}),
 \msg{\II}{i} = \MSG{\II}{i}(\msgss{\nb \II \back i}{\II}) \}
$$
and denote the set of all such fixed points by $\WW$.
Each BP fixed point corresponds to an approximation to the $\qq(\emptyset)$,
which we denote by $\QQ(\msgss{\cdot}{\cdot})(\emptyset)$ -- using this functional form
is to emphasize the dependence of this approximation on BP messages.
Recall that in the original problem, $\XX$ is the domain of assignments, 
$\qq(\xs)$ is the  expanded form and $\bpplus$-marginalization is (approximately) performed by BP.
In the case of survey propagation, $\WW$ is domain of assignments and the integral $\QQ(\msgss{\cdot}{\cdot})(\emptyset)$ 
evaluates a particular assignment $\msgss{\cdot}{\cdot}$ to all the messages -- \ie 
$\QQ(\msgss{\cdot}{\cdot})(\emptyset)$  is the new expanded form.

In this algebraic perspective, SP efficiently performs a second integral  using $\spplus$ over all fixed points:
\begin{align}\label{eq:sp_q_intmarg}
\QQ(\emptyset)(\emptyset) = \bigspplus_{\msgss{\cdot}{\cdot} \in \WW} \QQ(\msgssq{\cdot}{\cdot})(\emptyset)
\end{align}
\RefTable{table:sp} summarizes this correspondence.

\begin{table}
\caption{The correspondence between BP and SP}
\label{table:sp}
\begin{tabu}{ r|[2pt]l}
\textbf{Belief Propagation} & \textbf{Survey Propagation} \\
\multicolumn{2}{l}{domain:}\\
$\xs$ & $\msgss{\cdot}{\cdot}$ \\ 
$\forall i \quad \xx_i$ & $\msg{i}{\II}\;,\; \msg{\II}{i} \quad \forall i, \II \in \nb i$  \\ 
$\XX$ & $\WW$ \\
\multicolumn{2}{l}{expanded form:}\\
$\qq(\xs)$ & $\QQ(\msgss{\cdot}{\cdot})(\emptyset)$\\
\multicolumn{2}{l}{integration:}\\
 $\qq(\emptyset) = \bigbpplus_{\xs} \qq(\xs)$ & $\QQ(\emptyset)(\emptyset) = \bigspplus_{\msgss{\cdot}{\cdot}} \QQ(\msgss{\cdot}{\cdot})(\emptyset)$\\
\multicolumn{2}{l}{marginalization:}\\
 $\pp(\xx_i) \propto \bigbpplus_{\xs \back i} \pp(\xs)$ & $\PSP(\msg{\II}{i}) \propto \bigspplus_{\back \msgss{\II}{i}} \PP(\msgss{\cdot}{\cdot})$\\
\multicolumn{2}{l}{factors:}\\
$ \forall \II \quad \ff_\II(\xs_\II) $ & $\PHT_\II(\msgss{\nb \II}{\II})(\emptyset)$, $\PHT_i(\msgss{\nb i \back \II}{i})(\emptyset)$ and $\MSGBIT{i}{\II}(\msg{i}{\II},\msg{\II}{i})(\emptyset)^{-1} \quad \forall i, \II \in \nb i$
\end{tabu}
\end{table}


Our derivation requires $(\RR, \bptimes)$ to be an Abelian group (\ie every element of $\RR$ has an inverse w.r.t. $\bptimes$).\index{Abelian group}
The requirement for invertablity of $\bptimes$ is because we need to work with normalized BP
and SP messages.
In \refSection{sec:uniformsp} we introduce another variation of SP that simply counts the 
BP fixed points and relaxes this requirement.


\subsection{Decomposition of the integral}
\index{BP!integral}
\label{sec:decompose_integral}
In writing the normalized BP equations in \refSection{sec:bp}, we hid the normalization constant using $\propto$ sign.
Here we explicitly define the normalization constants or \magn{local integrals} by defining unnormalized messages, based on their normalized version
\begin{align}
\msgt{\II}{i}(\xx_i) \quad &\defeq \quad   \bigoplus_{ \xs_{\back i}} \ff_{{\II}}(\xs_{\II}) 
\bigotimes_{j \in \nb \II \back i} \msg{j}{{\II}}(\xx_{j}) \quad &\defeq \quad \MSGT{\II}{i}(\msgss{\nb \II \back i}{\II})(\xx_i) \label{eq:mIi_partition}\\
\msgt{i}{\II}(\xx_i)  \quad &\defeq \quad   \bigotimes_{\JJ \in \nb{i} \back \II} \msg{{\JJ}}{i}(\xx_i)  \quad &\defeq \quad  \MSGT{i}{\II}(\msgss{\nb i \back \II}{i})(\xx_i)\label{eq:miI_partition}\\
\pht_\II(\xs_\II)  \quad &\defeq \quad   \ff_{\II}(\xs_{\II}) \bigotimes_{i \in \nb \II} \msgss{i}{\II}(\xx_i)   \quad &\defeq \quad \PHT_\II(\msgss{\nb \II}{\II})(\xs_\II) \label{eq:marg_factor_partition}\\
\pht_i(\xx_i) \quad  &\defeq \quad  \bigotimes_{\II \in \nb i} \msg{\II}{i}(\xx_i)  \quad &\defeq \quad  \PHT_i(\msgss{\nb i}{i})(\xx_i)  \label{eq:marg_partition}
\end{align}
where each update also has a functional form on the r.h.s.
In each case, the local integrals are simply the integral of unnormalized messages or marginals -- \eg $\msgt{\II}{i}(\emptyset) = \bigoplus_{\xx_i} \msgt{\II}{i}(\xx_i)$.

Define the functional $\MSGBIT{i}{\II}(\msg{i}{\II},\msg{\II}{i})$ as the product of messages from $i$ to $\II$ and vice versa
\begin{align} \label{eq:bi_direction}
  \msgbit{i}{\II}(\xx_i) \quad \defeq \quad  \msg{i}{\II}(\xx_i) \otimes \msg{\II}{i}(\xx_i) \quad \defeq \quad \MSGBIT{i}{\II}(\msg{i}{\II},\msg{\II}{i})(\xx_i)
\end{align}


\begin{theorem}\label{th:integral_decompose} 
If the factor-graph has no loops and $(\RR, \otimes)$ is an Abelian group, the global integral decomposes to local BP integrals as
  \begin{align}\label{eq:partition_decomp_nofunc}
    \qq(\emptyset) \quad = \quad \bigotimes_{\II} \pht_\II(\emptyset) \;\bigotimes_i\pht_i(\emptyset)
\left (\bigotimes_{i, \II \in \nb i} \msgbit{i}{\II}(\emptyset) \right )^{-1}  
  \end{align}
or in other words $\qq(\emptyset)  =  \QQ(\msgss{\cdot}{\cdot})(\emptyset)$ where
  \begin{align}\label{eq:partition_decomp}
    \QQ(\msgss{\cdot}{\cdot})(\emptyset) = \bigotimes_{\II} \PHT_\II(\msgss{\nb \II}{\II})(\emptyset) \;\bigotimes_i\PHT_i(\msgss{\nb i}{i})(\emptyset)
 \left ( \bigotimes_{i, \II \in \nb i} \MSGBIT{i}{\II}(\msg{i}{\II},\msg{\II}{i})(\emptyset) \right)^{-1} 
  \end{align}
\end{theorem}
\begin{proof} 
For this proof we build a tree around an  root node $r$
that is connected to one factor. (Since the factor-graph is a tree such a node always exists.) 
Send BP messages from the leaves, up towards the root $r$ and back to the leaves. 
Here, any message $\msgq{i}{\II}(\xx_i)$,
can give us the integral for the sub-tree that contains all the nodes and factors up to node $i$ using $\msgq{i}{\II}(\emptyset) = \bigbpplus_{\xx_i}\msgq{i}{\II}(\xx_i)$.
Noting that the root is connected to exactly one factor, the global integral is 
\begin{align}\label{eq:genesis}
\bigbpplus_{\xx_r} \qq(\xx_r) = \bigbpplus_{\xx_r} \bigbptimes_{\II \in \nb r} \msgq{\II}{r}(\xx_r) = 
\msgq{\II}{r}(\emptyset)
\end{align}

On the other hand, We have the following relation between $\msgq{i}{\II}$ and $\msg{i}{\II}$ (also corresponding factor-to-variable message)
\begin{align}
  \msgq{i}{\II}(\xx_i) &= \msg{i}{\II}(\xx_i) \bptimes \msgq{i}{\II}(\emptyset) \quad \forall i, \II \in \nb i \label{eq:integral_subs1}\\
  \msgq{\II}{i}(\xx_i) &= \msg{\II}{i}(\xx_i) \bptimes \msgq{\II}{i}(\emptyset) \quad \forall i, \II \in \nb i \label{eq:integral_subs2}
\end{align}

Substituting this into BP \refEqs{eq:miI_semiring}{eq:mIi_semiring} we get
\begin{align}
\msgq{i}{{{\II}}}(\xx_i) \quad & = \quad  \bigotimes_{\JJ \in \nb{i} \back \II} \msgq{\JJ}{i}(\emptyset) \msg{{\JJ}}{i}(\xx_i) \label{eq:step1_1}\\
\msgq{{{\II}}}{i}(\xx_i) \quad & = \quad  \bigoplus_{ \xs_{\back i}} \ff_{{\II}}(\xs_{\II}) 
\bigotimes_{j \in \nb \II \back i} \msgq{j}{{\II}}(\emptyset) \msg{j}{{\II}}(\xx_{j})  
\label{eq:step1_2}
\end{align}
By summing over both l.h.s and r.h.s in equations above and substituting from \refEq{eq:miI_partition} we get
\begin{align}
\bigbpplus_{\xx_i} \msgq{i}{{{\II}}}(\xx_i) \quad & = \quad  \left ( \bigotimes_{\JJ \in \nb{i} \back \II} \msgq{\JJ}{i}(\emptyset) \right ) \otimes \left (\bigbpplus_{\xx_i} \bigotimes_{\JJ \in \nb{i} \back \II}  \msg{{\JJ}}{i}(\xx_i) \right ) \rightarrow \notag \\
\msgq{i}{{{\II}}}(\emptyset) \quad & = \quad  \msgt{i}{\II}(\emptyset)\; \bigotimes_{\JJ \in \nb{i} \back \II} \msgq{\JJ}{i}(\emptyset) \label{eq:recurse_1}
\end{align}
and similarly for \refEq{eq:step1_2} using integration and substitution from \refEq{eq:mIi_partition} we have
\begin{align}
\bigbpplus_{\xx_i} \msgq{{{\II}}}{i}(\xx_i) \quad & = \quad  \left ( \bigotimes_{j \in \nb \II \back i} \msgq{j}{{\II}}(\emptyset) \right) \otimes \left( \bigoplus_{ \xs_{\II}} \ff_{{\II}}(\xs_{\II}) 
\bigotimes_{j \in \nb \II \back i} \msg{j}{{\II}}(\xx_{j})  \right ) \rightarrow \notag \\
\msgq{{{\II}}}{i}(\emptyset) \quad & = \quad   \msgt{\II}{i}(\emptyset)
\bigotimes_{j \in \nb \II \back i} \msgq{j}{{\II}}(\emptyset)   \label{eq:recurse_2}
\end{align}

\RefEqs{eq:recurse_1}{eq:recurse_2} are simply recursive integration on a tree, where the integral
up to node $i$ (\ie $\msgq{i}{{{\II}}}(\emptyset)$ in \refEq{eq:recurse_1}) is reduced to integral in its sub-trees.
By unrolling this recursion we see that $\msgq{i}{{{\II}}}(\emptyset)$ is simply the product of all $\msgt{\II}{i}(\emptyset)$ and $\msgt{\II}{i}(\emptyset)$
in its sub-tree, where the messages are towards the root. \RefEq{eq:genesis}
tells us that the global integral is not different.
Therefore, \refEqs{eq:recurse_1}{eq:recurse_2} we can completely expand the recursion
for the global integral.
For this, let $\uparrow i$
restrict the $\nb i$ to the factor that is higher than variable $i$ in the tree (\ie closer to the root $r$). Similarly let $\uparrow \II$ be the variable that is closer to the root than $\II$.
We can write the global integral as
 \begin{align}\label{eq:partition_by_message_partition}
   \qq(\emptyset) = \bigbptimes_{i, \II = \uparrow i} \msgt{i}{\II}(\emptyset) \bigbptimes_{\II, i = \uparrow \II} \msgt{\II}{i}(\emptyset)
 \end{align}

\RefProposition{th:simplify_sp} shows that these local integrals can be written in terms of local integrals of interest -- \ie
  \begin{align*}
    \msgt{\II}{i}(\emptyset) \quad = \quad \frac{\pht_\II(\emptyset)}{\msgbit{i}{\II}(\emptyset)} \quad \text{and} \quad
    \msgt{i}{\II}(\emptyset) \quad = \quad \frac{\pht_i(\emptyset)}{\msgbit{i}{\II}(\emptyset)}
  \end{align*}
Substituting from the equations above into \refEq{eq:partition_by_message_partition} we get
the equations of \refTheorem{th:integral_decompose}. 
\end{proof}

\begin{proof}(\pcref{th:simplify_sp})
 By definition of $\pht_\II(\xs_\II)$ and $\msg{i}{\II}(\xx_i)$ in \refEqs{eq:mIi_partition}{eq:marg_factor_partition}
\begin{align*}
  \pht_\II(\xx_i) =  \msgt{\II}{i}(\xx_i) \bptimes \msg{i}{\II}(\xx_i) \quad &\rightarrow \quad \bigbpplus_{\xx_i}\pht_\II(\xx_i) = \bigbpplus_{\xx_i} \msgt{\II}{i}(\xx_i) \bptimes \msg{i}{\II}(\xx_i) 
&\rightarrow\\
\pht_{\II}(\emptyset) = \msgt{\II}{i}(\emptyset) \bptimes \left (\bigbpplus_{\xx_i} \msg{\II}{i}(\xx_i) \bptimes \msg{i}{\II}(\xx_i) \right ) \quad &\rightarrow \quad \pht_{\II}(\emptyset) = \msgt{\II}{i}(\emptyset) \bptimes \msgbit{i}{\II}(\emptyset)
\end{align*}
where in the last step we used \refEq{eq:bi_direction}.

Similarly for the second statement of the proposition we have
\begin{align*}
  \pht_i(\xx_i) =  \msgt{i}{\II}(\xx_i) \bptimes \msg{\II}{i}(\xx_i) \quad &\rightarrow \quad \bigbpplus_{\xx_i}\pht_i(\xx_i) = \bigbpplus_{\xx_i} \msgt{i}{\II}(\xx_i) \bptimes \msg{\II}{i}(\xx_i) 
&\rightarrow\\
\pht_{i}(\emptyset) = \msgt{i}{\II}(\emptyset) \bptimes \big (\bigbpplus_{\xx_i} \msg{\II}{i}(\xx_i) \bptimes \msg{i}{\II}(\xx_i) \big ) \quad &\rightarrow \quad \pht_{i}(\emptyset) = \msgt{i}{\II}(\emptyset) \bptimes \msgbit{i}{\II}(\emptyset)
\end{align*}
\end{proof}


\subsection{The new factor-graph and semiring}\label{sec:newsemiring}


\index{SP!factor graph}
The decomposition of integral in \refTheorem{th:integral_decompose} means $\QQ(\msgss{\cdot}{\cdot})(\emptyset)$
has a factored form. Therefore, a factor-graph with 
$\msgss{\cdot}{\cdot}$ as the set of variables and 
\index{variable!SP}
three different types of factors corresponding to different
terms in the decomposition -- \ie $\PHT_\II(\msgss{\nb \II}{\II})(\emptyset)$, $\PHT_i(\msgss{\nb i \back \II}{i})(\emptyset)$ and $\MSGBIT{i}{\II}(\msg{i}{\II},\msg{\II}{i})(\emptyset)^{-1}$ can represent $\QQ(\msgss{\cdot}{\cdot})(\emptyset)$. 

\begin{figure}
\centering
\includegraphics[width=1\textwidth]{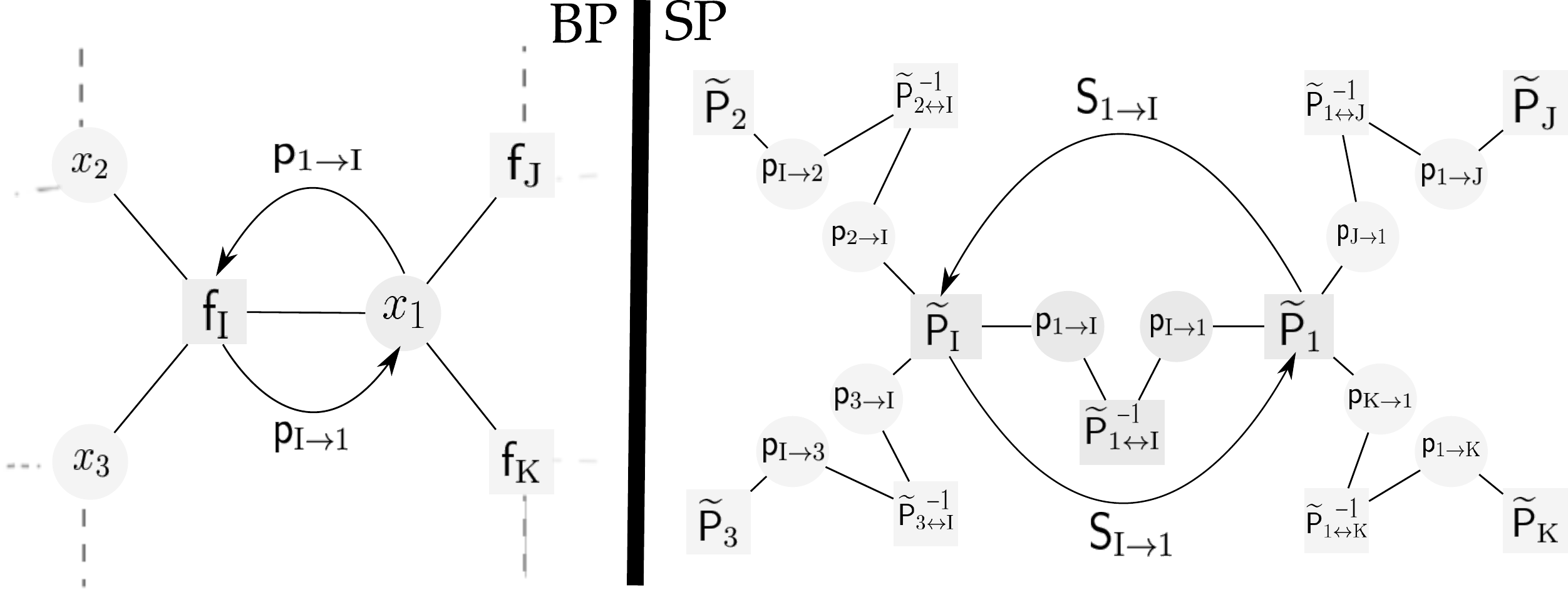}
\caption[Survey propagation factor-graph.]{Part of a factor-graph (left) and the corresponding SP factor-graph on the right.
The variables in SP factor-graph are the messages in the original graph. The SP factor-graph
has three type of factors: (I)$\PHT_\II(.)(\emptyset)$, (II) $\PHT_i(.)(\emptyset)$ and (III)${\MSGBIT{i}{\II}(.)(\emptyset)}^{-1}$. 
As the arrows suggest, SP message updates
are simplified so that only two type of messages are exchanged: $\msp{i}{\II}$ and $\msp{\II}{i}$ between factors of type (I) and (II).
}
\label{fig:spfg}
\end{figure}

\RefFigure{fig:spfg} shows a simple factor-graph 
and the corresponding SP factor-graph. The new factor-graph has one variable per each message in the
original factor-graph and three types of factors as discussed above.
Survey propagation is simply belief propagation applied to the  this new factor-graph using
the new semiring.
As before BP messages are exchanged between variables and factors.
But here, we can simplify BP messages by substitution and only keep two types of factor-to-factor messages. 
We use $\msp{i}{\II}$ and $\msp{\II}{i}$ to denote these two types of SP messages.
These messages are exchanged between two types of factors, namely $\PHT_\II(\msgss{\nb \II}{\II})(\emptyset)$ and $\PHT_i(\msgss{\nb i \back \II}{i})(\emptyset)$.
Since the third type of factors $\MSGBIT{i}{\II}(\msg{i}{\II},\msg{\II}{i})(\emptyset)^{-1}$ 
are always connected to only two variables, $\msg{i}{\II}$ and $\msg{\II}{i}$, 
we can simplify their role in the SP message update to get 
\begin{align}
  \msp{i}{\II}(\msg{i}{\II}, \msg{\II}{i}) \; &\propto \; \bigspplus_{\back \msg{i}{\II},\msg{\II}{i}} \left ( \frac{\PHT_i(\msgss{\nb i}{i})(\emptyset)}{\MSGBIT{i}{\II}(\msg{i}{\II},\msg{\II}{i})(\emptyset)} \;
 \bigbptimes_{\JJ \in \nb i \back \II} \msp{\JJ}{i}(\msg{i}{\JJ},\msg{\JJ}{i}) \right ) \label{eq:spiI_first}\\
  \msp{\II}{i}(\msg{i}{\II}, \msg{\II}{i}) \; &\propto \; \bigspplus_{\back \msg{i}{\II},\msg{\II}{i} } \left (\frac{\PHT_\II(\msgss{\nb \II}{\II})(\emptyset)}{\MSGBIT{i}{\II}(\msg{i}{\II},\msg{\II}{i})(\emptyset)} \;
 \bigbptimes_{j \in \nb \II \back i} \msp{j}{\II}(\msg{j}{\II},\msg{\II}{j}) \right ) \label{eq:spIi_first}
\end{align}
where in all cases we are assuming the messages $\msgss{\cdot}{\cdot} \in \WW$ are consistent with each other -- \ie satisfy BP equations on the original factor-graph.
Note that, here again we are using the normalized BP message update and the normalization
factor is hidden using $\propto$ sign. This is possible because we assumed $\bptimes$ has an inverse.  
We can further simplify this update using the following proposition.

\begin{proposition} \label{th:simplify_sp} for $\msgss{\cdot}{\cdot} \in \WW$
  \begin{align} 
    \frac{\PHT_i(\msgss{\nb i}{i})(\emptyset)}{\MSGBIT{i}{\II}(\msg{\II}{i}, \msg{i}{\II})(\emptyset)} \; &= \; \MSGT{i}{\II}(\msgss{\nb i \back \II}{i})(\emptyset) \\
\quad &\text{and} \quad \notag \\
    \frac{\PHT_\II(\msgss{\nb \II}{\II})(\emptyset)}{\MSGBIT{i}{\II}(\msg{\II}{i}, \msg{i}{\II})(\emptyset)} \; &= \;  \MSGT{\II}{i}(\msgss{\nb \II \back i}{\II})(\emptyset)
  \end{align}
\end{proposition}
\begin{proof}
 By definition of $\pht_\II(\xs_\II)$ and $\msg{i}{\II}(\xx_i)$ in \refEqs{eq:mIi_partition}{eq:marg_factor_partition}
\begin{align*}
  \pht_\II(\xx_i) =  \msgt{\II}{i}(\xx_i) \bptimes \msg{i}{\II}(\xx_i) \quad &\rightarrow \quad \bigbpplus_{\xx_i}\pht_\II(\xx_i) = \bigbpplus_{\xx_i} \msgt{\II}{i}(\xx_i) \bptimes \msg{i}{\II}(\xx_i) 
&\rightarrow\\
\pht_{\II}(\emptyset) = \msgt{\II}{i}(\emptyset) \bptimes \left (\bigbpplus_{\xx_i} \msg{\II}{i}(\xx_i) \bptimes \msg{i}{\II}(\xx_i) \right ) \quad &\rightarrow \quad \pht_{\II}(\emptyset) = \msgt{\II}{i}(\emptyset) \bptimes \msgbit{i}{\II}(\emptyset)
\end{align*}
where in the last step we used \refEq{eq:bi_direction}.

Similarly for the second statement of the proposition we have
\begin{align*}
  \pht_i(\xx_i) =  \msgt{i}{\II}(\xx_i) \bptimes \msg{\II}{i}(\xx_i) \quad &\rightarrow \quad \bigbpplus_{\xx_i}\pht_i(\xx_i) = \bigbpplus_{\xx_i} \msgt{i}{\II}(\xx_i) \bptimes \msg{\II}{i}(\xx_i) 
&\rightarrow\\
\pht_{i}(\emptyset) = \msgt{i}{\II}(\emptyset) \bptimes \big (\bigbpplus_{\xx_i} \msg{\II}{i}(\xx_i) \bptimes \msg{i}{\II}(\xx_i) \big ) \quad &\rightarrow \quad \pht_{i}(\emptyset) = \msgt{i}{\II}(\emptyset) \bptimes \msgbit{i}{\II}(\emptyset)
\end{align*}
\end{proof}

The term on the l.h.s. in the proposition above appear in \refEqs{eq:spiI_first}{eq:spIi_first} and the terms on the r.h.s  are local message integrals given by \refEqs{eq:mIi_partition}{eq:miI_partition}.
We can enforce $\msgss{\cdot}{\cdot} \in \WW$, by enforcing BP updates 
$\msg{i}{\II} = \MSG{i}{\II}(\msgss{\nb i \back \II}{i})$ and 
$\msg{\II}{i} = \MSG{\II}{i}(\msgss{\nb \II \back }{\II})$ ``locally'', 
during the message updates in the new factor-graph.
Combining this constraint with the simplification offered by
\refProposition{th:simplify_sp} gives us the SP message updates 
{\small
\begin{align}
  \msp{i}{\II}(\msg{i}{\II}) &\propto  \underset{\msgss{\nb i \back \II}{i}}{\bigspplus} \left (  
\ident\big ( \msg{i}{\II} = \MSG{i}{\II}(\msgss{\nb i \back \II}{i}) \big) \bptimes 
\MSG{i}{\II}(\msgss{\nb i \back \II}{i})(\emptyset) \;  
 \bigbptimes_{\JJ \in \nb i \back \II} \msp{\JJ}{i}(\msg{\JJ}{i}) \right )\label{eq:spiI_semiring}\\
  \msp{\II}{i}(\msg{\II}{i})  &\propto  \underset{\msgss{\nb \II \back i}{\II}}{\bigspplus} \left ( 
\ident\big ( \msg{\II}{i} = \MSG{\II}{i}(\msgss{\nb \II \back i}{\II}) \big) \bptimes 
\MSG{\II}{i}(\msgss{\nb \II \back i}{\II})(\emptyset) \;
 \bigbptimes_{j \in \nb \II \back i} \msp{j}{\II}(\msg{j}{\II}) \right ) \label{eq:spIi_semiring}
\end{align}
}%
where $\ident(.)$ is the identity function on the SP semiring, where $\ident(\truemath) = \identt{\bptimes}$ and
$\ident(\falsemath) = \identt{\spplus}$.

Here each SP message is a functional over all possible BP messages between the same variable and factor. However, in updating the SP messages, the identity functions
ensure that only the messages that locally satisfy BP equations are taken into account.
Another difference from the updates of \refEqs{eq:spiI_first}{eq:spIi_first} is that SP messages
have a single argument. This is because the new local integrals either depend on
$\msg{i}{\II}$ or $\msg{\II}{i}$, and not both.

\begin{example} In variational approach, survey propagation comes in two variations: entropic $\mathrm{SP}(\xi)$ and energetic $\mathrm{SP}(\mathbf{y})$ \cite{Mezard09}. For the readers familiar
\index{Parisi parameter}
with variational derivation of SP, here we express the relation to the algebraic approach.
According to the variational view, the partition function of the \textit{entropic SP}  is  $\sum_{\msgss{\cdot}{\cdot}} e^{\xi \log(\QQ(\msgss{\cdot}{\cdot})(\emptyset))}$, where $\QQ(\msgss{\cdot}{\cdot})(\emptyset)$ is the partition function for the sum-product semiring.
The entropic SP has an inverse temperature parameter, \aka \textit{Parisi parameter}, $\xi \in \Re$. It is easy to see that $\xi = 1$ corresponds to $\spplus = +, \bpplus=+$ and $\bptimes = \times$ in our algebraic approach. 
\index{energetic SP}
\index{SP!energetic}
\index{entropic SP}
\index{SP!entropic}
\index{1RSB}
The limits of $\xi \to \infty$ corresponds to $\spplus = \max$. On the other hand, the limit of
$\xi \to 0$ amounts to ignoring $\QQ(\msgss{\cdot}{\cdot})(\emptyset)$ and corresponds to the counting SP; see \refSection{sec:uniformsp}.

The \textit{energetic $\mathrm{SP}(\mathbf{y})$} is different only in the sense that 
$\QQ(\msgss{\cdot}{\cdot})(\emptyset)$ in $\sum_{\msgss{\cdot}{\cdot}} e^{-\mathbf{y} \log(\QQ(\msgss{\cdot}{\cdot})(\emptyset))}$ is the ground state energy.
This corresponds to $\spplus = +, \bpplus=\max$ and $\bptimes = \sum$, and the limits of
the inverse temperature parameter $\mathbf{y} \to \infty$ is equivalent to $\spplus = \min, \bpplus=\min$ and $\bptimes = \sum$.
By taking an algebraic view we can choose between both operations and domains. For instance,
an implication of algebraic view is that all the variations of SP can be applied to the domain of complex numbers $\RR = \Ce$.
\end{example}

\subsection{The new integral and marginals}
Once again we can use \refTheorem{th:integral_decompose}, this time to approximate the \emph{SP integral}
\index{SP!integral}
$\QQ(\emptyset)(\emptyset) = \bigspplus_{\msgss{\cdot}{\cdot}} \QQ(\msgss{\cdot}{\cdot})(\emptyset)$
using local integral of SP messages.

The \emph{SP marginal} over each BP message $\msg{i}{\II}$ or $\msg{\II}{i}$ 
\index{SP!marginal}
is the same as the corresponding SP message -- \ie $\PSP(\msg{i}{\II}) = \msp{i}{\II}(\msg{i}{\II})$.
To see this in the factor-graph of \refFigure{fig:spfg},
note that each message variable is connected to two factors, and both of these factors are already
contained in calculating one SP messages.

Moreover, from the SP marginals over messages we can recover the SP marginals over BP marginals
which we denote by $\PSP(\ph)(\xx_i)$.
For this, we simply need to enumerate all combinations of BP messages that produce a particular marginal 
\begin{align}\label{eq:sp_marg}
  \PSP(\ph)(\xx_i) \; \propto \; \bigspplus_{\msgss{\nb i}{i}} \ident(\ph(\xx_i) = \PP(\msgss{\nb i}{i})(\xx_i)) \bigbptimes_{\II \in \nb i} \msp{\II}{i}(\msg{\II}{i}) 
\end{align}



\subsection{Counting survey propagation}\label{sec:uniformsp}
\index{counting SP}
Previously we required the $\otimes$ operator to have an inverse, so that we can decompose the BP integral $\qq(\emptyset)$ into local integrals.
Moreover, for a consistent decomposition of the BP integral, SP and BP semiring previously shared the $\otimes$ operation.\footnote{This is because if the expansion operation $\sptimes$ was different from the expansion operation of BP, $\bptimes$, the expanded form $\QQ(\msgss{\cdot}{\cdot})$ in the SP factor-graph
would not evaluate the integral $\qq(\emptyset)$ in the BP factor-graph, even in factor-graphs without any loops.}
 
Here, we lift these requirements by discarding the BP integrals altogether. This means SP semiring could be completely distinct from BP semiring and $(\RR, \otimes)$ does not have to be an Abelian group. This setting is particularly interesting when the SP semiring is sum-product over real domain 
\begin{align}
  \msp{i}{\II}(\msg{i}{\II}) \; &\propto \; \sum_{\msgss{\nb i \back \II}{i}}  
\ident\big ( \msg{i}{\II} = \MSG{i}{\II}(\msgss{\nb i \back \II}{i}) \big) 
 \prod_{\JJ \in \nb i \back \II} \msp{\JJ}{i}(\msg{\JJ}{i}) \label{eq:spiI_counting}\\
  \msp{\II}{i}(\msg{\II}{i}) \; &\propto \; \sum_{\msgss{\nb \II \back i}{\II}} 
\ident\big ( \msg{\II}{i} = \MSG{\II}{i}(\msgss{\nb \II \back i}{\II}) \big) 
 \prod_{j \in \nb \II \back i} \msp{j}{\II}(\msg{j}{\II})  \label{eq:spIi_counting}
\end{align}

Here, the resulting SP integral $\QQ(\msgss{\cdot}{\cdot}) = \sum_{\msgss{\cdot}{\cdot}} \ident(\msgss{\cdot}{\cdot} \in \WW)$ simply ``counts'' the number of BP fixed points and
SP marginals over BP marginals (given by \refEq{eq:sp_marg}) approximates the frequency of a particular
marginal.
The original survey propagation equations in \cite{braunstein_survey_2002}, that are very
successful in solving satisfiability
correspond to counting SP applied to the or-and semiring.
\begin{example}
Interestingly, in all min-max problems with discrete domains $\XX$, min-max BP messages can only take the values that are in the range of factors -- \ie $\RR = \YY$.
This is because any ordered set is closed under min and max operations.
Here, each counting SP message $\msp{i}{\II}(\msg{i}{\II}): \YY^{|\XX_i|} \to \Re$
is a discrete distribution over all possible min-max BP messages.
This means counting survey propagation where the BP semring is min-max is  
computationally ``tractable''.
In contrast (counting) SP, when applied to sum-product BP over real domains is not tractable.
This is because in this case each SP message is a distribution over a uncountable set:
$\msp{i}{\II}(\msg{i}{\II}): \Re^{|\XX_i|} \to \Re$.

In practice, (counting) SP is only interesting if it remains tractable. The most
well-known case corresponds to counting SP when applied to the or-and semiring. In this case
the factors are constraints and the domain of SP messages is $\{\truemath, \falsemath\}^{|\XX_i|}$.
Our algebraic perspective extends this set of tractable instances. For example, it show that 
counting SP can be used to count the number of fixed points of BP when applied to xor-and or min-max semiring.  
\end{example}

%% file: hybrid_methods.tex

\index{inference!stochastic}
\index{hybrid methods}
The contrasting properties of stochastic and deterministic
approximations make a general hybrid method desirable. 
After reviewing the basics of MCMC in \refSection{sec:mcmc},
we discuss some particle-based approaches to message passing and 
introduce our hybrid inference method that combines message passing and Gibbs sampling in 
 \refSection{sec:pbp}.
The discussions of this section are limited to sum-product inference.

\subsection{Markov Chain Monte Carlo}\label{sec:mcmc}
\input{mcmc.tex}

\subsection{Hybrid methods}\label{sec:hybridreview}
Stochastic methods are slow in convergence but they are guaranteed to converge.
Even if the kernel is reducible, samples will cover a subset of the true
support -- \ie MCMC still converges to a single sub-measure when the Gibbs measure is
not unique. 

On the
other hand, deterministic approximations are fast but non-convergent in 
difficult regimes. Modifications that result in convergence are either 
generally intractable (\eg SP), slow (\eg loop corrections and the methods that tighten a bound over the free energy)
 and/or degrade the quality of solutions.

Moreover, sampling methods are flexible in representing 
distributions. 
\index{non-parametric BP}
\index{particle BP}
\index{BP!particle}
\marnote{non-parametric \& particle BP }
This has motivated growing interest in nonparametric approach to
variational inference \cite{Gershman2012} and in particular
variations of Belief Propagation 
\cite{Koller1999,noorshams13a,Song2010,Song2011,AlexanderIhler2003,Ihler2009,Isard2003}.
However, in the sense that
these methods do not rely on a Markov chain for inference, they are closer to variational inference
than MCMC methods.

To better appreciate this distinction, consider two closely related 
methods: Gibbs Sampling and \magn{hard-spin mean-field} \cite{amit1992modeling}, 
\index{mean-field!hard-spin}
that uses the following update equation
\marnote{hard-spin mean-field}
\begin{align*}
\ph(\xx_i) \quad \propto \quad 
\sum_{\xs_{\back i}}\prod_{{\II} \in \nb i} \ff_{\II}(\xs_{\II \back i} , \xx_i)
\prod_{j \in \II \back i} \ph(\xx_j) 
\end{align*} 
Interestingly, the \magn{detailed balance} condition of \cref{eq:transition}
for Gibbs sampler gives us the same equation:
\begin{align*}
\ph(\xx_i) \quad \propto \quad 
\sum_{\xs_{\back i}}\prod_{\II \in \nb i} \ph(\xx_i \mid \xs_{\II \back i}) \prod_{j
\in \II \back i} \ph(\xx_j) 
\end{align*}
 However, given enough iterations, Gibbs Sampling can be much more accurate than  hard-spin mean field method. 
Here the difference is that, with Gibbs sampler, this equation is enforced by the
chain rather than explicit averaging of distributions, which means the correlation information 
is better taken into account.

\input{pbp.tex}

\input{psp.tex}

%% file: mcmc.tex
\marnote{MCMC}
\index{MCMC}
Markov Chain Monte Carlo (MCMC) is a technique to produce samples from a target distribution $\pp$, 
by
exploring a Markov Chain which is constructed such that more probable areas
are visited more often \cite{casella1999monte,Andrieu2003,neal1993probabilistic}.

A Markov Chain is a stochastic process 
$\xs\tst{0},\ldots,\xs\tst{t}$ in which:
\begin{align}
\pp(\xs\tst{t} \mid \xs\tst{t-1} ,\ldots,\xs\tst{1}) \quad = \quad \kernel_t(\xs\tst{t} , \xs\tst{t-1})
\end{align}  
that is the current state $\xs\tst{t}$ is independent of all the history,
given only the previous state $\xs\tst{t-1}$. 
\index{transition kernel}
\index{Markov chain}
\marnote{transition kernel}
For a homogeneous Markov chain, the \magn{transition kernel} $\kernel_t(\xs\tst{t} , \xs\tst{t-1})$ 
 is the same for all $t$.
In this case and under some assumptions\footnote{
The assumptions are: (I) Irreducibility: There is a non-zero
  probability of reaching all states starting with any arbitrary state and
(II) Aperiodicity: The chain does not trap in cycles.}, starting from any arbitrary distribution
$\xs\tst{0} \sim \pp\tst{0}(\xs)$ after at least $T_{\mathrm{mix}}$ transitions by the chain,  we have $\xs\tst{T_{mix}} \sim \pp(x)$. 
Given a set of particles $\xs\nn{1}, \ldots, \xs\nn{L}$ sampled from $\pp(\xs)$,
we can estimate the marginal probabilities (or any other expectation) as 
\begin{align}\label{eq:empavg}
  \ph(\xx_i) \quad \propto \quad \frac{1}{L} \sum_{n=1}^{L}
  \ident(\xx\nn{n}_i = \xx_i)
\end{align}

For a given transition kernel, the following condition, known as detailed
balance, identifies the stationary distribution $\pp$:
\marnote{detailed balance}
\begin{align}
\pp(\xs\tst{t}) \kernel(\xs\tst{t}, \xs\tst{t-1}) \quad &= \quad  \pp(\xs\tst{t-1}) \kernel(\xs\tst{t-1}, \xs\tst{t})
\quad \Rightarrow \notag \\
\pp(\xs\tst{t}) \quad &= \quad \sum_{\xs} \pp(\xs) \; \kernel(\xs\tst{t}; \xs) 
\label{eq:transition}
\end{align}
which means that $\pp(.)$ is the left
eigenvector of $\kernel(.,.)$  with eigenvalue $1$. 
\marnote{mixing time}
All the other
\index{mixing time}
eigenvalues are less than one and the \magn{mixing time}, $T_{\mathrm{mix}}$, of the chain depends on the second
largest eigenvalue; the smaller it is, the faster consecutive transition by 
$\kernel(.,.)$ shrinks the corresponding components, retaining only $\pp$.

\subsubsection{Metropolis-Hasting Algorithm and Gibbs sampling}
Many important MCMC algorithms can be interpreted as a
special case of Metropolis-Hasting (MH) \cite{Metropolis1953,Hastings1970}.
\index{Metropolis-Hasting}
Similar to importance sampling, MH uses proposal distribution $\mm(\xs\tst{t} \mid
\xs\tst{n-1})$, but in this case, the proposal distribution is to help with the design of
transition kernel $\kernel(.,.)$.
After sampling $\xs\tst{t}$ from the proposal $\xs\tst{t} \sim \mm(\xs)$, it is accepted with probability 
\begin{align}\label{eq:mhacceptance}
\min \left \{ 1 \; ,\; \frac{\pp(\xs\tst{t})/\pp(\xs\tst{t-1}) }{ \mm(\xs\tst{t}\mid\xs\tst{t-1}) /
\mm(\xs\tst{t-1}\mid\xs\tst{t})} \right \} 
\end{align}
where, if the proposed sample is not accepted, $\xs\tst{t} = \xs\tst{t-1}$.

The kernel resulting from
this procedure admits the detailed balance condition \wrt the stationary distribution
$\pp(\xs)$. 
An important feature of MCMC, which allows for its application in graphical models, is 
the possibility of building valid transition kernels as the \magn{mixtures and
 cycles} of other transition kernels. If $\pp$, is the stationary distribution
for $\kernel_1$ and $\kernel_2$, then it is also the stationary distribution for $\kernel_1 \kernel_2$
(cycle) and $\lambda \kernel_1 \; +\; (1 - \lambda) \kernel_2, \;\; 0 \leq \lambda \leq 1$ 
(mixture) \cite{Tierney1994}.

\index{Gibbs sampling}
Cycling of kernels gives us \magn{Gibbs
sampling} in graphical models \cite{Geman1984}, when kernels are
\marnote{Gibbs sampling}
\begin{align}
\kernel_i(\xx_i\tst{t}, \xs\tst{t-1}_{\mb i}) \quad = \quad \pp(\xx_i\tst{t} \mid \xs\tst{t-1}_{\mb i}) \quad \forall i
\end{align}
where as before $\mb i$ is the Markov blanket of node $i$.
It is also possible to use block MH-kernels with graphical models. In MH-samplers, 
when highly correlated variables are blocked together, mixing properties
improve.
In fact, the Gibbs sampler is such a method, with the proposal 
distribution that results in acceptance with probability $1$.

Similar to the general MH-methods, Gibbs sampling can fail if the kernel does
not mix properly. This could happen if variables are strongly correlated. 
 In principle one can assemble neighbouring variables into blocks and
update them as one \cite{Jensen1995}. However in difficult
regimes the number of variables that should be flipped to move from one local
optima to another, is in the order of total number of variables
\cite{Mezard09}, which makes this approach intractable.

Mixture of kernels can be used to combine a global proposal with a local
proposal (\eg \cite{DeFreitas2001,Earl2005}).
In fact if we could view a message passing operator as a transition kernel 
(at least when message passing is exact), then the mixture of kernels -- \eg
with Gibbs sampling -- could produce interesting hybrid methods.
In \refSection{sec:pbp}, by combining BP and Gibbs sampling 
operator (when rephrased as message update) we introduce a new hybrid method.

%% file: pbp.tex
\subsection{Perturbed Belief Propagation}\label{sec:pbp}

\index{perturbed BP}
\index{BP!perturbed}
Consider a single particle $\xsh = \xs\nn{1}$ in Gibbs Sampling. 
At any time-step $t$, $\xxh_i$ is updated according to
\begin{align}\label{eq:gs}
  \xxh\tst{t}_i \; \sim \; \ph(\xx_i) \quad \propto \quad \prod_{\II
    \in \nb i} \ff_{\II}(\xx_i , \xsh\tst{t-1}_{\nb I \back i})
\end{align}

Here we establish a
correspondence between a particle in Gibbs Sampling and a set of variable-to-factor 
messages in BP --\ie $\xsh \Leftrightarrow \{\msg{i}{\II}(.)\}_{i, I
  \in \nb i}$, by defining all the messages leaving variable $\xx_i$ as a delta-function
\index{Gibbs sampling!operator}
\marnote{Gibbs sampling operator}
\begin{align}\label{eq:gsop}
  \msg{i}{\II}(\xx_i) \quad \defeq \quad \ident(\xx_i = \xxh_i) \quad \defeq \quad \gsiI{i}{\II}(\msgss{\mb i}{\nb i})(\xx_i)\quad \forall \II
  \in \nb i
\end{align}
where $\gsiI{i}{\II}(\msg{\mb i}{\nb i})$ is the Gibbs sampling operator that defines variable-to-factor
message $\msg{i}{\II}$ as a function of all the messages from Markov blanket of $i$ ($\mb i$) to its adjacent factors ($\nb i$).
To completely define this random operator, note that $\xxh_i$
is a sample from the conditional distribution of Gibbs sampling 
\begin{align}
  \xxh_i \; \sim \; \ph(\xx_i)  \quad &\propto \quad \prod_{\JJ \in \nb i}
  \ff_{\II}(\xx_i , \xsh_{\nb \II \back i})\notag\\
  & \propto \quad \prod_{\II \in \nb i } \sum_{\xx_{\back i}} \ff_{\II}(\xs_\II) \prod_{j \in \nb \II \back i}
  \msg{j}{\II}(\xx_j)\label{eq:gsnbp}
\end{align}

\subsubsection{Combining the operators}\label{sec:pbpop}
Now, lets write the BP updates for sum-product semiring once again;
by substituting the factor-to-variable messages (\cref{eq:mIi_semiring_norm}) into variable-to-factor messages and the marginals (\cref{eq:miI_semiring_norm,eq:marg_semiring_norm}) we get  \marnote{BP operator}
\begin{align}
  \msg{i}{\II}(\xx_i) \quad &\propto \quad \prod_{\JJ \in \nb i \back \II}
  \sum_{\XX_{\nb \JJ \back i}} \ff_{\JJ}(\xs_\JJ) \prod_{j \in \nb \JJ \back i} \msg{j}{\JJ}(\xx_j)
  \quad \propto \quad \MSG{i}{\II}(\msgss{\mb i}{\nb i})(\xx_i) \label{eq:bpop}\\
  \ph(\xx_{i}) \quad &\propto \quad \prod_{\II \in \nb i }
  \sum_{\XX_{\nb \II \back i}} \ff_{\II}(\xs_I) \prod_{j \in \nb \II \back i} \msg{j}{\II}(\xx_j) \label{eq:bpmarg_merged}
\end{align}
where, similar to \cref{eq:miI_semiring_norm}, $\MSG{i}{\II}(.)$ denotes the message
update operator, with the distinction that here, the arguments are also variable-to-factor messages (rather than factor-to-variable messages). 

By this rewriting of BP updates, the BP marginals \cref{eq:bpmarg_merged} are identical in form to the Gibbs sampling distribution of \cref{eq:gsnbp}. 
This similar form allows us to combine the operators linearly to get perturbed BP operator:
\marnote{perturbed BP operator}
\begin{align}
\pbpiI{i}{\II}(\msgss{\mb i}{\nb i}) \defeq \gamma \; \gsiI{i}{\II}(\msgss{\mb i}{\nb i}) + (1 - \gamma) \MSG{i}{\II}(\msg{\mb i}{\nb i}) \quad \forall i, \II \in \nb i
\end{align}

The Perturbed BP operator $\pbpiI{i}{\II}(.)$ updates each message 
by calculating the outgoing message according to BP and GS operators and linearly combines
them to get the final massage.
During $T$ iterations of Perturbed BP, 
the parameter $\gamma$ is gradually and linearly 
changed from $0$ towards $1$.\footnote{Perturbed BP updates can also be used with any fixed $\gamma \in [0,1]$.} 
Algorithm~\ref{alg:pbp} summarizes this procedure. Note that the updates of perturbed BP are compatible with
variable synchronous message update (see \refSection{sec:bpcomplexity}). 

\begin{algorithm}[ht]
\SetKwInOut{Input}{input}\SetKwInOut{Output}{output}
\SetKwFunction{concomp}{ConnectedComponents}
\DontPrintSemicolon
 \Input{a factor graph, number of iterations T.}
 \Output{a sample $\xsh$.}
\DontPrintSemicolon
Initialize messages\;
$\gamma \leftarrow 0$\;
\Repeat{$T$ iterations}{
  \For{\textbf{each} variable $\xx_i$}{
    calculate $\ph(\xx_i)$ using \refEq{eq:gsnbp}\;
    calculate BP messages $\msg{i}{I}(.)$ using \refEq{eq:bpop}\quad $\forall
    \II \in \nb i$\;
    sample $\xxh_i \sim \ph(\xx_i)$ \;
    combine BP and Gibbs sampling messages:
    \begin{align}\label{eq:pbp}
    \msg{i}{\II}(\xx_i) \; \leftarrow \;  \gamma \  \msg{i}{\II}(\xx_i)\  + \ (1 - \gamma)\ \ident(\xx_i =  \xxh_i)
    \end{align}\;
   }
   $\gamma \leftarrow \gamma + \frac{1}{T-1}$\;
}
return $\xsh$
\caption{Perturbed Belief Propagation}\label{alg:pbp}
\end{algorithm}

\subsubsection{Experiments}
Perturbed BP is most successful in solving CSPs; see \refSection{sec:csp}. 
However we can also use it for marginalization  by sampling (\refEq{eq:empavg}).
Here we use the spin-glass Ising model on $8\times8$ grids and Erd\H{o}s-R\'{e}ny (ER) random graphs with $N=50$ and $150$ edges. 
We sampled local fields independently from $\mathcal{N}(0,
1)$ and interactions from $\mathcal{N}(0, \theta^2)$, where we change $\theta$
to control the problem difficulty -- higher values correspond to more difficult inference problems. 
We then compared the average of the logarithm (base $10$) of mean (over $N$ variables)
marginal error 
\index{Ising}
\begin{align}\label{eq:errmeasure}
\log( \frac{1}{N}\sum_{i,\xx_i} \vert \pp(\xx_i) - \ph(\xx_i) \vert) 
\end{align}
\begin{figure}[pht!]
  \begin{subfigure}[t]{.45\columnwidth}
\centering
\includegraphics[width=1\columnwidth]{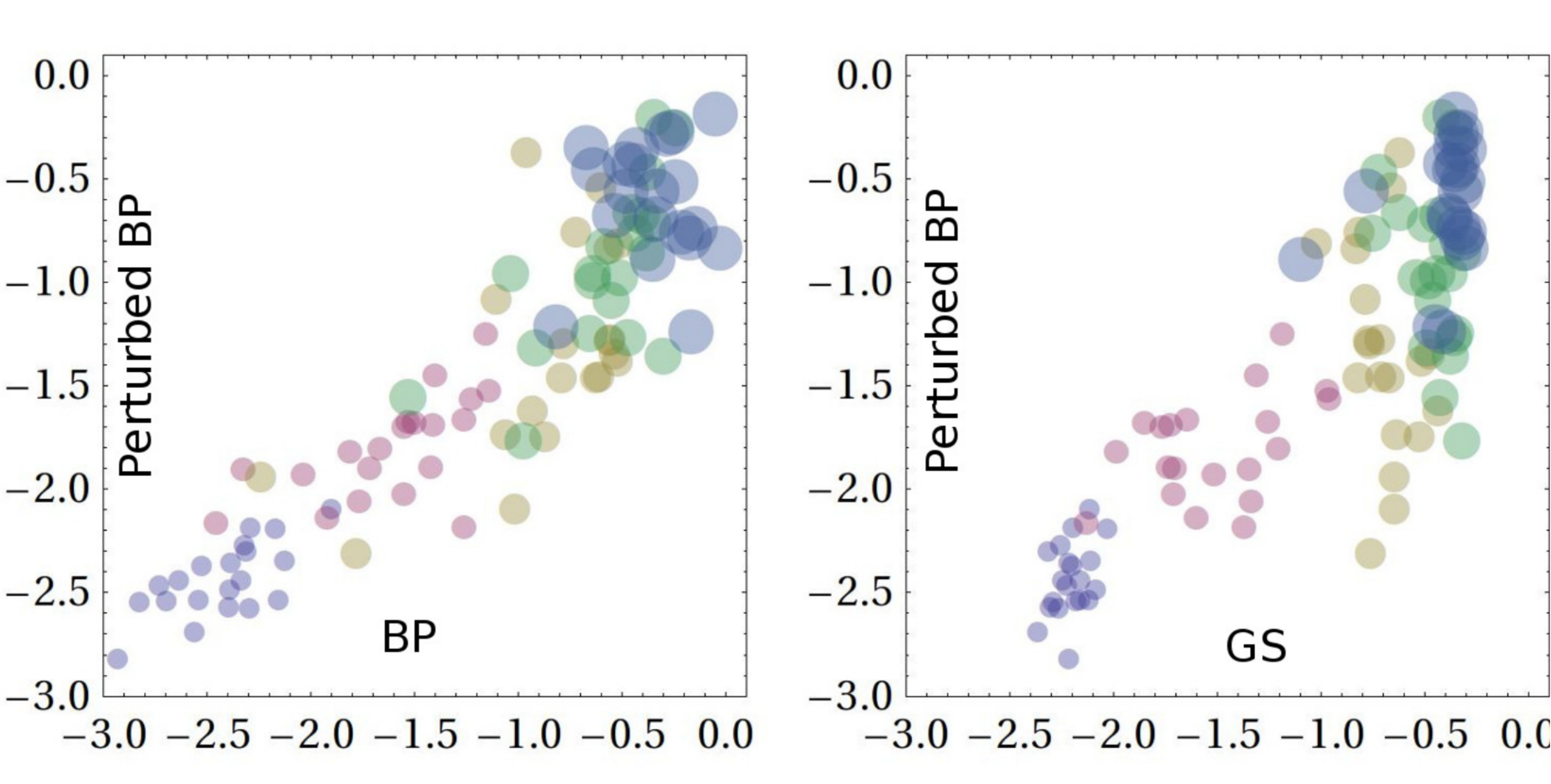}
    \caption{Ising grid}
  \end{subfigure}
~
  \begin{subfigure}[t]{.45\columnwidth}
\centering
\includegraphics[width=1\columnwidth]{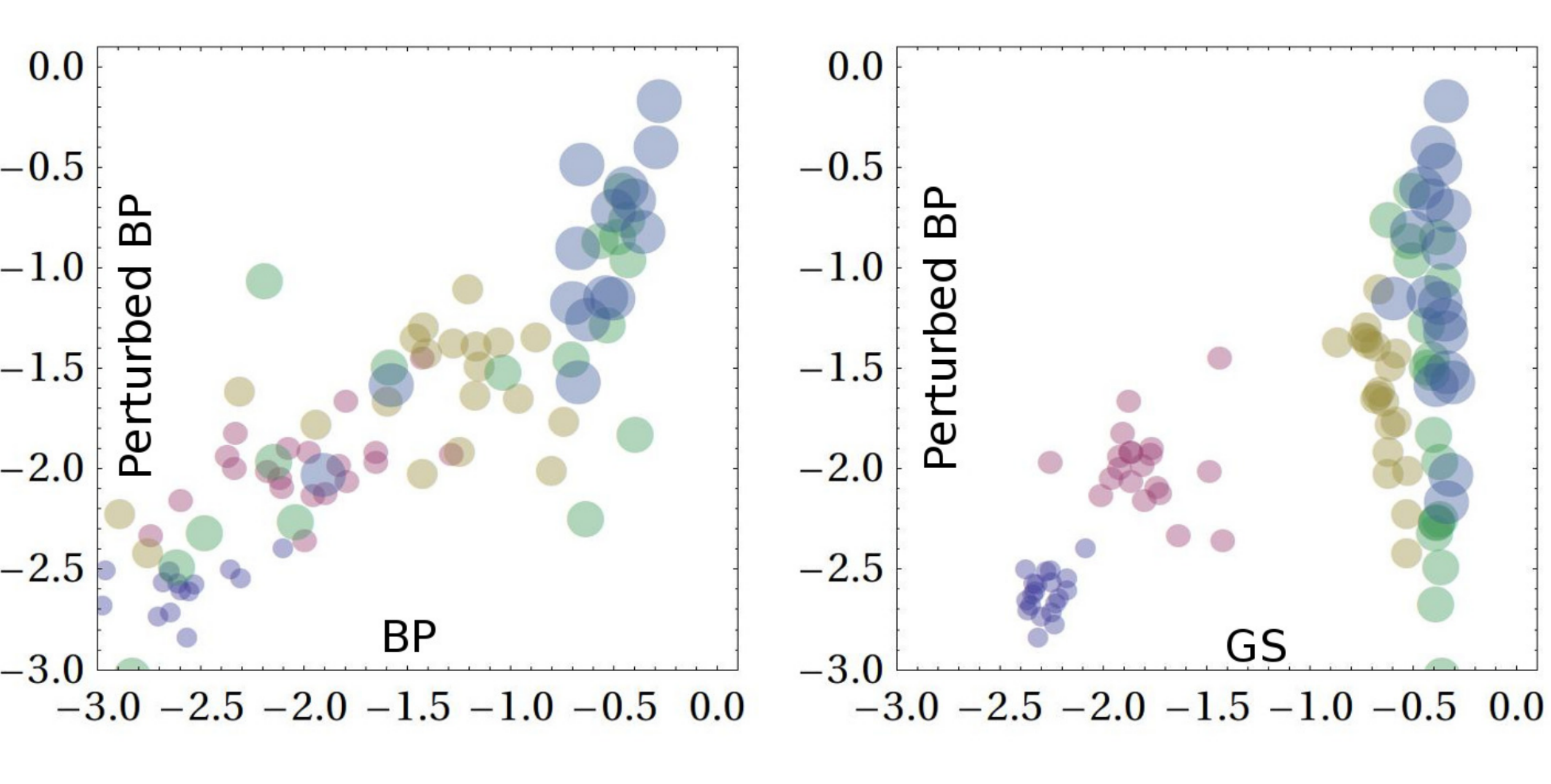}
    \caption{Random ER graph}
  \end{subfigure}
\caption[Comparison of marginalization accuracy for BP, Gibbs sampling and Perturbed BP.]{
Log mean marginal error (x and y axes) comparison between Perturbed BP, BP and Gibbs sampling for 
\textbf{(left)} 8x8 periodic Ising grid; \textbf{(right)}
random graph with 50 variables, 150 edges and spin-glass interactions. The size of each
circle is proportional to the difficulty of that problem.
}
\label{fig:perturbedbp}
\end{figure} 

\RefFigure{fig:perturbedbp} compares perturbed BP, 
Gibbs sampling and BP where larger
circles correspond to more difficult instances $\theta \in \{0.5, 1, 2,
4, 8\}$. All methods are given a maximum of $10,000$ iterations.
Perturbed BP and Gibbs sampling use $T = 100$ iterations to obtain each sample, while BP is ran once until 
convergence or until the maximum number of iterations is reached.

The results shows that Perturbed BP as a sampling method is generally better than Gibbs Sampling.
Also, for some cases in which BP's result is very close to
random (\ie $\sim -.3$ in log marginal error), Perturbed BP produces relatively
better results. 
Note that for difficult instances, increasing $T$ even by 100 folds does not significantly 
improve the results for either Gibbs sampling or perturbed BP. For Gibbs sampling, this can be explained by formation of
 pure states that result in exponential mixing time~\cite{Mezard09,levin2009markov}.

%% file: psp.tex

\subsection{Perturbed survey propagation}\label{sec:psp}
\index{SP!perturbed}
When the SP operators are sum-product -- \ie $\spplus = \sumop$ and $\sptimes = \prodop$ --
we can apply a perturbation scheme similar to perturbed BP.

Recall that sum-product SP defines a distribution over BP fixed point. Therefore
sampling from this distribution amounts to randomly selecting a single BP fixed point.
This corresponds to sampling a single message $\msg{i}{\II}\nn{1} \sim \MSP{i}{\II}(\msg{i}{\II})$\footnote{Recall that the SP marginal over each message $\msg{i}{\II}$
is identical to the corresponding message $\MSP{i}{\II}(\msg{i}{\II})$.
}
and bias the SP message $\MSP{i}{\II}(.)$ towards this random choice 
-- \ie (analogous to \cref{eq:pbp} in \cref{alg:pbp})
\begin{align}
  \MSP{i}{\II}(\msg{i}{\II})\; \quad &  \leftarrow \quad \gamma \; \MSP{i}{\II}(\msg{i}{\II}) \; + \; (1 - \gamma) \ident(\msg{i}{\II} = \msg{i}{\II}\nn{1}) \label{eq:psp_good}\\
\text{where} \quad & \msg{i}{\II}\nn{1} \sim  \MSP{i}{\II}(\msg{i}{\II}) \notag
\end{align}




An alternative form of perturbation is to perturb SP messages using
 implicit SP marginals.
Recall that in using counting SP, the SP marginals over BP marginals ($\PSP(\ph)(\xx_i)$; see \refEq{eq:sp_marg})
are simply the frequency of observing a particular marginal in BP fixed points.
This implicitly defines SP marginal over the original domains $\XX_i; \forall i$, 
which we denote by $\PSP(\xx_i)$
\index{SP!marginal}
\begin{align}\label{eq:sp_implicit_marg}
\PSP(\xx_i) \propto \sum_{\ph} \PSP(\ph)(\xx_i)
\end{align}

After obtaining a sample $\xxh_i \sim \PSP(\xx_i)$,
 we bias all the outgoing SP messages accordingly 
 \begin{align}
   \MSP{i}{\II}(\msg{i}{\II}) \quad &\leftarrow \quad  \gamma \; \MSP{i}{\II}(\msg{i}{\II}) \; + \; (1 - \gamma) \; \ident \big (\msg{i}{\II}(.) = \ident(\xxh_i, .) \big) \quad \forall \II \in \nb i \label{eq:pspc}\\
 \text{where} \quad & \xxh_i \sim \PSP_i(\xx_i)  \notag
 \end{align}
where, similar to perturbed BP,  $\gamma$ is gradually increased from
$0$ to $1$ during $T$ iterations of Perturbed SP. We use this form of perturbation
in \refSection{sec:sat_col} to obtain a satisfying assignment $\xsh$, to CSPs. We show that
 although computationally more expensive than perturbed BP, this method often outperforms all
the other well-known methods in solving random CSPs.

%% file: chap4.tex
\part{Combinatorial problems}\label{chapter:combinatorial}\label{part2}
Message Passing algorithms of different semirings are able to solve a variety of combinatorial problems:
(I)~To solve \magn{constraint satisfaction problems (CSPs)} the sum-product message passing is often used, where $\pp(\xs)$ defines a uniform distribution over solutions
and the objective is to produce a single assignment $\xs^*$ s.t. $\pp(\xs^*) > 0$.
(II)~Here the estimates of the partition function, either using the approximation given by the Bethe free energy (\cref{sec:sumprod_n_friends}) or the decomposition of integral in \refSection{sec:decompose_integral}, is used for \magn{approximate counting} of the number of solutions. 
This estimate to the partition function
is also used for integration problems such approximating the permanent of a matrix (see \refChapter{sec:permutations}). 
(III)~The min-sum semiring is often used for \magn{(constrained) optimization} and
we formulate 
(IV) \magn{bottleneck problems}  as min-max inference.


This part of the thesis studies the message passing solutions to combinatorial problems under three broad categories.
(1) \RefChapter{sec:csp} studies the constraint satisfaction problems, where we use perturbed message passing (\refChapter{sec:pbp} and \ref{sec:psp})
to produce state-of-the-art results in solving random instances of satisfiability and coloring problems in \refChapter{sec:sat_col}.
This chapter then studies  several other \NP-hard problems including set-cover, independent set, max-clique, clique-cover and 
packing for construction of non-linear codes. 
(2) \RefChapter{sec:cluster} studies variations of clustering problems including k-median, k-center, k-clustering, hierarchical clustering and modularity optimization. 
(3) In \refChapter{sec:permutations} we study problems that involve enumeration, constraint satisfaction or constrained optimization over permutations. This includes  
 (bottleneck) travelling salesman problem,
matching, graph alignment, graph isomorphism and finding symmetries.

Note that this classification of combinatorial problems into three categories is superficial and is made solely to provide some organization.
In several places we violate this categorization in favour of better flow.
For example we study some constraint satisfaction problems such as (sub)-graph isomorphism and Hamiltonian cycle in \refSection{sec:permutations}  
rather than \refSection{sec:csp}. We investigate the ``optimization'' counterpart of some CSPs in \refSection{sec:csp} and
review message passing solutions to finding trees rather than clusters in \refSection{sec:cluster}.
Moreover, many of the graphical models presented here are proposed by other researchers and they are
included here only for completeness. As a final remark, we note that many of the statements in the following are assuming $\Poly \neq \NP$.

\chapter{Constraint satisfaction}\label{sec:csp}\label{chapter:csp}
\input{csp_intro.tex}

\section{Satisfiability and coloring}\label{sec:sat_col}
\input{sat.tex}

\input{other_csps.tex}

\chapter{Clustering}\label{sec:cluster}\label{chapter:clustering}
\input{clusters.tex}

\input{minmax_clustering.tex}

\section{Modularity maximization}\label{sec:modularity}
\input{modularity.tex}

\chapter{Permutations}\label{sec:permutations}\label{chapter:permutations}
\section{Matching and permanent}\label{sec:matching}
\input{matching.tex}

\section{Traveling salesman problem}\label{sec:tsp}
\input{tsp.tex}

\section{Graph matching problems}\label{sec:graphongraph}
\input{symm.tex}

%% file: csp_intro.tex

\index{constraint satisfaction problem}
We saw in \refSection{sec:gdl} that ``any'' semiring can formulate Constraint Satisfaction Problems (CSPs).
In particular, as we saw in \refSection{sec:gdl}, several semirings are isomorphic to
the and-or $(\{\falsemath, \truemath \}, \vee, \wedge)$ semiring
 and therefore result in equivalent BP procedures. \marnote{warning propagation}
The BP message update over the and-or semiring is called \magn{warning propagation} (WP).
\index{warning propagation}
\index{message passing!warning propagation}
WP marginals indicate whether or not a particular assignment to each variable is allowed, and therefore 
indicate a \emph{cluster of solutions}.  
However, the success of warning propagation highly depends on initialization of messages. In contrast, if convergent, 
the fixed points of BP on the sum-product semiring $(\Re^{\geq 0}, +, \times)$ are less dependent on initialization.

\begin{example}\label{example:kcoloring}\magn{$K$-coloring:} \marnote{K-coloring}
Given a graph $\GG = (\VV, \EE)$, the K-coloring (K-COL) problem asks whether it
is possible to assign one color (out of K) to each node s.t. no two adjacent 
nodes have the same color. 
  Here, $\xx_i \in \XX_i = \{1,\ldots,q\}$ is a K-ary variable for
  each $i \in \NN$, and we have $M = \vert \EE \vert$ constraints; each constraint
  $\ff_{i,j}(\xx_{i}, \xx_j) = \ident(\xx_i \neq \xx_j) $ depends only on
  two variables and is satisfied iff the two variables have different
  values. Here the identity function $\ident(\xx_i \neq \xx_j) $ depends on the semiring (see \refSection{sec:gdl}).


\end{example}

\begin{figure}

\hbox{
  \centering
\hspace{.2\textwidth}
    \includegraphics[width=.3\textwidth]{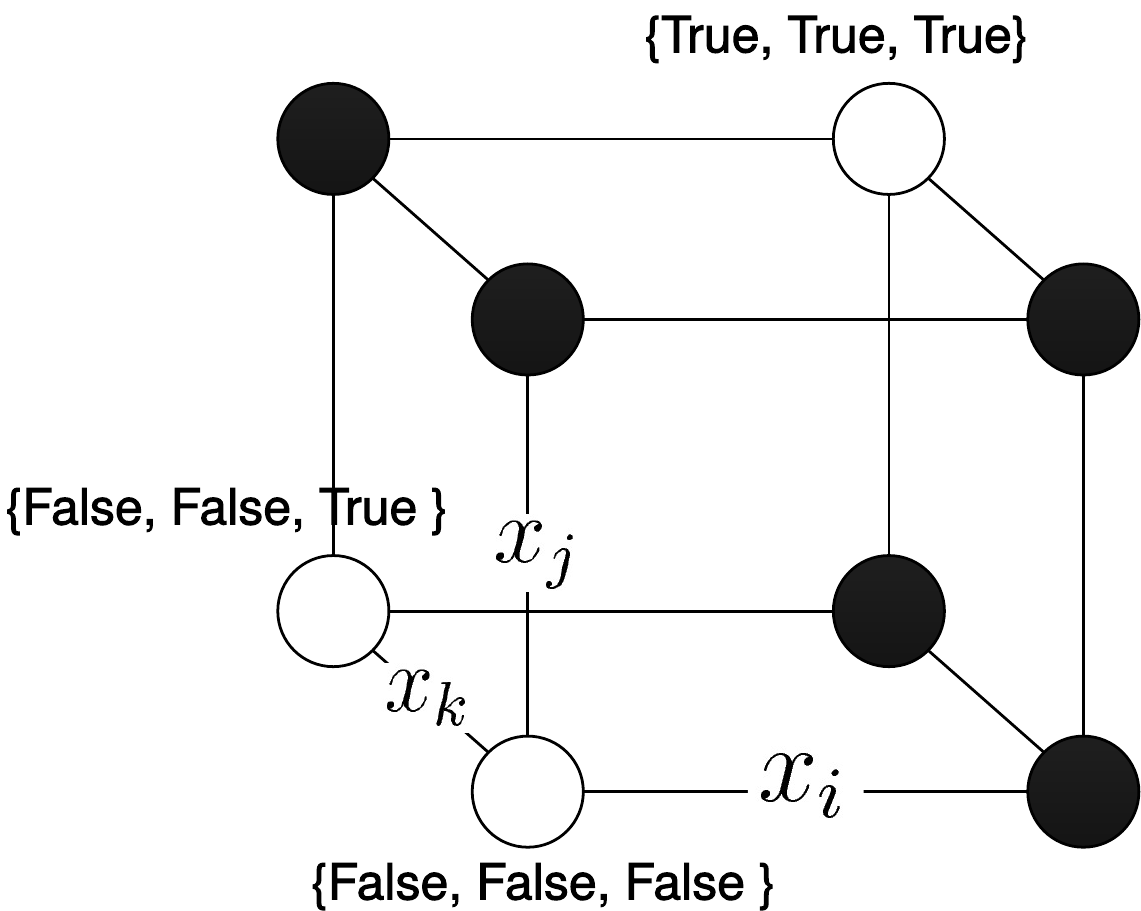}
    \includegraphics[width=.2\textwidth]{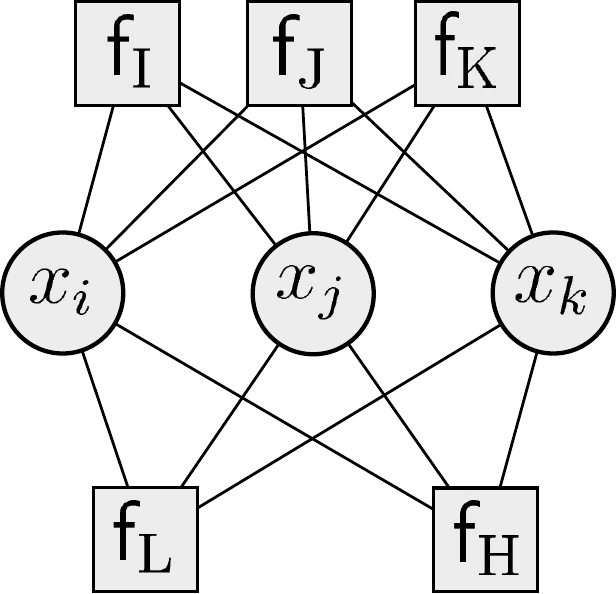}
}
  \caption[Factor-graph and set of solutions for a simple 3-SAT problem.]{\small{(a) The set of all possible assignments to 3
      variables. The solutions to the 3-SAT problem of
      \cref{eq:examplesat} are in white
      circles. 
      (b) The factor-graph corresponding to the 3-SAT problem of
      \cref{eq:examplesat}.
 Here each factor prohibits a single
      assignment.  }}
  \label{fig:simple3sat}
\end{figure}

\index{satisfiability! factor graph}
\begin{example} \magn{$K$-Satisfiability (K-SAT):} \marnote{K-Satisfiability}
Given a conjunction of disjunctions with  K  literals, K-satisfiability
seeks an assignment that evaluates to $\truemath$.
Here, all variables are binary ($\XX_i =
  \{ \truemath,\falsemath \}$) and each clause (factor $\ff_{\II}$) depends on $K =
  \mid \partial \II \mid$ variables. A clause evaluates to zero only for a single
  assignment out of $2^{K}$ possible assignment of variables~\cite{garey_computers_1979}.


  Consider the following (or-and semiring formulation of) 3-SAT problem over 3 variables with 5 clauses:
  \begin{align}\label{eq:examplesat}
    \qq(\xs)  =  &\underbrace{( \neg \xx_i \vee \neg \xx_j \vee \xx_k)}_{\ff_\II} \wedge \underbrace{( \neg
    \xx_i \vee \xx_j \vee \xx_k)}_{\ff_\JJ} \wedge \underbrace{( \xx_i \vee \neg \xx_j \vee \xx_k)}_{\ff_{\KK}}
    \wedge\\
 &\underbrace{( \neg \xx_i \vee \xx_j \vee \neg \xx_k)}_{\ff_{\LL}} \wedge \underbrace{( \xx_i \vee
    \neg \xx_j \vee \neg \xx_k)}_{\ff_\HH} \notag
  \end{align}


  The factor $\ff_\II$ corresponding to the first clause takes the value
  $\identt{\bptimes}$ (for or-and semiring this corresponds to $\identt{\wedge} = \truemath$), except for $\xs_\II = (\truemath, \truemath, \falsemath )$, in which case it is
  equal to $\identt{\bpplus}$ ($\identt{\vee}= \falsemath$). \RefFigure{fig:simple3sat} shows this factor-graph and its set of solutions:\\
  $\solutions = \big \{(\truemath, \truemath, \truemath), (\falsemath, \falsemath, \falsemath),
  (\falsemath, \falsemath, \truemath) \big\}$.  
\end{example}
\marnote{number of solutions}
When using sum-product semiring, where $\ff_\II(\xs) \in \YY_\II = \{0,1\}$, $\pp(\xs)$ (\refEq{eq:semiring_marginalization}) defines a uniform distribution over the set of solutions and 
the partition function $\qq(\emptyset)$ counts the number of solutions.
The challenge is then to sample from this distribution
(or estimate $\qq(\emptyset)$). 
The common approach to sample from $\pp(\xs)$ is to use decimation (see \refSection{sec:decimation}).
\index{decimation}
\marnote{BP-dec}
Here one repeatedly applies sum-product BP to estimate marginals $\ph(\xx_i)$. Then one fixes a subset of variables $\xs_{\AAA}$
according to their marginals. For this one may sample $\xx^*_i \sim \ph(\xx_i)$ or select $\xx_i$ with maximum marginal $\xx^*_i = \arg_{\xx_i} \max \ph(\xx_i)$.
Sum-product BP is then applied to the reduced factor-graph in which $\ff_\II(\xs_\II)$ is replaced by
$\ff_{\II}(\xs_{\II}) \ident(\xs_{\II \cap \AAA} = \xs^*_{\II \cap \AAA})$.
This process, called \magn{sum-product BP-guided-decimation} (BP-dec), is repeated to obtain a complete joint assignment $\xs^*$.

However, using BP-guided-decimation is solving a more difficult problem of marginalization.
In fact, in \refSection{sec:decimation} we showed how using decimation one may estimate the partition function (which is a \sharpP\ problem).
This suggests that decimation may not be the most efficient approach to solving \NP-complete CSPs.
Here, instead we consider using the perturbed belief propagation (\refSection{sec:pbp}) to sample from the set of solutions, where the semiring
used by perturbed BP is the same as sum-product BP-dec.

To better understand warning propagation, sum-product BP-dec, and perturbed BP, when applied to CSPs, consider the following examples.

\begin{example} Here we apply three different message passing methods to solve the simple
3-SAT example of \refFigure{fig:simple3sat}.\\

\noindent {\magn (I) Warning Propagation:} 
\\\noindent
We use the max-product semiring $(\{0,1\}, \max, \prodop)$ version of warning propagation for this example.
As \refFigure{fig:simple3sat} suggests, the set of solutions $\solutions$ clusters into two subsets\\
$\{\{\truemath, \truemath, \truemath \} \}$ and $\{\{\falsemath, \falsemath, \falsemath \},
  \{ \falsemath, \falsemath, \truemath\} \big\}$.
Here, each of the clusters is a fixed point for WP -- \eg the cluster with two solutions corresponds to the following fixed point
  \begin{align*}
    &\msg{i}{\AAA}(\truemath) = \ph(\xx_i = \truemath) = 0 & \\
    &\msg{i}{\AAA}(\falsemath) = \ph(\xx_i = \falsemath) = 1  &\forall \AAA \in \nb i\\
    &\msg{j}{\AAA}(\truemath) = \ph(\xx_j = \truemath) = 0 &\\
    &\msg{j}{\AAA}(\falsemath) = \ph(\xx_j = \falsemath) = 1 &\forall \AAA \in \nb j\\
    &\msg{k}{\AAA}(\truemath) = \ph(\xx_k = \truemath) = 1 &\\ 
    &\msg{k}{\AAA}(\falsemath) = \ph(\xx_k = \falsemath) = 1 & \forall \AAA \in \nb k
  \end{align*}
  where the messages indicate the allowed assignments within this
  particular cluster of solutions. Depending on the initialization, WP messages may converge to any
of its fixed points that also include the trivial cluster, where all (alternatively none) of the assignments are allowed.\\ 

\noindent\magn{(II) BP-dec:}\\\noindent
  Applying BP to this 3-SAT problem (starting from uniform messages) 
  takes 20
  iterations to converge -- \ie for the maximum change in the
  marginals to be below $\epsilon = 10^{-9}$.  
Here the message, $\msg{\II}{i}(\xx_i)$, from $\ff_\II$ to $\xx_i$ is:
\begin{align*}
\msg{\II}{1}(\xx_i) \quad \propto \quad \sum_{\xs_{j,k}} \ff_{\II}(\xs_{1,2,3})\; \msg{j}{\II}(\xx_j)\; \msg{k}{\II}(\xx_k)
\end{align*}

Similarly, the message in the opposite direction, $\msg{k}{\II}(\xx_i)$ is
\begin{align*}
\msg{i}{\II}(\xx_i) \quad \propto \quad \msg{\JJ}{i}(x_i)\; \msg{\KK}{i}(\xx_i)\; \msg{\LL}{i}(\xx_i)\; \msg{\HH}{i}(\xx_i)
\end{align*}

Here BP gives us the
  following approximate marginals: $\ph(\xx_i = \truemath) = \ph(\xx_j = \truemath) = .319$
  and $\ph(\xx_k = \truemath) = .522$.  From the set of solutions, we know that
  the correct marginals are $\pp(\xx_i = \truemath) = \pp(\xx_j = \truemath) = 1/3$ and
  $\pp(\xx_k = \truemath) = 2/3$.  The error of BP is caused by influential loops
  in the factor-graph of \refFigure{fig:simple3sat}(b). Here the error
  is rather small; it can be arbitrarily large in some instances; sometimes it ca
prevent converging at all.

  By fixing the value of $\xx_i$ to $\falsemath$, the
  SAT problem of \cref{eq:examplesat} collapses to:
  \begin{align}
    SAT(\xs_{\{j,k\}}\mid \xx_i = \falsemath) \; = \; ( \neg \xx_j \vee \xx_k)
    \wedge ( \neg \xx_j \vee \neg \xx_k)
  \end{align}

  BP-dec applies BP again to this reduced problem, which give
  $\ph(\xx_j = \truemath) = .14$ (note here that $\pp(\xx_j = \truemath) = 0$) and
  $\ph(\xx_k = \truemath) = 1/2$. By fixing $\xx_j$ to $\falsemath$, another round of
  decimation yields a solution $\xs^* = \{\falsemath, \falsemath, \truemath \}$.\\

\noindent
\magn{(III) Perturbed Belief Propagation:}\\\noindent
Perturbed BP can find a solution in $T = 4$ iterations (see \pcref{alg:pbp}).
Our implementation shuffles the order of updates for variables in each iteration. 

In the first iteration, $\gamma = 0$, which means updates are the same as that of sum-product BP. 
In the second iteration, the order of updates is $\xx_j$, $\xx_k$, $\xx_i$ and $\gamma = 1/3$. At the 
end of this iteration  $\ph\tst{t=2}(\xx_j = \truemath) = .38$. Perturbed BP then samples $\xxh_j = \falsemath$
from this marginal. This sample influences the outgoing message according to the perturbed BP update \cref{eq:pbp},
which in turn influences the beliefs for $\xx_i$ and $\xx_k$. At the end of this iteration $\ph\tst{t=2}(\xx_i = \truemath) = .20$ and $\ph\tst{t=2}(\xx_k = \truemath) = .53$.
At the final iteration $\gamma = 1$ and the order of updates is $\xx_i, \xx_j$ and $\xx_k$.
At this point $\ph\tst{t=3}(\xx_i  = \truemath) = .07$ and the sample $\xxh_i = \falsemath$.
This means the outgoing message is deterministic (\ie $\msg{i}{\AAA}(\falsemath) = 1$ and $\msg{i}{\AAA}(\truemath) = 0$, for all $\AAA \in \nb i$).
This choice propagates to select $\xxh_j = \falsemath$. Finally $\ph\tst{t=3}(\xx_k = \truemath) = \ph\tst{t=3}(\xx_k = \truemath) = .5$, which correctly
shows that both choices for $\xxh_k$ produce a solution.
 


\end{example}

To compare the performance of sum-product BP-dec and 
Perturbed BP on general CSPs, we considered all CSP instances from XCSP repository
\cite{roussel2009xml, xcsp}, that do not include global
constraints or complex domains. All instances with intensive
constraints (\ie functional form) were converted into extensive format
for explicit representation using dense factors.  We further removed
instances containing constraints with more that $10^6$ enteries in
their tabular form.  We also discarded instances that collectively had
more than $10^8$ enteries in the dense tabular form of their
constraints.\footnote{Since our implementation represents all factors
  in a dense tabular form, we had to remove many instances
  because of their large factor size. We anticipate that 
Perturbed BP and BP-dec could probably solve many of these instances 
using a sparse representation of factors.}

\begin{figure}
  \centering
\hbox{
    \includegraphics[width=.5\textwidth]{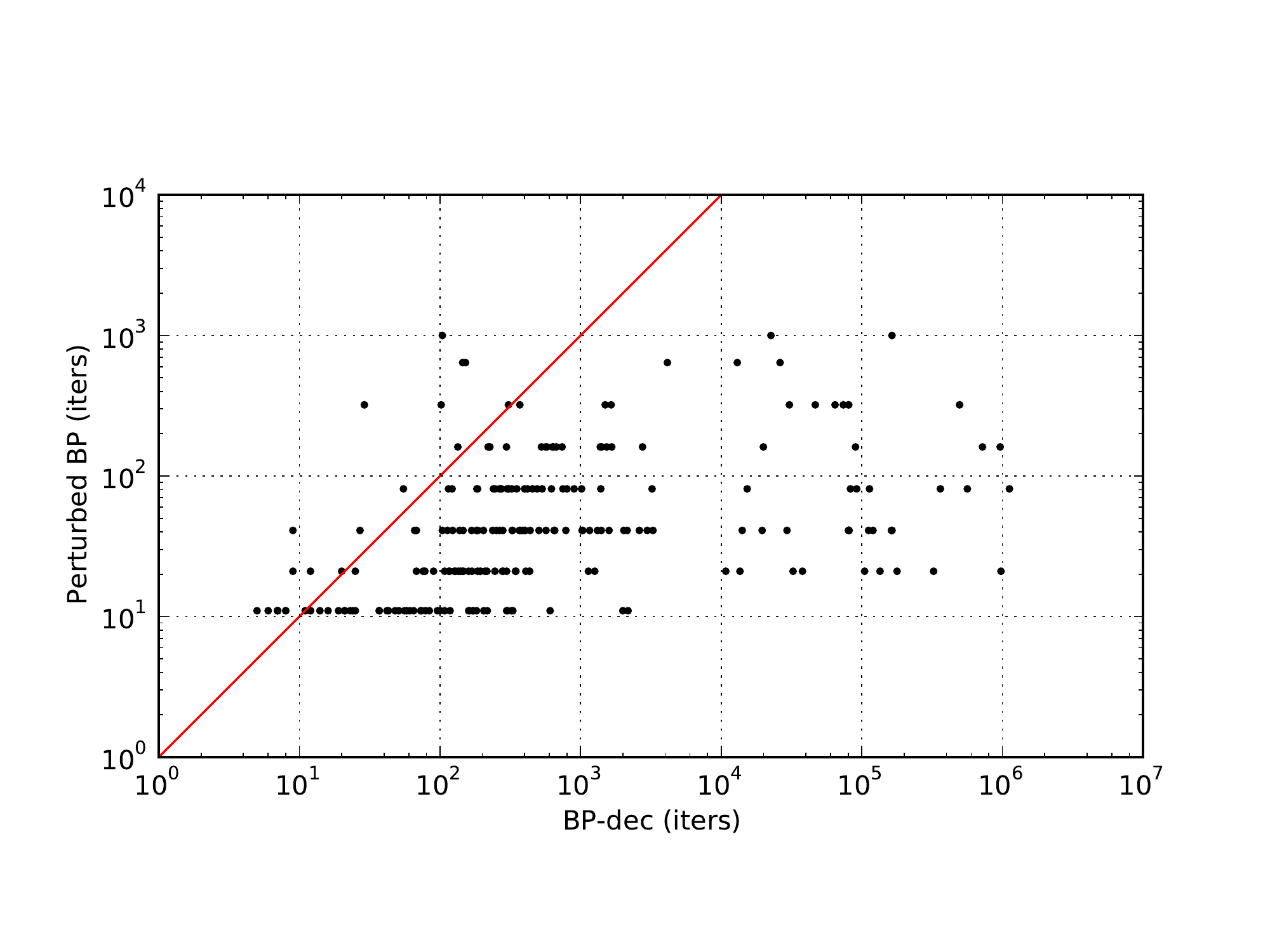}
    \includegraphics[width=.5\textwidth]{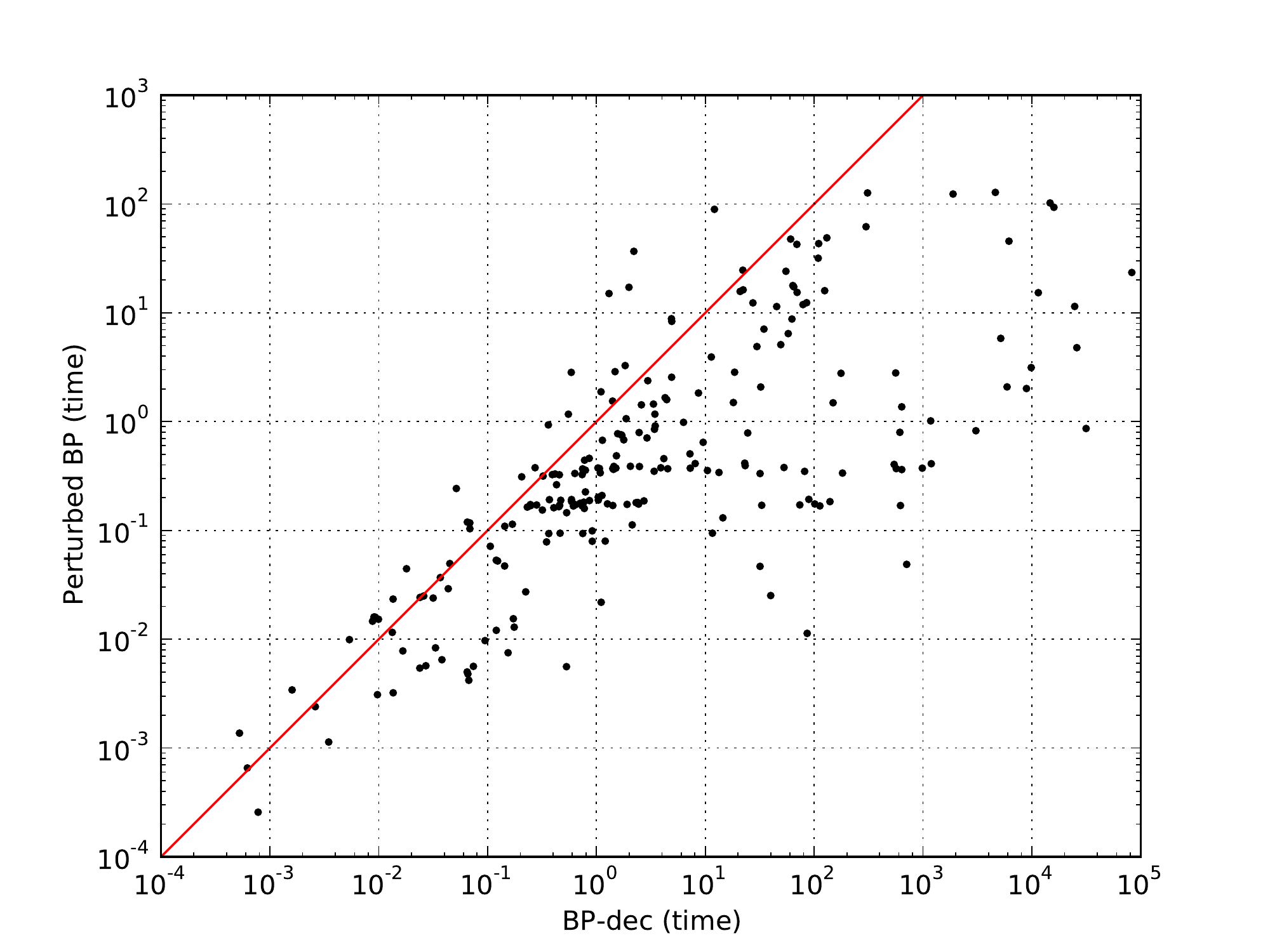}
}

  \caption[Comparison of perturbed BP and BP-dec on real-world benchmark CSPs.]{Comparison of  number of iterations (left) and time (right) used by BP-dec and Perturbed BP
in benchmark instances where both methods found satisfying assignments. 
}
\label{fig:benchmark}
\end{figure}

\RefFigure{fig:benchmark}(a,b) compares the time and iterations of BP-dec and
Perturbed BP for successful attempts where both methods satisfied an instance.
\footnote{
  We used a convergence threshold of $\epsilon = .001$ for BP and
terminated if the threshold was not reached after $T = 10
\times 2^{10} = 10,240$ iterations. To perform decimation, we sort the
variables according to their bias and fix $\rho$ fraction of the most biased
variables in each iteration of decimation. This fraction, $\rho$,
was initially set to $100\%$, and it was divided by $2$ each time
BP-dec failed on the same instance. BP-dec was repeatedly applied
using the reduced $\rho$, at most $10$ times, unless a solution was
reached -- \ie $\rho = .1\%$ at final attempt.

For Perturbed BP, $T = 10$ at the starting attempt, which was
increased by a factor of $2$ in case of failure. This was repeated at
most $10$ times which means Perturbed BP used $T = 10,240$ at its final
attempt. Note that Perturbed BP at most uses the same number of
iterations as the maximum iterations per single iteration of decimation in
BP-dec.
}

Overall Perturbed BP, with $284$ solved instances, is more successful
than BP-dec with $253$ successful runs. On the other hand, the average
number of iterations for successful instances of BP-dec is $41,284$,
compared to $133$ iterations for Perturbed BP.  This makes Perturbed
BP $300$ times more efficient than BP-dec.
\footnote{
We also ran BP-dec on all the benchmarks with maximum number of
iterations set to $T = 1000$ and $T = 100$ iterations.  This reduced
the number of satisfied instances to $249$ for $T = 1000$ and $247$
for $T = 100$, but also reduced the average number of iterations to
$1570$ and $562$ respectively, which are still several folds more
expensive than Perturbed BP. see Appendix xyz for more details on these results. 
}

\section{Phase transitions in random CSPs}
\index{phase transition}
\index{rCSP}
Random CSP (rCSP) instances have been extensively used in order to \marnote{rCSP}
study the properties of combinatorial problems
\cite{fu_application_1986,mitchell1992hard,achioptas2000optimal,krzakala_gibbs_2007}
as well as in analysis and design of algorithms
\cite{selman1994noise,mezard_analytic_2002}.  

\index{critical phenomena}
Studies of rCSP, as a critical phenomena, focus on the geometry of the
solution space as a function of the problem's difficulty, where
rigorous \cite{achlioptas_algorithmic_2008,cocco_rigorous_2002}
and non-rigorous  
\cite{mezard_bethe_2000,mezard_cavity_2002} analyses
have confirmed the same geometric picture.

\marnote{control parameter}
\index{control parameter}
When working with large random instances, a scalar $\alpha$ associated
with a problem instance, \aka \magn{control parameter} -- \eg the clause to
variable ratio in SAT-- can characterize that instance's difficulty
(\ie larger control parameter corresponds to a more difficult
instance) and in many situations it characterizes a sharp transition
from satisfiability to unsatisfiability \cite{cheeseman_where_1991}.  

\index{K-SAT!random}
\begin{example}\label{example:sat_generate}\magn{Random $K$-satisfiability} \marnote{random K-SAT}
  Random $K$-SAT instance with $N$ variables and $M = \alpha N$
  constraints are generated by selecting $K$ variables at random
   for each constraint.  Each constraint is set to
  zero (\ie unsatisfied) for a single random assignment (out of
  $2^{K}$).  Here $\alpha$ is the control parameter.
\end{example}
\index{K-coloring!random}
\begin{example}\label{example:col_generate}\magn{Random $K$-coloring} \marnote{random K-COL}
  The control parameter for a random $K$-COL instances with $N$
  variables and $M$ constraints is its average degree $\alpha =
  \frac{2M}{N}$.  We consider Erd\H{o}s-R\'{e}ny random graphs and
  generate a random instance by sequentially selecting two distinct variables out of $N$ at
  random to generate each of $M$ edges. For large $N$, this is
  equivalent to selecting each possible factor with a fixed
  probability, which means the nodes have Poisson degree distribution
  $\Pr(\vert\nb i\vert = d) \propto e^{-\alpha} \alpha^{d}$.
\end{example}

\begin{figure}[pth!]
  \centering
\hbox{
  \centering
    \includegraphics[width=.3\textwidth]{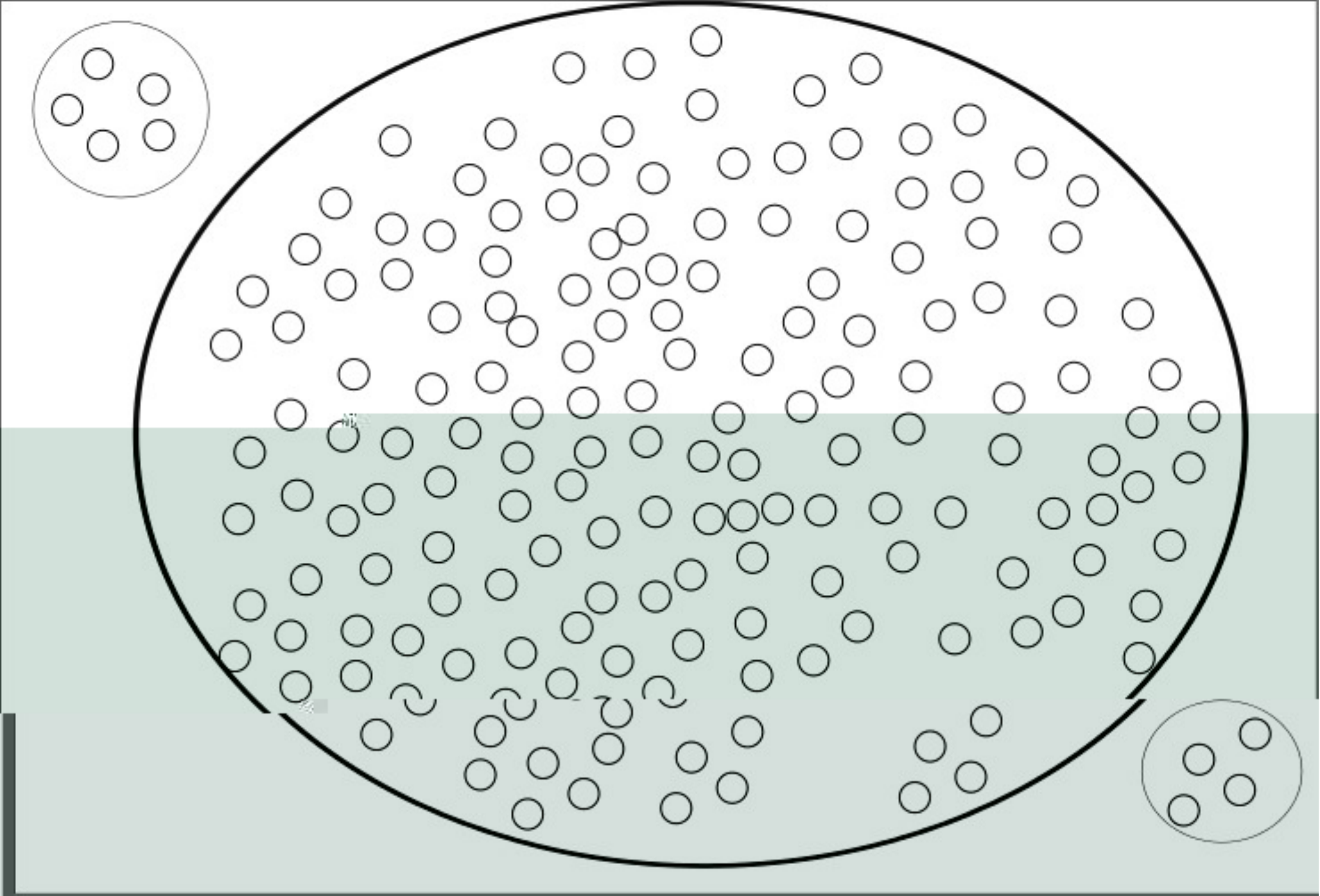}

    \includegraphics[width=.3\textwidth]{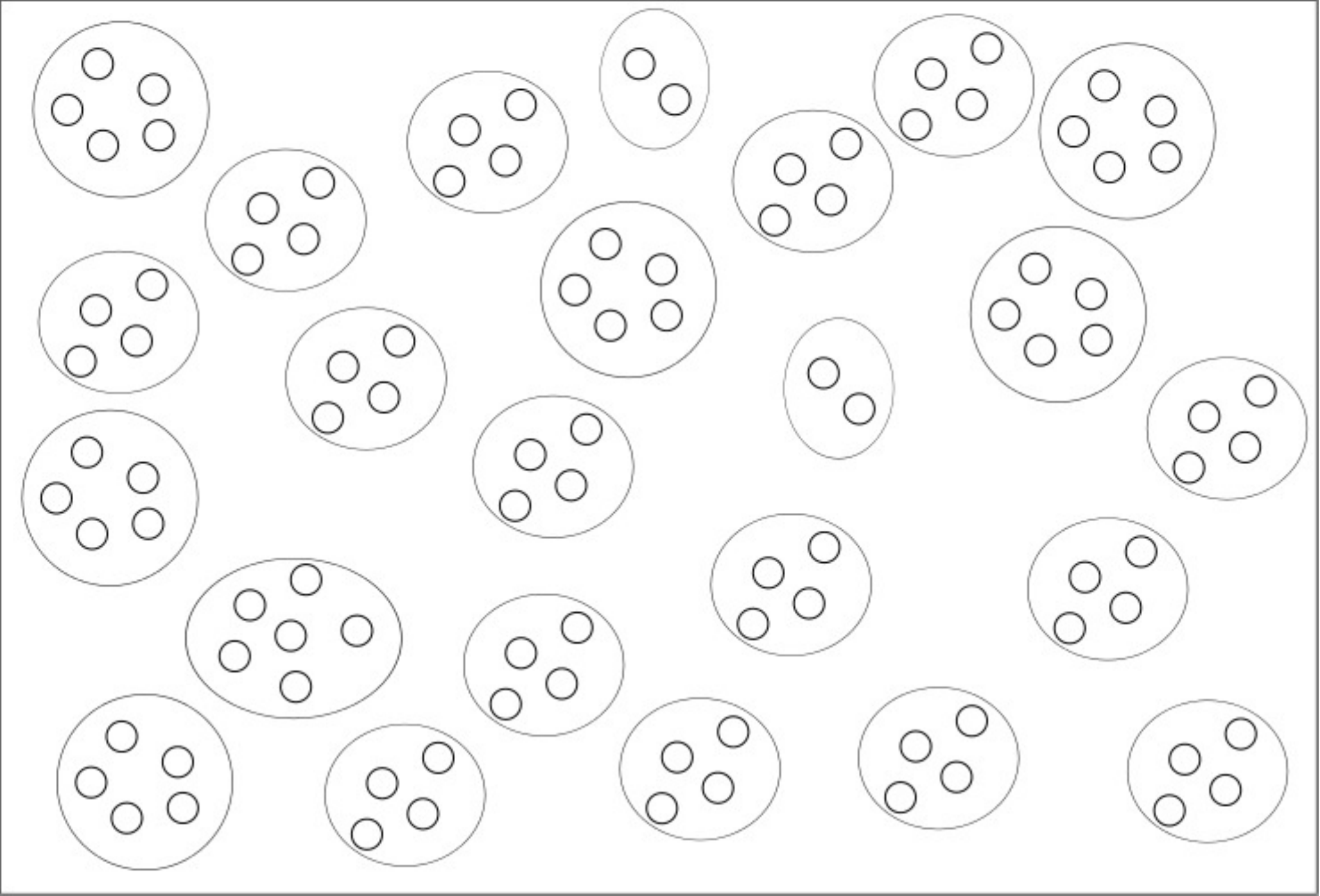}
  %
    \includegraphics[width=.3\textwidth]{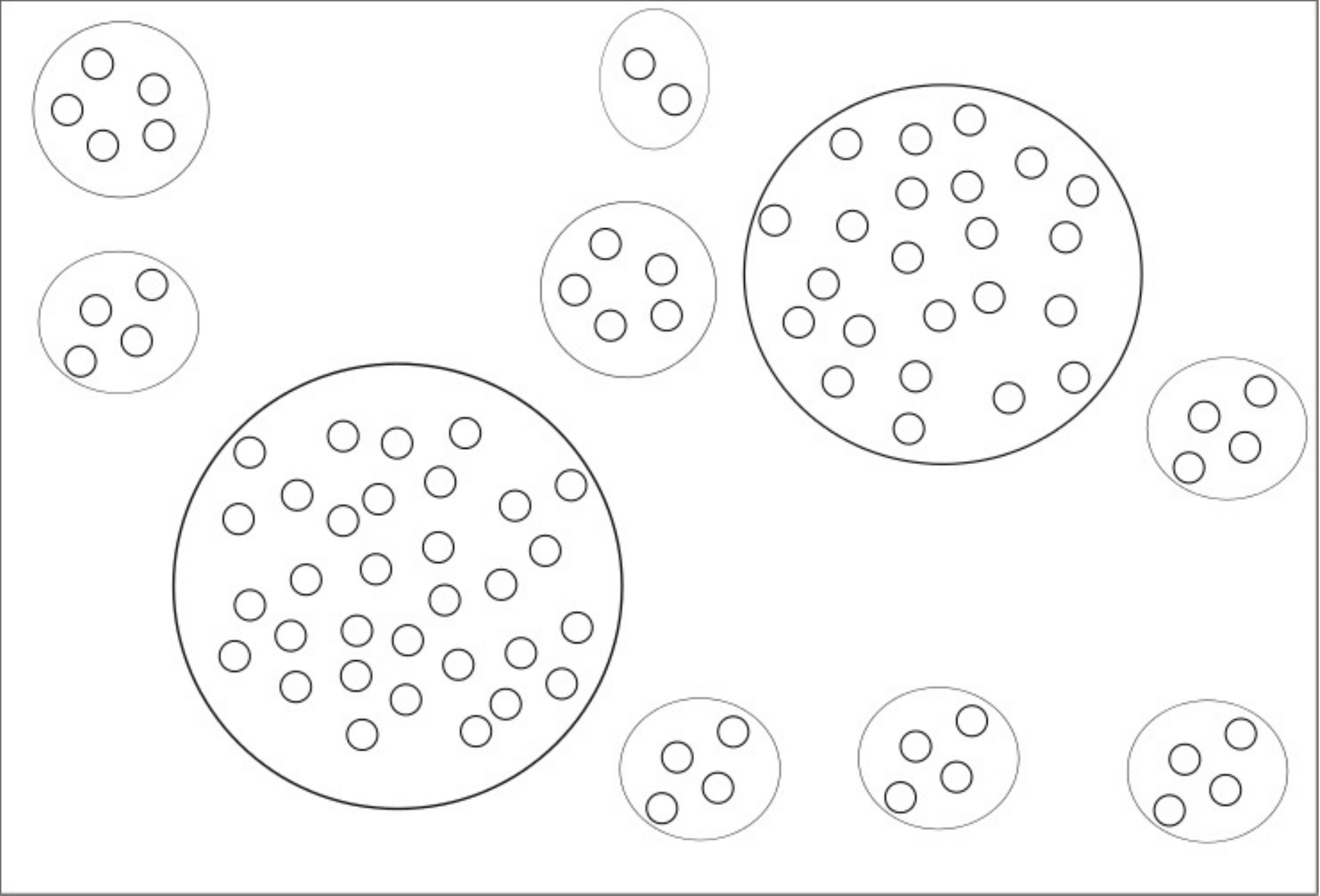}
}
  \caption[A schematic view of phase transitions in rCSPs]{A 2-dimensional schematic view of how the set of
      solutions of CSP varies as we increase the control parameter
      $\alpha$ from (left)~replica symmetric phase to (middle)~clustering
      phase to (right)~condensation phase.  Here small circles represent
      solutions and the bigger circles represent clusters of
      solutions.   Note that this view is very simplistic in many ways
      -- \eg the total number of solutions and the size of clusters
      should generally decrease from left to right.}  
  \label{fig:schematic}
    \label{fig:RS}
\label{fig:dynamical}
\label{fig:condensation}
\end{figure}

While there are tight bounds
for some problems \cite{achlioptas_rigorous_2005}, finding
the exact location of this transition for different CSPs is still an
open problem.  Besides transition to unsatisfiability, these analyses
have revealed several other (phase) transitions
\cite{krzakala_gibbs_2007}.  
\RefFigure{fig:schematic}(a)-(c)
shows how the geometry of the set of solutions changes by increasing
the control parameter.

Here we enumerate various phases of the problem for increasing values 
\marnote{replica symmetric phase}
\index{replica symmetric}
of the control parameter: \magn{(a)} In the so-called \magn{Replica 
  Symmetric} (RS) phase, the symmetries of the set of solutions (\aka
ground states) reflect the trivial symmetries of problem wrt variable
domains. For example, for $K$-COL the set of solutions is symmetric
wrt swapping all red and blue assignment.  In this regime, the set of \marnote{clusters of solution}
\index{clusters of solution}
solutions form a \magn{giant cluster} (\ie a set of neighboring solutions),
where two solutions are considered neighbors when their Hamming
distance is one \cite{achlioptas_algorithmic_2008} (or
non-divergent with number of variables
\cite{mezard_cavity_2002}.  Local search methods (\eg \cite{selman1994noise}) and BP-dec can often
efficiently solve random CSPs that belong to this phase.

\marnote{clustering phase}
\index{clustering phase}
\magn{(b)} In \magn{clustering} or \magn{dynamical} transition 
(1dRSB\footnote{1st order dynamical
  RSB.  Symmetry Breaking is a general term indicating a phenomenon
   during which a system is breaking the symmetry that governs its
   behaviour by selecting a particular branch.
  The term Replica Symmetry Breaking (RSB) originates from the
  technique --\ie Replica trick (\cite{Mezard1987})-- that was
  first used to analyze this setting. According to RSB, the trivial
  symmetries of the problem do not characterize the clusters of
  solution.}), the set of solutions decomposes into an exponential
number of distant clusters.  Here two clusters are distant if the
Hamming distance between their respective members is divergent (\eg
linear) in the number of variables.  
\magn{(c)} In the \magn{condensation}
\index{condensation phase}
phase transition (1sRSB\footnote{1st order static RSB.}), the set of
solutions condenses into a few dominant clusters. Dominant clusters
\marnote{condensation phase}
have roughly the same number of solutions and they collectively
contain almost all of the solutions. While SP can be used even within
the condensation phase, BP usually fails to converge in this regime.
However each cluster of solutions in the clustering and condensation
phase is a valid fixed-point of BP.  
\index{rigidity phase}
\magn{(d)} A \magn{rigidity} transition (not
\marnote{rigidity phase}
included in \refFigure{fig:schematic}) identifies a phase in which a
finite portion of variables are fixed within dominant clusters.  This
transition triggers an exponential decrease in the total number of
solutions, which leads to \magn{(e)} unsatisfiability transition.\footnote{In
  some problems, the rigidity transition occurs before condensation
  transition.}  
\marnote{1RSB}
\index{1RSB}
This rough picture summarizes first order Replica 
Symmetry Breaking's (1RSB) basic assumptions \cite{Mezard09}.

\subsection{Pitfalls of decimation}
Previously we gave an argument against decimation, based on the complexity of marginalization and integration. 
Some recent analyses draw similarly negative conclusions on the effect of decimation
\cite{coja-oghlan_belief_2010,montanari_solving_2007,ricci-tersenghi_cavity_2009}.
The general picture is that at some
point during the decimation process, variables form long-range
correlations such that fixing one variable may imply an assignment for
a portion of variables that form a loop, potentially leading to
contradictions. Alternatively the same long-range correlations result
in BP's lack of convergence and error in marginals that may lead to
unsatisfying assignments.

\begin{figure}
\centering
\includegraphics[width=.4\textwidth]{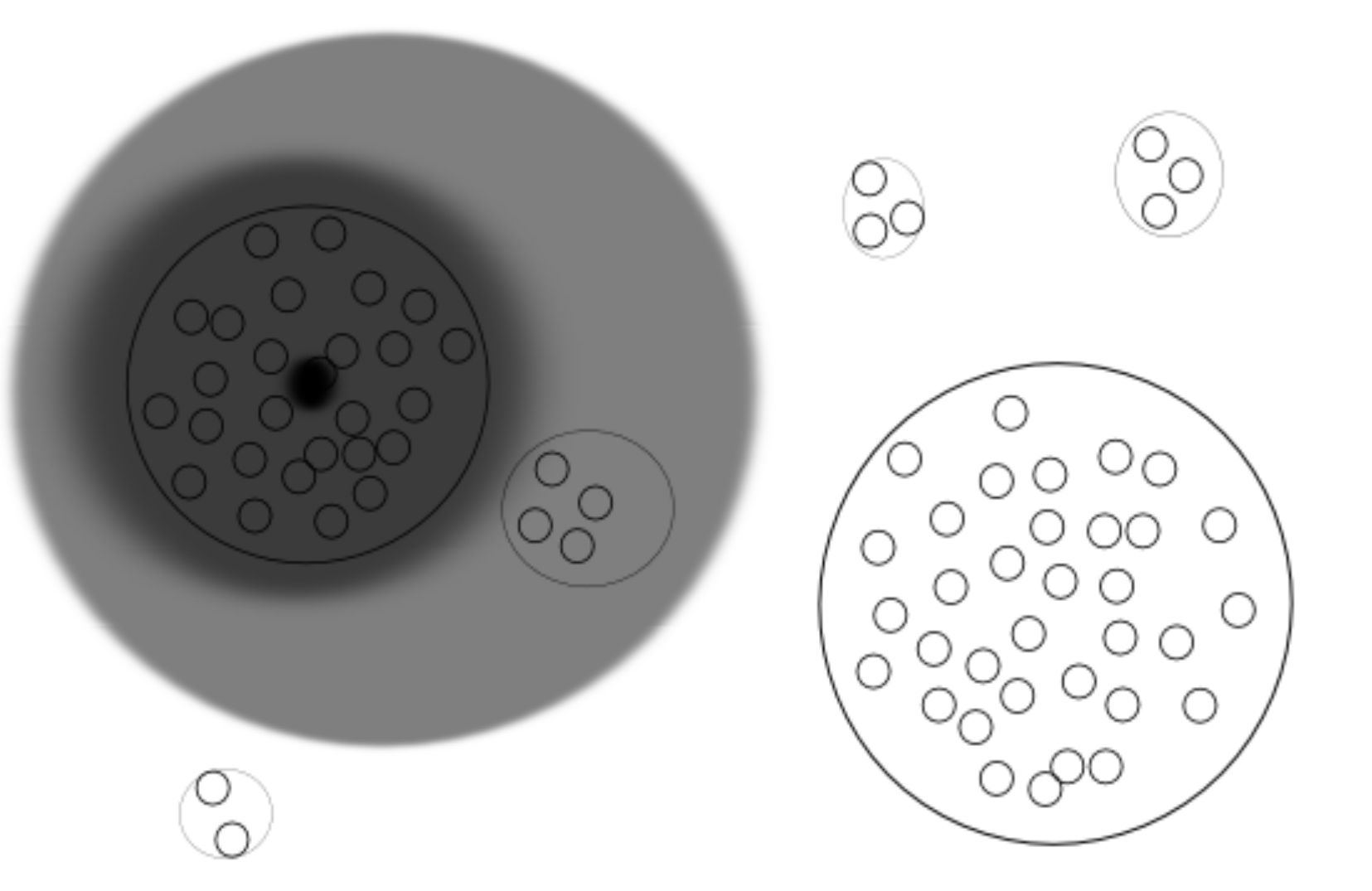}
\caption[Condensation phase and perturbed BP's behaviour.]{This schematic view demonstrates the clustering
      during condensation phase.  Here assume $x$ and $y$ axes
      correspond to $\xx_i$ and $\xx_j$.  Considering the whole space of
      assignments, $\xx_i$ and $\xx_j$ are highly correlated. The formation of this correlation between distant variables on a factor-graph breaks BP. Now
      assume that Perturbed BP messages are focused on the largest
      shaded ellipse.  In this case the correlation is significantly
      reduced.}\label{fig:1sRSB}
\end{figure}

Perturbed BP avoids the pitfalls of BP-dec in two ways: (I) Since many
configurations have non-zero probability until the final iteration,
perturbed BP can avoid contradictions by adapting to the most recent
choices. This is in contrast to decimation in which variables are \marnote{backtracking search}
fixed once and are unable to change afterwards. Some backtracking schemes 
\cite{parisi_backtracking_2033} attempt to fix this problem with decimation.  
(II) We speculate that simultaneous bias of
all messages towards sub-regions, 
prevents the formation of long-range correlations between
variables that breaks BP in 1sRSB; see \refFigure{fig:1sRSB}.

\section{Revisiting survey propagation}\label{sec:sp_revisit}

SP is studied on random (hyper) graphs representing CSPs at thermodynamic limit (\ie as $N \to \infty$).
Large random graphs are locally tree-like, which means the length of short \marnote{correlation decay}
\index{correlation decay}
loops are typically in the order of $\log(N)$
\cite{janson_random_2001}.  This ensures that, in the absence of
long-range correlations, BP is asymptotically exact, as the set of
messages incoming to each node or factor are almost independent.
Although BP messages remain uncorrelated until the condensation
transition \cite{krzakala_gibbs_2007}, the BP equations do not
completely characterize the set of solutions after the clustering
transition.  This inadequacy is indicated by the existence of a set of
\index{fixed point}
several valid fixed points (rather than a unique fixed-point) for WP as an instance of BP.  
For a better intuition,
consider the cartoons of~\refFigure{fig:schematic}(middle) and
(right). During the clustering phase (middle), $\xx_i$ and $\xx_j$ (corresponding to the $x$ and $y$ axes) 
are not highly
correlated, but they become correlated during and after condensation
(right). This correlation between variables that are far apart in the factor-graph
results in correlation between BP messages. This is because it implies that even if loops are
long, they remain influential.
This violates  
 BP's assumption that messages are uncorrelated, which results in BP's failure in this regime.

This is where survey propagation comes into the picture in solving CSPs. 
Going back to our algebraic notation for SP,
using counting SP with warning propagation semiring 
$(\{0, 1\}, \max, \prodop)$ as the initial semiring
and sum-product $(\Re^{\geq 0},  \sumop,  \prodop)$ as the SP semiring, is computationally tractable. 
This is because $\RR = \{0,1\}$ in the initial semiring is finite, and therefore each message
can have finite number of  $2^{\vert \XX_i\vert}$ values
{\small
\begin{align*}
 \msg{i}{\II}(\xx_i), \msg{\II}{i}(.) \in \big \{ (0,\ldots,0), (0,\ldots,0,1),(0,\ldots,1,0), \ldots,(1,\ldots,0),\ldots,(1,\ldots,1) \big \} \quad \forall i, \II \in \nb i
\end{align*}
}%

This means each SP message is a distribution over these possibilities $\msp{i}{\II}(\msg{i}{\II}) \in \Re^{2^{\vert \XX_i\vert}}$.
However since $(0,\ldots,0)$ indicates an unfeasible case, where no assignment is allowed, we explicitly ignore it in SP message updates.
This gives us the following update equations and marginals for SP when applied to CSPs
\index{SP!CSP}
\marnote{SP for CSPs}
\begin{align}
  \msp{i}{\II}(\msg{i}{\II}) \; &\propto \; \sum_{\msgss{\nb i \back \II}{i}}  
\ident\big ( \msg{i}{\II}(.) = \prod_{\JJ \in \nb i \back \II}\msg{\JJ}{i}(.) \big) 
\big ( \prod_{\JJ \in \nb i \back \II} \msp{\JJ}{i}(\msg{\JJ}{i}) \big ) \quad \forall i, \II \in \nb i \label{eq:spiI_csp}\\
  \msp{\II}{i}(\msg{\II}{i}) \; &\propto \; \sum_{\msgss{\nb \II \back i}{\II}} 
\ident \big ( \msg{\II}{i}(.) =  \sum_{\xx_{\back i}} \ff_{\II}(\xs_\II) \prod_{j \in \nb \II \back i}\msg{j}{\II}(.) \big)
\big ( \prod_{j \in \nb \II \back i} \msp{j}{\II}(\msg{j}{\II}) \big ) \label{eq:spIi_csp}\\
  \PSP(\ph_i) \; &= \; \sum_{\msgss{\nb i}{i}} \identt{\spsign}(\ph_i(.) = \prod_{\II \in \nb i} \msg{\II}{i}(.)) \prod_{\II \in \nb i} \msp{\II}{i}(\msg{\II}{i}) \label{eq:sp_marg_csp}\\
&\msp{\II}{i}((0,\ldots,0)) \; = \; 0 \quad \text{and} \quad \msp{i}{\II}((0,\ldots,0)) \; = \; 0\notag
\end{align}

\begin{example}
Consider the SP message $\msp{i}{\II}(\msg{i}{\II})$ in factor-graph of \pcref{fig:simple3sat}.
Here the summation in \cref{eq:spiI_csp} is over all possible combinations of max-product BP messages
 $\msg{\JJ}{i}$$\msg{\KK}{i}$,$\msg{\LL}{i}$,$\msg{\HH}{i}$. Since each of these messages can assume one of the
three valid values -- \eg $\msg{\JJ}{i}(\xx_i) \in \{(0,1),(1,0),(1, 1) \}$ -- 
for each particular assignment of $\msg{i}{\II}$, 
a total of $3^{4}$ 
possible combinations are enumerated in the summations of \cref{eq:spiI_csp}.
However only the combinations that form a valid max-product message update have non-zero
contribution in calculating $\msp{i}{\II}$.
\end{example}

\subsection{Flavours of SP-guided decimation}\label{sec:flavours}
The SP-marginal over WP marginals (\refEq{eq:sp_marg_csp}) also implies a distribution $\PSP(\xx_i)$ over the original
domain (see \refEq{eq:sp_implicit_marg}).
 Similar to BP-dec we can use either the implicit marginals of \cref{eq:sp_implicit_marg} or the SP marginals of \cref{eq:sp_marg_csp}  \marnote{SP-dec(S) \& SP-dec(C)}
to perform decimation. In the former case, which we call SP-dec(S) we select $\xx^*_i = \arg_{\xx_i} \max \PSP(\xx_i)$ during decimation, and in the later case, 
which we call SP-dec(C), we clamp $\ph^*_i \; =\; \arg_{\ph_i} \max \PSP(\ph_i) $. This means all the outgoing messages from this variable node in the factor-graph
are clamped in the same way -- \ie 
$\msp{i}{\II}(\msg{i}{\II}) = \ph^*_i \quad \forall \II \in \nb i$.

In the first case, SP-dec(S), we expect a single assignment $\xs^*$, while for SP-dec(C) at the end of decimation we should 
obtain a cluster of solutions, where a subset of assignments is allowed for each $\xx_i$.
However, during the decimation process (in both SP-dec(S) and SP-dec(C) ), usually after fixing a subset of variables,
SP marginals, $\PSP(\xx_i)$, become close to uniform, indicating that clusters of solution
have no preference over particular assignment of the remaining variables.
The same happens when we apply SP to random instances in
 RS phase (\refFigure{fig:schematic}(left)). At this point (\aka paramagnetic phase) \marnote{paramagnetic phase} 
solutions form a giant cluster and a local search method or BP-dec can often efficiently find an assignment to the 
variables that are not yet fixed by decimation. 

The original decimation procedure
for $K$-SAT \cite{braunstein_survey_2002} corresponds to  SP-dec(S).  SP-dec(C) for  
 CSP with Boolean variables is only slightly different, as SP-dec(C) can choose to fix a cluster to 
$\ph_i = ( 1, 1)$ in addition to the options of 
$\ph_i = (1,0)$ and $\ph_i = (0,1)$ (corresponding to $\xx_i = 0$ and $\xx_i = 1$ respectively), available to SP-dec(S). 
However, for larger domains (\eg $K$-COL), 
SP-dec(C) has a clear advantage. For example, 
in $3$-COL, SP-dec(C) may choose to fix a variable to $\msg{i}{\II} = (0,1,1)$ (\ie the first color is not allowed) while SP-dec(S) can only choose
between $\ph_i \in \{ (0,0,1),(0,1,0), (1,0,0)\}$. This significant difference is also reflected in their comparative success-rate on $K$-COL.
\footnote{Previous applications of SP-dec to $K$-COL by \cite{braunstein2003polynomial} used a heuristic for decimation that is similar SP-dec (C).} 
(See \refSection{sec:sat_col})

\subsection{Computational Complexity}
\index{SP!complexity}
\index{complexity!SP}
The computational complexity of each SP update of \refEq{eq:spIi_csp} is
$\OO(2^{|\XX_i|} - 1)^{\vert\nb \II\vert}$ as for each particular value $\msg{i}{\II}$, SP needs to consider
every combination of incoming messages, each of which can take
$2^{\vert \XX_i \vert}$ values (minus the empty set). Similarly, using a naive approach the cost of update of \refEq{eq:spiI_csp} is
$\OO(2^{\vert \XX_i \vert} - 1)^{\vert \nb i \vert}$. However by considering incoming messages one at a time, 
we can perform the same exact update in $\OO(\vert \nb i \vert \; 2^{2  \vert \XX_i \vert})$.
In comparison to the cost of
BP updates (\ie $\OO(\vert \nb i \vert\; \vert
\XX_i\vert)$ and $\OO(\vert \XX_{\II}\vert)$ for two types of message update; see \refSection{sec:bp}), 
we see that SP updates are substantially more expensive
for large domains $\vert \XX_i \vert$ and higher order factors with large $\vert \nb \II \vert$.

\subsection{Perturbed survey propagation for CSP}
Similar to SP, we use perturbed SP with $(\{0,1\}, \max,  \prodop)$ as the first semiring
and $(\Re,  \sumop,  \prodop)$ as the second semiring. 
\marnote{perturbed SP \& non-random CSP}
Since perturbed SP seeks a single assignment, rather than a cluster
of solutions, it can find satisfying solutions to paramagnetic instances.
This is in contrast to SP-dec, which in paramagnetic cases returns a trivial WP fixed 
point in which all assignment are allowed.
This means as opposed to SP-dec, which is mostly applied to random CSPs in the
clustering and condensation phase,
perturbed SP can be used to solve non-random and also random instances in RS phase.

To demonstrate this, we applied perturbed SP to benchmark CSP instances of 
\refFigure{fig:benchmark}, in which the maximum number of elements in the factor was less than $10$.\footnote{The number of iterations and other settings 
for perturbed SP were identical 
to the ones used to compare BP-dec and perturbed BP.}
Here perturbed SP solved $80$ instances out of $202$ cases, in comparison to perturbed BP that solved $78$ instances, making perturbed SP slightly better, also in solving real-world problems.

%% file: sat.tex

In \cref{example:sat_generate,example:col_generate}, we introduced the random procedures that are often 
used to produce instances of $K$-satisfiability and $K$-coloring problems.

\begin{figure}
  \centering
\hbox{
    \includegraphics[width=.507\textwidth]{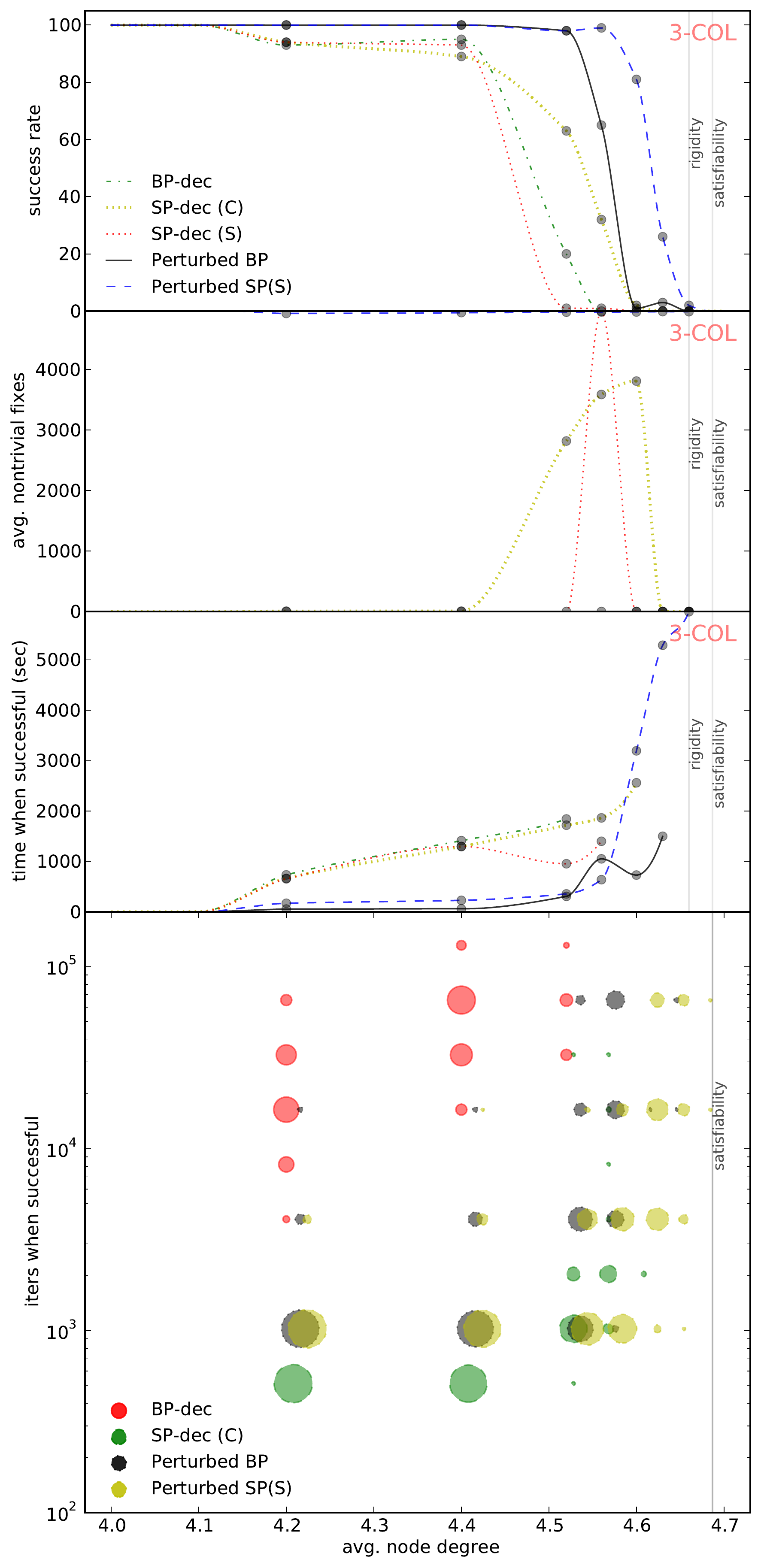}
    \includegraphics[width=.5\textwidth]{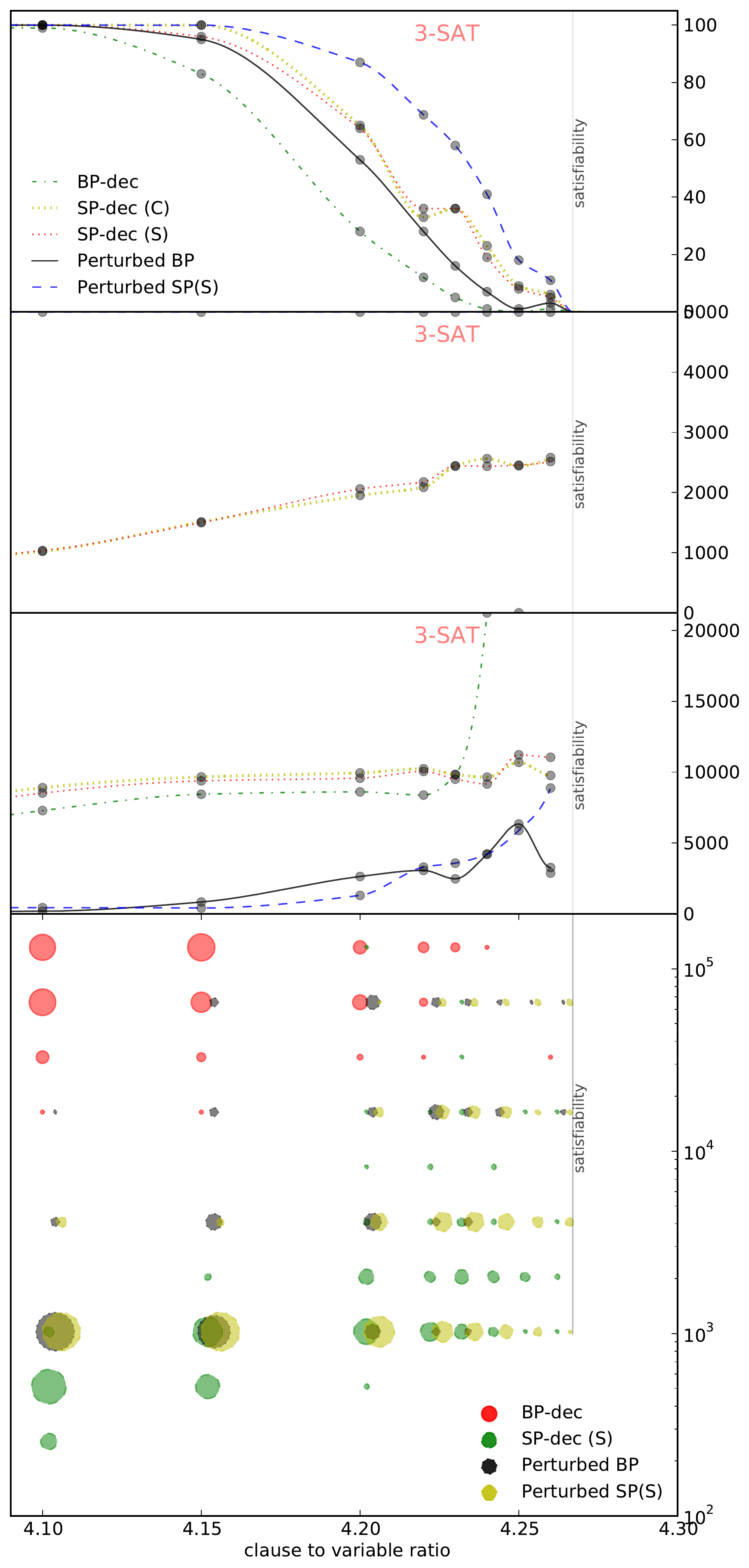}
}
\caption[comparison of different methods on K-satisfiability and K-coloring]{\textbf{(first row)}~Success-rate of different methods for 3-COL and 3-SAT for various control parameters.
\textbf{(second row)}~The average number of variables (out of $N = 5000$) that are fixed using SP-dec (C) and (S) before calling local search, averaged over 100 instances.
\textbf{(third row)}~The average amount of time (in seconds) used by the successful setting of each method to find a satisfying solution. For SP-dec(C) and (S) this includes the time used by local search. 
\textbf{(forth row)}~The number of iterations used by different methods at different control parameters, when the method
was successful at finding a solution. The number of iterations
for each of 100 random instances is rounded to the closest power of 2. This does not include the iterations used by local search
after SP-dec.
}
  \label{fig:all3sat3col}
\end{figure}




Here we report the results on $K$-SAT for $K \in \{3,4\}$
and $K$-COL for $K \in \{3,4,9\}$.  We used the procedures to
 produce $100$ random
instances with $N = 5,000$ variables for each control parameter
$\alpha$ and here report the probability of finding a satisfying assignment
for different methods -- \ie the portion of $100$ instances that were satisfied by each method.\footnote{For coloring instances, to help 
decimation, we break the initial symmetry 
of the problem by fixing a single variable to an arbitrary value.
For BP-dec and SP-dec, we use a convergence threshold of $\epsilon= .001$ and 
fix $\rho = 1\%$ of variables per iteration of decimation.
Perturbed BP and Perturbed SP use $T = 1000$ iterations. Decimation-based methods use a maximum of $T = 1000$ iterations per iteration of decimation.
If any of the methods failed to find a solution in the first attempt, $T$ was increased by a factor of $4$ at most $3$ times -- \ie in the final 
attempt $T = 64,000$. To avoid blow-up in run-time, for BP-dec and SP-dec, only the maximum iteration, $T$, during the first iteration of decimation, was increased  
(this is similar to the setting of \cite{braunstein_survey_2002} for SP-dec).
For both variations of SP-dec (see \refSection{sec:flavours}) after each decimation step, if $\max_{i,x_i} \ph(\xx_i) - \frac{1}{\vert\XX_i\vert} < .01$ (\ie marginals are close to uniform) 
we consider the instance para-magnetic,
and run BP-dec (with $T = 1000$, $\epsilon = .001$ and $\rho = 1\%$) on the simplified instance.
}

\RefFigure{fig:all3sat3col}(first row) visualizes the success rate of different methods on 3-SAT (right) and 3-COL (left).
\refFigure{fig:all3sat3col}(second row) reports the number of variables that are fixed by SP-dec(C) and (S) before calling BP-dec as local search.
The third row shows the average amount of time that is used to find a satisfying solution. This does not include the failed attempts. For SP-dec variations, this time includes the time used by local search.
The final row of \refFigure{fig:all3sat3col} shows the number of iterations used by each
method at each level of difficulty over the successful instances. Note that this does not
include the iterations of local search for SP-dec variations.
Here the area of each disk is proportional
to the frequency of satisfied instances with particular number of iterations for each control parameter and inference method\footnote
{The number of iterations are rounded to the closest power of two.}.

Here we make the following observations:

(I) Perturbed BP is much more effective than BP-dec, while remaining ten to hundreds of time 
 more efficient.
(II) As the control parameter grows larger,
the chance of requiring more iterations to satisfy the instance increases for all methods.
(III) Although computationally very inefficient, BP-dec is able to find solutions for instances 
with larger control parameter than suggested by previous results (\eg \cite{Mezard09}).
(IV) For many instances where SP-dec(C) and (S) use few iterations, 
the variables are fixed to a trivial cluster 
$\ph_i = (1,1,\ldots,1)$, in which all assignments are allowed. This is particularly pronounced for 3-COL. 
For instances in which non-trivial fixes are zero, the success rate is solely due to local search (\ie BP-dec). 
(V) While SP-dec(C) and SP-dec(S) have a similar performance for 3-SAT, SP-dec(C) significantly 
outperforms SP-dec(S) for 3-COL.

\RefTable{table:rcsp} reports the
success-rate as well as the average of total iterations in the
\emph{successful} attempts of each method, where the number of iterations
for SP-dec(C) and (S) 
is the sum of iterations 
used by the method plus the iterations of the following BP-dec.
Here we observe that perturbed BP can solve most of easier instances using only $T=1000$ iterations (\eg see perturb BP's result for 3-SAT at $\alpha = 4.$, 3-COL at $\alpha = 4.2$ and 9-COL at $\alpha = 33.4$). 
The results also show that 
most difficult instances (that require more time/iterations) for each method approximately correspond to the control parameter
for which half of the instances are satisfied. Larger control parameters usually result in early failure in satisfiability.


\Cref{table:rcsp} suggests that, as we speculated in \refSection{sec:sp_revisit}, SP-dec(C) is in general preferable to SP-dec(S), in particular when applied to the coloring problem.
The most important advantage of Perturbed BP over SP-dec and Perturbed SP is that it can be applied to instances with large factor cardinality (\eg $10$-SAT) and
variable domains (\eg $9$-COL). For example for $9$-COL, the cardinality of each SP message is $2^9=512$, which makes SP-dec and Perturbed SP impractical.
Here BP-dec is not even able to solve a single instance around the dynamical transition (as low as $\alpha = 33.4$) while perturbed BP satisfies all instances up to $\alpha = 34.1$.\footnote{
Note that for $9$-COL condensation transition happens after rigidity transition. So if we were able to find solutions after
rigidity, it would have implied that condensation transition marks the onset of difficulty. However, this did not occur and similar to all other cases, Perturbed BP failed
before rigidity transition.}

\begin{table}[!pht]
  \caption[Comparison of different methods for K-satisfiability and K-coloring.]{Comparison of different methods on  \textbf{$\{3,4\}$-SAT} and \textbf{$\{3,4,9\}$-COL}.
For each method the success-rate and the average  number of iterations (including local search) on successful attempts are reported. 
The approximate location of phase transitions are from \cite{montanari_clusters_2008,zdeborova_phase_2007-4}
.}\label{table:rcsp}
\centering
\begin{tikzpicture}
\node (table) [inner sep=0pt] {
    \scalebox{.6}{
    \begin{tabu}{c  c |[2pt] r  l |[2pt] r  l |[2pt] r  l|[2pt] r  l|[2pt] r  l}
      \cline{3-12}
      & &\multicolumn{2}{ c |[2pt]}{BP-dec}& \multicolumn{2}{ c|[2pt] }{SP-dec(C)}&\multicolumn{2}{ c|[2pt] }{SP-dec(S)}&
      \multicolumn{2}{ c|[2pt] }{Perturbed BP}& \multicolumn{2}{ c }{Perturbed SP}\\
      \cline{3-12}
      \begin{sideways}Problem \end{sideways}&
      \begin{sideways}ctrl param $\alpha$\end{sideways}&
      \begin{sideways}avg. iters.\end{sideways} &
      \begin{sideways}success rate\end{sideways} &
      \begin{sideways}avg. iters.\end{sideways} &
      \begin{sideways}success rate\end{sideways} &
      \begin{sideways}avg. iters.\end{sideways} &
      \begin{sideways}success rate\end{sideways} &
      \begin{sideways}avg. iters.\end{sideways} &
      \begin{sideways}success rate\end{sideways} &
      \begin{sideways}avg. iters.\end{sideways} &
      \begin{sideways}success rate\end{sideways} 
      \\\tabucline[2pt]{-}
      \multirow{10}{*}{\textbf{3-SAT}}&
      3.86 & \multicolumn{10}{l}{dynamical and condensation transition}\\\cline{2-12}
&4.1 & 85405 & 99\% &102800 & 100\% &96475 & 100\% &1301 & 100\%  &1211 & 100\% 
\\\cline{2-12}
&4.15 & 104147 & 83\% &118852 & 100\% &111754 & 96\% &5643 & 95\%  &1121 & 100\%
\\\cline{2-12}
&4.2 & 93904 & 28\% &118288 & 65\% &113910 & 64\% &19227 & 53\%  &3415 & 87\%
\\\cline{2-12}
&4.22 & 100609 & 12\% &112910 & 33\% &114303 & 36\% &22430 & 28\%  &8413 & 69\%
\\\cline{2-12}
&4.23 & 123318 & 5\% &109659 & 36\% &107783 & 36\% &18438 & 16\%  &9173 & 58\%
\\\cline{2-12}
&4.24 & 165710 & 1\% &126794 & 23\% &118284 & 19\% &29715 & 7\%  &10147 & 41\%
\\\cline{2-12}
&4.25 & N/A & 0\% &123703 & 9\% &110584 & 8\% &64001 & 1\%  &14501 & 18\%
\\\cline{2-12}
&4.26 & 37396 & 1\% &83231 & 6\% &106363 & 5\% &32001 & 3\%  &22274 & 11\%
\\\cline{2-12}
      &4.268 & \multicolumn{10}{l}{satisfiability transition}
      \\\tabucline[2pt]{-}
      \multirow{7}{*}{\textbf{4-SAT}}&
      9.38 & \multicolumn{10}{l}{dynamical transition}\\\cline{2-12}
      &9.547& \multicolumn{10}{l}{condensation transition}\\\cline{2-12}
&9.73 & 134368 & 8\% &119483 & 32\% &120353 & 35\% &25001 & 43\%  &11142 & 86\% \\\cline{2-12}
&9.75 & 168633 & 5\% &115506 & 15\% &96391 & 21\% &36668 & 27\%  &9783 & 68\% \\\cline{2-12}
&9.78 & N/A & 0\% &83720 & 9\% &139412 & 7\% &34001 & 12\%  &11876 & 37\% \\\cline{2-12}
      &9.88& \multicolumn{10}{l}{ rigidity transition}\\\cline{2-12}
      &9.931& \multicolumn{10}{l}{ satisfiability transition}
      \\\tabucline[2pt]{-}
      \multirow{10}{*}{\textbf{3-COL}}&
      4 & \multicolumn{10}{l}{dynamical and condensation transition}\\\cline{2-12}
&4.2 & 24148 & 93\% &25066 & 94\% &24634 & 94\% &1511 & 100\%  &1151 & 100\%
\\\cline{2-12}
&4.4 & 51590 & 95\% &52684 & 89\% &54587 & 93\% &1691 & 100\%  &1421 & 100\%
\\\cline{2-12}
&4.52 & 61109 & 20\% &68189 & 63\% &54736 & 1\% &7705 & 98\%  &2134 & 98\%
\\\cline{2-12}
&4.56 & N/A & 0\% &63980 & 32\% &13317 & 1\% &28047 & 65\%  &3607 & 99\%
\\\cline{2-12}
&4.6 & N/A & 0\% &74550 & 2\% &N/A & 0\% &16001 & 1\%  &18075 & 81\%
\\\cline{2-12}
&4.63 & N/A & 0\% &N/A & 0\% &N/A & 0\% &48001 & 3\%  &29270 & 26\%
\\\cline{2-12}
     &4.66& \multicolumn{10}{l}{rigidity transition}\\\cline{2-12}
&4.66 & N/A & 0\% &N/A & 0\% &N/A & 0\% &N/A & 0\%  &40001 & 2\%
\\\cline{2-12}
      &4.687& \multicolumn{10}{l}{ satisfiability transition}
      \\\tabucline[2pt]{-}
      \multirow{6}{*}{\textbf{4-COL}}&
      8.353 & \multicolumn{10}{l}{dynamical  transition}\\\cline{2-12}
&8.4 & 64207 & 92\% &72359 & 88\% &71214 & 93\% &1931 & 100\% &1331 & 100\% \\\cline{2-12}
&8.46 & \multicolumn{10}{l}{dynamical  transition}\\\cline{2-12}
&8.55 & 77618 & 13\% &60802 & 13\% &62876 & 9\% &3041 & 100\%  &5577 & 100\% \\\cline{2-12}
&8.7 & N/A & 0\% &N/A & 0\% &N/A & 0\% &50287 & 14\% &N/A  & 0\% \\\cline{2-12}
      &8.83& \multicolumn{10}{l}{rigidity transition}\\\cline{2-12}
      &8.901& \multicolumn{10}{l}{ satisfiability transition}
      \\\tabucline[2pt]{-}
      \multirow{9}{*}{\textbf{9-COL}}&
33.45 & \multicolumn{10}{l}{dynamical  transition}\\\cline{2-12}
&33.4 & N/A & 0\%& N/A&N/A&N/A&N/A &1061 & 100\%&N/A&N/A \\\cline{2-12}
&33.9 & N/A & 0\%& N/A&N/A&N/A&N/A &3701 & 100\%&N/A&N/A \\\cline{2-12}
&34.1 & N/A & 0\%& N/A&N/A&N/A&N/A &12243 & 100\%&N/A&N/A \\\cline{2-12}
&34.5 & N/A & 0\%& N/A&N/A&N/A&N/A &48001 & 6\%&N/A&N/A \\\cline{2-12}
&35.0 & N/A & 0\%& N/A&N/A&N/A&N/A & N/A & 0\%&N/A&N/A \\\cline{2-12}
      &39.87& \multicolumn{10}{l}{rigidity transition}\\\cline{2-12}
      &43.08& \multicolumn{10}{l}{condensation transition}\\\cline{2-12}
      &43.37& \multicolumn{10}{l}{ satisfiability transition}
    \end{tabu}
}
};
\draw [rounded corners=.5em] (table.north west) rectangle (table.south east);
\end{tikzpicture}
\end{table}

%% file: other_csps.tex


\section{Clique-cover problem}\label{sec:clique-cover}
\index{clique-cover|see {K-clique-cover}}
\index{K-clique-cover}
The \magn{K-clique-cover} $\CC = \{\CC_1,\ldots,\CC_K\}$ for a \marnote{K-clique-cover}
graph $\GG = (\VV, \EE)$ is a partitioning of $\VV$ to
at most $K$ cliques -- \ie $\forall i,j,k\;\;\;\; i,j \in \CC_k \Rightarrow (i,j) \in \EE$.

\NP-completeness of K-clique-cover can be proved by reduction from K-coloring~\cite{karp1972reducibility}: 
A K-clique-cover for $\GG'$, the complement of $\GG$ (\ie $\GG' = (\VV, \EE' = \{(i,j) \mid (i,j) \notin \EE \})$),
is a K-coloring for $\GG$, where all the nodes in the same clique of $\GG'$
are allowed to have the same color in $\GG$.

The relation between K-clique-cover and K-coloring extends to their factor-graphs.
While in K-coloring, factors $\ff_{\{i,j\}}(\xx_i, \xx_j) = \ident(\xx_i \neq \xx_j) \quad \forall (i,j) \in \EE$ ensure that the connected nodes have different colors,
for k-clique-cover factors $\ff_{\{i,j\}}(\xx_i, \xx_j) = \ident(\xx_i \neq \xx_j) \quad \forall (i,j) \notin \EE$ ensure that nodes that are not connected \marnote{factor-graph}
can not belong to the same clique. Here $\xx_i \in \{1,\ldots,K\}$ represents the 
clique of node $i$.

The factors, $\ff_{\{i,j\}}(\xx_i, \xx_j) = \ident(\xx_i \neq \xx_j)$, in both K-clique-cover and K-coloirng 
are inverse Potts factors that allow efficient $\OO(K)$ calculation (see \refSection{sec:hop}). 
Using $\EE(i, \cdot) = \{ (i,j) \in \EE\}$ to denote
the set of edges adjacent to node $i$, the following claim states the complexity of BP updates.
\marnote{complexity}
\begin{claim}\label{th:K_clique_cover_cost}
Each iteration of BP with variable-synchronous message update for K-clique-cover factor-graph is $\OO(K (N^2 - \vert \EE \vert))$,
while asynchronous message update is $\OO(K \sum_{i \in \VV} ( N - \vert\EE(i,\cdot) \vert)^2)$.
\end{claim}
\begin{proof}
Here the complexity of calculating factor-to-variable message ($\msg{\{i,j\}}{i}$) is $\OO(K)$. Since there are 
$N^2 - \EE$ factors (one for each edge in $\GG'$) the total cost of factor-to-variable messages becomes $\OO(K (N^2 - \vert \EE \vert))$.

The time complexity of each variable-to-factor message ($\msg{i}{\{i,j\}}$) is $\OO(K \vert \mb i\vert )$, where $\mb i$, 
Markov blanket of $i$ in the factor-graph, is the set of nodes in $\VV$
that are \emph{not} adjacent to $i$ in $\GG$ -- \ie $\vert \mb i\vert =  N - \EE(i,\cdot)$. 
Using variable-synchronous update the total cost of variable-to-factor messages becomes $\OO(K \sum_{i \in \VV} N - \vert \EE(i,\cdot)\vert ) = \OO(K (N^2 - 2 \vert \EE \vert))$.
This means the cost of all messages in BP update is in the order of $\OO(K (N^2 - \vert \EE \vert))$.

However, using asynchronous update, at each node $i$, we have to calculate $N - \vert \EE(i,\cdot) \vert$ messages. 
Since each of them is $\OO(N - \vert \EE(i,\cdot) \vert)$, the total cost of variable-to-factor dominates the cost of BP update
which is $\OO(K \sum_{i \in \VV} ( N - \vert\EE(i,\cdot) \vert)^2)$. 
\end{proof}

Our experimental results for K-clique-cover are within the context of a binary-search scheme, as
the sum-product reduction of the min-max formulation of K-clustering.




\section{Dominating set  and  set cover}\label{sec:set-cover}
\index{K-dominating-set}
\index{dominating-set}
\index{set-cover}
\index{K-set-cover}
\marnote{K-dominating set}
The \magn{$K$-dominating set} of graph $\GG = (\VV, \EE)$ 
is a subset of nodes $\DD \subseteq \VV$ of size $\vert \DD \vert = K$ such that any node in
$\VV \back \DD$ is adjacent to at least one member of $\DD$ --
\ie $\forall i \in \VV \back \DD \quad \exists j\in \DD \;
s.t. \; (i,j) \in \EE$.
The dominating set problem is \NP-complete\ \cite{garey_computers_1979} and has 
simple reductions to and from set cover problem \cite{kann1992approximability}.
As we see, the factor-graph formulations of these problems are also closely related.

\begin{figure}[pth!]
    \centering
    \includegraphics[width=.7\linewidth]{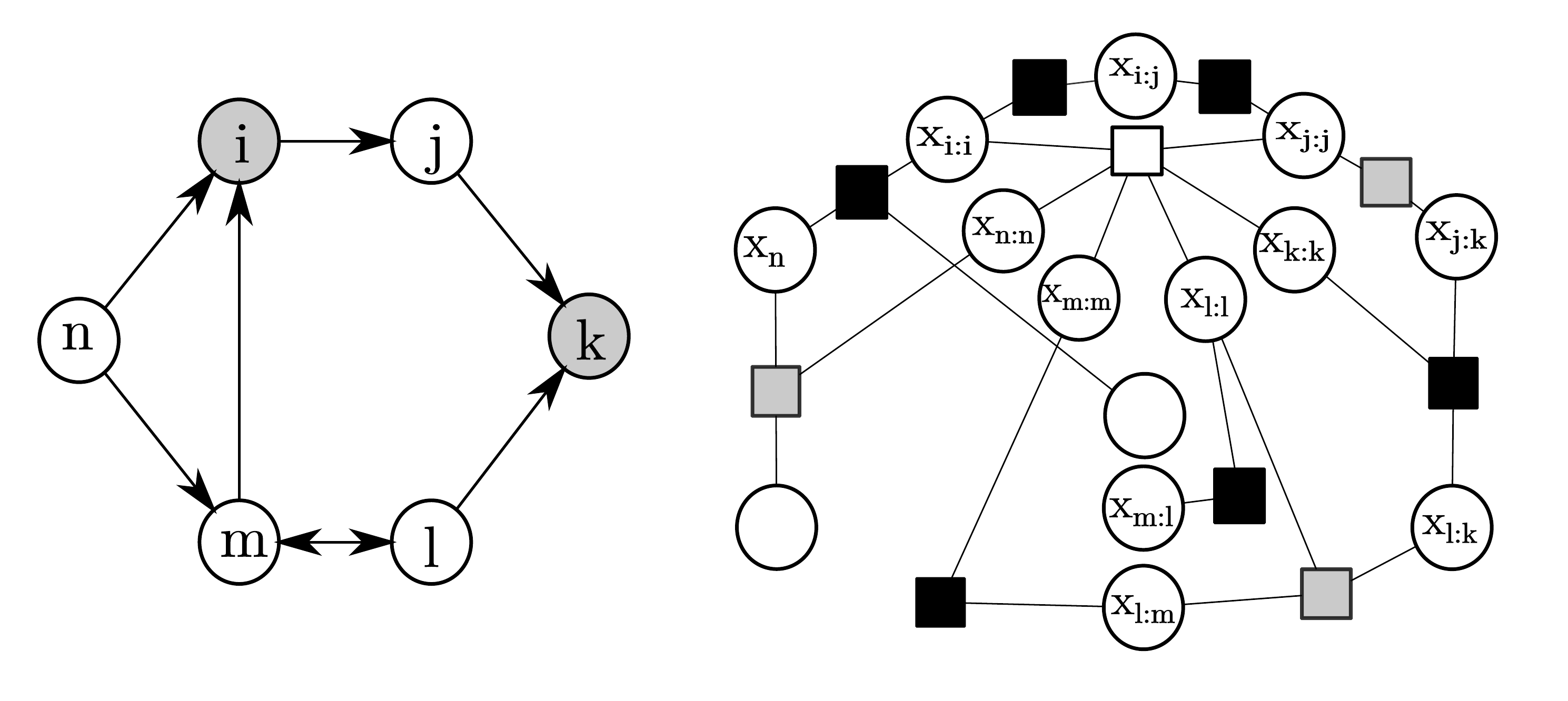}
\caption[An example of induced set cover and its factor-graph.]{\textbf{(left)} an induced 2-set-cover problem and the solution $\DD = \{ i, k\}$.
\textbf{(right)} The factor-graph representation of the same problem, where leader factors are grey squares,
consistency factors are in black and the K-of-N factor is in white.
}\label{fig:set_cover}
\end{figure}

Given universe set $\VV$ and a set of its subsets $\settype{S} = \{\VV_1,\ldots,
\VV_M \} \;s.t.\;\; \VV_m \subseteq \VV$, we say $\CC
\subseteq \settype{S}$ \magn{covers} $\VV$ iff each member of
$\VV$ is present in at least one member of $\CC$ --\ie
$\bigcup_{\VV_m \in \CC} \VV_m = \VV$.  
Now we
consider a natural set-cover problem induced by any \magn{directed} graph. \marnote{induced set cover}
\index{set-cover!induced}
Given a directed-graph $\GG = (\VV, \EE)$, for each node
$i \in \VV$, define a subset $\VV_i = \{ j \in \VV\; \mid \;
(j,i) \in \EE \}$ as the set of all nodes that are connected
\emph{to} $i$. Let $\settype{S} = \{\VV_1,\ldots,\VV_N\}$ denote
all such subsets.  An induced $K$-set-cover of $\GG$ is a
set $\CC \subseteq \settype{S}$ of size $K$ that covers $\VV$.
Equivalently,
The \magn{induced $K$-set-cover} of $\GG = (\VV, \EE)$ is a subset of vertices $\DD \subseteq \VV$, with 
$\vert \DD \vert  = K$,
such that every node not in $\DD$ is connected to at least one node in $\DD$.
For example in \refFigure{fig:set_cover}(left), $\settype{S} = \{\{n,m,i\},\{i,j\},\{j,k,l\},\{l,m,n\},\{n\}\}$
and its induced solution $\CC = \{\{n,m,i\},\{j,k,l\}\}$ is indicated by grey nodes $\DD = \{i,k\}$.

\marnote{set-cover $Leftrightarrow$ dominating set}
If we consider an undirected graph $\GG$ as a directed graph with edges in both directions,
then K-dominating set of $\GG$ is equivalent to an induced K-set-cover problem on $\GG$.
Moreover given any K-set-cover problem instance $\settype{S} = \{\VV_1,\ldots,\VV_m\}$, we can construct a directed graph $\GG$ such that the ``induced'' K-set-cover on $\GG$ is equivalent to the given K-set-cover problem. 
For this, let $\VV = (\bigcup_{\VV_m \in \settype{S}} \VV_m ) \cup \{ u_1,\ldots,u_M\}$ be the collection of nodes in $\settype{S}$ plus one node $u_m$ per each subset $\VV_m \in \settype{S}$.
Now define the directed edges in $\EE$ to connect every $i \in \VV_m$ to its representative $u_m$. Moreover connect all representatives to each other in both
directions -- \ie $\EE = \{ (i, u_m)  \mid \forall m, i \in \VV_m\} \cup \{ (u_{m}, u_{m'}) \mid \forall m \neq m'\}$.
It is easy to show that the induced K-set-cover on this directed graph defines a set-cover for $\settype{S}$.



\subsection{Factor-graph and complexity}
For both problems we have one variable per edge $\xxx{i}{j} \in \{0,1\} \;\forall (i,j) \in \EE$.
Note that the $\GG$ for induced K-set-cover problem is a directed graph, while the $\GG$
for the K-dominating-set is undirected. 
This is the only difference that affects the factor-graph
representation of these two problems.
Here, $\xxx{i}{j} = 1$ indicates that node $j \in \DD$ and node $i$ is associated with node $j$.
Three types of constraint factors ensure the assignments to $\xxx{i}{j}$ 
define a valid solution to K-dominating-set and induced K-set-cover:  


\index{factor!leader}
\marnote{leader factors}
\noindent \bullitem \magn{Leader factors} ensure that each node $i$ is associated with at least one node $j$ (where $j = i$ is admissible).
Let $\EE^+(i, \cdot) = \{(i,j) \in \EE \} \cup \{(i,i)\}$ be the set of edges leaving node $i$ plus $(i,i)$.
Then 
\begin{align}
\ff_{\EE^+(i, \cdot)}(\xs_{\EE^+(i, \cdot)}) = \ident((\sum_{(i,j) \in \EE^+(i, \cdot)} \xxx{i}{j}) \geq 1) \quad \forall i \in \VV
\end{align}
is the leader factor associated with node $i$. 
\\

\index{factor!consistency}
\noindent \bullitem \magn{Consistency factors}\marnote{consistency factors} ensure that if node $j$ is 
selected as the leader by node $i$, node $j$ also selects itself as leader:
\begin{align}
\ff_{\{i:j,j:j\}}(\xxx{i}{j}, \xxx{j}{j}) = \ident(\xxx{i}{j} = 0 \vee \xxx{j}{j} = 0) \quad  \forall (i,j) \in \EE
\end{align}

An alternative form of this factor is a high-order factor that allows efficient $\OO(\vert \EE(\cdot, i) \vert)$ factor-to-variable update
\begin{align}\label{eq:consistency_hop}
\ff_{\EE(\cdot, i)}(\xs_{\EE(\cdot, i)}) = \ident(\xxx{i}{i} = 1 \vee \sum_{(i,j) \in \EE(\cdot, i)} \xxx{j}{i} = 0) \quad \forall i \in \VV
\end{align}
\index{factor!K-of-N}
\marnote{at most K-of-N factors}
\noindent \bullitem \magn{At most K-of-N factor} ensures that at most 
$K$ nodes are selected as leaders ($\vert \DD \vert \leq K$):
\begin{align}
\ff_{\{i:i,j:j,\ldots,l:l\}}(\xs_{\{i:i,j:j,\ldots,l:l\}}) = \ident(\sum_{i \in \VV} \xxx{i}{i} \leq  K) 
\end{align}




\RefFigure{fig:set_cover} shows an example of induced K-set-cover problem and its corresponding factor-graph.
In \refSection{sec:hop}
we saw that it is possible to calculate sum-product factor-to-variable BP messages for leader factors
in $\OO(\vert \EE(i, \cdot) \vert)$
and each at-most-K-of-N factor in $\OO(K N)$. 
This cost for consistency factors is $\OO(1)$ for the pairwise and $\OO(\vert \EE(\cdot, i) \vert)$ for the alternative formulation.

\marnote{complexity}
\begin{claim}\label{th:set_cover_cost}
The time-complexity of message passing for the factor-graph above depending on the update schedule is 
$\OO(\vert \EE \vert + K N)$ for factor-synchronous ($\ff$-sync; see \refSection{sec:bpcomplexity}) update and 
$\OO(K N^2 + \sum_{i \in \VV} \vert \EE(i,\cdot) \vert^2 + \vert \EE(\cdot,i) \vert^2)$ for asynchronous update.
\end{claim}
\begin{proof}
We assume the consistency factors are in the higher order form of \refEq{eq:consistency_hop}.
Here, each variable $\xxx{i}{j}$ is connected to at most three factors and therefore the cost 
of variable-to-factor messages is $\OO(1)$. If we calculate factor-to-variable messages
simultaneously, the cost is $\OO(\sum_{i \in \VV} \vert \EE(i,\cdot) \vert )$ for leader and $\OO(\sum_{i \in \VV} \vert \EE(\cdot,i) \vert )$,
giving a total of $\OO(\vert \EE \vert )$. Adding this to the cost of K-of-N factor the total cost per iteration of 
BP is $\OO(\vert \EE \vert + K N)$.

On the other hand, if we update each factor-to-variable separately, the previous costs are multiplied by $\vert \nb \II \vert$,
which gives $\OO(K N^2 + \sum_{i \in \VV} \vert \EE(i,\cdot) \vert^2 + \vert \EE(\cdot,i) \vert^2)$.
\end{proof}




\section{Clique problem, independent set and sphere packing}\label{sec:independent-set}
\index{clique problem}
\index{K-clique problem}
\index{independent-set}
\index{K-independent-set}

Given \marnote{K-clique problem}
graph $\GG$, the \magn{K-clique problem} asks whether $\GG$ contains
a clique of size at least $K$.
The K-clique problem is closely related to \magn{K-independent-set}.
Given graph $\GG = (\VV, \EE)$, \marnote{K-independent-set}
the K-independent set problem asks whether $\VV$
contains a subset of size at least $K$, s.t. there is no connection between nodes in $\DD$ -- 
\ie $\forall i,j \in \DD \;\; (i,j) \notin \EE$.
The relation between K-clique problem and K-independent-set
is analogous to the connection between K-coloring and K-clique-cover problems:
the K-clique problem on $\GG$ is equivalent to
K-independent-set problem on its complement $\GG'$.

\index{vertex-cover}
K-independent-set is in turn equivalent to (N-K)-\magn{vertex cover} problem. 
A vertex cover $\DD' \subseteq \VV$ is a subset of nodes such that each
edge is adjacent to at least one vertex in $\DD'$. It is easy to see that
 $\DD$ is an independent set iff $\DD' = \VV \back \DD$ is a vertex cover.
Therefore our solution here for independent set directly extends to vertex cover.

\magn{K-packing} is a special case of sub-graph isomorphism, which asks whether\marnote{K-packing \& sub-graph isomorphism}
a graph $\GG_1$ is a sub-graph of $\GG_2$. For packing, $\GG_2$ is the main graph ($\GG$)
and $\GG_1$ is the complete graph of size $K$ (see \refSection{sec:graphongraph})
K-independent-set and K-clique problems are also closely related to \magn{sphere packing} and finding nonlinear codes.
To better motivate these problems, here we start with the problem of K-packing
formulated as min-max problem in (several) factor-graphs.
We then show that the sum-product reduction of K-packing is K-independent-set
and K-clique problems. By doing this we simultaneously introduce message passing
solutions to all three problems. 

\index{K-packing}
\index{packing}
\marnote{K-packing}
Given a symmetric distance matrix $\Dn \in \Re^{N \times N}$ between $N$ 
data-points and a number of \magn{code-words} $K$, the \magn{K-packing} problem is
to choose a subset of $K$ points such that the minimum distance
$\Dn_{i,j}$ between any two code-words is maximized.  
Here we introduce
two different factor-graphs such that min-max
inference obtains the K-packing solution.

In order to establish the relation between the min-max problem and the CSPs above,
we need the notion of $\yy$-neighbourhood graph
\marnote{$y$-neighbourhood graph}
\index{y-neighbourhood graph}
\index{neighbourhood graph}
\begin{definition}\label{def:y-neighbourhood} 
The \magn{y-neighborhood graph} for (distance) matrix $\Dn \in
\Re^{N \times N}$ is defined as the graph $\GG(\Dn, \yy) =
(\VV,\EE)$, where $\VV = \{1,\ldots,N\}$ and $\EE =
\{(i,j) \mid \; \Dn_{i,j} \leq \yy\}$.  
\end{definition}

\subsection{Binary variable factor-graph}\label{sec:packing_binary}
Let binary variables $\xs =  \{ \xx_1,\ldots,\xx_N \} \in \{0,1\}^{N}$
indicate a subset of variables of size $K$ that are selected as
code-words. We define the factors such that the min-max assignment $$\xs^* =  \arg_{\xs} \min \; \max_{\II \in \FF} \; \ff_{\II}(\xs_\II)$$ is the K-packing solution.
Define the factor-graph with the following two types of factors:
\index{K-packing!binary-variable-model}

\index{factor!K-of-N}
\marnote{K-of-N factor}
\noindent
\bullitem \magn{K-of-N factor}  (\refSection{sec:hop})
\begin{align*}
  \ff_{\NN}(\xs) \; = \; \ident((\sum_{i \in \NN} \xx_i) = K)
\end{align*}
ensures that $K$ code-words are selected. Recall that definition of $\ident(.)$
depends on the semiring (\refEq{eq:identity}). The K-packing problem is defined on the min-max semiring. However, since we plan to solve the min-max inference by sum-product reductions, it is important to note that the $\pp_\yy$ reduction of 
$\ident(.)$ for any $\yy$ is $\ident(.)$ as defined for sum-product ring. 

\index{factor!paiwise}
\marnote{pairwise factors}
\noindent\bullitem \magn{Pairwise factors} are only effective if both 
$\xx_i$ and $\xx_j$ are non-zero
\begin{align}\label{eq:pairwise_packing}
  \ff_{\{i,j\}}(\xx_i, \xx_j) = \min \left ( \ident(\xx_i = 0 \vee \xx_j = 0) ,\; \max \big (\ident(\xx_i = 1 \wedge \xx_j = 1) ,\; -\Dn_{i,j} \big ) \right )  \quad \forall i,j
\end{align}
where the tabular form is simply
\vspace{.1in}
\begin{center}
\scalebox{.8}{
\begin{tabu}{r r |[2pt] c | c }
\multicolumn{2}{c}{}&\multicolumn{2}{c }{$\xx_j$}\\
\multicolumn{2}{c}{}&0&1\\\tabucline[2pt]{3-4}
\multirow{2}{*}{$\xx_i$}&0&$-\infty$&\multicolumn{1}{c |[2pt]}{$-\infty$}\\\cline{2-4}
&1&$-\infty$&\multicolumn{1}{c |[2pt]}{$-\Dn_{i,j}$}\\\tabucline[2pt]{3-4}
\end{tabu}
}
\end{center}
\vspace{.1in}

Here the use of $-\Dn_{i,j}$ is to convert the initial max-min 
objective to min-max.\footnote{The original objective is max-min because it aims to maximize the minimum distance between any two code-words.}
Recall that $\GG(\Dn, \yy)$ defines a graph based on the distance matrix $\Dn$, s.t. two nodes are connected iff
their distance is ``not larger than'' $\yy$. This means $\GG(-\Dn, -\yy)$ corresponds to a graph in which the
connected nodes have a distance of ``at least'' $\yy$.

\index{K-clique!binary-variable-model}
\marnote{K-packing \& K-clique}
\begin{proposition}\label{th:packing}
  The $\pp_{\yy}$-reduction of the $K$-packing factor-graph above for the distance
  matrix $\Dn \in \Re^{N \times N}$ defines a uniform distribution over
  the cliques of $\GG(-\Dn, -\yy)$ of size $K$.
\end{proposition}
\begin{proof}
Since $\pp_\yy(\xs)$ is uniform over its domain it is enough to show that:

\noindent \bullitem \emph{Every clique of size $K$ in $\GG(-\Dn, -\yy)$  corresponds to a unique assignment $\xs^*$ with $\pp_\yy(\xs^*) > 0$:}\\
\noindent Given a clique $\CC \subseteq \VV$ of size K in $\GG(-\Dn, -\yy) = (\VV, \{(i,j) \mid \Dn_{i,j} \geq y\})$, 
define $\xs^* = \{\xx_i = ident(i \in \CC) \mid i \in \VV\}$. It is easy to show that $\pp_\yy(\xs^*) > 0$.
For this we need to show that all the constraint factors in $\pp_\yy$ are satisfied. The K-of-N factor is trivially satisfied as $\vert \CC \vert  = K$.
The $\pp_\yy$-reduction of the pairwise factor of \cref{eq:pairwise_packing} becomes
\begin{align}\label{eq:pairwise_packing_reduced}
  \ff_{\{i,j\}}(\xx_i, \xx_j) =  \ident(\xx_i = 0 \vee \xx_j = 0)  +   \big (\ident(\xx_i = 1 \wedge \xx_j = 1) \ident(-\Dn_{i,j} \leq -y) \big )   \quad \forall i,j
\end{align}
where we replaced $\min$ and $\max$ with $+$ and $.$ operators of the sum-product semiring and thresholded $-\Dn_{i,j}$ by $-y$.
To see that all pairwise constraint factors are satisfied consider two cases: (I) nodes $i, j \in \CC$, and therefore $\xx^*_i = \xx^*_j = 1$. 
This also means $i$ and $j$ are connected and 
 definition of $\GG(-\Dn, -\yy)$, this implies $\Dn_{i,j} \geq y$. Therefore the second term in factor above evaluates to one.
(II) either $i$ or $j$ are not in $\CC$, therefore the first term evaluates to one.
Since both pairwise factors and the K-of-N factors for $\xs^*$ are non-zero $\pp_\yy(\xs^*) > 0$.

\noindent \bullitem \emph{every assignment $\xs^*$ with $\pp_\yy(\xs^*) > 0$ corresponds to a unique clique of size $K$ in $\GG(-\Dn, -\yy)$:}\\
\noindent 
\cref{eq:pairwise_packing_reduced} implies $\xx_i = 1 \wedge \xx_j = 1 \quad \Rightarrow \quad \Dn_{i,j} \geq y$. On the other hand, 
$\pp_\yy(\xs) > 0$ means K-of-N factor is satisfied and therefore exactly K variables $\xx_i$ are nonzero.
Therefore the index of these variables identifies subset of nodes in $\GG(-\Dn, -y)$ that are connected (because $\Dn_{i,j} \geq y$),
forming a clique.
\end{proof}

\begin{figure}
    \centering
    \includegraphics[width=.4\linewidth]{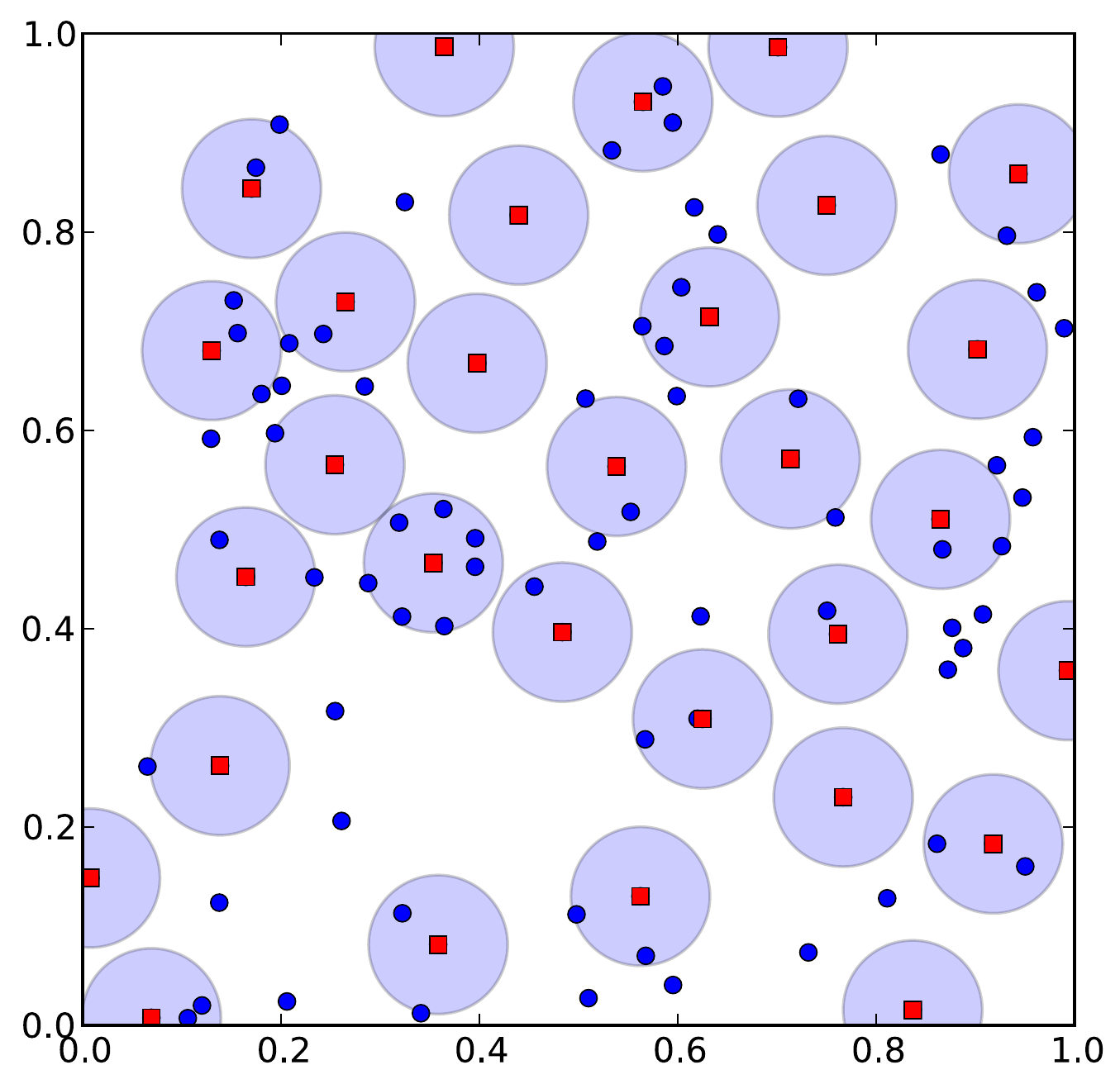}
\caption[Example of message passing for K-packing.]{Using message passing to choose $K =
      30$ out of $N = 100$ random points in the Euclidean plane to
      maximize the minimum pairwise distance (with $T = 500$
      iterations for PBP). Touching circles show the minimum distance}\label{fig:packing_euclidean}
\end{figure}

\marnote{complexity}
\begin{claim}\label{th:bin_packing_cost} 
The time-complexity of each iteration of sum-product BP for the factor-graph above is 
\begin{easylist}
& $\OO(N K + \vert \EE \vert)$ for variable and factor synchronous update.
& $\OO(N^2 K)$ for factor-synchronous ($\ff$-sync) message update.
& $\OO(N^3)$ for completely asynchronous update.
\end{easylist}
\end{claim}
\begin{proof}
The cost of $\ff$-sync calculation of factor-to-variable messages for the K-of-N factor is $\OO(N K)$
and the cost of factor-to-variable messages for pairwise factors is $\OO(1)$. Since there are $\vert \EE \vert$
such factor, this cost evaluates to $\OO(N K + \vert \EE \vert)$.
Since each node is adjacent to $\vert \EE(i, \cdot) \vert$ other nodes, the variable-synchronous update of 
variable-to-factor messages is $\OO(\sum_{i \in \VV} \vert \EE(i, \cdot) \vert ) = \OO(\vert \EE \vert)$, which gives 
a total time-complexity of $\OO(N K + \vert \EE \vert)$.

In asynchronous update, the cost of factor-to-variable messages for the K-of-N factor is $\OO(N^2 K)$
as we need to calculate each message separately. 
Moreover, updating each variable-to-factor message is $\OO(\vert \EE(i,\cdot))$, resulting a total of 
$\OO(\sum_{i\in \VV} \vert \EE(i, \cdot) \vert^2) = \OO(N^3)$. Since $K \leq N$, when all the updates
are asynchronous, this cost subsumes the factor-to-variable cost of $\OO(N^2 K)$.
\end{proof}

A corollary is that the complexity 
of finding the approximate min-max solution by sum-product reduction is 
$\OO( (N K + \vert \EE \vert) \log(N))$ using synchronous update.
\RefFigure{fig:packing_euclidean} shows an example of the solution found by message passing for K-packing with Euclidean distance.

\subsection{Categorical variable factor-graph}\label{sec:packing_categorical}
\index{K-packing!categorical-model}
Define the K-packing factor-graph as follows:
Let $\xs = \{
\xx_1,\ldots, \xx_K\}$ be the set of $K$ variables where $\xx_i \in 
 \{1,\ldots,N\}$. For every two
distinct points $1 \leq i <j \leq K$ define the factor \marnote{pairwise factors}
\begin{align}\label{eq:packing_pairwise_cat}
\ff_{\{i,j\}}(\xx_i,\xx_j) =  \max \big (-\Dn_{\xx_i,\xx_j}, \ident(\xx_i \neq \xx_j) \big ) 
\end{align}
Here each variable represents a code-word and this
 factor ensures that code-words are distinct. Moreover if $\xx_i$ and $\xx_j$  are distinct,  $\ff_{\{i,j\}}(\xx_i,\xx_j) = -\Dn_{\xx_i, \xx_j}$
is the distance between the nodes that $\xx_i$ and $\xx_j$ represent.

The tabular form of this factor is 
\vspace{.1in}
\begin{center}
\scalebox{.8}{
\begin{tabu}{r r |[2pt] c | c | c | c | c  }
\multicolumn{2}{c}{}&\multicolumn{5}{c }{$\xx_j$}\\
\multicolumn{2}{c}{}& 1 & 2 & $\cdots$ & $N-1$ & $N$ \\\tabucline[2pt]{3-7}
\multirow{5}{*}{$\xx_i$}
        &1&$\infty$ & $-\Dn_{1,2}$  & $\cdots$  & $-\Dn_{1,N-1}$ &\multicolumn{1}{c |[2pt]}{ $-\Dn_{1,N}$} \\\cline{2-7}
        &2&$-\Dn_{2,1}$ & $\infty$ & $\cdots$ & $-\Dn_{2,N-1}$ & \multicolumn{1}{c |[2pt]}{$-\Dn_{2,N}$} \\\cline{2-7}
        &$\cdots$&$\vdots$ & $\vdots$ & $\ddots$  & $\vdots$ & \multicolumn{1}{c |[2pt]}{$\vdots$}  \\\cline{2-7}
        &$N-1$&$-\Dn_{N-1,1}$ & $-\Dn_{N-1,2}$ & $\cdots$ & $\infty$ & \multicolumn{1}{c |[2pt]}{$-\Dn_{N-1,N}$}\\\cline{2-7}
        &$N$&$-\Dn_{N,1}$ & $-\Dn_{N,2}$ &  $\cdots$ & $-\Dn_{N, N-1}$& \multicolumn{1}{c |[2pt]}{$\infty$}\\\tabucline[2pt]{3-7}
\end{tabu}
}
\end{center}
\vspace{.2in}

The following proposition relates this factor-graph to K-clique problem.
\marnote{K-packing \& K-clique}
\index{K-clique!categorical model}
\begin{proposition}\label{th:packing_2}
  The $\pp_{\yy}$-reduction of the $K$-packing factor-graph above for the distance
  matrix $\Dn \in \Re^{N \times N}$ defines a uniform distribution over
  the cliques of $\GG(-\Dn, -\yy)$ of size $K$.
\end{proposition}
\begin{proof}
 Since $\pp_\yy$ defines a
  uniform distribution over its support, it is enough to show that any
  clique of size $K$ over $\GG(-\Dn,-y)$ defines a unique set of
  assignments all of which have nonzero probability ($\pp_\yy(\xs) > 0$)
  and any assignment $\xs$ with $\pp_\yy(\xs) > 0$ defines a unique clique
  of size at least $K$ on $\GG(-\Dn, -y)$.  First note that the
  basic difference between $\GG(\Dn, y)$ and $\GG(-\Dn, -y)$ is
  that in the former all nodes that are connected have a distance of
  at most $y$ while in the later, all nodes that have a distance of at
  least $y$ are connected to each other.  Consider the
  $\pp_\yy$-reduction of the pairwise factor of \refEq{eq:packing_pairwise_cat}
  \begin{align}\label{eq:packing_pairwise_cat_reduced}
    \ff_{\{i,j\}}(\xx_i,\xx_j) &=    \ident \left (\max (-\Dn_{\xx_i,\xx_j},  \ident(\xx_i \neq \xx_j)) \leq y \right) \\
     & = \ident(-\Dn_{\xx_i,\xx_j} \leq y \wedge \xx_i \neq \xx_j) \notag
  \end{align}
where, basically, we have replaced $\max$ from min-max semiring with $\otimes$ operator of the sum-product semiring and thresholded $\Dn_{\xx_i, \xx_j}$.
 
\noindent \bullitem \emph{ Any clique of size $K$ in $\GG(-\Dn, -y)$, defines $K\!$
    unique assignments, such that for any such assignment $\xs^*$,
    $\pp_\yy(\xs^*) > 0$:}\\
\noindent  For a clique $\CC = \{c_1,\ldots,c_K \} \subseteq \VV$ of size $K$, define $\xx^*_i
  = c_{\pi(i)}$, where $\pi: \{1,\ldots,K \} \to \{1,\ldots,K \}$ is a
  permutation of nodes in clique $\CC$. Since there are $K\!$ such
  permutations we may define as many assignments $\xs^*$.  Now consider
  one such assignment.  For every two nodes $\xx^*_i$ and $\xx^*_j$, since
  they belong to the clique $\CC$ in $\GG(-\Dn, -y)$, they are
  connected and $\Dn_{\xx_i, \xx_j} \geq y$. This means that all the
  pairwise factors defined by \cref{eq:packing_pairwise_cat_reduced} have
  non-zero values and therefore $\pp_\yy(\xs) > 0$.

\noindent \bullitem  \emph{ Any assignment $\xs^*$ with $\pp_\yy(\xs^*) > 0$ corresponds to a
    unique clique of size $K$ in $\GG(-\Dn,-y)$:} Let $\CC =
  \{\xx^*_1, \ldots, \xx^*_K\}$. Since $\pp_\yy(\xs^*) > 0$, all pairwise factors
  defined by \cref{eq:packing_pairwise_cat_reduced} are non-zero.  Therefore
  $\forall i,j\neq i\;\; \Dn_{\xx_, \xx_j} \geq y$, which means all $\xx_i$
  and $\xx_j$ are connected in $\GG(-\Dn,-y)$, forming a clique of
  size $K$.
\end{proof}

\index{Py-reduction}
To put simply, in acquiring the $\pp_\yy$-reduction, we set the values in the table form above to zero if their value is less than $\yy$
and set them to one otherwise. The resulting factor-graph, defines a distribution $\pp_\yy(\xs)$, s.t. $\pp_\yy(\xs) > 0$ means $\xs$ defines
a clique of size $K$ in a graph $\GG(-\Dn, -\yy)$ which connects nodes with distance larger than $\yy$.

\marnote{complexity}
\begin{claim}\label{th:packing_categorical_cost}
Each iteration of BP for this factor-graph with pairwise factors is $\OO(N^2 K^2)$,
for synchronized update and $\OO(N^2 K^3)$ for asynchronous update. Using the sum-product reduction
of min-max inference this suggests a $\OO(N^2 K^2 \log(N))$ (for sync. update) and $\OO(N^2 K^3 \log(N))$ (for asynchronous update) 
procedure for K-packing problem.
\end{claim}
\begin{proof} (\pcref{th:packing_categorical_cost})
Since the factors are not sparse, the complexity of calculating a single factor-to-variable
message is $\OO(\vert \XX_{\II} \vert) = \OO(N^2)$, resulting in $\OO(N^2K^2)$ cost per
iteration of variable-synchronous update for BP. However if we update each message separately,
since each message update costs $\OO(N^2 K)$, the total cost of BP per iteration is $\OO(N^2 K^3)$

Since the diversity of pairwise
distances is that of elements in $\Dn$ -- \ie $\vert \YY \vert = \OO(N^2)$ -- the general cost of finding
an approximate min-max solution by message passing is $\OO(N^2K^2
\log(N))$ for sync. message update and $\OO(N^2K^3\log(N))$ for async. update. 
\end{proof}

The $\pp_{\yy}$-reduction of our second formulation was first proposed by \cite{ramezanpour2012cavity} to find
non-linear binary codes.  The authors consider the Hamming distance
between all binary vectors of length $n$ (\ie $N = 2^n$) to obtain
binary codes with known minimum distance $\yy$. As we saw, this method
is $\OO(N^2K^2) = \OO(2^{2n}K^2)$ --\ie grows exponentially
in the number of bits $n$. In the following section, we introduce a
factor-graph formulation specific to categorical variables with
Hamming distance whose message passing complexity is polynomial in $n = \log(N)$.
Using this
formulation we are able find optimal binary and ternary codes where both $n$ and $y$ are
large.

\subsection{Efficient Sphere packing with Hamming distance}\label{sec:hamming}
\index{sphere-packing}
\index{Hamming}
Our factor-graph defines a distribution over the $K$ binary vectors of 
length $n$ such that the distance between every pair of binary vectors
is at least $\yy$. \footnote{ For convenience we restrict this
  construction to the case of binary vectors.  A similar procedure may
  be used to find maximally distanced ternary and $q$-ary vectors, for
  arbitrary $q$.}  

\begin{figure}
\centering
      \includegraphics[width=.5\textwidth]{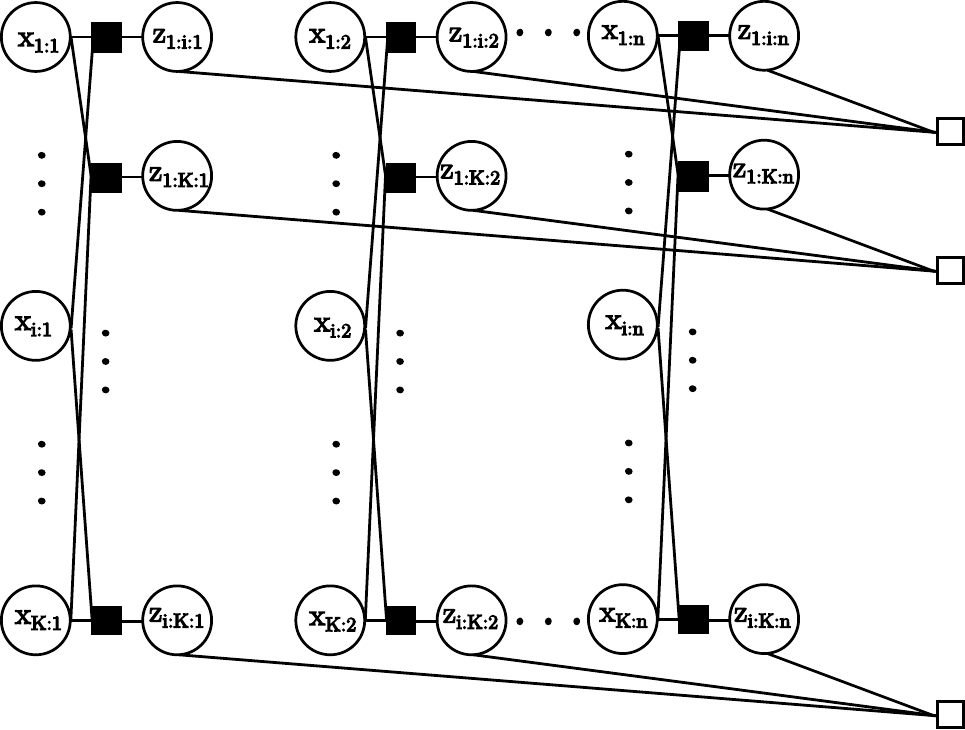}
      \caption{The factor-graph for sphere-packing with Hamming distances, where the distance factors are white squares 
and z-factors are in black.}
      \label{fig:spherepack_fg}
\end{figure}

To better relate this to K-clique problem, consider $2^n$ binary code-words of length $n$
as nodes of a graph and connect two nodes iff their Hamming distance is at least $\yy$.
Finding a K-clique in this graph is equivalent to discovery of   \marnote{nonlinear binary codes}
\index{nonlinear binary codes}
so-called \magn{nonlinear binary codes} --  a
fundamental problem in information theory (\eg see
\cite{bincodes,Cover1991}).
Here assuming $\yy$ is an odd number, if at most $\frac{\yy-1}{2}$ digits of a code-word are corrupted (\eg in communication),
since every pair of codes are at least $\yy$ digits apart, we can still recover the uncorrupted code-word. 
The following is a collection of $K=12$ ternary code-words of length $n=16$, obtained using the factor-graph that we discuss in this section,
where every pair of code-words are different in at least $\yy = 11$ digits. 

\begin{center}
    \scalebox{.7}{
      \begin{tabu}[b]{cccccccccccccccc}
        \tabucline[2pt]{-}
        2& 1& 2& 1& 0& 1& 1& 2& 2& 1& 2& 1& 2& 1& 0& 2\\
        1& 1& 1& 2& 1& 0& 0& 2& 2& 1& 1& 1& 0& 2& 1& 0\\
        0& 0& 1& 2& 0& 1& 2& 0& 2& 1& 2& 2& 1& 0& 2& 0\\
        0& 0& 0& 2& 1& 0& 1& 2& 1& 0& 1& 0& 2& 1& 0& 1\\
        2& 2& 0& 2& 2& 2& 1& 1& 2& 0& 2& 2& 1& 2& 1& 2\\
        0& 2& 0& 0& 0& 2& 0& 2& 0& 0& 2& 1& 0& 0& 0& 0\\
        1& 1& 2& 0& 0& 1& 2& 2& 0& 0& 1& 0& 1& 2& 2& 1\\
        2& 0& 2& 0& 2& 1& 0& 1& 1& 2& 0& 1& 1& 1& 2& 0\\
        0& 2& 1& 1& 1& 1& 1& 1& 0& 1& 0& 0& 0& 0& 1& 2\\
        1& 0& 0& 1& 0& 2& 0& 0& 1& 2& 0& 0& 2& 2& 1& 2\\
        1& 2& 2& 1& 1& 0& 2& 1& 1& 2& 2& 2& 2& 0& 2& 1\\
        1& 0& 1& 0& 2& 2& 2& 1& 0& 1& 1& 2& 2& 1& 0& 2\\
        \tabucline[2pt]{-}
      \end{tabu}
    }
\end{center}

The construction of this factor-graph is more involved than our previous constructions.
The basic idea to avoid the exponential blow up is to have one variable per \textit{digit} of each code-word (rather than one variable per code-word).
Then for each pair of code-words we define an auxiliary binary vector of the same length, that indicates if the two code-words are different in each digit.
Finally we define an at-least-$\yy$-of-$n$ constraint over each set of auxiliary vectors that ensures every two pair of code-words are at least different in $\yy$ digits.
\RefFigure{fig:spherepack_fg} shows this factor-graph.

More specifically,
let $\xs = \{\xxs{1}{\cdot}, \ldots, \xxs{K}{\cdot} \}$ be a set of binary
vectors, where $\xxs{i}{\cdot} = \{\xxx{i}{1},\ldots,\xxx{i}{n}\}$ represents the
$i^{th}$ binary vector or code-word.  Additionally for each two code-words $ 1\leq i
< j \leq K$, define an auxiliary binary vector $\zzs{i:j}{\cdot} = \{ \zzz{i:j}{1} ,\ldots,\zzz{i:j}{n}\}$ 
of length $n$.

For each distinct pair of binary vectors $\xxs{i}{\cdot}$ and $\xxs{j}{\cdot}$, 
and a particular digit $1 \leq k \leq n$, the auxiliary variable is constrained to be $\zzz{i:j}{k} = 1$
iff $\xxx{i}{k} \neq \xxx{j}{k}$.  Then we define an
at-least-y-of-n factor over $\zzs{i:j}{\cdot}$ for every pair of code-words,
to ensure that they differ in at least $\yy$ digits.

\index{variable!auxiliary}
\index{factor!Z-factor}
The factors are defined as follows\\\marnote{Z-factors}
\noindent \bullitem \magn{Z-factors:} For every $ 1\leq i < j \leq K$ and $1 \leq k
\leq n$, define
\begin{align*}
  \ff_{\{i:k, j:k, i:j:k\}}(\xxx{i}{k},\xxx{j}{k},\zzz{i:j}{k}) &= 
                                   \ident\left( \big( (\xxx{i}{k} \neq \xxx{j}{k}) \wedge \zzz{i:j}{k} = 1 \big ) \vee 
\big ( (\xxx{i}{k} = \xxx{j}{k}) \wedge \zzz{i:j}{k} = 0 \big )\right )\\
\end{align*}
This factor depends on three binary variables, therefore we can
explicitly define its tabular form containing $2^3 = 8$ possible inputs.
Here the only difference between binary and ternary (and $q$-ary) codes in general
is in  the tabular form of this factor.
For example for ternary codes, the tabular for of this factor is a $3 \times 3 \times 2$ array ($\zzz{i:j}{k}$ is always binary).

\index{factor!distance}
\marnote{distance factors}
\noindent \bullitem \magn{Distance-factors:} For each $\zzs{i:j}{\cdot}$ define 
at-least-y-of-n factor (\refSection{sec:hop}):
\begin{align*}
  \ff_{i:j:\cdot}(\zzs{i:j}{\cdot}) = \ident(\sum_{1\leq k \leq n} \zzz{i:j}{k} \geq \yy)
\end{align*}

\marnote{complexity}
\begin{claim}\label{th:hamming_cost}
Each iteration of sum-product BP over this factor-graph is 
\begin{easylist}
& $\OO(K^2 n y)$ for variable and factor synchronous update.
& $\OO(K^2 n^2 y)$ for variable-sync update.
& $\OO(K^3 n + K^2 n^2 y)$ for completely asynchronous BP update.
\end{easylist}
\end{claim}
\begin{proof}(\pcref{th:hamming_cost})
We first consider the complexity of variable-to-factor updates:
Each auxiliary variable $\zzz{i:j}{k}$ is connected to three factors and each
$\xxx{i}{j}$ is connected to $\OO(K)$ z-factors. Since there are $\OO(n K)$, $\xxx{i}{j}$
variables a variable-sync. update of variable-to-factor messages is $\OO(n K^2)$, while 
async. update is $\OO(n K^3)$.

Next we consider two possibilities of factor-to-variable updates:
We have $\OO(n K^2)$ z-factors, and factor-to-variable update for each of them is $\OO(1)$,
this cost is subsumed by the minimum variable-to-factor cost.
The factor-graph also has $\OO(K^2)$ distance factors, where the cost of each factor-to-variable
update is $\OO(n y)$. Since $\vert \nb \II \vert = K$ for a distance factor, a factor-sync.
update is $\OO(K^2 n y)$ in total, while an async. update is $\OO(K^2 n^2 y)$.

Adding the cost of variable-to-factor and factor-to-variable in different scenarios we
get the time-complexities stated in the claim.
\end{proof}

The following table reports some optimal binary codes (including codes
with large number of bits $n$) from \cite{bincodes}, recovered using this factor-graph.
We used Perturbed BP with variable-sync update and $T = 1000$ iterations to find an assignment $\xs^*,\zs^*$ with $\pp(\xs^*,\zs^*) > 0$.
\vspace{.1in}
\marnote{optimal binary codes}
\begin{center}
  \scalebox{.7}{
    \begin{tabu}{c c c|[2pt] c c c|[2pt]c c c |[2pt] c c c }
      \tabucline[2pt]{-}
      n & K & y & n & K & y &n & K & y &n & K & y\\      \tabucline[2pt]{-}
      8 & 4& 5 &11 &4 &7 &14 &4 &9 &16 &6 &9 \\\hline
      17 & 4 & 11 &19 &6 &11 &20 &8 &11 &20 &4 &13 \\
      23 & 6 & 13 &24 &8 &13 & 23&4 &15 &26 & 6& 15\\
      27 & 8 & 15 &28 &10 &15 &28 &5 &16 &26 &4 &17 \\
      29 & 6 & 17 &29 &4 &19 &33 &6 & 19&34 &8 &19 \\
      36 & 12 & 19 &32 &4 &21 &36 &6 &21 &38 &8 &21 \\
      39 &10  &21  &35 & 4& 23& 39& 6&23 & 41& 8&23 \\
      39 &4  &25  & 43& 6&23 &46 &10 &25 &47 &12 &25 \\
      41 & 4 &27  & 46& 6& 27&48 &8 &27 &50 &10 &27 \\
      44 & 4 &29  &49 & 6& 29& 52&8 &29 &53 &10 &29 \\\tabucline[2pt]{-}
    \end{tabu}
  }
\end{center}
\vspace{.1in}

\section{Optimization variations of CSPs}\label{sec:csp_opt}
\index{constraint satisfaction problem!optimization}
This section briefly reviews the optimization variations of the CSPs, we have 
studied so far.
The optimization version of satisfiability is known as \magn{max-SAT} or maximum weighted SAT, where each clause has a weight, and the objective is to maximize the
\index{satisfiability!max}
\index{max-SAT}
weighted sum of satisfied clauses. Here, simply using factors 
$\ff_{\II}(\xs_\II): \XX_\II \to \{ 0, -w_\II \}$, where $w_\II$ is the positive weight of clause
$\II$, min-sum BP will attempt to find the max-SAT solution. Note that here
$\ff_\II$ is not a ``constraint'' factor anymore (see \refDefinition{def:constraint}).
Alternative approaches
using variations of energetic survey propagation has also been used to improve max-SAT results~\cite{chieu2007cooled,chieu2008relaxed}. 
\index{SP!energetic}
\index{adversarial SAT}
\index{satisfiability!adversarial}
A less studied optimization variation of satisfiability is adversarial SAT which corresponds to min-max-sum inference \cite{castellana2011adversarial}.

For minimum coloring -- \aka \magn{chromatic number} -- and \magn{minimum clique-cover} problem, since the optimal value $1 \leq K^* \leq K_{\max}$ is bounded,
\index{chromatic number}
\index{coloring}
by access to an oracle for the decision version (or an incomplete solver \cite{kautz2009incomplete} such as message passing), we can use \magn{binary search} 
to find the minimum $K$ in $\OO( \tau \log(K_{\max}))$ time, where the decision problem has a $\OO(\tau)$ time-complexity. 
In particular, since the chromatic number is bounded by the maximum degree \cite{brooks1941colouring},
approximating the chromatic number using binary search and message passing 
\index{binary search}
gives a $\OO(\log(\max_i(|\EE(i,\cdot)|)) \ \max_i(|\EE(i,\cdot)|)\  |\EE|)$ time procedure.

\index{K-clique!maximum}
The same binary-search approach can be used
for minimum  dominating-set, minimum set-cover, maximum clique, maximum independent set and minimum vertex cover. However, these optimization variations also allow a more efficient and direct approach.
Note that both min-max variation  and the minimization (or maximization)
variations of these problems use binary search.
For example both minimum clique-cover and K-clustering (see \refSection{sec:k_clustering}) can be solved using binary-search over K-clique-cover 
decision problem. The parameter of interest for binary search is $K$ in the former case, and $\yy$ in the later case, 
where $\yy$ is a threshold distance that defines connectivity in $\GG$ (see \cref{def:y-neighbourhood}). However, often both variations (\ie min-max and minimization or maximization over $K$) also allow 
direct message passing solutions.

\index{dominating-set!minimum}
For \magn{minimum dominating-set and set-cover},
we replace the sum-product semiring with min-sum semiring and drop the K-of-N factor.
Instead, a local factor $\ff_{i:i}(\xx_{i:i}) = -\xx_{i:i} w_{i}$ gives a weight to each node $i \in \VV$.
Here, the min-sum inference seeks a subset of nodes that form a dominating set and have the largest
sum of weights $\sum_{i \in \DD}  w_{i}$. 
\index{set-cover!minimum}
\index{dominating-set!minimum}
Also note that by changing the semiring to min-sum, the identity functions $\ident(.)$ change accordingly so that the leader factor and consistency constraints remain valid.
This gives an efficient $\OO(\vert \EE \vert)$ synchronous procedure for minimum set-cover and minimum dominating-set.
This problem and the resulting message passing solution are indeed a variation of K-medians and affinity propagation respectively (see \refSection{sec:k-median}).

The same idea applies to \magn{maximum clique and maximum independent set}:
as an alternative to fixing or maximizing the ``size'' of a clique or an independent
set, we may associate each node with a weight $w_i \; \forall i \in \VV$ and seek
a subset of nodes that form an independent set (or a clique) with maximum
weight. \citet{sanghavi2007message} study the max-product message passing solution
to this problem and its relation to its LP-relaxation.
In particular they  show that starting from uniform messages,
if BP converges, it finds the solution to LP relaxation.

\index{independent-set!maximum}
Here we review their factor-graph for maximum independent set using min-sum inference.
Let $\xs = \{ \xx_1,\ldots,\xx_N \} \in \{0,1\}^{N}$ be a set of binary variables,
one for each node in $\VV$, where $\xx_i = 1$ means $i \in \DD$, the independent set.

\noindent \bullitem \magn{Local factors} capture the cost (negative weight) for each node and is equal to $-w_i$ if $\xx_i = 1$ and zero otherwise
\begin{align*}
  \ff_{i}(\xx_i) \quad = \quad \min(-w_i, \ident(\xx_i = 0)) \quad \forall i \in \VV
\end{align*}
\noindent \bullitem \magn{Pairwise factors} ensure that if $(i,j) \in \EE$, then 
either $\xx_i = 0 $ or $\xx_j = 0$
\begin{align*}
  \ff_{\{i,j\}}(\xs_{\{i,j\}}) \quad = \quad  \ident(\xx_i = 0 \vee \xx_j = 0) \quad \forall (i,j) \in \EE
\end{align*}

It is easy to see that using a variable synchronous update, 
message passing can be performed very efficiently in $\OO(\vert \EE \vert)$.

\index{vertex-cover!minimum}
\index{SP!energetic}
\citet{weigt2006message} (also see \cite{zhou2003vertex}) propose an interesting approach to \magn{minimum vertex cover} using energetic survey propagation.
Here, again $\xs = \{ \xx_1,\ldots,\xx_N \} \in \{0,1\}^{N}$ has one binary variable per node $i \in \VV$ and a pairwise factor ensure that all edges are covered by at least one node in the cover
\begin{align*}
  \ff_{\{i,j\}}(\xs_{\{i,j\}}) \quad = \quad  \ident(\xx_i = 1 \vee \xx_j = 1) \quad \forall (i,j) \in \EE
\end{align*}
\index{warning propagation}
Using the xor-and semiring, the resulting fixed points of warning propagation reveal minimal (but not necessarily optimal) vertex covers. The authors then suggest using survey propagation, and decimation to find the warning propagation fixed points with lowest energy (\ie smallest size of cover).

%% file: clusters.tex

\index{clustering}
Clustering of a set of data-points is a central problem in machine learning and data-mining  \cite{Rosvall08Infomap,Lancichinetti09Comparison,frey2007clustering,krzakala2013spectral,leskovec2010empirical,palla2005uncovering,pons2005computing,raghavan2007near,xu2007scan,ahn2010link,white2005spectral,ng2002spectral}.
However many interesting clustering objectives, including the problems that we consider in this section are \NP-hard.

In this section we present message passing solutions to several well-known clustering objectives 
including  K-medians, K-clustering, K-centers and modularity optimization. 
Message passing \marnote{stochastic block models}
 has also been used within Expectation Maximization to obtain some of the
best results in learning  stochastic block models (a hidden variable model for clustering) \cite{decelle2011asymptotic}.
The message passing solution to K-medians  and its generalization, clustering by shallow trees 
are proposed by other
researchers. However, for completeness, we review their factor-graphs in \cref{sec:k-median,sec:shallowtrees}.
expressing K-clustering and K-centers problems as min-max inference problems on factor-graphs in \cref{sec:k_clustering,sec:k_center}.

\section{K-medians}\label{sec:k-median}
\index{K-medians}
\index{clustering!K-medians}
Given a symmetric matrix of pairwise distances \marnote{K-medians}
$\Dn \in \Re^{N \times
  N}$ between $N$ data-points, and a number of clusters $K$, K-medians
 seeks a partitioning of data-points into $K$ clusters, 
each associated with a cluster center,
s.t. the sum of distances from each data-point 
to its cluster center is minimized.

This problem is \NP-hard; however, there exists several approximation algorithms for 
metric distances \cite{arora1998approximation,charikar1999constant}. 
Here we present the binary variable factor-graph \cite{givoni2009binary} for a slightly different version of this objective, \marnote{affinity propagation}
proposed by~\citet{frey2007clustering}. The simplified form of min-sum BP messages in this factor-graph is known as \magn{affinity propagation}.
Here, instead of fixing the number of clusters, $K$, the objective is modified to incorporate each cluster's cost to become a center. This cost is added to the sum of distances between the 
data-points and their cluster centers and used to decide the number of clusters at run-time.

Let $\Dn$ be the distance matrix of a directed graph $\GG = (\VV, \EE)$, 
where $(i,j) \notin \EE \Leftrightarrow \Dn_{i,j} = \infty$. Moreover let $\Dn_{i,i}$
denote the willingness of node $i$ to be the center of a cluster. A simple heuristic
is to set this value uniformly to $\mathrm{median}_{i \neq j} \Dn_{i,j}$. 

Define $\xs = \{ \xxx{i}{j} , \forall \; (i,j) \in \EE\}$ 
as a set of binary variables -- one per each directed edge $(i,j)$ -- where $\xxx{i}{j} \in \{0,1\}$ indicates
whether  node $i$ follows node $j$ as its center. Here node $i$ can follow itself as center.
The following factors define the cost and constraints of affinity
propagation:

\index{affinity propagation}
\index{factor!leader}
\marnote{leader factors}
\noindent $\bullet$ \magn{Leader factors} ensure that each node selects exactly one cluster center. 
\begin{align*}
  \ff_{\EE^+(i, \cdot)}(\xs_{\EE(i,\cdot)}) = \ident((\sum_{(i,j) \in \EE^+(i,\cdot)} \xxx{i}{j}) = 1) \quad \forall i \in \VV
\end{align*}
where as before $\EE^+(i,.) = \{(i,j) \in \EE\} \cup \{(i,i)\}$ is the set of edges leaving node $i$.

\marnote{consistency factors}
\index{factor!consistency}
\noindent $\bullet$ \magn{Consistency factors} as defined by \refEq{eq:consistency_hop} ensure that if any node $i \in \EE(., j)$ selects node $j$ as its center of cluster, node $j$ also selects itself as the center

At this point we note that we used both of these factor-types for set-cover and dominating set problem in \refSection{sec:set-cover}. The only addition (except for using a different semiring) is the following factors.

\marnote{local factors}
\noindent $\bullet$ \magn{Local factors} 
take distances and the willingness to become a center into account
\begin{align*}
\ff_{i:j}(\xx_{(i,j)}) = \min\big( \Dn_{i,j}, \ident(\xxx{i}{j} = 0) \big ) \quad \forall (i,j) \in \EE \cup \{(i,i) \mid i \in \VV \}
\end{align*}
where in effect $\ff_{i:j}(\xxx{i}{j})$ is equal to $\Dn_{i,j}$ if $\xxx{i}{j} = 1$ and $0$ otherwise

See \refEq{eq:identity} for definition of $\ident(.)$ in min-sum semiring and note the fact that $\bpplus = \min$ in this case. This means extensions to other semirings (\eg min-max) need only to consider a different $\ident(.)$ and use their own $\bpplus$ operator. However, direct application of min-max inference to this factor-graph is problematic for another reason: since the number of clusters $K$ is not enforced, as soon as node $i$ becomes center of a cluster, all nodes $j$ with $\Dn_{j,j} \leq \Dn_{i,i}$ can become their own centers without increasing the min-max value. In \refSection{sec:k_center}, we resolve
this issue by enforcing the number of clusters $K$ and use inference on min-max semiring to solve the corresponding problem known as K-center problem.

\marnote{complexity}
The complexity of min-sum BP with variable and factor synchronous message update is $\OO(\vert \EE \vert)$ as 
each leader and consistency factor allows efficient $\OO(\vert \EE(.,j)\vert)$ and $\OO(\vert \EE(i,.) \vert)$ calculation of factor-to-variable messages respectively. Moreover, all variable-to-factor
messages leaving node $i$ can be calculated simultaneously in $\OO(\vert \EE(i,.) \vert)$ using variable-synchronous update of \refEq{eq:belief_update}.

\subsection{Facility location problem}\label{sec:facility}
\index{facility location}
A closely related problem to K-medians is the facility location problem, \marnote{facility location}
where a matrix $\Dn \in \Re^{\vert \VV_1\vert \times \vert \VV_2\vert}$ specifies the pairwise
distance between two parts of  
bipartite graph $\GG = (\VV = (\VV_1, \VV_2), \EE)$. The goal is to select 
a sub-set of facilities $\DD \subset \VV_1$ to minimize the sum of distances 
of each customer $i \in \VV_2$ to its associated facility in $\DD$.

The uncapacitated version of this problem (no restriction on $\vert \DD \vert$)
can be solved as special K-median problem where $\Dn_{i,i} = \infty \; \forall i \in \VV_2$.
Message passing solution for min-sum variations of this problem are discussed in \cite{lazic2010solving,lazic2011message}. We discuss the min-max facility location
as a special case of k-center problem in \refSection{sec:k_center}.

\section{Hierarchical clustering}\label{sec:shallowtrees}
\index{clustering!hierarchical}
\index{hierarchical clustering}
By adding a dummy node $*$ and connecting all the cluster centers to this node \marnote{shallow-trees \& K-median}
with the cost $\Dn_{i,*} = \Dn_{i,i}$, we can think of the K-median clustering 
of previous section as finding a minimum cost tree of depth two with the dummy node as its root.
The \magn{shallow trees} of \citet{bailly2009clustering} generalize this notion by allowing
more levels of hierarchy.
The objective is to find a tree of maximum depth $d$,
that minimizes the sum over all its edges.
An alternative approach to hierarchical clustering based on nested application of affinity propagation is discussed in \cite{tan2007joint,givoni2012beyond}. 

Previously we presented the binary-variable model for affinity propagation.
However, it is possible to obtain identical message updates using categorical
variables \cite{frey2007clustering}. Here $\xs = \{\xx_1,\ldots,\xx_N\}$, where $\xx_i \in \XX_i = \{ j \mid (i,j) \in \EE \}$ selects one of the neighbouring nodes
as the center of the cluster for node $i$. In this case we can ignore the 
leader factors and change the consistency and local factors accordingly.

The idea in building hierarchies is to allow each node $i$ to follow another
node $j$ even if $i$ itself is followed by a third node $k$ (\ie dropping the
consistency factors). Also $\xx_i = i$ is forbidden.  However, this creates the risk of forming loops.
\index{factor!depth}
\index{clustering!shallow trees}
\index{shallow trees}
\marnote{depth factors}
The trick used by \citet{bailly2009clustering} is to add an auxiliary ``depth'' variable $\zz_i \in \{0,\ldots,d\}$ at each node. Here, the depth of the dummy node $*$ is zero and the \magn{depth
factors} ensure that if $i$ follows $j$ then $\zz_i = \zz_j + 1$:
\begin{align*}
\ff_{i,j}(\xs_{i}, \zs_{i,j}) \; = \; \ident \bigg ( \zz_i = \zz_j + 1 \vee \xx_i \neq j \bigg)
\end{align*}

\subsection{Spanning and Steiner trees}\label{sec:trees}
\index{spanning tree}
Although finding trees is not a clustering problem, 
since one could use the same techniques used for clustering by shallow trees, we include it in this section.
\marnote{prize-collecting Steiner tree}
Given a graph $\GG = (\VV, \EE)$, 
a penalty $w_{i:j} < 0$ per edge $(i,j) \in \EE$ 
and a prize $w_i >0$ per node $i \in \VV$, the \magn{prize-collecting Steiner tree}'s objective is to select a connected sub-graph
$\GG' = (\VV' \subseteq \VV, \EE'\subseteq \EE)$ with maximum sum of prizes $\sum_{i \in \VV'} w_i + \sum_{(i,j) \in \EE'} w_{i:j}$. 
Since the optimal sub-graph is always a tree, a construction similar to that of 
shallow trees (with min-sum BP) finds high-quality solutions to depth-limited versions of this problem in $\OO(d \vert \EE \vert )$
\cite{bailly2011finding,biazzo2012performance,bailly2009prize,bayati2008statistical}.

\index{Steiner tree}
The main difference with factor-graph of shallow trees is that here the root node is a pre-determined member of $\VV$ and several tricks are used to find best (set of) root(s). However, since the $\vert \VV' \vert$ may be smaller than $\vert \VV \vert$, a different dummy node $*$ is introduced, with zero cost
of connection to all nodes $i \in \VV$ 
\marnote{min. spanning tree}
s.t. $\xx_i = *$ means node $i$ is not a part of the Steiner tree ($i \in \VV \back \VV'$).
\index{dummy node}

Alternatively, if we do not introduce this dummy node such that $\VV' = \VV$ and moreover, set the node penalties to zero, 
the result is a \magn{depth limited spanning tree}.
\citet{bayati2008rigorous} show that if the maximum depth is large enough (\eg $d = N$) and BP is convergent, it will find the minimum spanning tree.

%% file: minmax_clustering.tex
\section{K-clustering problem}\label{sec:k_clustering}
\index{K-clustering}
\index{clustering!K-clustering}
\marnote{K-clustering}
Given a symmetric matrix of pairwise distances $\Dn \in \Re^{N \times
  N}$ between $N$ data-points, and a number of clusters $K$, K-clustering (\aka \magn{min-max
clustering}) seeks a partitioning of data-points that minimizes the
maximum distance between all the pairs in the same partition.
\index{min-max clustering}

We formulate this problem as \magn{min-max} inference problem in a factor-graph. 
Let $\xs = \{ \xx_1,\ldots, \xx_N\}$ with $\xx_i \in 
\{1,\ldots,K\}$ be the set of variables, where $\xx_i= k$ means, point
$i$ belongs to cluster $k$.
The Potts factor 
\marnote{pairwise factors}
\begin{align}\label{eq:kclustering_factor}
\ff_{\{i,j\}}(\xx_i,\xx_j)
= \min \big ( \ident(\xx_i \neq \xx_j), \Dn_{i,j} \big ) \quad \forall 1 \leq i < j \leq N
\end{align}
is equal to $\Dn_{i,j}$ if two nodes are in the same cluster and $-\infty$ otherwise. 
It is easy to see that using min-max inference on  this factor graph, the min-max solution $\xs^*$ 
(\refEq{eq:min-max}) defines a clustering that minimizes the
maximum of all inter-cluster distances.

Recall that the y-neighborhood graph (\cref{def:y-neighbourhood}) for a distance matrix $\Dn$ 
is the graph $\GG(\Dn, \yy) =
(\VV,\EE(\Dn, \yy) = \{ (i,j) \mid \Dn_{i,j} \leq \yy \})$.
\marnote{K-clustering \& K-clique cover}
\begin{claim}\label{th:k_clustering}
  The $\pp_{\yy}$-reduction of the min-max clustering factor-graph above   is identical to $K$-clique-cover factor-graph of \refSection{sec:clique-cover} for $\GG(\Dn, \yy)$.
\end{claim}
\index{Py-reduction}
\index{min-max}
\begin{proof}
The $\pp_\yy$-reduction of the K-clustering factor (\cref{eq:kclustering_factor}) is 
\begin{align*}
\ff_{\{i,j\}}(\xx_i,\xx_j)
= \ident\left (\min \big ( \ident(\xx_i \neq \xx_j), \Dn_{i,j} \big ) \leq \yy\right ) = \ident(\xx_i \neq \xx_j \vee \Dn_{i,j} \leq y) 
\end{align*}

Recall that the K-clique-cover factor-graph, defines a pairwise factor between any two nodes $i$ and $j$ whenever $(i,j) \notin \EE(\Dn, \yy)$,
-- \ie whenever $\Dn_{i,j} \leq y$ it does not define a factor.
However,  $\Dn_{i,j} \leq y$ means that the reduced constraint factor above is satisfied and therefore we only need to consider the cases 
where $\Dn_{i,j} > y$. This gives $\ff_{\{i,j\}}(\xx_i,\xx_j) = \ident(\xx_i \neq \xx_j)$, which is the K-clique-cover factor between two nodes
$i$ and $j$ s.t. $(i,j) \notin \EE(\Dn, \yy)$.
\end{proof}

\marnote{complexity}
We use binary-search over sum-product reductions to approximate the min-max solution (see \refSection{sec:minmax2sumprod}). 
Considering the $\OO(N^2K)$ cost of 
message passing for K-clique-cover, this gives $\OO(N^2K \log(N^2)) = \OO(N^2K \log(N))$
cost for K-clustering, where the $\log(N^2)$ is the cost of binary search over the set of all possible pairwise distance values in $\Dn$.

\begin{figure}[htp!]
  \hbox{
 \includegraphics[width=.45\textwidth]{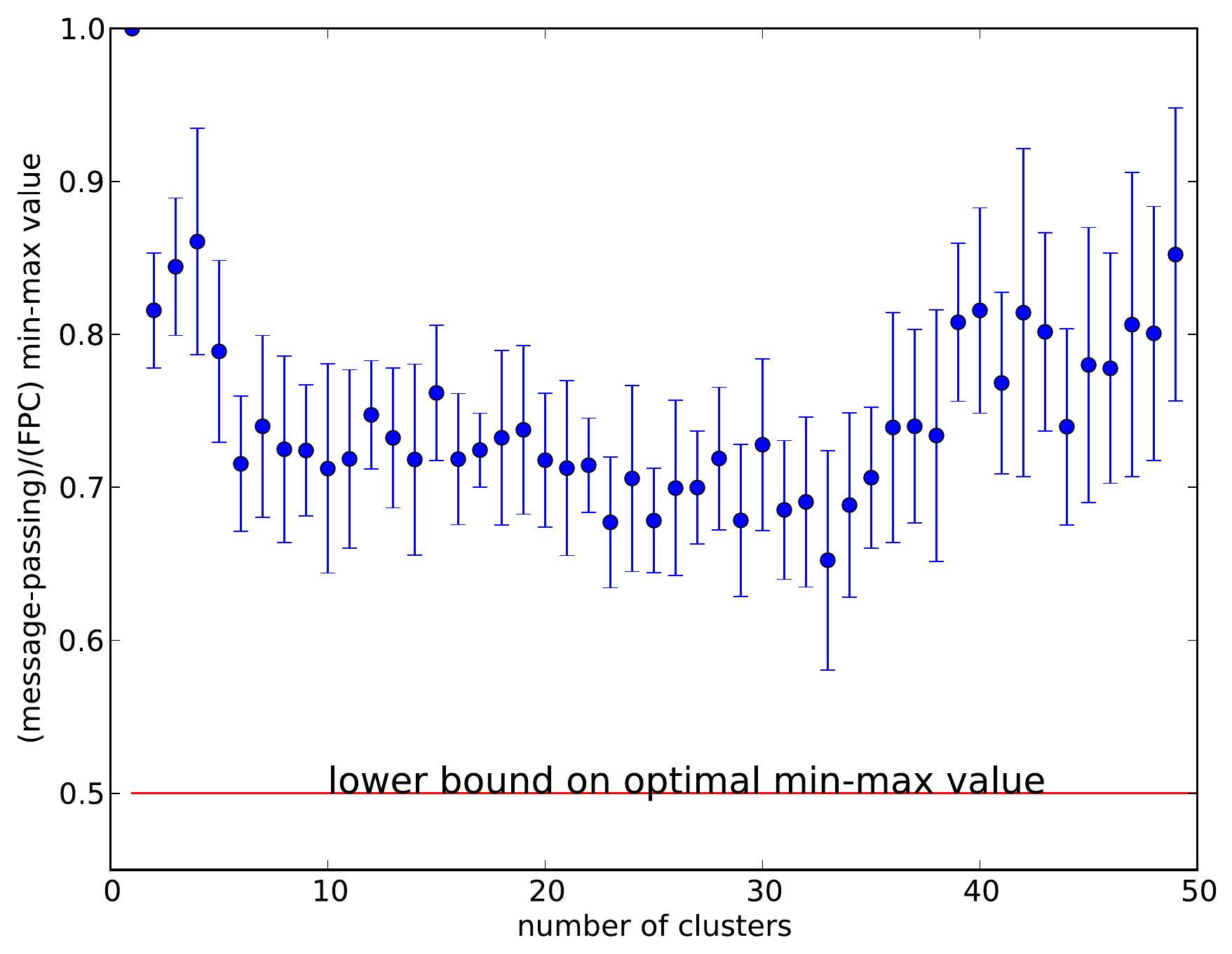}
 \includegraphics[width=.45\textwidth]{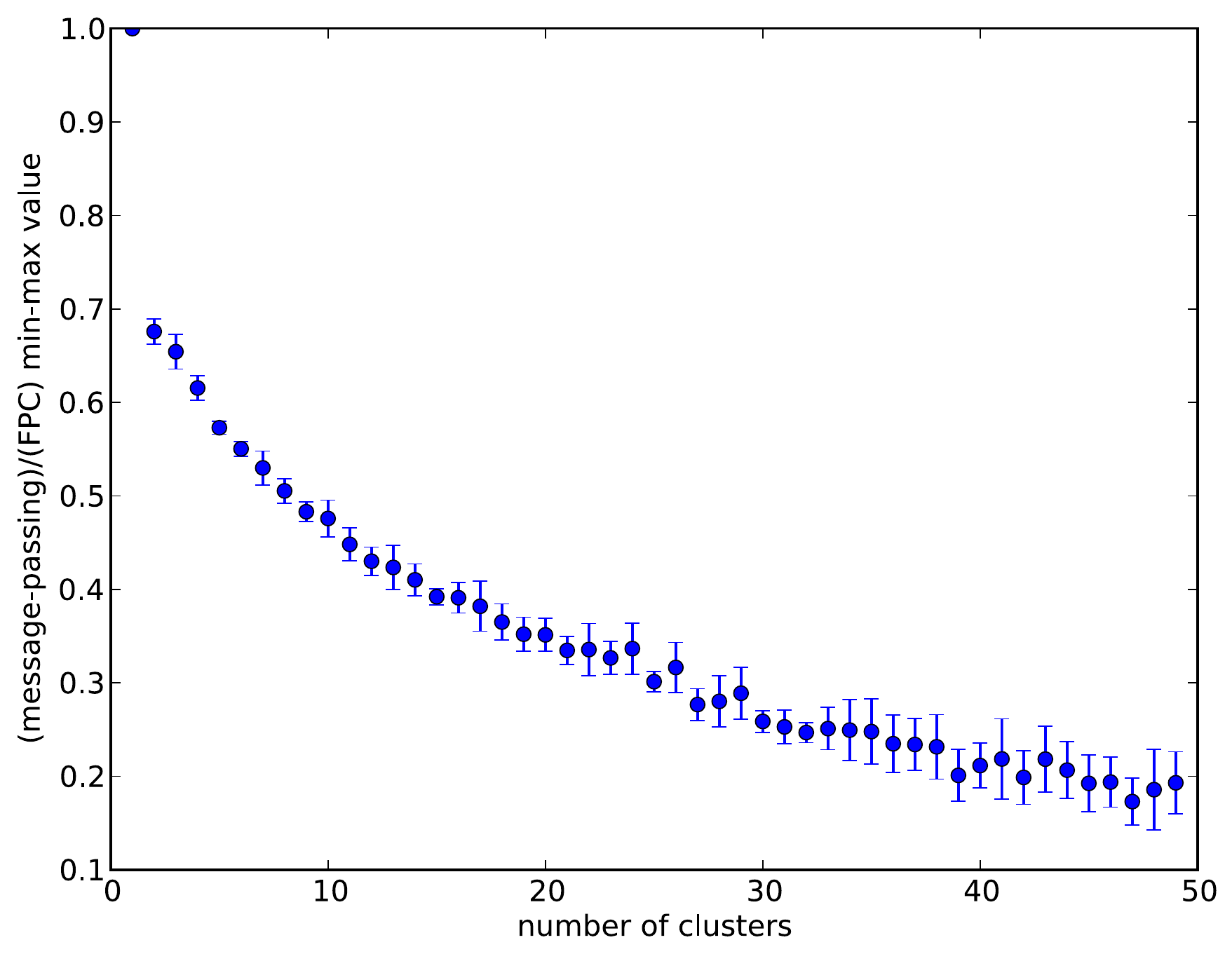}
  } 
  \caption[experimental results for K-clustering.]{Min-max clustering of 100 points with varying
      numbers of clusters (x-axis).  Each point is an average over 10
      random instances.  The y-axis is the ratio of the min-max value
      obtained by sum-product reduction (and using perturbed BP with $T = 50$ to find
satisfying assignments ) divided by
      the min-max value of Furthest Point Clustering (FPC).  
\textbf{(left)} Clustering of random
      points in 2D Euclidean space.  The red line is the lower bound
      on the optimal result based on the worst case guarantee of FPC.
      \textbf{(right)} Using symmetric random distance matrix where
      $\Dn_{i,j} = \Dn_{j,i} \sim U(0,1)$.  }\label{fig:clusteringres}
\end{figure}

\index{furthest point clustering}
\marnote{furthest point clustering}
\RefFigure{fig:clusteringres} compares the performance of min-max
clustering using message passing to that of Furthest Point Clustering
(FPC) \cite{gonzalez1985clustering} which is $2$-optimal when the
triangle inequality holds.  Note that message passing solutions are
superior even when using Euclidean distance.

\section{K-center problem}\label{sec:k_center}
\marnote{K-center}
\index{K-center problem}
\index{clustering!K-center}
Given a pairwise distance matrix $D \in \Re^{N \times N}$, the
K-center problem seeks a partitioning of nodes, with one center per
partition such that the maximum distance from any node to the center
of its partition is minimized.  

This problem is known to be \NP-hard,
even for Euclidean distance matrices
\cite{masuyama1981computational}, however K-center has some approximation 
algorithms that apply when the distance matrix satisfies the triangle inequality
 \cite{dyer1985simple,hochbaum1985best}. The method of \citet{dyer1985simple} 
 is very similar to furthest point clustering and extends to
 weighted K-center problem in which the distance from any point to
 all the other points is scaled by its weight. The more general case
 of asymmetric distances does not allow any constant-factor
 approximation (however $o(\log(N))$-approximation exists \cite{panigrahy1998log,chuzhoy2005asymmetric}).

Here we define a factor-graph, whose min-max inference results in the 
optimal solution for K-center problem. For this consider the graph $\GG(\VV, \EE)$
induced by the distance matrix $\Dn$ s.t. $\VV = \{1,\ldots,N\}$ and
 $\Dn_{i,j} = \infty \Leftrightarrow (i,j) \notin \EE$.

\begin{figure}
\centering
      \includegraphics[width=.6\textwidth]{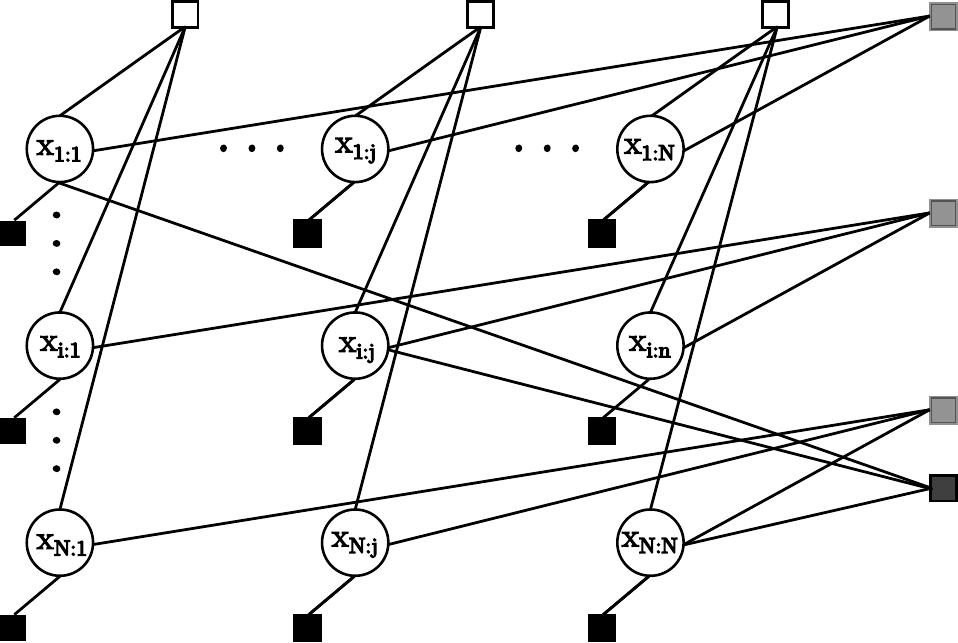}
      \caption[The factor-graph for K-center clustering.]{The factor-graph for K-center problem, where the local factors are black squares, the leader factors
      are light grey, consistency factors are white and K-of-N factor is dark grey.}
      \label{fig:k-center}
\end{figure}

Let $\xs = \{\xxx{i}{j} \mid (i,j) \in \EE\}$ be the set of variables, where $\xxx{i}{j} \in \{0,1\}$ indicates whether $j$ is the center of cluster for $i$.
Define the following factors:

\noindent \bullitem \magn{Local factors:}
\begin{align*}
\ff_{(i,j)}(\xxx{i}{j}) = \min\big( \Dn_{i,j}, \ident(\xxx{i}{j} = 0) \big ) \quad \forall (i,j) \in \EE 
\end{align*}

\index{factor!K-of-N}
\index{factor!leader}
\index{factor!consistency}
\noindent \bullitem \magn{Leader, Consistency and at-most-K-of-N factors} as defined for the sum-product case of induced K-set-cover and K-independent set (\refSection{sec:set-cover}). Here we need to replace sum-product $\ident(.)$ with the min-max version (\refEq{eq:identity}).

For variants of this problem such as the capacitated K-center, additional
constraints on the maximum/minimum points in each group may be added as
the at-least/at-most K-of-N factors.

\begin{claim}\label{th:k_center}
\marnoteloc{K-center \& K-set cover}{-1}  The $\pp_{\yy}$-reduction of the K-center clustering factor-graph above is identical to $K$-set-cover factor-graph of \refSection{sec:set-cover} for $\GG(\Dn, \yy)$.
\end{claim}
\begin{proof}
Leader, consistency and at-most-K-of-N factors are identical to the factors used for $K$-set-cover (which is identical to their $\pp_\yy$-reduction), 
where the only difference is that
here we have one variable per each $(i,j) \in \EE$ whereas in $K$-set-cover for $\GG(\Dn, \yy)$ we have one variable per $(i,j) \in \EE  \mid \Dn_{i,j} \leq \yy $.
By considering the $\pp_\yy$-reduction of the local factors 
\begin{align*}
  \ff_{i:j}(\xxx{i}{j}) = \ident \left (\min\big( \Dn_{i,j}, \ident(\xxx{i}{j)} = 0) \big ) \leq y \right ) = \ident (\Dn_{i,j} \leq y \vee \xxx{i}{j} = 0)
\end{align*}
we see that we can assume that for $\Dn{i}{j} > 0$, $\xxx{i}{j} = 0$ and we can drop these variables from the factor-graph. After this we can also omit the
$\pp_\yy$-reduction of the local factors as they have no effect. This gives us the factor-graph of \refSection{sec:set-cover} for set-cover.
\end{proof}

\marnote{complexity}
Similar to the K-clustering problem, we use the sum-product reduction to find 
near-optimal solutions to this problem. The binary search seeks the optimal $\yy \in \YY$ (where $\YY$ is the collective range of all the factors). Here, since only local factors 
take values other than $\pm \infty$, the search is over their range, which is basically
all the values in $\Dn$. This adds an additional $\log(N)$ multiplicative factor to the complexity of K-set-cover (which depends on the message update).

\begin{figure}
  \centering 
\hspace{.07\textwidth}
\hbox{ 
\includegraphics[width=.4\textwidth]{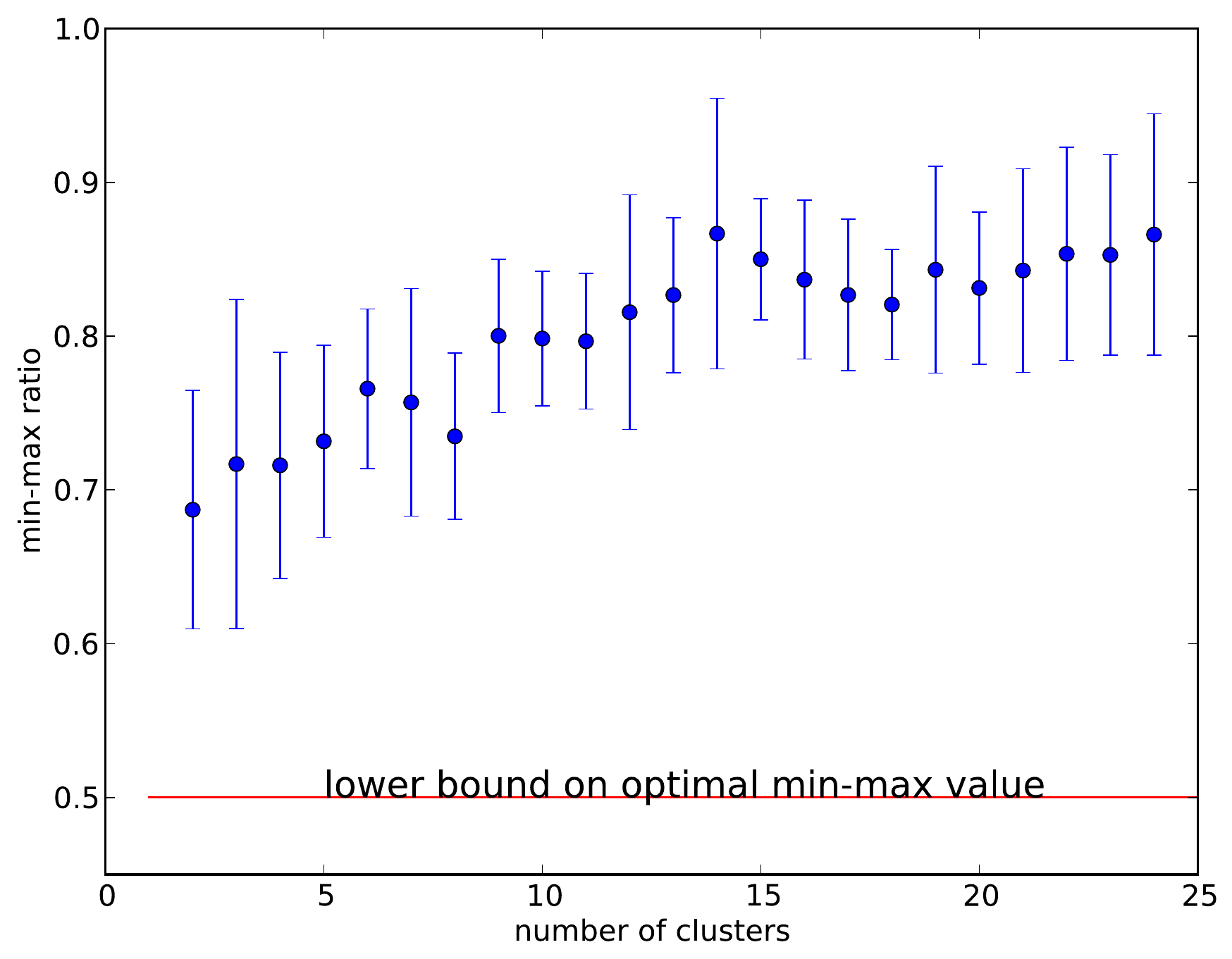}
     \includegraphics[width=.3\textwidth]{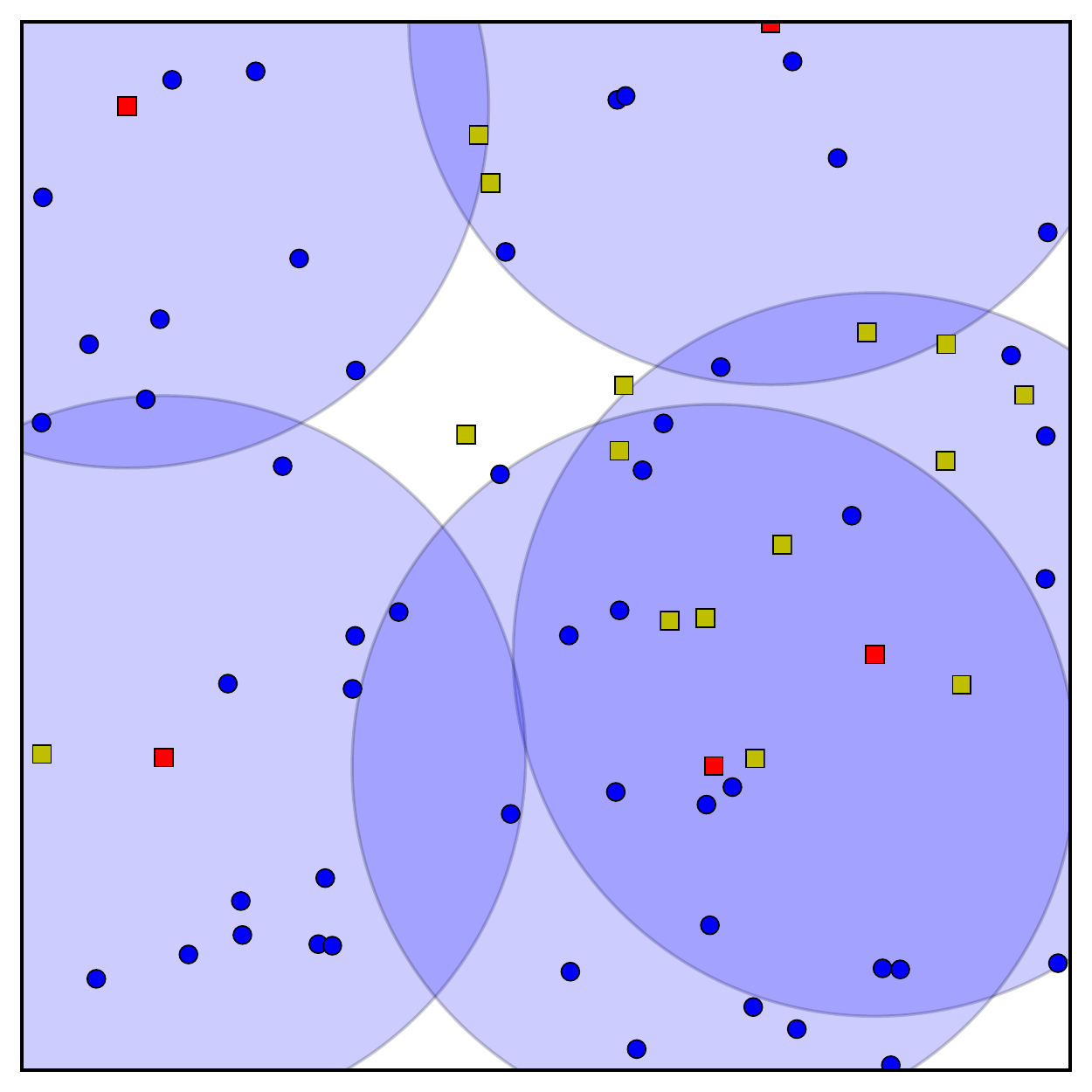}
}
\caption[Experimental results for K-center clustering and min-max facility location.]{
  \textbf{(left)} K-center clustering of 50 random
      points in a 2D plane with various numbers of clusters (x-axis). The
      y-axis is the ratio of the min-max value obtained by sum-product reduction 
      ($T = 500$ for perturbed BP) over the min-max value of
      2-approximation of \cite{dyer1985simple}.  \textbf{(right)}
      Min-max K-facility location formulated as an asymmetric K-center
      problem and solved using message passing.  Yellow squares indicate $20$
      potential facility locations and small blue circles indicate $50$
      customers.  The task is to select $5$ facilities (red squares)
      to minimize the maximum distance from any customer to a
      facility. The radius of circles is the min-max value.
}
      \label{fig:facilitylocation}
     \label{fig:kcenter_euclidean} 
\end{figure}

\marnote{using the upper bound}
We can significantly reduce the number of variables and the complexity
by bounding the distance to the center
of the cluster $\overline{\yy}$.  Given an upper bound $\overline{\yy}$,
we may remove all the variables $\xxx{i}{j}$ where $\Dn_{i,j} > \overline{\yy}$
from the factor-graph. Assuming that at most $R$ nodes are at distance
$\Dn_{i,j} \leq \overline{\yy}$ from every node $j$, the complexity of
min-max inference with synchronous update drops to $\OO(N R^2 \log(N))$. This upper bound can be obtained
for example by applying approximation algorithms.

\RefFigure{fig:kcenter_euclidean}(left) compares the performance of message-passing
and the $2$-approximation of \cite{dyer1985simple} when triangle
inequality holds. The min-max facility location problem can also be
formulated as an asymmetric $K$-center problem where the distance
\emph{to} all customers is $\infty$ and the distance from a facility
to another facility is $-\infty$ (\refFigure{fig:facilitylocation}(right)).

%% file: modularity.tex

A widely used objective for clustering (or community mining) is Modularity maximization~\cite{Newman04}. 
However, exact optimization of Modularity is \NP-hard \cite{brandes2008modularity}.
Modularity is closely related to fully connected Potts graphical models \cite{reichardt2004detecting,zhang2014scalable}. 
Many have proposed various other heuristics for modularity optimization 
\cite{Clauset05RModularity,newman2006finding,reichardt2004detecting,
ronhovde2010local,blondel2008fast}.
Here after a brief review of the Potts model in \refSection{sec:mod_potts}, we introduce a factor-graph representation of this problem 
that has a large number of factors in \refSection{sec:clique_model}.
We then use the augmentation technique (\refSection{sec:augmentation}) to incrementally incorporate violated constraints in these models.
\index{clustering!Modularity}
\index{Modularity}

\subsection{The Potts model}\label{sec:mod_potts}
\index{Modularity!Potts-model}
Let $\GG = (\VV, \EE)$ be an undirected graph, with $M = \vert \EE \vert$, $N = \vert \VV \vert$ and the adjacency matrix $\Dt \in \Re^{ N \times N}$,
where $ \Dt_{i,j} \neq 0 \Leftrightarrow  (i,j) \in \EE$.
Let $\Dn$ be the normalized adjacency matrix where $\sum_{i < j} \Dn = 1$. 
 Also let 
$\Dn_{\EE(i,\cdot)} \defeq \sum_{j} \Dn_{i,j}$ denote the normalized degree of node $v_i$.
Graph clustering using \magn{modularity} optimization
seeks a partitioning of the nodes into unspecified number of clusters $\CC = \{\CC_1,\ldots,\CC_K\}$, maximizing
\marnoteloc{Modularity}{0}
\begin{align}\label{eq:modularity}
\modul(\CC) \; \defeq \; \sum_{\CC_i \in \CC} \;\;\sum_{i,j \in \CC_i} \bigg ( \Dn_{i,j}\ -\ \zeta \Dn_{\EE(i, \cdot)}\, \Dn_{\EE(j, \cdot)} \bigg )
\end{align}
The first term of modularity is proportional to within-cluster
edge-weights. The second term is proportional to the expected number of within cluster edge-weights for a null model with the same weighted node degrees for each node $i$. 
Here the null model is a fully-connected graph. The \magn{resolution} parameter $\zeta$ -- which is by default set to one -- influences the size of communities, where higher resolutions
motivates a larger number of clusters.

In the Potts model, each node $i \in \VV$ is associated with a variable $\xx_i \in \{1,\ldots,K_{\max}\}$, where $K_{\max}$ is an upper bound on the number of clusters. Here, each pair of variables have a pairwise interaction 
\begin{align}\label{eq:mod_factor}
\ff_{\{i,j\}}(\xs_{\{i,j\}}) \quad = \quad  \left \{ 
\begin{array}{l l}
 \zeta (\Dn_{\EE(i, \cdot)}\, \Dn_{\EE(j, \cdot)}) - \Dn_{i,j}& \xx_i =\xx_j \\
0 & \xx_i \neq \xx_j
\end{array}   
\right .
\end{align}
where min-sum inference on this fully connected factor-graph gives an assignment of each node
to a cluster so as to maximize the modularity.

\subsection{Clique model} \label{sec:clique_model}
\index{Modularity!clique-model}
Here, we introduce an alternative factor-graph for modularity optimization.
Before introducing our factor-graph representation, we suggest a procedure to stochastically
approximate the null model using a sparse set of interactions.

We generate a random \textit{sparse null model} with $M^{\nulll} < \alpha M$ weighted edges ($\EE^{\nulll}$), 
by randomly sampling two nodes, each drawn independently from $\prob(i) \propto \sqrt{\Dn_{\EE(i, \cdot)}}$, and connecting them with a 
weight proportional to $\Dt^{\nulll}_{i,j} \propto \sqrt{\Dn_{\EE(i, \cdot)} \Dn_{\EE(j, \cdot)}}$. If they have been already
connected, this weight is added to their current weight.
We repeat this process $\alpha M$ times, however since some of the edges are repeated, the total number of edges in the sparse null model may be less than $\alpha M$.
Finally the normalized edge-weight in the sparse null model is
\begin{align*}
\Dn^{\nulll}_{i,j} \defeq \frac{\Dt^{\nulll}_{i,j}}{\sum_{k,l} \Dt^{\nulll}_{k,l}}.
\end{align*}

It is easy to see that this generative process in expectation produces the fully connected null model.\footnote{The choice of using square root of 
weighted degrees for both sampling and weighting is to reduce the variance. One may also use pure importance sampling (\ie use
the product of weighted degrees for sampling and set the edge-weights in the null model uniformly), or uniform sampling of edges, where the edge-weights of
the null model are set to the product of weighted degrees.} 


\subsubsection{Factor-graph representation}
Here we use the following binary-valued factor-graph formulation. 
Let $\xs = \{ \xxx{i}{j} \in \{ 0,1\} \mid (i,j) \in \EE \cup \EE^{\nulll} \} $ be a set of binary variables, 
and let $L$ denote the cardinality of $\EE \cup \EE^{\nulll}$. The variable $\xxx{i}{j}$ is equal to one means the corresponding edge, $(i,j)$, is
present in the final model, where our goal is to define the factor-graph such that the final model consists of cliques.
For this, define the factors as follows: 

\marnote{local factors}
\noindent \bullitem \emph{Local factor} for each variable are equal to the difference between the weighted adjacency and the null model
if an edge is present (\ie $\xxx{i}{j} = 1$)
\begin{align}
\ff_{i:j}(\xxx{i}{j}) =  \min \bigg (\ident(\xxx{i}{j} = 0), \Dn^{\nulll}_{i,j} - \Dn_{i,j} \bigg)
\end{align}

By enforcing the formation of cliques, while minimizing the sum of local factors
the negative sum of local factors evaluates to modularity (\cref{eq:modularity}):

\marnote{clique factors}
\noindent \bullitem For each three edges $(i,j),(j,k),(i,k) \in \EE \cup \EE^{\nulll}, i < j< k$ that form a triangle, define a \textbf{clique factor} as 
\begin{align}
\ff_{\{i:j, j:k, i:k\}}(\xxx{i}{j}, \xxx{j}{k}, \xxx{i}{k}) =  \ident(\xxx{i}{j} + \xxx{j}{k} + \xxx{i}{k} \neq 2)
\end{align}

These factors ensure the formation of cliques -- 
\ie if two edges that are adjacent to the same node are present (\ie $\xxx{i}{i} = 1$ and $\xxx{i}{k} = 1$), the third edge in the triangle should also be present (\ie $\xxx{j}{k} = 1$).
The computational challenge here is the large number of clique constraints.
For the fully connected null model, we need $\OO(N^3)$ such factors and 
even using
the sparse null model -- assuming a random edge probability \aka Erdos-Reny graph -- there are $\OO(\frac{L^3}{N^6} N^3) = \OO(\frac{L^3}{N^3})$
triangles in the graph (recall that $L = \vert \EE \cup \EE^{\nulll} \vert$).

Brandes \etal \cite{brandes2008modularity} first introduced an LP formulation with similar form of constraints. 
However, since they include
all the constraints from the beginning and the null model is fully connected, their method is only applied to small toy problems.

\subsubsection{Simplified message update and augmentation}
Here, we give technical details as how to simplify the min-sum BP message update for this factor-graph.
The clique factor is satisfied only if either zero, one, or all three of the variables in its domain are non-zero.
Therefore, in order to derive message updates $\msg{\{i:j, j:k, i:k\}}{i:j}$ from the clique factor $\ff_{\{i:j, j:k, i:k\}}$
to variable $\xxx{i}{j}$ for a particular value of $\xxx{i}{j}$ (\eg $\xxx{i}{j} = 0$), 
we apply $\bpplus$ operator (\ie minimize) over all the valid cases of incoming messages 
(\eg when $\xxx{i}{k}$ and $\xxx{j}{k}$ in clique factor $\ff_{\{i:j, j:k, i:k\}}$ are zero).
This gives the simplified factor-to-variable messages
\begin{align}
 &\msg{\{i:j, j:k, i:k\}}{i:j}(0) = 
\min \{0 ,\; \msg{j:k}{\{i:j, j:k, i:k\}} ,\; \msg{i:k}{\{i:j, j:k, i:k\}}\} \notag\\
&\msg{\{i:j, j:k, i:k\}}{i:j}(1) = 
\min \{0 ,\; \msg{j:k}{\{i:j, j:k, i:k\}} \;+\; \msg{i:k}{\{i:j, j:k, i:k\}}\}\label{eq:mIi_clustering_1}
\end{align}
where
for $\xxx{i}{j} = 0$, the minimization is over three feasible cases
(a)~$\xxx{j}{k} = \xxx{i}{k} = 0$,
(b)~$\xxx{j}{k} = 1, \xxx{i}{k} = 0$ and 
(c)~$\xxx{j}{k} = 0, \xxx{i}{k} = 1$.
For $\xxx{i}{j} = 1$, there are two feasible cases (a)~$\xxx{j}{k} = \xxx{i}{k} = 0$
and (b)~$\xxx{j}{k} = \xxx{i}{k} = 1$. 

Here we work with normalized messages, such that $\msg{\II}{i}(0) = 0$ \footnote{Note that this is different from the standard normalization for min-sum semiring in which $\min_{\xx_i} \msg{\II}{i}(\xx_i) = 0$.} and use $\msg{\II}{i}$ to denote $\msg{\II}{i}(1)$. The same applies to the marginal $\ph_{i:j}$, which is a scalar called bias. 
Here $\ph_{i:j} > 0$ means $\ph_{i:j}(1) > \ph_{i:j}(0)$ and shows a 
preference for $\xxx{i}{j} = 1$.
Normalizing clique-factor messages above we get the following form of simplified factor-to-variable messages for clique constraints
\begin{align}\label{eq:mIi_clique}
\msg{\{i:j, j:k, i:k\}}{i:j} = 
&\min \{0 ,\; \msg{j:k}{\{i:j, j:k, i:k\}} \;+\; \msg{i:k}{\{i:j, j:k, i:k\}}\} - \\
&\min \{0 ,\; \msg{j:k}{\{i:j, j:k, i:k\}} ,\; \msg{i:k}{\{i:j, j:k, i:k\}}\}.\notag
\end{align}
\index{augmentation}
\begin{figure} 
\hbox{
\includegraphics[width=.5\textwidth]{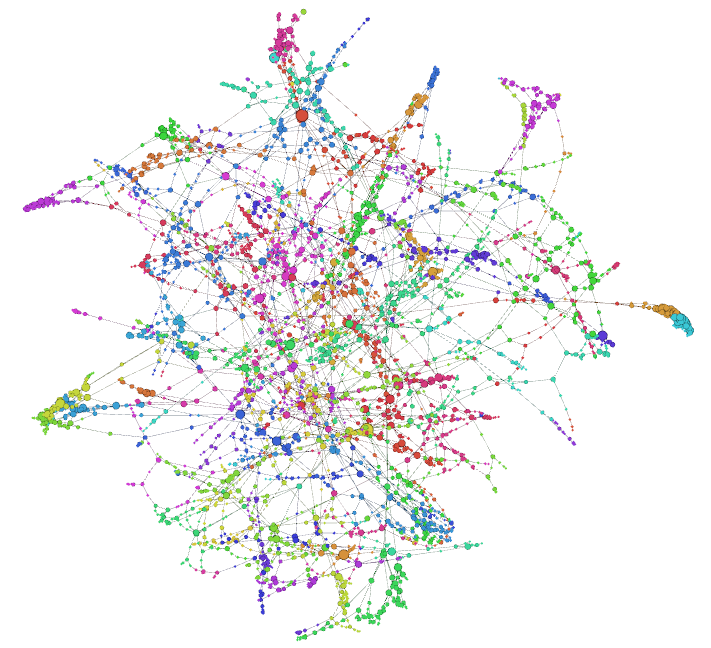}
\includegraphics[width=.5\textwidth]{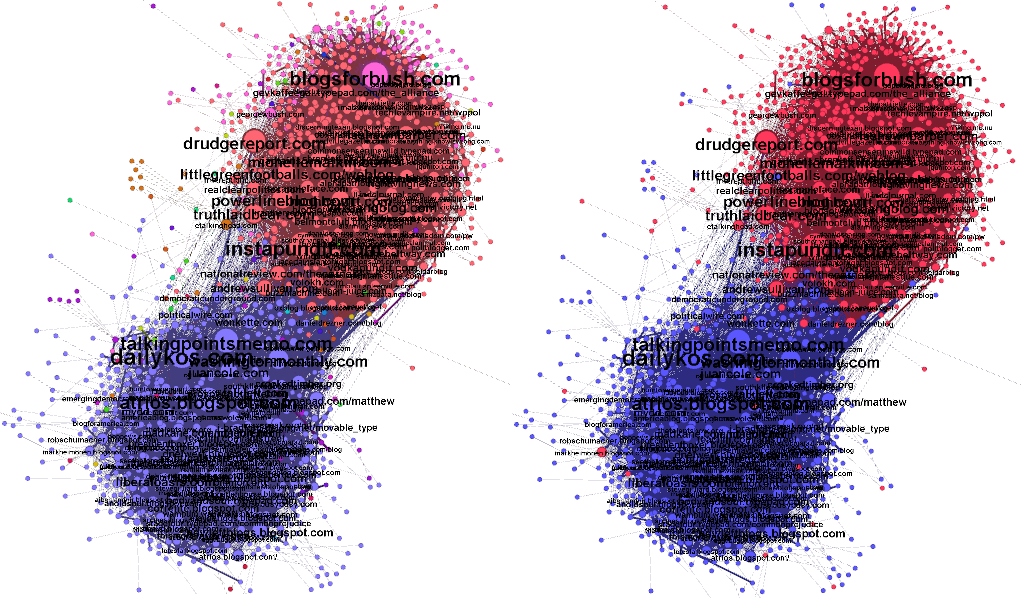}
}
\caption[Example of maximum Modularity clustering by message passing.]{
(left) Clustering of power network ($N = 4941$) by message passing. Different clusters have different colors and the nodes are scaled by their degree.
(right) Clustering of politician blogs network ($N = 1490$) by message passing and by meta-data -- \ie liberal or conservative.
}
\label{fig:power}
\label{fig:polblogs}
\end{figure}

In order to deal with large number of factors in this factor-graph, we use the augmentation approach of \refSection{sec:augmentation}.
We start with no clique-factors, and run min-sum BP to obtain a solution (which may even be unfeasible).
We then find a set of clique constraints that are violated in the current solution and augment the factor-graph with factors to enforce these constraints. 
In order to find violated constraints in the current solution,
we simply look at pairs of positively fixed
edges ($\xxx{i}{j} = 1$ and $\xxx{i}{k} = 1$) around each node $i$ and if the third edge of the triangle is not positively fixed ($\xxx{j}{k} = 0$),
we add the corresponding clique factor ($\ff_{\{i:j, j:k, i:k\}}$) to the factor-graph. See \cref{alg:clustering} (in the appendix) for details of our message-passing for Modularity maximization.

\begin{table}[pht!]
  \caption{Comparison of different modularity optimization methods.}\label{table:clustering}
  \begin{center}
    \scalebox{.5}{
    \begin{tabu}{c | c | c | c |[2pt] r | c | c | l |[2pt] r |c | c | l |[2pt] r| l|[2pt] r| l |[2pt] r| l |[2pt] r| l |}
      \cline{5-20}
      & & & & \multicolumn{4}{ |c |[2pt]}{message passing (full)}&\multicolumn{4}{ |c |[2pt]}{message passing (sparse)}& \multicolumn{2}{ c|[2pt] }{Spin-glass}&\multicolumn{2}{ c|[2pt] }{L-Eigenvector}
      &   \multicolumn{2}{ c|[2pt] }{FastGreedy}&   \multicolumn{2}{ c|[2pt] }{Louvian}\\
      \cline{5-20}
      \begin{sideways}Problem \end{sideways}&
      \begin{sideways} Weighted? \end{sideways}&
     \begin{sideways} Nodes \end{sideways}&
      \begin{sideways} Edges \end{sideways}&
      \begin{sideways} $L$ \end{sideways} &
      \begin{sideways} Cost \end{sideways} &
      \begin{sideways} Modularity \end{sideways} &
      \begin{sideways} Time \end{sideways} &
      \begin{sideways} $L$ \end{sideways} &
      \begin{sideways} Cost \end{sideways} &
      \begin{sideways} Modularity \end{sideways} &
      \begin{sideways} Time \end{sideways} &
      \begin{sideways} Modularity \end{sideways} &
      \begin{sideways} Time \end{sideways} &
      \begin{sideways} Modularity \end{sideways} &
      \begin{sideways} Time \end{sideways} &
      \begin{sideways} Modularity \end{sideways} &
      \begin{sideways} Time \end{sideways} &
 	  \begin{sideways} Modularity \end{sideways} &
      \begin{sideways} Time \end{sideways} 
      \\\tabucline[2pt]{-}

polbooks&y&105&441 
&5461 & 5.68$\%$&0.511&.07
&3624 & 13.55$\%$ &0.506&.04
&0.525&1.648
&0.467&0.179	      
&0.501&0.643
&0.489&0.03 \\

football&y&115&615
&6554&27.85$\%$&0.591&0.41
&5635&17.12$\%$&0.594&0.14
&0.601&0.87
&0.487&0.151
&0.548&0.08
&0.602&0.019 \\
wkarate&n&34&78
&562&12.34$\%$&0.431&0
&431&15.14$\%$&0.401&0
&0.444&0.557
&0.421&0.095
&0.410&0.085
&0.443&0.027 \\
netscience&n&1589&2742
&NA&NA&NA&NA
&53027&\textbf{.0004$\%$}&0.941&2.01
&0.907&8.459
&0.889&0.303
&0.926&0.154
&0.948&0.218 \\
dolphins&y&62&159
&1892&14.02$\%$&0.508&0.01
&1269&6.50$\%$&0.521&0.01
&0.523&0.728
&0.491&0.109
&0.495&0.107
&0.517&0.011 \\
lesmis&n&77&254
&2927&5.14$\%$&0.531&0
&1601&1.7$\%$&0.534&0.01
&0.529&1.31
&0.483&0.081
&0.472&0.073
&0.566&0.011 \\
 celegansneural&n&297&2359
&43957&16.70$\%$&0.391&10.89
&21380&3.16$\%$&0.404&2.82
&0.406&5.849
&0.278&0.188
&0.367&0.12
&0.435&0.031 \\
polblogs&y&1490&19090
&NA&NA&NA&NA
&156753&\textbf{.14$\%$}&0.411&32.75
&0.427&67.674
&0.425&0.33
&0.427&0.305
&0.426&0.099 \\
karate&y&34&78
&562&14.32$\%$&0.355&0
&423&17.54$\%$&0.390&0
&0.417&0.531
&0.393&0.086
&0.380&0.079
&0.395&0.009

        \\\tabucline[2pt]{-}

    \end{tabu}
		}
  \end{center}
\end{table}

\subsubsection{Experiments}
We experimented with a set of classic benchmarks%
\footnote{Obtained form Mark Newman's website:
\url{http://www-personal.umich.edu/~mejn/netdata/}
}.
Since the optimization criteria is modularity,
we compared our method only against best known
``modularity optimization'' heuristics: 
(a)~FastModularity\cite{Clauset05RModularity}, (b)~Louvain~\cite{blondel2008fast}, (c)~Spin-glass ~\cite{reichardt2004detecting} and (d)~Leading eigenvector~\cite{newman2006finding}. 
\footnote{For message passing, we use $\lambda = .1$, $\epsilon_{\max} =  \mathrm{median}\{\vert \Dn_{i,j} - \Dn^{\nulll}_{i,j}\vert \}_{(i,j) \in \EE \cup \EE^{\nulll}}$ and
$T_{\max} = 10$. Here we do not perform any decimation and directly fix the variables based on their bias $\ph_{i:j} > 0 \Leftrightarrow \xxx{i}{j} = 1$. }

\refTable{table:clustering} summarizes our results (see also Figure~\ref{fig:power}). 
Here for each method and each data-set, we report the \textit{time} (in seconds) and the \textit{Modularity} of the communities found by each method.
The table include the results of message passing for both full and sparse null models, 
where we used a constant $\alpha = 20$ to generate our stochastic sparse null model. For message passing, we also included $L = \vert \EE + \EE^{\nulll}\vert$ and the saving in the \textit{cost} using augmentation. This column shows
the percentage of the number of all the constraints considered by the augmentation.
For example, the cost of {\tt .14\% } for the polblogs data-set shows that
augmentation and sparse null model meant using
{\tt .0014} times fewer clique-factors, compared to the full factor-graph.

Overall, the results suggest that our method is comparable to state-of-the-art
in terms both time and quality of clustering. Although, we should note that the number of triangle constraints in large and dense graphs
increases very quickly, which deteriorates the performance of this approach despite using the augmentation.
Despite this fact, our results confirm the utility of augmentation by showing that
it is able to find feasible solutions using
a very small portion of the constraints.

%% file: matching.tex
The integration and maximization problems over unrestricted permutations define several important combinatorial problems. 
Two notable 
examples of integration problems are \magn{permanent and determinant} of a matrix.
\index{determinant}
\index{permanent}
Determinant of matrix $\Dn \in \Re^{N \times N}$ is defined as 
  $$\mathsf{det}(\Dn) \; = \; \sum_{\xs \in \sn} \mathsf{sign}(\xs) \prod_{i = 1}^{N} \Dn_{i, \xx_i}$$
\marnoteloc{symmetric group}{-1.5}
where $\sn$ is the set of all permutations of $N$ elements (\aka symmetric group) and $\xx_i \in \{1,\ldots, N\}$ is the index of 
$i^{th}$ element in particular permutation $\xs$.
Here the $\mathsf{sign}(.)$ classifies permutations as even ($\mathsf{sign}(\xs) = 1$) and odd ($\mathsf{sign}(\xs) = -1$), where   
we can perform an even (odd) permutation by even (odd) number of pairwise exchanges. The only difference in definition of permanent is removal of the sign function
\marnote{permanent}
\begin{align*}
  \mathsf{perm}(\Dn) \; = \; \sum_{\xs \in \sn} \prod_{i = 1}^{N} \Dn_{i, \xx_i} 
\end{align*}

Here, we see that both permanent and determinant are closely related with two 
easy combinatorial problems on graphs -- \ie perfect matching and spanning sub-tree. 
While calculating the permanent for $\Dn \in \{0,1\}^{N\times N}$ is \sharpP-hard  \cite{valiant1979complexity}, the determinant can be obtained in $\OO(N^3)$ \cite{golub2012matrix}.

\marnote{matrix tree theorem}
\index{matrix-tree theorem}
The \magn{matrix-tree theorem} states that the number of spanning trees in a graph with adjacency
matrix $\Dn$ is equal to $\mathsf{det}(\mathbf{L}(i,i))$ for an arbitrary $1 \leq i \leq N$.
Here $\mathbf{L} = \Dn - \mathbf{D}$ is the Laplacian of $\Dn$, 
\index{Laplacian}
where $\mathbf{D}$ is a diagonal matrix with degree of each node on the diagonal (\ie $\mathbf{D}_{i,i} = \sum_{j} \Dn_{i,j}$) and $\mathbf{L}(i,i)$ is the $(N-1)\times(N-1)$ sub-matrix of $\mathbf{L}$
in which row and column $i$ are removed.

An intermediate step in representing the permutation as a graphical model is to use a \magn{bipartite graph} $\GG = (\VV = (\VV_1, \VV_2), \EE)$, where $\vert \VV_1 \vert = \vert \VV_2 \vert = N$
and the edge-set $\EE = \{ (i,j) \mid i \in \VV_1, j \in \VV_2 \}$. 
A \magn{perfect matching} is a one to one mapping of elements of $\VV_1$ to $\VV_2$ and can be represented using the corresponding edges $\EE' \subset \EE$.
It is easy to see that any perfect matching $\EE' \subset \EE$ identifies a permutation $\xs$. 
\marnoteloc{bipartite matching}{-3}
Here the maximum weighted matching (\aka assignment problem) problem is to find a perfect matching  $\xs^* = \arg_{\xs \in \sn}\max\;\prod_{i = 1}^{N} \Dn_{i, \xx_i}$, while the 
\textbf{bottleneck} assignment problem seeks $\xs^* = \arg_{\xs \in \sn}\min\;\max_{i = 1}^{N} \Dn_{i, \xx_i}$.
The factor-graph representation of the next section shows that bipartite matching  and bottleneck assignment problems correspond to the max-product (min-sum) and min-max inference,
and computation of permanent corresponds to sum-product inference over the same factor-graph.

Interestingly min-sum and min-max inference in this setting are in \Poly~\cite{kuhn1955hungarian,gross1959bottleneck} while sum-product is in \sharpP.
Indeed the application of max-product BP to find the maximum weighted matching \cite{bayati2005maximum} (and its generalization to maximum weighted $b$-matching~\cite{huang2007loopy}) 
is one of the few cases in which loopy BP is guaranteed to be optimal.
Although MCMC methods (\refSection{sec:mcmc}) can provide polynomial time approximation schemes for permanent~\cite{jerrum2004polynomial} (and many other combinatorial integration problems~\cite{jerrum1996markov}),
\index{MCMC}
they are found to be slow in practice~\cite{huang2009approximating}. 
This has motivated approximations using deterministic variational techniques~\cite{watanabe2010belief,Chertkov2013} and in particular BP~\cite{huang2009approximating},
which is guaranteed to provide a lower bound on the permanent~\cite{vontobel2013bethe}.

\subsection{Factor-graph and complexity}\label{sec:matching_factor}
\index{bipartite matching}
Here we review the factor-graph of~\citet{bayati2005maximum} for maximum bipartite matching.
Given the bipartite graph $\GG = ((\VV_1,\VV_2), \EE)$ and the associated matrix $\Dn \in \Re^{N \times N}$ with non-negative entries,
define two sets of variables $\xs = \{\xx_i \in \VV_2 \mid i \in \VV_1\}$ and $\zs = \{ \zz_j \in \VV_1 \mid j \in \VV_2\}$, where $\xx_i = j$ and $\zz_j = i$
both mean  node $i$ is connected to node $j$.
Obviously this representation is redundant and for $(i,j) \in \EE$ a pairwise factor should ensure $\xx_i$ and $\zz_i$ are consistent:
\index{variable!auxiliary}

\begin{figure}[pth!]
\centering
\includegraphics[width=.5\textwidth]{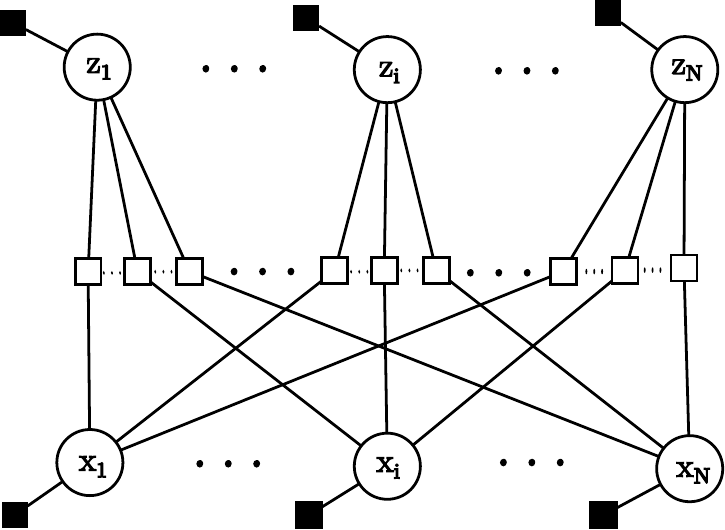}
\caption[The factor-graph for max-weighted matching]{The factor-graph for matching where the local factors are black and consistency factors are white squares.}
\label{fig:bipartite}
\end{figure}

\index{factor!consistency}
\marnoteloc{consistency factors}{-.5}
\noindent \bullitem \magn{Consistency factors} ensure the consistency of $\xs$ and $\zs$
\begin{align*}
  \ff_{\{i,j\}}(\xx_i, \zz_j) \; = \; \ident \big ( (\xx_i =j \wedge \zz_j = i) \vee (\xx_i \neq j \wedge \zz_j \neq i) \big ) \quad \forall (i,j) \in \EE
\end{align*}

\marnote{local factors}
\noindent \bullitem \magn{Local factors} represent the cost of a matching
\begin{align*}
  \ff_{\{i\}}(\xx_i) \;& = \; \Dn_{i, \xx_i} \quad& \forall i \in \VV_1\\
\end{align*}
where if $i$ and $j$ are connected in a matching, two local factors $\ff_{\{i\}}(\xx_j) = \sqrt{\Dn_{i, j}}$ and $\ff_{\{j\}}(\xx_i) = \sqrt{\Dn_{i, j}}$ account for $\Dn_{i,j}$.
\RefFigure{fig:bipartite} shows this factor-graph.


It is easy to see that the joint form $\qq(\xs, \zs)$  is equal to $\bigotimes_{i = 1}^{N} \Dn_{i, \xx_i}$ for any consistent assignment to $\xs,\zs$ and it is equal to $\identt{\oplus}$ otherwise.
Therefore sum-product and max-product (\ie $\sum_{\xs, \zs}\qq(\xs, \zs)$ and $\max_{\xs, \zs}\qq(\xs, \zs)$) produce the permanent and max-weighted matching respectively.

The cost of each iteration of both max-product and sum-product BP in the factor-graph above is $\OO(N^2)$.
Moreover, for max-product BP, if the optimal solution is unique, BP is guaranteed to converge
to this solution after $\OO(\frac{N y^*}{\epsilon})$ iterations, where $\epsilon$ is the difference
between the cost of first and second best matchings and $y^*$ is the cost of best matching~\cite{bayati2005maximum}.

An alternative is to use a \magn{binary variable model} in which each edge of the bipartite graph
is associated with a binary variable and replace the consistency factors with  \magn{degree constraint} to ensure that
each node is matched to exactly one other node.
However this model results in BP updates equivalent to the one above (the simplification of updates discussed in \cite{bayati2008max} is exactly the updates
of binary variable model).

\marnoteloc{bottleneck assignment}{-1}
For min-max semiring, $\qq(\xs, \zs) = \max_{i = 1}^{N} \Dn_{i, \xx_i}$ for a consistent assignment to $\xs, \zs$ and it evaluates to $\identt{\min} = \infty$. Therefore min-max inference here seeks an assignment that minimizes the maximum matching cost -- \aka \magn{bottleneck assignment} problem.

\subsection{Arbitrary graphs}\label{sec:matching_arbitrary}

\index{matching}
We can also use message passing to solve max-weighted matching 
in an arbitrary graph $\GG = (\VV, \EE)$ with adjacency matrix $\Dn$.
\citet{zdeborova2006number} proposed a factor-graph for the related task of counting of the perfect matchings in an arbitrary graph.
For this, each edge $(i,j) \in \EE$ is assigned to one binary variable $\xxx{i}{j} \in \{0,1\}$ and
 the \emph{degree factors} on each node restrict the number of non-zero values to one
\index{factor!degree}
 \begin{align*}
   \ff_{\EE(i,.)}(\xs_{\EE(i,.)}) \quad = \quad \ident((\sum_{(i,j) \in \EE(i,.)} \xxx{i}{j}) \leq 1)
 \end{align*}
where $\EE(i,.)$ is the set of all the edges adjacent to node $i$.

\citet{sanghavi2007equivalence} consider the problem of maximum weighted b-matching in arbitrary graphs by changing the degree factor to 
 \begin{align*}
   \ff_{\EE(i,.)}(\xs_{\EE(i,.)}) \quad = \quad \ident((\sum_{(i,j) \in \EE(i,.)} \xxx{i}{j}) \leq b)
 \end{align*}
and also \emph{local factor} $\ff_{i:j}(\xxx{i}{j}) = \xxx{i}{j} \Dn_{i,j}$
that takes the weights into account.
They show that if the solution to the corresponding LP relaxation is integral
then BP converges to the optimal solution. Moreover, BP does not converge if
the LP solution is not integral. 

\marnote{cycle-cover \& perfect matching}
\index{cycle-cover}
\index{perfect matching}
Matchings in an arbitrary graph $\GG$ is also related to the permanent and also (vertex disjoint) \magn{cycle covers}; a set of directed cycles in $\GG$ that cover all of its nodes exactly once. The number of such cycle covers in an un-weighted graph is equal to 
$\mathsf{perm}(\Dn)$, which is in turn equal to the square of number of perfect matchings \cite[ch. 13]{moore2011nature}.

In fact a directed cycle cover with maximum weight is equivalent to the
maximum weighted bipartite matching in the construction of the previous section.
In \refSection{sec:tsp} below we will use message passing to obtain a minimum weighted ``undirected'' cycle cover and further restrict these covers to obtain a minimum weighted cover with a single cycle -- \ie a minimum tour for TSP.

%% file: tsp.tex
\index{traveling salesman problem}
A Traveling Salesman Problem (TSP) seeks the minimum 
length tour of $N$ cities that visits each city exactly once. 
TSP is \NP-hard and 
for general distances, no constant factor approximation to this problem is possible \cite{papadimitriou1977euclidean}. 
The best known exact solver,
\index{dynamic programming}
due to \citet{held_dynamic_1962}, uses dynamic programming to reduce the cost of enumerating all orderings from
$\OO(N!)$ to $\OO(N^2 2^N)$.
\marnote{optimization \& TSP}
The development of many (now) standard optimization techniques are closely linked with advances in solving TSP.
Important examples are simulated annealing \cite{vcerny1985thermodynamical,kirkpatrick1983optimization},
mixed integer linear programming \cite{gomory1958outline}, 
\index{linear program}
dynamic programming \cite{bellman1962dynamic},
ant colony optimization \cite{dorigo1997ant} and
genetic algorithms \cite{goldberg1985alleles, grefenstette1985genetic}.

\marnote{baranch \& bound}
\index{branch and bound}
Since \citet{dantzig1954solution} manually applied the cutting plane method to 49-city problem, 
a combination of more sophisticated cuts, used with branch-and-bound techniques~\cite{padberg1991branch,balas1985branch} 
has produced the state-of-the-art TSP-solver, Concorde \cite{concorde}.
\marnote{Lin-Kernigan} 
Other notable results on very large instances have been reported by Lin-Kernigan heuristic \cite{lkh} 
that continuously improves a solution by exchanging nodes in the tour. 
For a readable historical background of the state-of-the-art in TSP and its  applications, see \cite{applegate2006traveling}.

The search over the optimal tour is a search over all permutations of $N$ cities that contains no sub-tours -- that is the permutation/tour is constrained
such that we dot not return to the starting city without visiting all other cities.
Producing the permutation with minimum cost that may include sub-tours is called the (vertex disjoint) \magn{cycle-cover} and is in \Poly\ (see \refSection{sec:matching_arbitrary}).
\index{cycle-cover}

We provide two approaches to model TSP: \refSection{sec:augmentative_tsp} presents the first approach, which  ignores the subtour constraints -- \ie finds cycle covers -- and then ``augment''
the factor-graph with such constraints when they become violated. This augmentation process is repeated until a feasible solution is found.
 The second approach, presented in \refSection{sec:btsp}, is to use the variables that represent
the time-step in which a node is visited. By having the same number of time-steps as cities, the subtour constraint is
automatically enforced. 
\marnoteloc{Hamiltonian cycle \& subgraph isomorphism}{-1}
This second formulation, which is computationally more expensive, is closely related to our
factor-graph for sub-graph isomorphism  (see \cref{sec:graphongraph}).
This is because one can think of the problem of finding a \magn{Hamiltonian cycle} in $\GG$ as
 finding a sub-graph of $\GG$ that is isomorphic to a loop of size $\vert \VV \vert$. 

\subsection{Augmentative approach}\label{sec:augmentative_tsp}
\index{augmentation}
Let $\GG = (\VV, \EE)$ denote a graph of our problem with the positively weighted symmetric adjacency matrix $\Dn$, s.t. $\Dn_{i,j} = 0 \Leftrightarrow (i,j) \notin \EE$.
The objective is to select a subset of $\EE$ that identifies  shortest tour of $N$ cities.  
Let $\xs = \{ \xxx{i}{j} \mid (i,j) \in \EE\}$ be a set of $M$ binary variables (\ie $\xxx{i}{j}\in \{0,1\}$), one for each edge in the graph (\ie $M = \vert \EE \vert$) where 
$\xxx{i}{j} = 1$ means $(i,j)$ is in the tour. We use $\xxx{i}{j}$ and $\xxx{j}{i}$ to refer to the same variable.
Recall that for each node $i$, $\EE(i,\cdot)$ 
denotes the edges adjacent to $i$.
Define the factors of the factor-graph as follows

\marnote{local factors}
\noindent \bullitem \magn{Local factors} represent the cost associated with each edge
\begin{align}
\ff_{i:j}(\xxx{i}{j})\ =\  \min \big ( \ident(\xxx{i}{j} = 0), \Dn_{i,j}\big) \quad \forall (i,j) \in \EE
\end{align}
where $\ff_{i:j}(\xxx{i}{j})$ is 
either $\Dn_{i,j}$ or zero. 

\index{Held-Karp constraints}
\marnoteloc{Held-Karp constraints}{-1}
Any valid tour satisfies the following necessary and sufficient constraints -- \aka \magn{Held-Karp constraints} \cite{held1970traveling}:
\\[1ex]
\index{factor!degree}
\marnote{Degree factors}
\noindent \bullitem \magn{Degree factors} ensure that exactly two edges that are adjacent to each vertex are in the tour 
\begin{align}
\ff_{\EE(i, \cdot)}(\xs_{\EE(i, \cdot)}) \; = \; \ident\big ( (\sum_{(i,j) \in \EE(i, \cdot)} \xxx{i}{j}) = 2 \big )
\end{align}
\index{factor!K-of-N}

\index{subtour constraints}
\noindent \bullitem \magn{Subtour factors} 
ensure that there are no short-circuits -- 
\ie there are no loops that contain strict subsets of nodes. 
To enforce this, for each 
 $\SS \subset \VV$, define 
$\EE(\SS, .)\; \defeq \; \{(i,j) \in \EE  \mid  i \in \SS , j \notin \SS \}$
to be the set of edges, with one end in $\SS$ and the other end in $\VV \back \SS$.
We need to have at least two edges leaving each subset $\SS$. 
The following set of factors enforce these constraints
\marnote{subtour factors}
\begin{align}
\ff_{\EE(\SS, \cdot)}(\xs_{\EE(\SS, \cdot)})\ =\ \ident\big((\sum_{(i,j) \in \EE(\SS, \cdot)} \xxx{i}{j}) \geq 2 \big)\quad  
\forall\, \SS \subset \VV ,\ \  \SS \neq \emptyset
\end{align}
These three types of factors define a factor-graph, 
whose minimum energy configuration 
is the smallest tour for TSP. Therefore we can use min-sum inference to obtain the optimal tour.
Note that both subtour and degree constraints depend on large number of variables, however, due to sparsity they allow
efficient linear time calculation of factor-to-variable messages; see \refSection{sec:tsp_simplified}.
The more significant computational challenge is that the complete TSP factor-graph has $\OO(2^{N})$ subtour factors, one for each subset of variables.
In \refSection{sec:tsp_augmentation} we address this problem using factor-graph augmentation.



\subsubsection{Simplified messages}\label{sec:tsp_simplified}
In \refSection{sec:hop}, we introduced the K-of-N factors for min-sum inference.
Both degree and subtour factors are different variations of this types of factor.
For simplicity we work with
normalized message $\msg{\II}{i:j} = \msg{\II}{i:j}(1) - \msg{\II}{i:j}(0)$,
which is equivalent to assuming  $\msg{\II}{i:j}(0) = 0\  \forall \II, i:j \in \nb \II$.
The same notation is used for variable-to-factor message, and marginal belief. 
As before, we refer to the normalized marginal belief, $\ph_{i:j} = \ph(\xxx{i}{j} = 1) - \ph(\xxx{i}{j} = 0) $ as bias.


Recall that a \emph{degree constraint} for node $i$ ($\ff_{\EE(i, \cdot)}$) depends on all the variables $\xxx{i}{j}$ for edges $(i,j)$ that are adjacent to $i$.
Here we review the factor-to-variable message for min-sum BP 
\begin{align}\label{eq:mIi_degree_temp}
  \msg{\EE(i, \cdot)}{i:j}(\xxx{i}{j}) \quad = \quad \min_{\xs_{\back i:j}} \;\ff_{\EE(i, \cdot)}(\xs_{\EE(i,\cdot)}) \; \sum_{(i,k) \in \EE(i, \cdot) \back (i,j)} \msg{i:k}{\EE(i, \cdot)}(\xxx{i}{k}) 
\end{align}

We show that this message update simplifies to 
\begin{align}
  \msg{\EE(i, \cdot)}{i:j}(1)  \quad &= \quad \min \big ( \msg{i:k}{\EE(i, \cdot)} \mid (i,k) \in \EE(i, \cdot) \back (i,j)\big ) \quad \forall i \in \VV\\
  \msg{\EE(i, \cdot)}{i:j}(0) \quad &= \quad  \min \big ( \msg{i:k}{\EE(i, \cdot)} + \msg{i:l}{\EE(i, \cdot)} \mid (i,k), (i,l) \in \EE(i, \cdot) \back (i,j) \big )\label{eq:mIi_degree_both}
\end{align}
where 
for $\xxx{i}{j} = 1$,
in order to satisfy degree constraint $\ff_{\EE(i, \cdot)}(\xs_{\EE(i, \cdot)})$, only \textit{one} other $\xxx{i}{k}$ for  $(i,k) \in \EE(i, \cdot) \back (i,j)$ should be non-zero. 
On the other hand, we know that messages are normalized
such that $\msg{i:j}{\EE(i, \cdot)}(0) = 0$,
which means they can be ignored in the summation of \cref{eq:mIi_degree_temp}. 
For $\xxx{i}{j} = 0$, in order to satisfy the constraint factor, \textit{two} of the adjacent variables should have a non-zero value. 
Therefore we seek
two such incoming messages with minimum values. 
Let $\min^{k} \settype{A}$ denote the $k^{th}$ smallest value in the set $\settype{A}$ -- \ie 
$\min \settype{A} \equiv \min^1 \settype{A}$.
We combine the updates above 
to get a ``normalized message'', $\msg{\EE(i, \cdot)}{i:j}$, which is simply the negative of the second largest incoming message (excluding $\msg{\EE(i, \cdot)}{i:j}$) 
to the degree factor $\ff_{\EE(i, \cdot)}$: 
\begin{align}
\msg{\EE(i, \cdot)}{i:j} = \msg{\EE(i, \cdot)}{i:j}(1) - \msg{\EE(i, \cdot)}{i:j}(0) 
= - \min^{2} \{\msg{i:k}{\EE(i, \cdot)} \mid (i,k) \in \EE(i, \cdot) \back (i,j)\} \label{eq:mIi_degree}
\end{align}

Following a similar procedure, factor-to-variable messages for \emph{subtour factors} is given by
\begin{align}
  \msg{\EE(\SS, \cdot)}{i:j} = -\max \bigg (0 , \min^{2} \{\msg{i:k}{\EE(\SS, \cdot)} \mid (i,k) \in \EE(\SS, \cdot) \back (i,j)\} \bigg )\label{eq:mIi_subtour} 
\end{align}

While we are searching for the minimum incoming message, if we encounter two messages with negative or zero values, we can safely assume $\msg{\EE(\SS, \cdot)}{i:j} = 0$, 
and stop the search. This results in significant speedup in practice. 
Note that both \cref{eq:mIi_degree} and \cref{eq:mIi_subtour} only need to calculate the second smallest incoming message to their corresponding factors, less the current outgoing message.
In the asynchronous calculation of messages, this minimization should be repeated for each outgoing message. However in a factor-synchronous update,
by finding \emph{three} smallest incoming messages to each factor,
we can calculate all the factor-to-variable messages at the same time.

\begin{figure}[!pth] 
\includegraphics[width=1\textwidth]{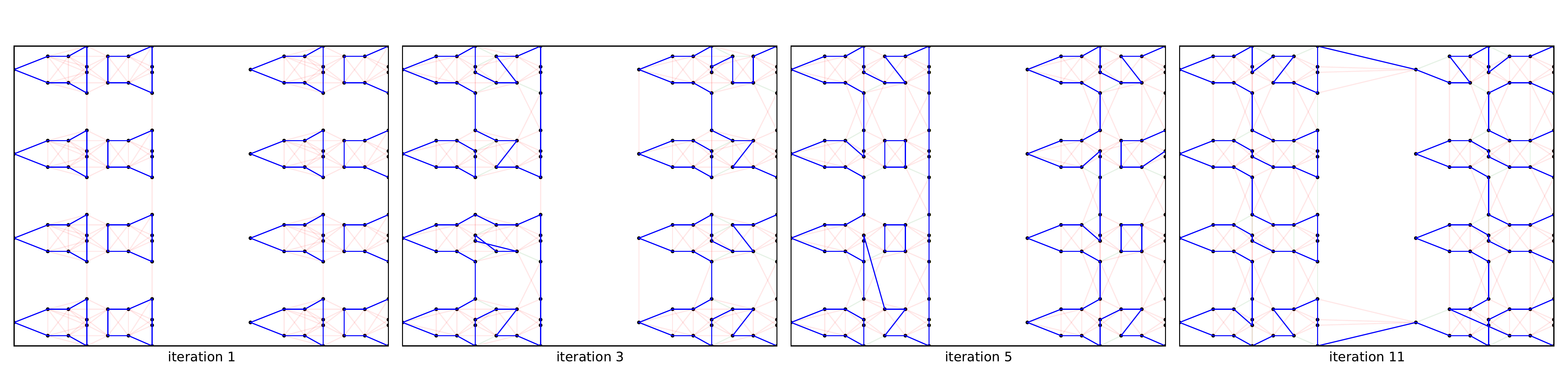}
\caption[An example of application of augmentative message passing to solve TSP.]{The message passing results after each augmentation step for the complete graph of printing board instance from \cite{tsplib}. 
The blue lines in each figure show the selected edges at the end of message passing. 
The pale red lines show the edges with the bias that,
although negative ($\ph_{i:j} < 0$), were close to zero.}\label{fig:tspexample}
\end{figure}

\subsubsection{Augmentation}\label{sec:tsp_augmentation}
To deal with the exponentially large number of subtour factors, we use the augmentation procedure of \refSection{sec:augmentation}.
Starting with a factor-graph with no subtour factor, we find a solution $\xs^*$ using min-sum BP. If the solution is feasible (has no subtours) we are done.
Otherwise, we can find all subtours in $\OO(N)$ by finding connected components.
We identify all the variables in each subtour as $\SS \subset \VV$ and add a subtour factor $\ff_{\EE(\SS, \cdot)}$
to ensure that this constraint is satisfied in the next iteration of augmentation.
Here to speed up the message passing we reuse the messages from the previous augmentation step.
Moreover in obtaining $\xs^*$ from min-sum BP marginals $\ph(\xx_i)$, we ensure that no degree constraint is violated (\ie each node has two neighbouring edges in $\xs^*$).
\RefFigure{fig:tspexample} shows iterations of augmentation over a print board TSP instance from \cite{tsplib}. 
\Cref{alg:tsp} in the appendix gives details of our message passing solution to TSP, which also uses several other minor tricks to speed up the BP message updates.

\begin{figure}
\centering
\includegraphics[width=1\textwidth]{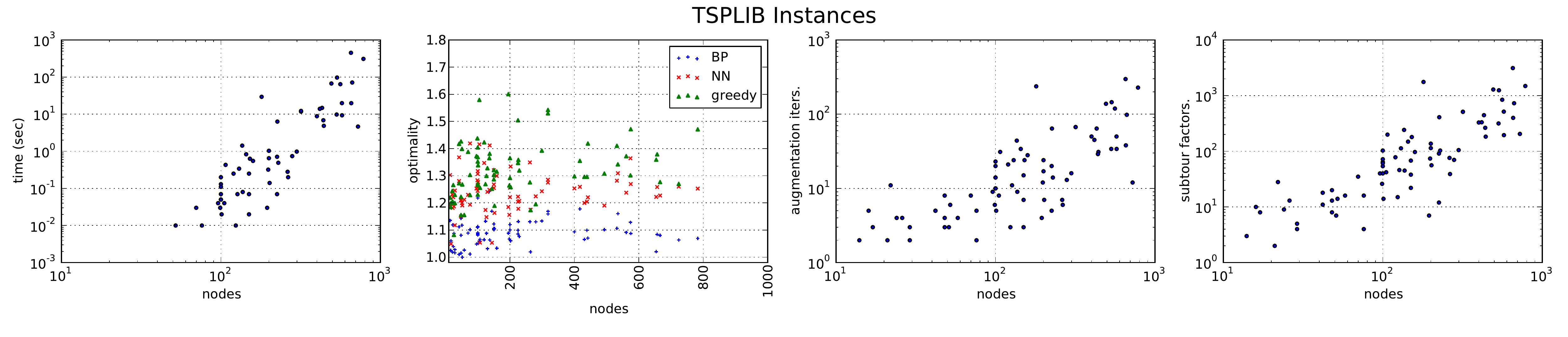}
\centering
\includegraphics[width=1\textwidth]{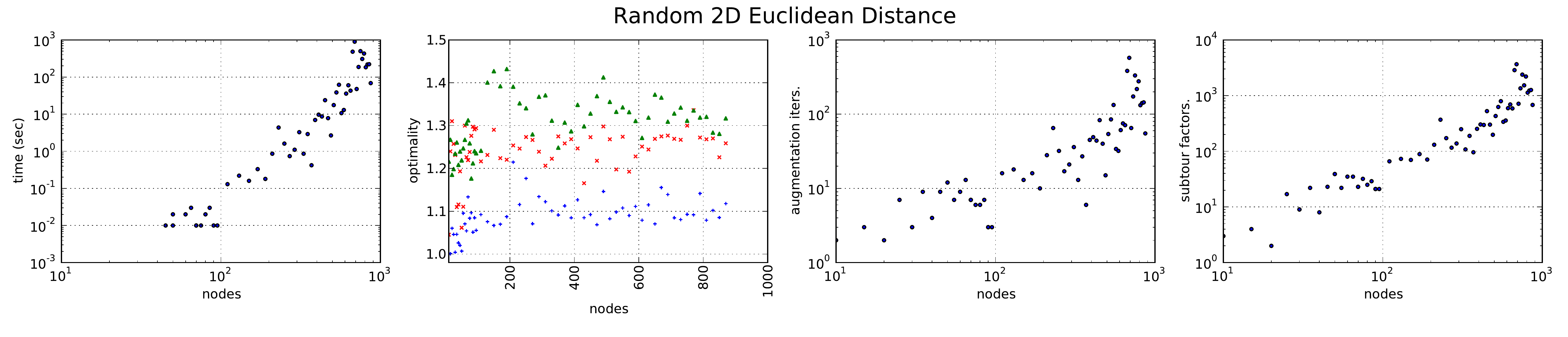}
\centering
\includegraphics[width=1\textwidth]{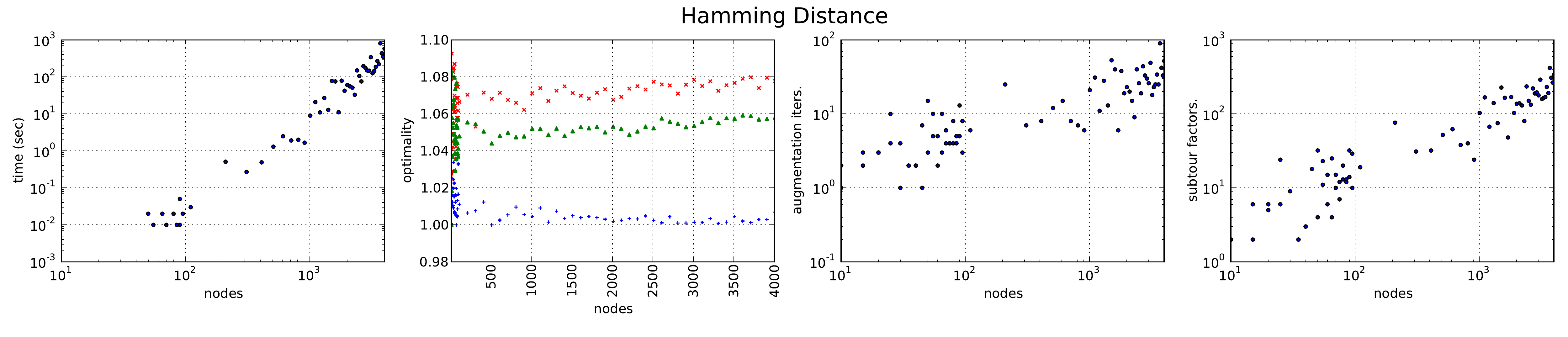}
\centering
\includegraphics[width=1\textwidth]{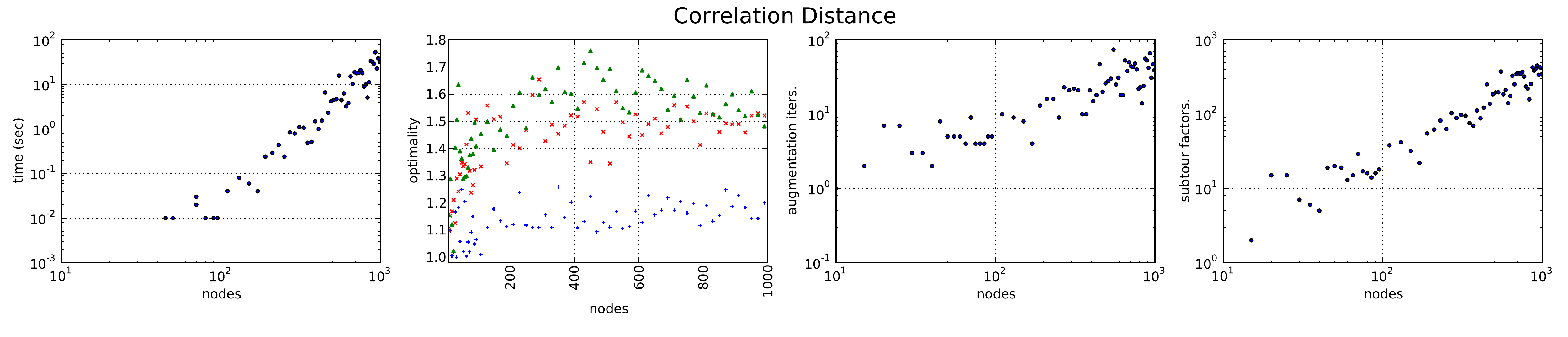}
\centering
\includegraphics[width=1\textwidth]{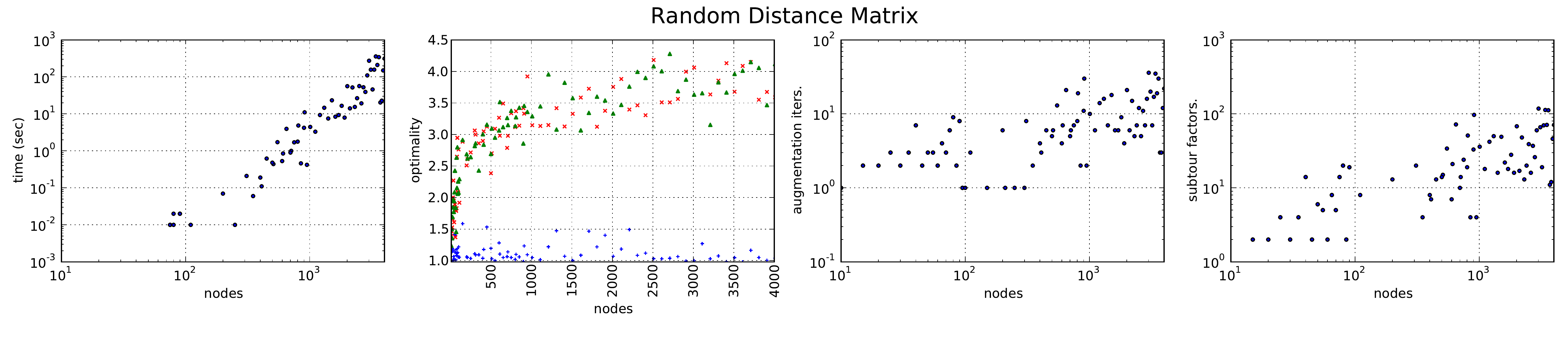}
\caption[Experimental results for TSP using message passing on various benchmark instances.]{Results of message passing for TSP on different benchmark problems.
From left to right, the plots show: (a)~running time, (b)~optimality ratio (compared to Concorde), (c)~iterations of augmentation and 
(d)~number of subtours constraints -- all as a function of number of nodes. 
The optimality is relative to the result reported by Concorde. Note that all plots except optimality are log-log plots where a linear trend shows a monomial relation ($y = a x^{m}$) between the values on the $x$ and $y$ axis, where the slope shows the power $m$.}
\label{fig:tspresults}
\end{figure}

\subsubsection{Experiments}\label{sec:experiments} 
Here we evaluate our method over five benchmark datasets\footnote{
In all experiments,
we used the full graph $\GG = (\VV, \EE)$,
which means each iteration of message passing is $\OO(N^2 \tau)$, 
where $\tau$ is the number of subtour factors.
All experiments  use $T_{\max} = 200$ iterations, $\epsilon_{\max} =  \mathrm{median}\{\Dn_{i,j} \mid i,j\}$ and damping with $\lambda = .2$. 
We used decimation, and fixed $10\%$ of the remaining variables (out of $N$) per iteration of decimation. Note that here we are only fixing the top $N$ variables with \textit{positive} bias. The remaining $M - N$ variables are automatically clamped to zero. This increases
the cost of message passing by an $\OO(\log(N))$ multiplicative factor, however it often produces better results.}:
\textbf{(I)}~TSPLIB, which contains a variety of real-world benchmark instances, the majority of which are 2D or 3D Euclidean or geographic distances.%
\footnote{Geographic distance is the distance on the surface of the earth as a large sphere.}
\textbf{(II)}~Euclidean distance between random points in 2D. 
\textbf{(III)}~Random (symmetric) distance matrices. 
\index{Hamming}
\textbf{(IV)}~Hamming distance between random binary vectors with fixed length (20 bits). This appears in applications such as data compression \cite{johnson2004compressing} and 
radiation hybrid mapping in genomics~\cite{ben1997constructing}. 
\textbf{(V)}~Correlation distance between random vectors with 5 random features (\eg using TSP for gene co-clustering \cite{climer2004take}).
In producing random points and features as well as random distances (in (III)),
we  used uniform distribution over $[0,1]$.

For each of these cases, we report the (a)~run-time, (b)~optimality,
(c)~number of iterations of augmentation and (d)~number of subtour factors at the final iteration.
In all of the experiments, we use Concorde~\cite{concorde} with its default settings to obtain the optimal solution.%
\footnote{For many larger instances, Concorde (with default setting and using CPLEX as LP solver) was not able to find the optimal solution. 
Nevertheless we used the upper bound on the optimal produced by Concord in evaluating our method.}
The results in \refFigure{fig:tspresults} (2nd column from left) 
reports the optimality ratio -- \ie ratio of the tour found by message passing, 
to the optimal tour. 
To demonstrate the non-triviality of these instance, we also report
the optimality ratio for two heuristics that have 
$(1 + \lceil \log_2(N) \rceil)/2$-
optimality guarantees for metric instances~\cite{johnson_traveling_1997}:
\marnoteloc{nearest neighbour \& greedy}{-.5}
(a) \textit{nearest neighbour} heuristic ($\OO(N^2)$), which incrementally adds  to any end of the current path the closest city that does not form a loop; 
\index{nearest neighbour}
(b) \textit{greedy} algorithm ($\OO(N^2 \log(N))$), which incrementally adds a lowest cost edge to the current edge-set, while avoiding subtours.

All the plots in \refFigure{fig:tspresults}, except for the second column, are in log-log format.
When using log-log plot, a linear trend shows a monomial relation between $x$ and $y$ axes --
\ie $y = a x ^m$.
Here $m$ indicates the slope of the line in the plot and the intercept corresponds to $\log(a)$. 
By studying the slope of the linear trend in the run-time (left column) in \refFigure{fig:tspresults},
we observe that, for almost all instances,
message passing seems to grow with $N^3$ (\ie slope of $\sim 3$). 
Exceptions are TSPLIB instances,
which seem to pose a greater challenge, and
random distance matrices which 
seem to be easier for message passing.
A similar trend is suggested by the number of subtour factors and iterations of augmentation, which has a slope of $\sim 1$, suggesting a linear dependence on $N$.
Again the exceptions are TSPLIB instances that grow faster than $N$ and random distance matrices that seem to grow sub-linearly.%

Overall, we observe that augmentative message-passing is able to find near-optimal solutions in polynomial time.
Although powerful branch-and-cut methods, such as Concorde, are able to exactly solve instances with several thousands of variables, their general run-time on random benchmark instances remains exponential~\cite[p495]{concorde}, while our approximation on random instances appears to be $\OO(N^3)$.

\subsection{Using pairwise factors}\label{sec:btsp}
Here we present an alternative factor-graph for finding permutations without subtours.
This formulation has $\OO(N^2)$ factors and therefore the complete factor-graph remains tractable.
However in practice, the min-sum inference over this factor-graph is not as effective as 
the augmentation approach. Therefore, here we use this factor-graph to solve min-max version
\index{bottleneck TSP}
\index{traveling salesman problem!bottleneck}
of TSP, known as \magn{bottleneck TSP} through sum-product reductions.\footnote{A binary-variable formulation with similar symantics is proposed in \cite{wangmessage}.
However the authors applied their
message passing solution to an instance with only five cities.}

Given an asymmetric distance matrix $\Dn \in \Re^{N\times N}$,
 the task in the Bottleneck
Traveling Salesman Problem (BTSP) is to find a tour of all $N$ points
such that the maximum distance between two consecutive cities in the
tour is minimized \cite{kabadi2004bottleneck}. Any
constant-factor approximation for arbitrary instances of this problem
is \NP-hard \cite{parker1984guaranteed}.

Let $\xs = \{ \xx_1,\ldots,\xx_N\}$ denote the set of variables where
$\xx_i \in  \XX_i = \{0,\ldots,N-1\}$ represents the time-step at
which node $i$ is visited. 
We assume modular arithmetic (module
$N$) on members of $\XX_i$ -- \eg $N \equiv 0 \mod N$ and $1-2
\equiv N-1 \mod N$.  For each pair $\xx_i$ and $\xx_j$ of variables,
define the factor 
\marnote{pairwise factor}
\begin{align}\label{eq:btspfactor}
  \ff_{\{i,j\}}(\xx_i,\xx_j) 
  = 
\min
\bigg (     
\ident(\vert \xx_i - \xx_j \vert > 1),
\max(\Dn_{i,j}, \ident(\xx_i = \xx_j-1)),
\max(\Dn_{j,i}, \ident(\xx_j = \xx_i-1))
\bigg ) 
\end{align}
where the tabular form is 
\vspace{.1in}
\begin{center}
\scalebox{.8}{
\begin{tabu}{r r |[2pt] c | c | c | c | c | c | c }
\multicolumn{2}{c}{}&\multicolumn{7}{c }{$\xx_j$}\\
\multicolumn{2}{c}{}& 0 & 1 & $\cdots$ & $N-2$ & $N-1$ \\\tabucline[2pt]{3-9}
\multirow{7}{*}{$\xx_i$}
        &0&$\infty$ & $\Dn_{i,j}$ & $-\infty$  & $\cdots$ & $-\infty$ & $-\infty$ &\multicolumn{1}{c |[2pt]}{ $D_{j,i}$} \\\cline{2-9}
        &1&$\Dn_{j,i}$ & $\infty$ & $\Dn_{i,j}$  & $\cdots$ & $-\infty$ & $-\infty$ &\multicolumn{1}{c |[2pt]}{ $-\infty$} \\\cline{2-9}
        &2&$-\infty$ & $\Dn_{i,j}$ & $\infty$ & $\cdots$ & $-\infty$ & $-\infty$ &\multicolumn{1}{c |[2pt]}{ $-\infty$} \\\cline{2-9}
        &$\vdots$&$\vdots$ & $\vdots$ & $\vdots$ & $\ddots$ & $\vdots$ & $\vdots$ & \multicolumn{1}{c |[2pt]}{$\vdots$} \\\cline{2-9}
        &$N-3$&$-\infty$ & $-\infty$ & $-\infty$ & $\cdots$ & $\infty$ & $\Dn_{i,j}$ &\multicolumn{1}{c |[2pt]}{ $-\infty$} \\\cline{2-9}
        &$N-2$&$-\infty$ & $-\infty$ & $-\infty$ & $\cdots$ & $\Dn_{j,i}$ & $\infty$ &\multicolumn{1}{c |[2pt]}{ $\Dn_{i,j}$} \\\cline{2-9}
        &$N-1$&$\Dn_{i,j}$ & $-\infty$ & $-\infty$ & $\cdots$ & $-\infty$ & $\Dn_{j,i}$ &\multicolumn{1}{c |[2pt]}{ $\infty$} \\\tabucline[2pt]{3-9}
\end{tabu}
}
\end{center}
\vspace{.2in}

Here, $\identt{\bptimes} = \infty$ on diagonal enteries ensures that $\xx_i \neq \xx_j$.
Moreover $\vert \xx_i - \xx_j \vert > 1$ means cities $i$ and $j$ are not visited consecutively,
so this factor has no effect ($\identt{\bpplus} = -\infty$).
However if two cities are visited one after the other, depending on whether $i$ was visited before $j$ or vicee-versa,
$\Dn_{i,j}$ or   $\Dn_{j,i}$ represent the distance between them.
This factor can be easily converted to min-sum domains by replacing the identity values $-\infty, +\infty$ with $-\infty, 0$ in the tabular
form above. Alternatively we can replace the identity, min and max operations in \cref{eq:btspfactor} with their corresponding min-sum functions. 

\index{Hamiltonian cycle}
Here we relate the min-max factor-graph above to a uniform distribution
over Hamiltonian cycles and use its sum-product reduction to find solutions to Bottleneck TSP.
Recall $\GG(\Dn, \yy)$ is a graph in which there is a connection between node $i$ and $j$ iff
$\Dn_{i,j} \leq \yy$.
\marnote{bottleneck TSP \& Hamiltonian cycle}
\begin{proposition}\label{th:btsp}
  For any distance matrix $\Dn \in \Re^{N \times N}$, the
  $\pp_\yy$-reduction of the BTSP factor-graph above, defines a
  uniform distribution over the (directed) Hamiltonian cycles of
  $\GG(\Dn, \yy)$.
\end{proposition}
\index{Py-reduction}
  \begin{proof}
  First note that $\pp_\yy$ defines a uniform distribution over its
  support as its unnormalized value is only zero or one.  Here w.l.o.g
  we distinguish between two Hamiltonian cycles that have a different
  starting point but otherwise represent the same tour.  Consider the
  $\pp_\yy$-reduction of the pairwise factor of \refEq{eq:btspfactor}
  \begin{align}\label{eq:tspreduction}
    &\ff_{\{i,j\}}(\xx_i,\xx_j) 
    =\; \ident( \vert \xx_i - \xx_j \vert > 1) 
          +\; \ident(\xx_i = \xx_j - 1 \wedge \Dn_{i,j} \leq \yy)  \\
    +\; &\ident(\xx_i = \xx_j + 1 \wedge \Dn_{j,i} \leq \yy)
  \end{align}
  
\noindent \bullitem \emph{Every Hamiltonian cycle over $\GG(\Dn, \yy)$, defines a
    unique assignments $\xs$ with $\pp_\yy(\xs) > 0$:} Given the
  Hamiltonian cycle $H = h_0,h_2,\ldots,h_{N-1}$ where $h_i \in
  \{1,\ldots,N\} $ is the $i^{th}$ node in the path, for each $i$
  define $\xx_i = j\;s.t.\; h_j = i$.  Now we show that all pairwise
  factors of \refEq{eq:tspreduction} are non-zero for $\xs$. Consider
  two variables $\xx_i$ and $\xx_j$. If they are not consecutive in the
  Hamiltonian cycle then $\ff_{\{i,j\}}(\xx_i,\xx_j) = \ident(\vert \xx_i -
  \xx_j \vert > 1) > 0$.  Now w.l.o.g. assume $i$ and $j$ are consecutive
  and $\xx_i$ appears before $\xx_j$.  This means $(i, j) \in \EE$ and
  therefore $\Dn_{i,j} \leq \yy$, which in turn means
  $\ff_{\{i,j\}}(\xx_i,\xx_j) = \ident(\xx_i = \xx_j - 1 \wedge \Dn_{i,j} \leq
  \yy) > 0$ Since all pairwise factors are non-zero, $\pp_\yy(\xs) > 0$.

\noindent \bullitem \emph{Every $\xs$ for which $\pp_\yy(\xx) > 0$, defines a unique
    Hamiltonian path over $\GG(\Dn,\yy)$:} Given assignment $\xs$,
  construct $H = h_0,\ldots,h_{N-1}$ where $h_i = j \; s.t. \xx_j = i$.
  Now we show that if $\pp(\xs) > 0$, $H$ defines a Hamiltonian path.  If
  $\pp(\xs) > 0$, for every two variables $\xx_i$ and $\xx_i$, one of the
  indicator functions of \refEq{eq:tspreduction} should evaluate to
  one.  This means that first of all, $\xx_i \neq \xx_j$ for $i \neq j$,
  which implies $H$ is well-defined and $h_i \neq h_j$ for $i \neq j$.
  Since all $\xx_i \in \{0,\ldots,N-1\}$ values are distinct, for each
  $\xx_i = s$ there are two variables $\xx_j = s - 1$ and $\xx_k = s + 1$
  (recall that we are using modular arithmetic) for which the pairwise
  factor of \refEq{eq:tspreduction} is non-zero.  This means $\Dn_{j,i}
  \leq \yy$ and $\Dn_{i,k} \leq \yy$ and therefore $(j,i), (i,k) \in
  \EE$ (the edge-set of $\GG(\Dn,\yy)$).  But by definition of
  $H$, $h_{s} = i$, $h_{s-1} = j$ and $h_{s+1} = k$ are consecutive
  nodes in $H$ and therefore $H$ is a Hamiltonian path. 
\end{proof}

This proposition implies that we can use the sum-product reduction of this factor-graph to solve Hamiltonian cycle problems.
The resulting factor-graph for Hamiltonian cycle problem is an special case of our graph-on-graph technique, where the adjacency matrix of one graph is directly used to build factors in a second graph. Here, as the tabular form of the factor above suggests, the second graph is a simple loop
of length $N$ (see \refSection{sec:graphongraph}).

As expected, due to sparse form of this pairwise factor, we can perform BP updates efficiently.
\begin{claim}\label{th:btsp_cost} \marnote{complexity}
The factor-to-variable BP messages for the sum-product reduction of factor of \refEq{eq:btspfactor}
can be obtained in $\OO(N)$. 
\end{claim}
\begin{proof}
The $\pp_\yy$-reduction of the min-max factors of \refEq{eq:btspfactor} is given by:
\begin{align}
  &\ff_{\{i,j\}}(\xx_i,\xx_j) = \ident(\ff_{\{i,j\}}(\xx_i,\xx_j) \leq \yy) \\
  =\; &\ident(\vert \xx_i - \xx_j \vert > 1) 
        +\; \ident(\xx_i = \xx_j - 1 \wedge \Dn_{i,j} \leq \yy)  \\
  +\; &\ident(\xx_i = \xx_j + 1 \wedge \Dn_{j,i} \leq \yy)
\end{align}

The matrix-form of this factor (depending on the order of $\Dn_{i,j}$,
$\Dn_{j,i}, \yy$) takes several forms all of which are band-limited.
Assuming the variable-to-factor messages are normalized (\ie $
\sum_{\xx_i} \msg{j}{\II}(\xx_i) = 1$) the factor-to-variable message is
given by
\begin{align*}
  \msg{\{i,j\}}{i}(\xx_i) = 1 - \msg{j}{\{i,j\}}(\xx_i) + \\ 
  \ident(\Dn_{i,j} \leq \yy)(1 - \msg{j}{\{i,j\}}(\xx_i - 1)) +\\ 
  \ident(\Dn_{j,i} \leq \yy) (1 - \msg{j}{\{i,j\}}(\xx_i + 1))
\end{align*}

Therefore the cost of calculating factor-to-variable message is that of normalizing 
the variable-to-factor message, which is $\OO(N)$. 
\end{proof}

Since there are $N^2$ pairwise factors,
this gives $\OO(N^3)$ time-complexity for each iteration of sum-product BP 
in solving the Hamiltonian cycle problem (\ie the sum-product reduction) and $\OO(N^3 \log(N))$ for bottleneck-TSP,
where the $\log(N)$ factor is the cost of binary search (see \refSection{sec:minmax2sumprod}).

\begin{figure}
  \centering 
  \hbox{ 
      \includegraphics[width=.4\textwidth]{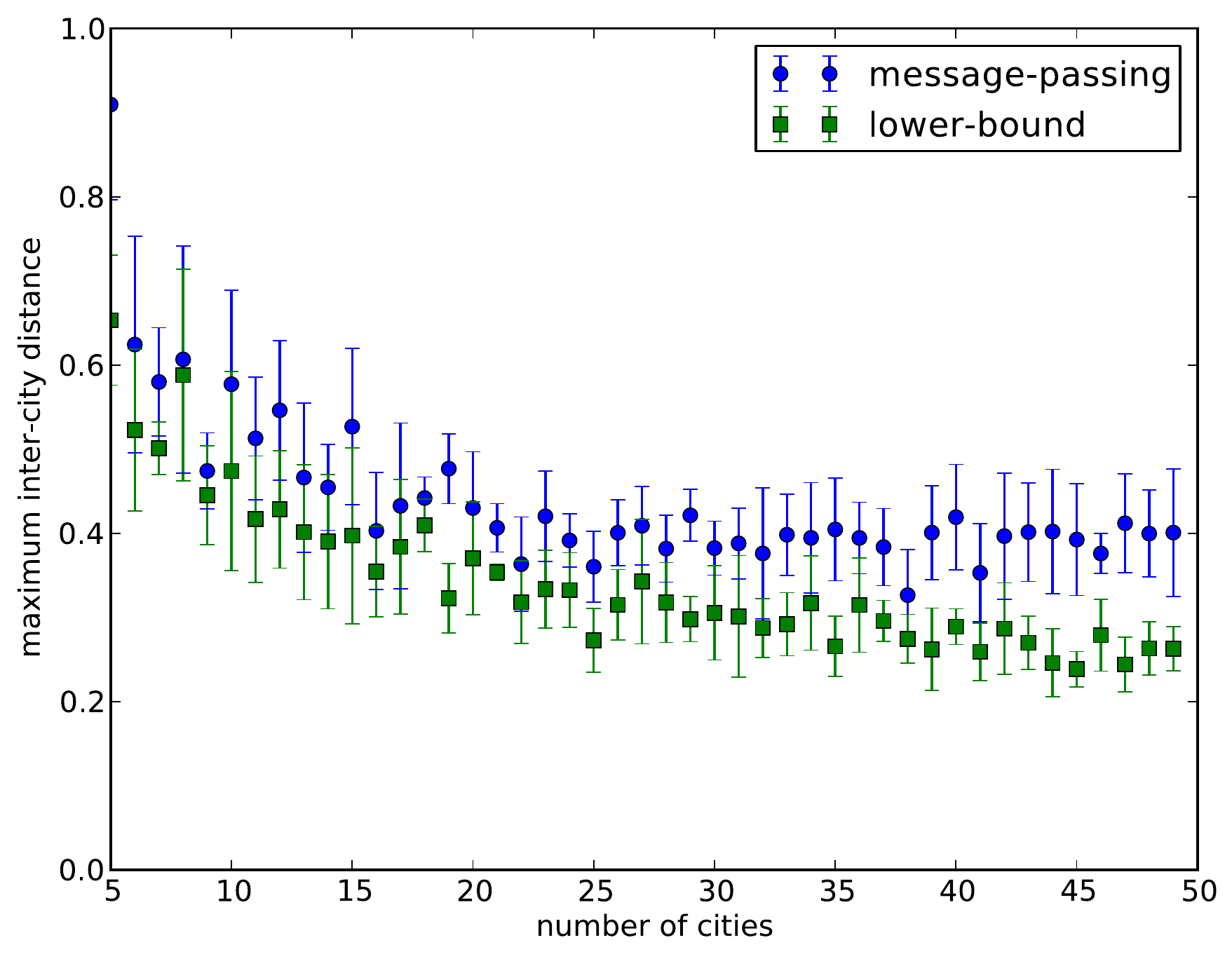}
      \includegraphics[width=.4\textwidth]{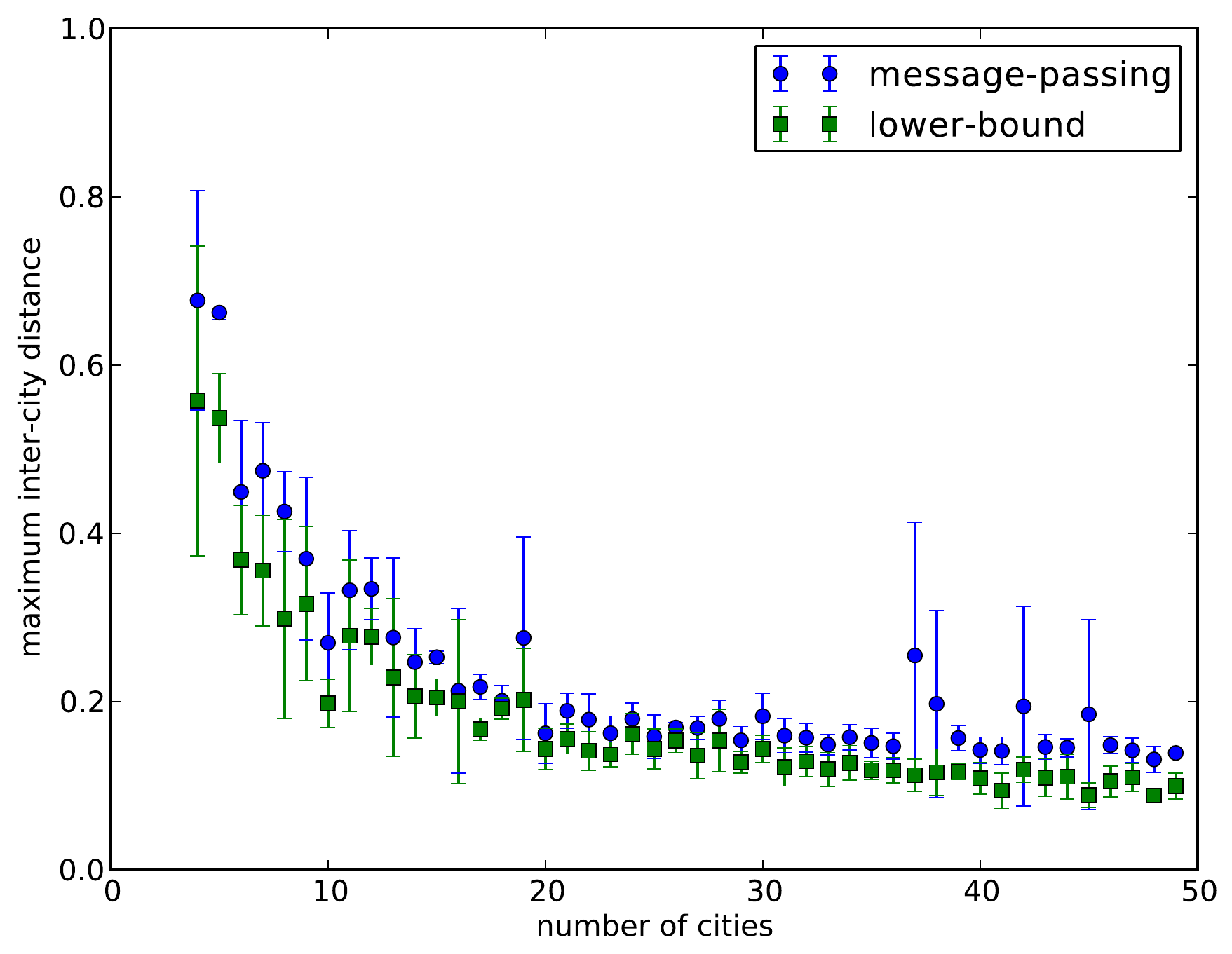}
   }
  \caption[Experimental results for bottleneck TSP.]{
      The min-max solution (using sum-product reduction) for Bottleneck TSP with
      different number of cities (x-axis) for 2D Euclidean space (left) as
      well as asymmetric random distance matrices (right) with $T = 5000$ for
      Perturbed BP.  The error-bars in all figures show one standard deviation
      over 10 random instances.}
  \label{fig:btspresults}
\end{figure}

\Cref{fig:btspresults} reports the average performance of message 
passing (over 10 instances) as well as a \emph{lower bound} on the optimal
min-max value for tours of different length ($N$). Here we report the
\marnoteloc{lower bound}{0}
results for random points in 2D Euclidean space as well as asymmetric
random distance matrices.  For Euclidean problems, the lower bound is the
maximum over $j$ of the distance of two closest neighbors to each node
$j$.  For the asymmetric random distance matrices, the maximum is over all
the minimum length incoming edges and minimum length outgoing edges
for each node.\footnote{If for one node the minimum length in-coming
  and out-going edges point to the same city, the second minimum length
  in-coming and out-going edges are also considered in calculating a
  tighter bound.}

%% file: symm.tex
Consider two graphs $\GG = (\VV, \EE)$ and $\GG' = (\VV', \EE')$, with (weighted) adjacency matrices $\Dn$, $\Dn'$ respectively.
Here we enumerate several of the most important problems over permutations based on two graphs~\cite{conte2004thirty}.
\RefSection{sec:iso} introduces the graphical model for the problems of (sub-graph) isomorphism and monomorphism. 
Then we study the message passing solution to graph homomorphism, use it to find symmetries in graphs. 
In \refSection{sec:alignment} we introduce a general factor-graph for graph alignment based on the previous work of \citet{bradde2010aligning} and show how it can model other problems such as quadratic assignment problem and maximum common subgraph.
The common idea in all these settings is that a ``variation'' of the adjacency matrix  $\Dn'$ of graph $\GG'$ 
is used as a pairwise factor over the edges (or/and non-edges) of graph $\GG$ (with adjacency $\Dn$), defining a Markov network (a factor-graph with pairwise factors). 

\subsection{Sub-graph  isomorphism}\label{sec:iso}
\index{isomorphism}
The graph \magn{isomorphism} problem asks whether  $\GG = (\VV, \EE)$ and $\GG' = (\VV', \EE')$ are identical up to a permutation of nodes (written as $\GG \cong \GG'$). That is, it seeks 
a one-to-one mapping $\pi: \VV \to \VV'$  such that $(i,j) \in \EE \; \Leftrightarrow (\pi(i), \pi(j)) \in \EE'$. With some abuse of notation we also write $\pi(\GG) = \GG'$.
Although, there have been polynomial time solutions to special instances of graph isomorphism problem~\cite{eppstein1995subgraph}, the general case
has remained elusive. So much so that we do not know whether the problem is \NP-complete (\eg see \cite{moore2011nature}).

\index{symmetric group}
A permutation $\pi \in \sn$ (recall $\sn$ is the symmetric group) such that  $\pi(\GG) \cong \GG$ is called an {automorphism}.
The  automorphisms of $\GG$, under composition form a group, called the \magn{automorphism group} $\auto(\GG)$.
\index{automorphism}
The automorphism group also defines a natural notion of symmetry on the nodes of graphs. Here, the \magn{orbit} 
of each node, is the set of nodes that are mapped to $i$ in any automorphism -- \ie $\orbit(i) \defeq \{ \pi(i) \mid \pi \in \auto(\GG)\}$. The orbits partition the set of nodes $\VV$ into group of nodes
that are in a sense symmetric, which makes them a prime candidate for defining symmetry in complex networks \cite{macarthur2008symmetry,xiao2008emergence}.

\magn{Sub-graph isomorphism} asks whether $\GG'$ is isomorphic to a vertex induced subgraph of $\GG$.
-- \ie it seeks an injective mapping $\pi: \VV' \to \VV$
where 
$$
(\pi(i),\pi(j)) \in \EE \; \Leftrightarrow \; (i,j) \in \EE' \quad \forall i,j \in \VV'
$$ 
When dealing with sub-graph morphisms, we assume that the mapping is from the smaller graph to the larger graph and therefore $|\VV| \leq |\VV'|$.

The factor-graph for subgraph isomorphism is defined as follows:
We have one variable per $i \in \VV$, where the domain of each variable $\xx_i$ is $\VV'$ -- \ie $\xs = \{ \xx_i \in \VV' \mid i\in \VV\}$. The factor-graph has two types of pairwise factors:

\index{factor!edge}
\noindent \bullitem\ \magn{Edge factors:} ensure that each edge in $\GG = (\VV, \EE)$ is mapped to an edge in $\GG' = (\VV', \EE')$
 \begin{align}\label{eq:edge_factor}
   \ff_{\{i,j\}}(\xs_{\{i,j\}}) = \ident((\xx_i, \xx_j) \in \EE') \quad \forall (i,j) \in \EE
\end{align}
where assuming the tabular form of this factor for sum-product semiring
is simply the adjacency matrix $\Dn' \in \{0,1\}^{|\VV'| \times |\VV'|}$ of $\GG'$.

\index{factor!non-edge}
\noindent \bullitem\ \magn{Non-edge factors:} ensure that each non-edge in $\GG = (\VV, \EE)$ is mapped to a non-edge in $\GG' = (\VV', \EE')$
 \begin{align}\label{eq:nonedge_factor}
   \ff_{\{i,j\}}(\xs_{\{i,j\}}) = \ident((\xx_i, \xx_j) \notin \EE' \wedge \xx_i \neq \xx_j) \quad \forall i,j \in \VV,  i\neq j ,  (i,j) \notin \EE
\end{align}
where again for sum-product semiring, the tabular form takes a simple form \wrt\ the binary valued adjacency matrix of $\GG'$ -- \ie $1 - \Dn'$.
Using sum-product semiring, this fully connected Markov network defines a uniform distribution over the (subgraph) isomorphisms from $\GG$ to $\GG'$. We could use sum-product BP with decimation or perturbed BP to sample individual assignments $\xs^* \sim \pp(\xs)$. Here, each assignment $\xs \equiv \pi$ is an (injective) mapping from $\VV$ to $\VV'$, where $\xx_i = j'$ means node $i \in \VV$ is mapped to node $j' \in \VV'$.

In particular, for $\GG = \GG'$,  the integral is equal to the cardinality of the automorphism group $\qq(\emptyset) = | \auto(\GG) |$ and two nodes $i,j \in \VV$ are in the same 
\index{orbit}
orbit iff the have the same marginals -- \ie
$$\pp(\xx_i) = \pp(\xx_j) \; \Leftrightarrow \; i \in \orbit(j)$$
This also suggests a procedure for finding (approximate) symmetries in graphs. We can use sum-product BP to find marginals and group the nodes based on the similarity of their marginals.  
However, the cost of message passing in this graphical model is an important barrier in practice.
\begin{claim}\label{claim:iso_cost}
Assuming $|\VV| \leq |\EE| \leq |\VV|^2$ and $|\VV'| \leq |\EE'| \leq |\VV'|^2$ and using variable synchronous update, the time complexity of each iteration of sum-product BP 
for subgraph isomorphism is $\OO(|\VV|^2\;|\EE'|)$.
\end{claim}
\begin{proof}
First, we calculate the cost of sending sum-product BP messages through the edge and non-edge factors.
For the edge factors ($\Dn'$), we can go through each row of the tabular form of the factor and multiply the non-zero entries
with corresponding message -- \ie $ \fg_{\{i,j\}}(\xx_i, \xx_j) = \msg{i}{\{i,j\}}(\xx_i) \ff_{\{i,j\}}(\xx_i, \xx_j)$.
Then we add the values in each column, to obtain the outgoing message -- \ie $\msg{\{i,j\}}{j}(\xx_j) = \sum_{\xx_i} \fg_{\{i,j\}}(\xx_i, \xx_j)$. This procedure depends on the number of non-zero entries -- \ie $\OO(|\EE'|)$.
The procedure is similar for non-edge factors.
Therefore the overall cost of sending sum-product BP messages through all $\OO(|\VV|^2)$ factors is $\OO(|\EE'|\; |\VV|^2)$.
Using variable synchronous update, calculating all variable-to-factor messages takes $\OO(|\VV|^2 |\VV'|)$.
Using the assumption of the claim, the overall cost is therefore $\OO(|\EE'|\; |\VV|^2)$.
\end{proof}

However it is possible to improve this complexity by 
considering sparse mappings. 
For example, we can restrict the domain of each variable $\xx_i$  to all the nodes $j' \in \VV'$ that have the same degree with node $i$, or furthermore to all the nodes that also have neighbours
with same degree as the neighbours of node $i$. 

\subsection{Subgraph  monomorphism and supermorphism}
\index{monomorphism}
Sub-graph monomorphism relaxes the constraint of the subgraph isomorphism to  
\begin{align}\label{eq:monomorphism}
(i,j) \in \EE  \; \Rightarrow \;  (\pi(i),\pi(j))\in \EE' \quad \forall i,j \in \VV  
\end{align}
where $\pi: \VV' \to \VV$ has to be injective -- \ie nodes in $\VV'$ are mapped to distinct nodes in $\VV$.
However, $\GG'$ is allowed to cover a ``subset'' of edges in an induced subgraph of $\GG$.
We note here that \textit{previous graphical models introduce in \cite{bradde2010aligning,bradde2010statistical} for isomorphism in fact define monomorphism.}
The only difference between this factor-graph and that of sub-graph isomorphism is that the non-edge
factors are replaced by the following uniqueness factors.

\index{factor!uniqueness}
\noindent \bullitem\ \magn{Uniqueness factors} are inverse Potts factors that ensure disconnected nodes are mapped to different nodes (-- \ie the mapping is injective)
 \begin{align}\label{eq:uniqueness_factor}
   \ff_{\{i,j\}}(\xs_{\{i,j\}}) = \ident(\xx_i \neq \xx_j) \quad \forall i.j \in \VV, j \neq i , (i,j) \notin \EE
\end{align}

Despite this difference, when  $\GG = \GG'$ -- that is when we are interested in automorphisms -- or more generally when $|\EE| = |\EE'|$ and $|\VV| = |\VV'|$, 
the distribution defined by the monomorphism factor-graph is identical to that of isomorphism factor-graph.

\begin{claim}\label{claim:mono_cost}
Assuming $|\VV| \leq |\EE| \leq |\VV|^2$ and $|\VV'| \leq |\EE'| \leq |\VV'|^2$, using variable synchronous update, the time complexity of each iteration of sum-product BP for 
sub-graph monomorphism $\OO(|\EE|\;|\EE'|\; + |\VV'|\;|\VV|^2)$.\footnote{\citet{bradde2010aligning} suggest a trick to reduce this time-complexity to $\OO((|\EE|\; |\VV|) |\VV'|)$, but unfortunately details are omitted and we are unable to follow their route.}
\end{claim}
\begin{proof}
The complexity of sending sum-product BP messages through the edge factors is $\OO(|\EE'|)$ (see proof of \cref{claim:iso_cost}).
However, the uniqueness factors are inverse Potts factors and allow  $\OO(|\VV'|)$ calculation of messages.
This means the overall cost of sending messages through all the factors is $\OO(|\EE|\;|\EE'|\; + |\VV'|\;|\VV|^2)$.
The cost of calculating variable-to-factor messages using variable-synchronous update is also $\OO(|\VV|^2|\VV'|)$, which gives the overall complexity of $\OO(|\EE|\;|\EE'|\; + |\VV'|\;|\VV|^2)$
per BP iteration.
\end{proof}

So far we have seen two variations of mapping a graph $\GG$ to a subgraph of $\GG'$. In subgraph isomorphism, the image of $\GG$ strictly agrees with a sub-graph of $\GG'$.
In monomorphism, the image is contained within a sub-graph. A third possibility is to ask for a mapping such that the image of $\GG$ ``contains'' a sub-graph $\GG'$:
\begin{align}\label{eq:supermorphism}
(i,j) \in \EE  \; \Leftarrow \;  (\pi(i),\pi(j))\in \EE' \quad \forall i,j \in \VV  
\end{align}
where again $\pi: \VV' \to \VV$ has to be injective.
Due to lack of a better name, we call this mapping \magn{subgraph supermorphism}.
\index{supermorphism}

The factor-graph for sub-graph supermorphism has two types of factors, both of which we have seen before: 1) The non-edge factors between non-edges 
 (\refEq{eq:nonedge_factor}) ensure that the image contains a sub-graph of $\GG'$ while 2) uniqueness factors between the edges $(i,j) \in \EE$ ensure
that the mapping is injective. 
\begin{claim}\label{claim:super_cost}
Assuming $|\VV| \leq |\EE| \leq |\VV|^2$ and $|\VV'| \leq |\EE'| \leq |\VV'|^2$, using variable synchronous update, the time complexity of each iteration of sum-product BP 
for subgraph supermorphism is $\OO((|\VV^2| - |\EE|)\;|\EE'|\; + |\VV'|\;|\VV|^2)$
\end{claim}
\begin{proof} 
The complexity of sending sum-product BP messages through the non-edge factors is $\OO(|\EE'|)$ (see proof of \cref{claim:iso_cost}) and each uniqueness factor require $\OO(|\VV'|)$ computation.
This means the overall cost of sending messages through all the factors is $\OO((|\VV^2| - |\EE|)\;|\EE'|\; + |\VV'|\;|\EE|)$.
The cost of calculating variable-to-factor messages using variable-synchronous update is also $\OO(|\VV|^2|\VV'|)$, which gives the overall complexity of $\OO((|\VV^2| - |\EE|)\;|\EE'|\; + |\VV'|\;|\VV|^2)$
per BP iteration.
\end{proof}

\subsection{Graph homomorphism}\label{sec:homo}
\index{homomorphism}
Graph homomorphism~\cite{hell2004graphs} further relaxes the constraint of the  sub-graph monomorphism 
such that a homomorphic mapping 
can map two distinct nodes in $\VV$ to the same node in $\VV'$.
However, \cref{eq:monomorphism} should hold, and therefore if two nodes in $\VV'$ are adjacent, 
they should still be mapped to distinct and adjacent nodes in $\VV$. 
Compared to other areas of graph theory, the study of rich structure and surprising properties of homomorphisms is relatively recent. Study of graph homomorphisms covers 
\index{property testing}
diverse areas including property testing \cite{alon2006homomorphisms}, graph sequences and limits~\cite{borgs2006graph, lovasz2006limits, borgs2008convergent,borgs2012convergent} and constraint satisfaction problems~\cite{mugnier2000knowledge}. 
Many of these areas are interested in ``counting'' the number of homomorphisms~\cite{borgs2006counting} from a graph $\GG$ to $\GG'$. 
As we see shortly, graph homomorphism reduces to many interesting CSPs and therefore it is not hard to show that it is \NP-complete.
Moreover, counting the number of homomorphisms is \sharpP-complete~\cite{dyer2000complexity}.
The set of all homomorphisms $\pi$ of a graph $\GG$ to itself --\ie its endomorphisms -- under composition form \magn{endomorphism monoid} (see \refDefinition{def:semigroup} for  monoid).  Here, the identity element simply maps each element to itself.
Our interest in endomorphism is because through the \cref{conj:homo}, we can use it (instead of automorphism) to approximate symmetries in graphs.

The graphical model representation of homomorphism has been investigated in different contexts~\cite{brightwell1999graph,bulatov2005complexity}, but to our knowledge, message passing has not been previously used for counting and finding homomorphisms.
The Markov network for homomorphism resembles that of isomorphism and monomorphism: 
The variables are $\xs =\{ \xx_i \mid i \in \VV\}$ where $\xx_i \in \VV'$, and the factor-graph \textit{only contains edge-factors} of \refEq{eq:edge_factor}.
Assuming $|\VV'| \leq |\EE'|$ and $|\VV| \leq |\EE|$, it is easy to see that the complexity of variable-synchronous message passing is $\OO(|\EE|\;|\EE'|)$, which makes this method very efficient for sparse graphs.
This graphical model can be extended to represent homomorphism in weighted graphs~\cite{borgs2006counting}. For this define the edge-factors as
\begin{align*}
   \ff_{\{i,j\}}(\xs_{\{i,j\}}) =  \Dn_{i,j}^{\Dn'_{\xx_i, \xx_j}}\quad \forall i,j \in \VV, i\neq j, \Dn_{i,j} > 0
\end{align*}

\begin{figure}
  \centering 
  \hbox{ 
      \includegraphics[width=.45\textwidth]{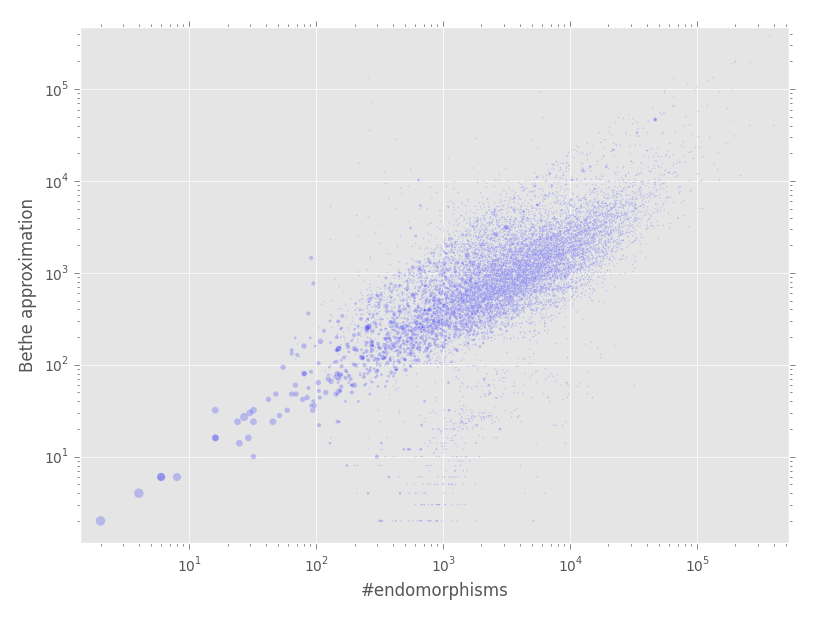}
      \includegraphics[width=.45\textwidth]{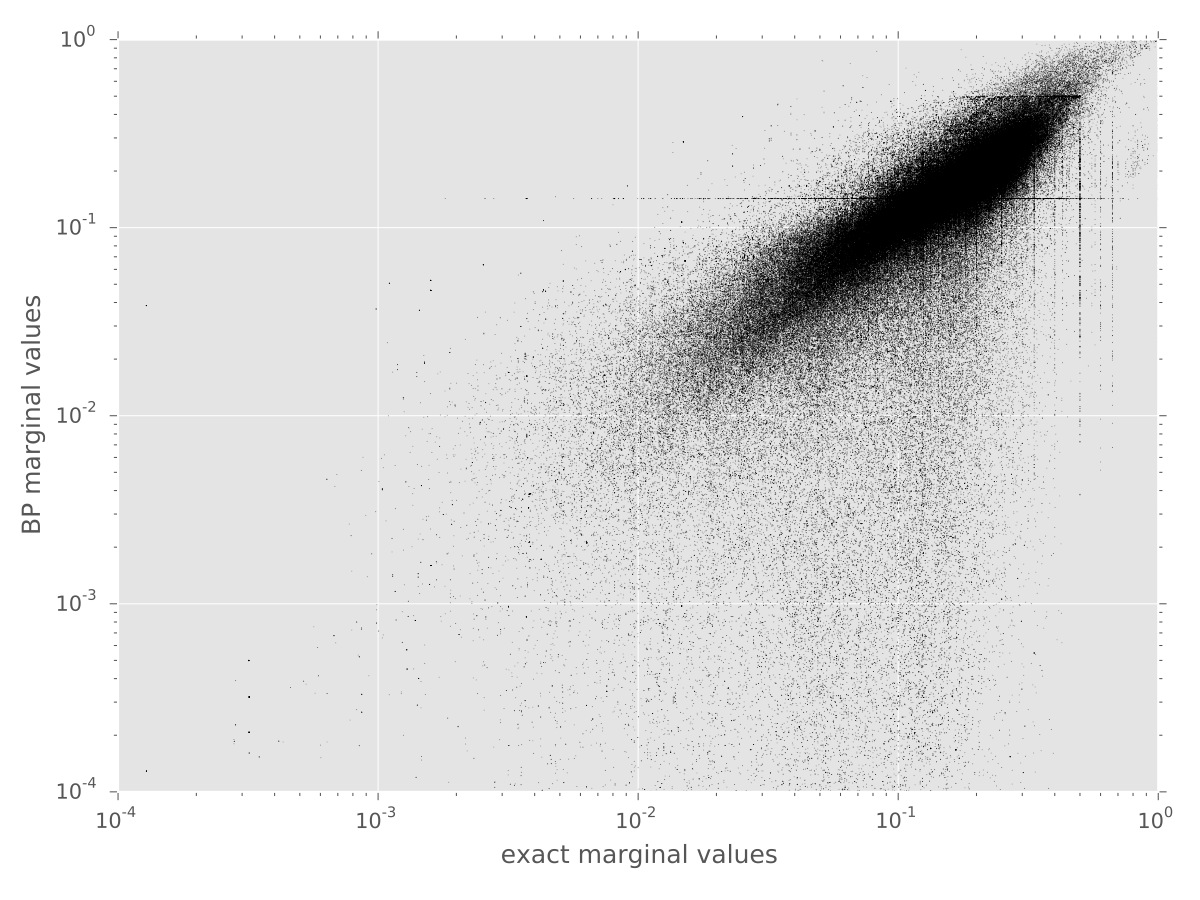}
   }
  \caption[Quality of BP marginals and integral in graph endomorphism.]{ \textbf{(left)} The number of endomorphisms for all distinct graphs up to 8 nodes compared to the BP integral. Here, larger disks represent graphs with smaller number of nodes. \textbf{(right)} Comparison of normalized BP marginals and the exact marginals for endomorphism graphical model.} 
  \label{fig:homo_bp}
\end{figure}

For small graphs we can exactly count the number of homomorphisms and endomorphisms.
To evaluate the accuracy of message passing, we compared the BP estimates with the exact number of 
endomorphisms, for all isomorphically distinct graphs with $|\VV| < 9$ (\ie $> 13,000$ instances); \RefFigure{fig:homo_bp} reports this result as well as the accuracy of BP marginals, 
suggesting that despite existence of short loops in these graphs BP estimates are relatively accurate. 

\subsubsection{Reduction to other CSPs}\label{sec:morphism_reduction}
The relation between homomorphism and other CSPs is well-known~\cite{mugnier2000knowledge}.
Here, we review this relation between the factor-graphs of this section and other CSP factor-graphs we encountered in this thesis.

\noindent \magn{Coloring and clique cover:} (see \refExample{example:kcoloring} and \refSection{sec:clique-cover}) K-coloring of $\GG$ corresponds to finding a homomorphism from $\GG$ to $\KG_K$, the complete graph of order $K$. Similarly the homomorphism from the complement of $\GG$ to $\KG_k$ corresponds to K-clique-cover.
The relation is also reflected in the corresponding factor-graphs as the adjacency matrix for $\KG_K$ is the inverse Potts model, used in K-coloring.

\index{independent-set!reduction}
\index{Clique problem!reduction}
\index{coloring!reduction}
\noindent \magn{Clique problem and independent-set:} (see \refSection{sec:independent-set})
Recall that K-clique problem corresponds to finding a clique of size $K$ in $\GG$. This is a special case of sub-graph isomorphism
from $\KG_K$ to $\GG$ which is in this case identical to sub-graph monomorphism and homomorphism from $\KG_K$ to $\GG$.
The (sum-product reduction of the) categorical-variable model of \refSection{sec:packing_categorical}
is a fully connected Markov network with edge factors that we used for isomorphism, monomorphism and homomorphism \refEq{eq:edge_factor}. Similarly, the K-independent set problem is equivalent to finding homomorphisms from 
$\KG_K$ to the complement of $\GG$. Independent set has an alternative relation with graph homomorphism: any homomorphism from $\GG$ to a graph with two connected nodes and a self-loop on one of the nodes defines an independent set for $\GG$.

\index{Hamiltonian cycle!reduction}
\noindent \magn{Hamiltonian cycle} problem corresponds to sub-graph monomorphism from $\CG_{|\VV|}$, the cycle of length $|\VV|$, to $\GG$. 
Alternatively, we can formulate it as subgraph supermorphism from $\GG$ to $\CG_{|\VV|}$. The sum-product reduction of our min-max formulation for bottleneck TSP in \refSection{sec:btsp} 
is indeed the factor-graph of sub-graph supermorphism from $\GG$ to $\CG_{|\VV|}$.

\subsection{Finding symmetries}
\index{graph symmetries}
One may characterize a graph $\GG$ using the ``number'' of homomorphism from/to
other  graphs $\GG'$. This characterization is behind the application of graph homomorphism in
property testing and definition of graph sequences.
Let $\homo(\HG, \GG)$ be the set of homomorphism from $\HG$ to $\GG$ -- \ie the set of all assignments $\xs$ where $\pp(\xs) > 0$.
Let $\HG_1,\ldots,\HG_M$ be the sequence of all graphs whose number of nodes is at most  $|\VV|$. Then, the \magn{Lov{\'a}sz vector} of $\GG$ which is defined as 
\begin{align}\label{eq:lovas}
\vs(\GG) \; = \; (|\homo(\HG_1,\GG)|, |\homo(\HG_2,\GG)|, \ldots, |\homo(\HG_M,\GG)| )
\end{align}
uniquely identifies $\GG$ up to an isomorphism~\cite{lovasz1967operations}.
\index{Lovasz vector}

Here, rather than identifying a particular graph $\GG$ within the set of all graphs, we are interested in identifying a node $i \in \VV$ of a single graph $\GG$ within the set of all nodes $\VV$. Note that here both identifications are up to an isomorphism. Our objective is equivalent to finding the orbits of $\GG$ and our approach in finding the orbits using graph homomorphism (rather than isomorphism)  is founded on the following conjecture.

\begin{figure}
  \centering 
      \includegraphics[width=.35\textwidth]{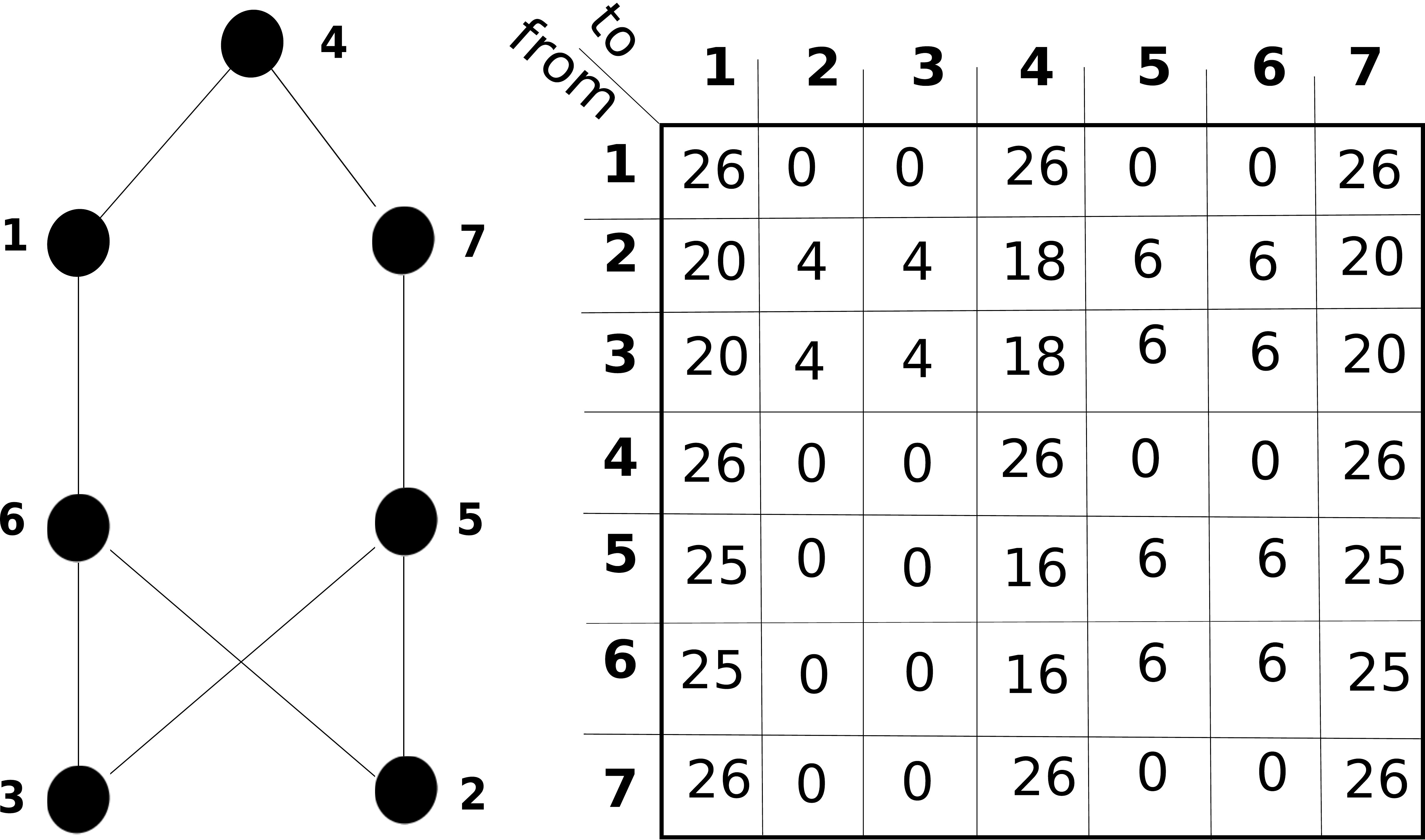}
  \caption[Counter-example for relaxed conditions of \cref{conj:homo}.]{The table on the right shows the unnormalized endomorphism marginals for the graph on the left. Here row $i$ corresponds to $\qq(\xx_i)$ (normalization of which gives $\pp(\xx_i)$), 
and the number at row $i$ and column $j$ is the number of times node $i$ is mapped to $j$ in an endomorphism. The total number of endomorphisms for this graph is $\qq(\emptyset) = 78$. Here, the orbits 
are $\{2,3\}, \{5,6\}, \{1,7\},\{4\}$. However, $\qq(\xx_1) = \qq(\xx_7) = \qq(\xx_4)$ -- that is node $4$ maps ``to'' other nodes with the same frequency as nodes $1$ and $7$. However, the mappings to node $4$ (\ie the $4^{th}$ column of the table) remains different from the mappings to $1$ and $7$. Here, as predicted by \cref{conj:homo}, nodes with similar rows and columns belong to the same orbit.}
  \label{fig:counter_example}
\end{figure}

\begin{conjecture}\label{conj:homo}
Given the uniform distribution over the endomorphisms of $\GG$:
$$
\pp(\xs) \propto \prod_{(i,j) \in \EE} \ident((\xx_i, \xx_j) \in \EE) 
$$
the necessary and sufficient condition for $i$ and $j$ to be in the same orbit is 
\begin{align*}
\pp(\xx_i) = \pp(\xx_j)\;\; \text{and} \;\; \forall k \in \VV:\; \pp(\xx_k = i) = \pp(\xx_k = j) \quad \Leftrightarrow \quad \orbit(i) = \orbit(j)
\end{align*}
\end{conjecture}

Note that $\pp(\xx_i = k)$ is simply the relative frequency of mapping of node $i$ to node $k$ in an endomorphism. Therefore this conjecture simply states that for node $i$ and $j$ to be equivalent up to an automorphism of $\GG$, it is necessary and sufficient for them to have the same frequency of mapping to/from all other nodes of $\GG$. While it is trivial to prove necessity ($\Leftarrow$), we found it difficult to prove the statement in the other direction.

Therefore, similar to other related conjectures~\cite{mckay1997small,kelly1957congruence}, 
we turn to \magn{computational verification}. For this, we experimented with distinct graphs with up to $9$ nodes (\ie $> 286,000$ instances). Here, we obtained the exact marginals $\pp(\xx_i)\; \forall i \in \VV$ and constructed the orbits as suggested by \cref{conj:homo}. We then obtained the exact orbits using the software of \citet{McKay201494}. In all cases two partitioning of the nodes to orbits were identical.

We also note that in \cref{conj:homo} restricting the condition to $\pp(\xx_i) = \pp(\xx_j)$ -- \ie similar frequency of mappings ``to'' other nodes -- is not sufficient. \RefFigure{fig:counter_example}
shows a counter-example for this weaker condition.
\begin{figure}
  \centering 
      \includegraphics[width=.35\textwidth]{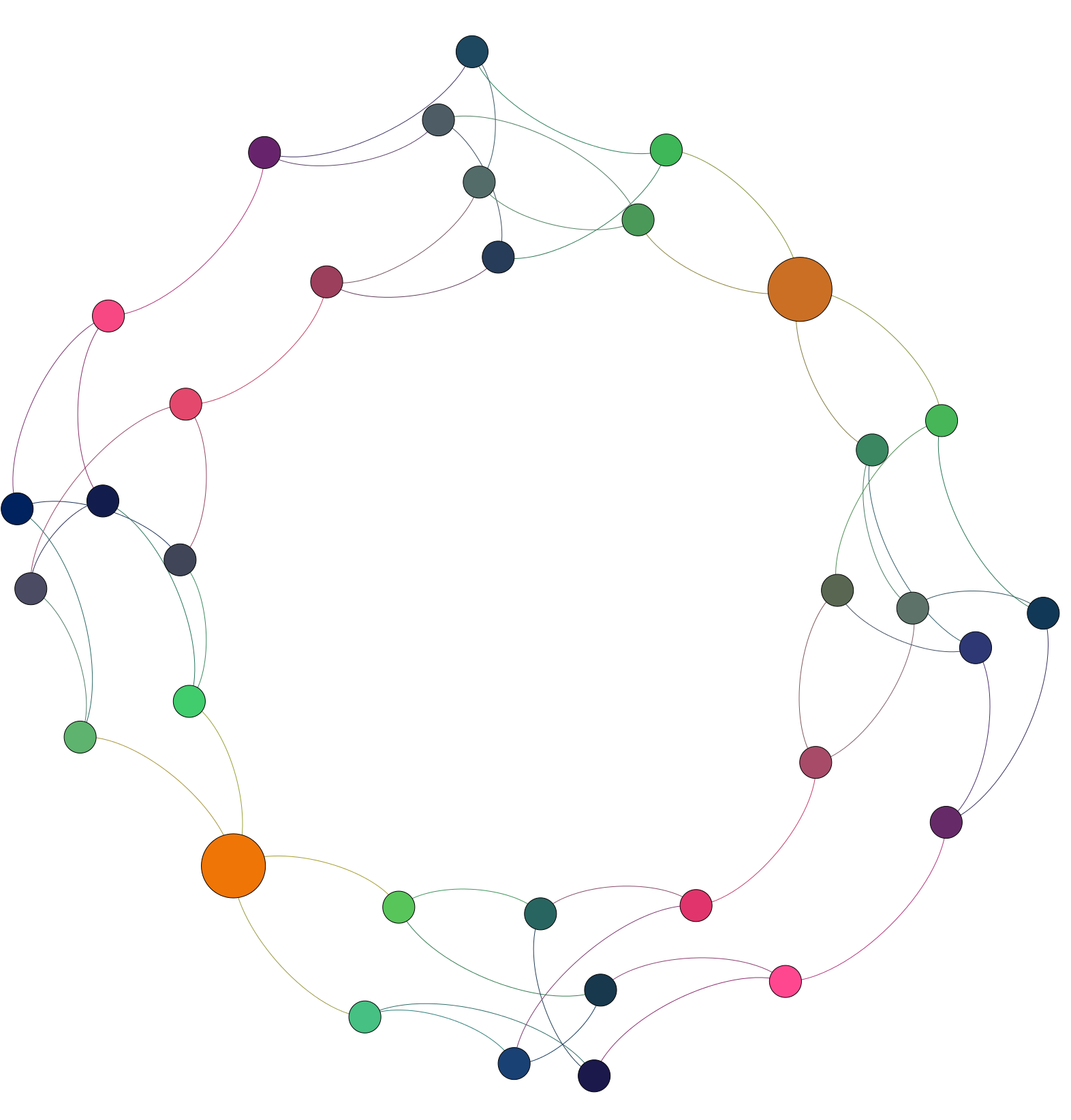}
      \includegraphics[width=.6\textwidth]{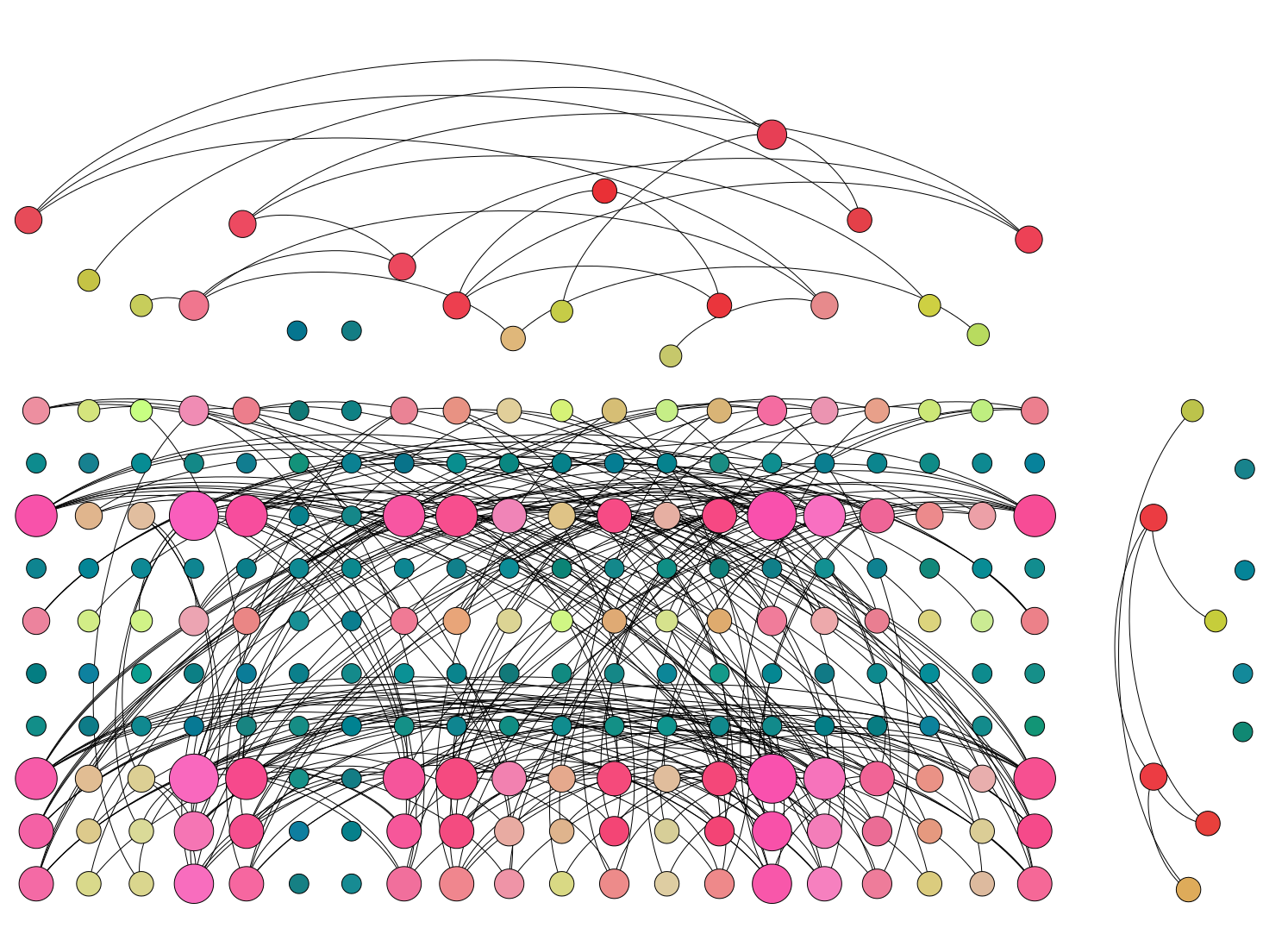}
  \caption[An example of finding approximate symmetry using endomorphism marginals.]{Coloring of nodes by reducing the dimensionality of marginals to three dimensions of RGB using PCA with whitening. 
\textbf{(left)} 
The marginals are calculated using sum-product BP. \textbf{(right)} The marginals of the Kronecker graph is obtained using Gibbs sampling and annealing. The graph in the middle is the product of two graphs on the top and right -- \ie $\GG = \GG_1 \times \GG_2$. Two nodes in $\GG$ are connected iff the corresponding nodes in $\GG_1$ and $\GG_2$ are connected. Note that the product of each two colors produces the same color in the product graph and similarly colored nodes have similar neighbours.}
  \label{fig:symmetry}
\end{figure}

\begin{example}
In \refFigure{fig:symmetry}(left) we used the homomorphism marginals to find approximate symmetries in a graph with visible symmetries, known as Thomassen graph. 
Here, after obtaining the marginals we used Principle Component Analysis (PCA) with whitening \cite{murphy2012machine} to extract three values for RGB colors. As the figure suggests, this approach is able to identify similar nodes with similar colors.

For dense graphs, message passing is no longer accurate. In \refFigure{fig:symmetry}(right) we use Gibbs sampling with annealing to estimate the endomorphism marginals
\index{Kronecker product}
in \magn{Kronecker product}\cite{leskovec2010kronecker} of two random graphs. Here again, we use PCA to color the nodes. Note that the algorithm is unaware of the product format. The choice of 
the product graph is to easily observe the fact that the product of similarly colored nodes produce similarly colored nodes in the product graph.

An alternative is to use spectral clustering on the matrix of marginals to obtain clusters.
In a related context \citet{krzakala2013spectral} use the matrix of non-backtracking random-walks to find symmetric clusters in stochastic block models.
\index{spectral clustering}
Note that the basic difference with our approach is the matrix used with the spectral clustering.
Other notable matrices that are used within this context are the Laplacian matrix~\cite{von2007tutorial}, the modularity
\index{Laplacian}
 matrix~\cite{newman2006finding} and the Bethe hessian~\cite{saade2014spectral}.
The relation between the clustering (in its conventional sense) and orbits of a graph is better understood in the extreme case: when clusters form isolated cliques, they are identical to orbits.
\end{example}

\subsection{Graph alignment}\label{sec:alignment}
\index{graph alignment}
The graph alignment problem can be seen as the optimization counterpart of the decision problems of isomprphism, monomorphism and homomorphism.
In the past, different methods have tried to optimize a variety of different objectives~\cite{conte2004thirty,bunke1997relation}.
In the context of message passing two distinct approaches have been used: 1) \citet{bayati2013message} propose a factor-graph for ``sparse'' graph alignment and show that it scales to very large instances. Here the term sparsity both refers to the number of edges of $\GG$ and $\GG'$
\index{sparse graph alignment}
 and also to the restricted possibility of matching nodes of $\VV$ to $\VV'$ -- \ie each node in $\VV$ can match only a few predetermined nodes in $\VV'$. The factor-graph used by the authors resembles the binary version of the maximum bipartite matching factor-graph of \refSection{sec:matching_factor}.
2) \citet{bradde2010aligning} used the min-sum BP to minimize the number of misalignment in a factor-graph similar to that of graph monomorphism above.
Here we follow their route, with the distinction that we suggest using the min-sum semiring with ``sub-graph isomorphism'' factor-graph and account for different matching costs using several tricks.

\index{graph matching}
Here we consider a general objective function that evaluates the mapping $\pi: \VV' \to \VV \cup \{\nulll\}$.
We then show how to optimize this objective using max-sum inference in a factor-graph.
\begin{easylist}[itemize]
& Node matching preference for matching node $i \in \VV$ to $j' \in \VV'$: $\varphi(i,j'): \VV \times \VV' \to \Re$. 
& Edge matching preference for matching $(i,j) \in \EE$ to $(i',j') \in \EE'$: $\varsigma((i,j),(k',l')): \EE \times \EE' \to \Re$. 
& Node merging preference for mapping nodes $i,j \in \VV$ to the same node $k \in \VV'$: $\vartheta(i,j,k'): \VV \times \VV \times \VV' \to \Re$.  
& Node deletion preference $\delta(i): \VV \to \Re$, is the  penalty for ignoring the node $i \in \VV$ -- \ie mapping it to the \nulll\ node. 
& Edge deletion  preference is the preference for dropping $(i,j) \in \EE$: $\varpi(i,j): \EE \to \Re$. 
& Edge insertion  preference is the preference for adding an edge $(i',j') \in \EE'$, when it is not matched against any edge in $\EE$: $\upsilon(i',j'): \EE' \to \Re$. 
\end{easylist}

We can define these preferences in such a way that the optimal solution is also a solution
to an interesting decision problem. 
\begin{example}\label{example:alignment_reduction}
The optimal alignments with the following parameters reproduce $\homo(\GG, \GG')$:
\begin{easylist}
& node matching $\varphi(i,j') = 0\quad \forall\; i \in \VV, j' \in \VV'$
& edge matching $\varsigma((i,j),(k',l')) = 1 \quad \forall\; (i,j) \in \EE, (k',l') \in \EE'$
& node merging  $\vartheta(i,j,k') = 0 \quad \forall\; i,j \in \VV, k' \in \VV'$
& node deletion $\delta(i) = -\infty \quad \forall\; i \in \VV$
& edge deletion $\varpi(i,j) = -\infty \quad \forall \; (i,j) \in \EE$
& edge insertion $\upsilon(k',l') = 0 \quad \forall \; (k',l') \in \EE'$
\end{easylist}

Alternatively, using positive and uniform node and edge matching preferences, 
\index{maximum common subgraph}
if we set the merge cost to $-\infty$ and allow node deletion at zero cost, the optimal solution will be the \magn{maximum common subgraph}.
In particular, for maximum-edge common subgraph we use
\begin{easylist}
& node matching $\varphi(i,j') = 0\quad \forall\; i \in \VV, j' \in \VV'$
& edge matching $\varsigma((i,j),(k',l')) = 1 \quad \forall\; (i,j) \in \EE, (k',l') \in \EE'$
& node merging  $\vartheta(i,j,k') = -\infty \quad \forall\; i,j \in \VV, k' \in \VV'$
& node deletion $\delta(i) = 0 \quad \forall\; i \in \VV$
& edge deletion $\varpi(i,j) = 0 \quad \forall \; (i,j) \in \EE$
& edge insertion $\upsilon(k',l') = 0 \quad \forall \; (k',l') \in \EE'$
\end{easylist}

\index{quadratic assignment problem}
Given two weighted adjacency matrices $\Dn$ and $\Dn'$ (for $\GG$ and $\GG'$ respectively), where $\Dn_{i,j}$ is the ``flow'' between the facilities $i$ and $j$, while $\Dn'_{k',l'}$ is the ``distance''
between the locations $k'$ and $l'$, the \magn{quadratic assignment problem}, seeks a one-to-one mapping $\pi^*: \VV \to \VV'$ of facilities to locations in order to optimize the flow
$$
\pi^* = \arg_{\pi} \max \;\; \sum_{i,j \in \VV} \Dn_{i,j} \Dn'_{\pi(i), \pi(j)} 
$$
Here w.l.o.g. we assume all weights are positive and set the alignment preferences so as to optimize the quadratic assignment problem:
\begin{easylist}
& node matching $\varphi(i,j') = 0\quad \forall\; i \in \VV, j' \in \VV'$
& edge matching $\varsigma((i,j),(k',l')) = \Dn_{i,j} \Dn'_{k', l'} \quad \forall\; (i,j) \in \EE, (k',l') \in \EE'$
& node merging  $ \vartheta(i,j,k') = -\infty \quad \forall\; i,j \in \VV, k' \in \VV'$
& node deletion $\delta(i) = -\infty \quad \forall\; i \in \VV$
& edge deletion $\varpi(i,j) = -\infty \quad \forall \; (i,j) \in \EE$
& edge insertion $\upsilon(k',l') = -\infty \quad \forall \; (k',l') \in \EE'$
\end{easylist}
\end{example}

Now we define the factors based on various alignment preferences.
The factor-graph has one variable per node $i \in \VV$: $\xs = \{ \xx_i \mid i \in \VV\}$, where 
$\xx_i \in \VV' \cup \{\nulll\}$. Here, $\xx_i = \nulll$ corresponds to ignoring this node in the mapping. The alignment factor-graph has three type of factors: 

\noindent \bullitem\ \magn{Local factors:} take the node matching preferences and node deletion preference into account:
\begin{align*}
\ff_i(\xx_i) \; = \; \left \{ 
\begin{array}{l l }
\delta(i) & \xx_i = \nulll \\
\varphi(i,\xx_i) & \text{otherwise}
\end{array}   
\right . \quad \forall i\in \VV
\end{align*}

\index{factor!edge}
\noindent \bullitem\ \magn{Edge factors:} are defined for each edge $(i,j) \in \EE$ and partly account for edge matching, node merging and edge deletion:
\begin{align*}
\ff_i(\xx_i, \xx_j) \; = \; \left \{ 
\begin{array}{l l }
0 & \xx_i = \nulll \; \vee \; \xx_j = \nulll\\
\vartheta(i,j,\xx_i) & \xx_i = \xx_j \\
\varsigma((i,j),(\xx_i,\xx_j)) & (\xx_i,\xx_j) \in \EE' \\
\varpi(i,j) & \text{otherwise}
\end{array}   
\right . \quad \forall (i,j)\in \EE
\end{align*}

\index{factor!non-edge}
\noindent \bullitem\ \magn{Non-edge factors:} are defined for non-existing edge $i,j\neq i \in \VV, (i,j) \notin \EE$ and partly account for node merging and edge insertion:
\begin{align*}
\ff_i(\xx_i, \xx_j) \; = \; \left \{ 
\begin{array}{l l }
0 & \xx_i = \nulll \; \vee \; \xx_j = \nulll\\
\vartheta(i,j,\xx_i) & \xx_i = \xx_j \\
\upsilon(\xx_i,\xx_j) & (\xx_i,\xx_j) \in \EE'\\
0 & \text{otherwise}
\end{array}   
\right . \quad \forall i,j\neq i \in \VV, (i,j) \notin \EE
\end{align*}

This factor-graph in its general form is fully connected. The cost of max-sum message-passing through each of these factors is $\OO(|\VV'|\;\log(\VV'|))$, which means each iteration of  variable synchronous max-sum BP is $\OO(|\VV|^2 \;|\VV'|\;\log(\VV'|))$. However, if the matching candidates are limited (\aka sparse alignment), and the graphs $\GG$ and $\GG'$ are sparse, this cost can be significantly reduced in practice.

\begin{figure}
  \centering 
      \includegraphics[width=.95\textwidth]{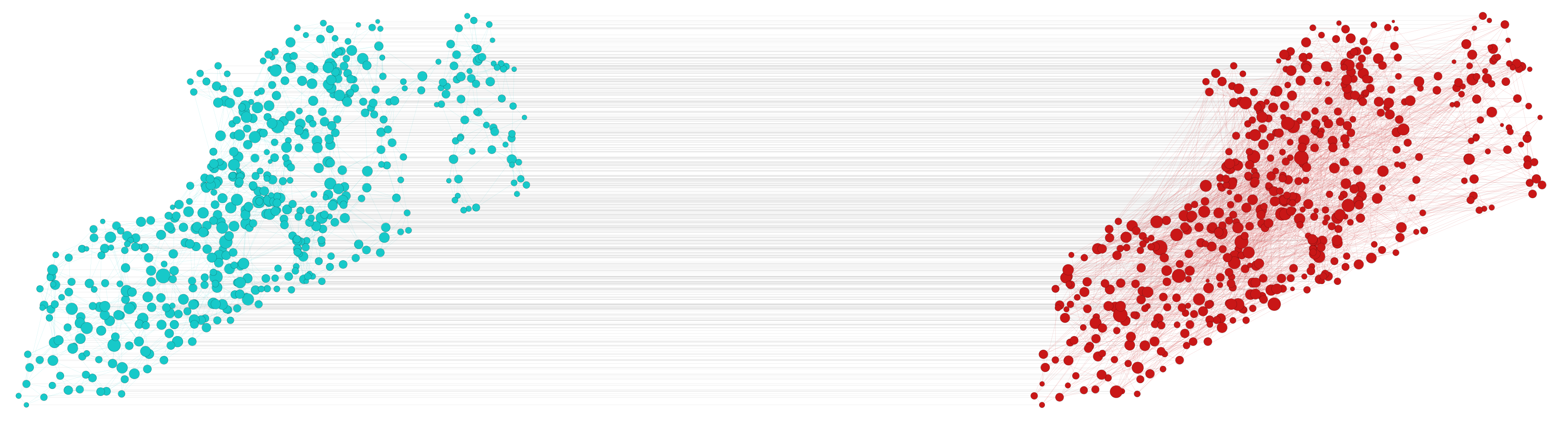}
  \caption[An example of sparse graph matching by message passing.]{Matching the E-coli metabolic network against a highly distorted version using message passing. Here $50\%$ of $|\EE| = 4306$ edges in the original network (left) are \textbf{removed} and the same number of random edges are \textbf{added}, to produce the distorted network (\ie $|\EE'| = |\EE|$). Each node had 10 random candidates for matching (including the correct choice) and message passing was able to identify the correct matching with $100\%$ accuracy.}
  \label{fig:matching}
\end{figure}

\begin{example}
\RefFigure{fig:matching} shows a matching of E-coli metabolic network against a distorted version, where 
$50\%$ of edges were removed and the same number of random edges were added. 
Then we generated 10  matching candidate for each node of the original graph, including the correct match and 9 other randomly selected nodes.
We used graph alignment with the following preferences to match the original graph against the distorted version\footnote{We used $T = 100$ initial iterations with damping parameter $\lambda = .2$.  After this, we used decimation and fixed 
$\rho = 1\%$ of variables after each $T = 50$ iterations. The total run-time was less than 5 minutes.}
\begin{easylist}
& node matching $\varphi(i,j') = 0 \quad \forall\; i \in \VV, j' \in \VV'$
& edge matching $\varsigma((i,j),(k',l')) = 1 \quad \forall\; (i,j) \in \EE, (k',l') \in \EE'$
& node merging  $ \vartheta(i,j,k') = -\infty \quad \forall\; i,j \in \VV, k' \in \VV'$
& node deletion $\delta(i) = -\infty \quad \forall\; i \in \VV$
& edge deletion $\varpi(i,j) = 0 \quad \forall \; (i,j) \in \EE$
& edge insertion $\upsilon(k',l') = 0 \quad \forall \; (k',l') \in \EE'$
\end{easylist}

We observed that message passing using our factor-graph was able to correctly match all the nodes in two graphs.
\end{example}

%% file: conclusion.tex

\chapter*{Conclusion}
\addcontentsline{toc}{chapter}{Conclusion}

This thesis studied a general form of inference in graphical models
with an emphasis on algebraic abstractions. We organized an important subset of these inference problems under an
inference hierarchy and studied the settings under which
distributive law allows efficient approximations in the form of message passing.
We investigated different methods to improve this approximation in loopy graphs using 1) variational formulation and loop correction; 2) survey propagation; 3) hybrid techniques. We then studied  graphical modelling of combinatorial optimization problems under different modes of inference.

As with any other inference and optimization framework, graphical modeling has its pros and cons.
The cons of using graphical models for combinatorial optimization are twofold
a) implementing message passing procedures, when compared to other standard techniques
such as using Integer Programming solvers, is more complex and time consuming.
This is further complicated by b) the fact that there is no standard guideline for designing a factor-graph 
representation, so as to minimize the computational complexity or increase
the quality of message passing solution. Indeed we used many tricks to efficiently approximate the solution to our problems; example include
 simplification of BP messages through alternative normalization, 
augmentation, variable and factor-synchronous message update, 
introduction of auxiliary variables, using damping and decimation \etc

On the other hand, when dealing with large scale and difficult optimization problems, one has to resort to conceptual and computational decomposition, and graphical modelling and message
passing techniques are the immediate candidates.  
Message passing is  mass parallelizable, scalable and often finds high-quality solutions.
By providing factor-graphs for a diverse set of combinatorial problems, this thesis also was an attempt to establish the 
 universality of message passing. Of course some of these problems better lend themselves to graphical modelling than some others,
resulting in better computational complexity and quality of results. \Cref{table:complexity} summarizes some important information
about the message passing solutions to combinatorial problems that are proposed or reviewed in this thesis. 

\begin{table}[!pht]
  \caption[Summary of message-passing solutions to combinatorial problems.]
{Summary of message-passing solutions to combinatorial problems. The time complexity is for one iteration of message passing.
We report different costs for different update schedules for the same problem. See the text for references. 
Here, $N$ is the number of nodes and $M$ is the number of constraints/factors.
}\label{table:complexity}
\centering
\begin{tikzpicture}
\node (table) [inner sep=1pt] {
    \scalebox{.6}{
    \begin{tabu}{ r  c  l  c l }
\textbf{Problem} & \textbf{Semiring ops.} & \textbf{Complexity} & \textbf{Schedule} & \textbf{relation to others} \\\tabucline[2pt]{-}
Belief Propagation & *  & $\OO(\sum_{i} \vert \XX_i \vert \vert \nb i \vert^2 + \sum _{\II}\vert \XX_\II \vert \vert \nb \II\vert )$ & async. & -  \\
                   & not \minmax\  & $\OO(\sum_{i} \vert \XX_i \vert \vert \nb i \vert + \sum _{\II}\vert \XX_\II \vert \vert \nb \II \vert)$ & v-sync. &    \\
                   & *  & $\OO(\sum_{i} \vert \XX_i \vert \vert \nb i \vert + \sum _{\II}\vert \XX_\II \vert )$ & $\ff$-sync. &  \\
Perturbed BP & \sumprod\  & $\OO(\sum_{i} \vert \XX_i \vert \vert \nb i \vert^2 + \sum _{\II}\vert \XX_\II \vert \vert \nb \II\vert )$ & async. & reduces to Gibbs samp. \& BP  \\
                   &    & $\OO(\sum_{i} \vert \XX_i \vert \vert \nb i \vert + \sum _{\II}\vert \XX_\II \vert \vert \nb \II \vert)$ & v-sync. &     \\
\tabucline[.1pt]{-}
Survey Propagation & *  & $\OO(\sum_{i} 2^{\vert \XX_i \vert} \vert \nb i \vert^2 + \sum _{\II}2^{\vert \XX_\II \vert} \vert \nb \II\vert )$ & async. & reduces to BP  \\
\tabucline[1pt]{-}
K-satisfiability & \sumprod & $\OO(K^2 M + \sum_i \vert \nb i \vert^2)$& async. & -  \\
                   &          & $\OO(K^2 M)$                            & v-sync. &   \\
                   &          & $\OO(K M)$                            & ($\ff$,v)-sync. &   \\
\tabucline[.1pt]{-}
K-coloring & \sumprod & $\OO(K \sum_i \vert \EE(i,\cdot) \vert^2)$ & async.&  $K$-clique-cover \\
             &          & $\OO(K \vert \EE \vert)$                 & v-sync. &   \\
\tabucline[.1pt]{-}
K-clique-cover & \sumprod & $\OO(K \sum_i (N - \vert \EE(i,\cdot)) \vert^2)$ & a-sync. & $\equiv$ to $K$-coloring on $\GG^c$ \\
                 & \sumprod & $\OO(K (N^2 -  \vert\EE\vert) )$ & v-sync. &           \\
\tabucline[.1pt]{-}
 K-dominating-set   & \sumprod\ & $\OO(K N^2 + \sum_{i \in \VV} \vert \EE(i,\cdot) \vert^2 + \vert \EE(\cdot,i) \vert^2 )$ & async. & - \\
 \& K-set-cover     &        & $\OO(K N + \vert \EE \vert )$ & $\ff$-sync. &  \\
min set-cover    & \minsum\ & $\OO( \vert \EE \vert )$ & ($\ff$,v)-sync. &  similar to K-median\\
 \tabucline[.1pt]{-}
 K-independent-set   & \sumprod\ & $\OO( N^3 )$ & async. & \textbf{binary} var. model  \\
 \& K-clique          &           & $\OO(K N^2 )$ & $\ff$-sync. &   $\equiv$ K-independent-set on $\GG^c$\\
      &           & $\OO(K N + \vert \EE \vert )$ & ($\ff$,v)-sync. &  \\
K-packing      & \minmax\   & $\OO(\log(N) (KN + \vert \EE \vert) )$ & ($\ff$,v)-sync. & \sumprod\ reduction $\equiv$ K-independent-set  \\
               & \minmax\   & $\OO(KN + \vert \EE \vert)$ & ($\ff$,v)-sync. & min-max BP \\
\tabucline[.1pt]{-}
 K-independent-set   & \sumprod\ & $\OO( K^3 N^2 )$ & async. &  \textbf{categorical} var. model  \\
 \& K-clique           &           & $\OO(K^2 N^2 )$ & v-sync. & $\equiv$ K-independent-set on $\GG^c$  \\
  K-packing        & \minmax\  & $\OO(K^2 N^2 \log(N) )$ & v-sync. & \sumprod\ reduction $\equiv$ K-independent-set  \\
 \tabucline[.1pt]{-}
max independent set& \maxsum\ & $\OO( \vert \EE \vert )$ & ($\ff$,v)-sync. &  \\
\& min vertex cover&   &  &  & $\equiv$ max independent-set \\
 \tabucline[.1pt]{-}
  sphere-packing & \sumprod\  & $\OO(K^2 2^{2n} )$ & v-sync. & \\
   (Hamming)       &    & $\OO(K^3 n + K^2 n^2 y)$ & async. & -  \\
   $n$: digits     &    & $\OO(K^2 n^2 y)$ & v-sync. &   \\
   $y$: min dist.                       &    & $\OO(K^2 n y)$ & ($\ff$,v)-sync. &  \\
 \tabucline[.1pt]{-}
 K-medians   & \minsum\ & $\OO( \vert \EE \vert )$ & $\ff$-sync. &  \aka\ affinity propagation  \\
 facility location & \minsum\ & $\OO( \vert \EE \vert )$ & $\ff$-sync. &    \\
 \tabucline[.1pt]{-}
 $d$-depth min span. tree & \minsum\ & $\OO( d \vert \EE \vert )$ & v-sync. &     \\
 prize-coll. Steiner tree & \minsum\ & $\OO( d \vert \EE \vert )$ & v-sync. &     \\
 \tabucline[.1pt]{-}
K-clustering & \minmax & $\OO(K N^2 \log(N) )$ & v-sync.  & \sumprod\ reduction $\equiv$ $K$-clique-cover  \\
 \tabucline[.1pt]{-}
 K-center   & \minmax\ & $\OO(\log(N)(K N + \vert \EE \vert ))$ & $\ff$-sync. & \sumprod\ reduction $\equiv$ K-set-cover  \\
            &          & $\OO(K N + \vert \EE \vert )$ & min-max BP &  \\
 \tabucline[.1pt]{-}
Modularity max   & \minsum\ & - &  & clique model, using augmentation  \\
                 &          & $\OO(K_{\max} N^2)$ &  & Potts model \\
 \tabucline[.1pt]{-}
max  matching   & \minsum\ & $\OO(N^2)$ & v-sync & - \\
\& cycle cover           &          &            &        &       \\
bottleneck assignment    & \minmax\ & - & - & - \\
max  $b$-matching & \minsum\ & $\OO(b N^2)$ & v-sync & - \\
 \tabucline[.1pt]{-}
TSP & \minsum\ & $\OO(N^2 \tau)$ or $\sim N^3$ & ($\ff$,v)-sync & - \\
bottleneck TSP & \minmax\ & $\OO(N^3 \log(N))$ & async & \sumprod\ reduction $\equiv$ Hamiltonian cycle \\
\tabucline[.1pt]{-}
subgraph isomorphism & \sumprod\ & $\OO(|\VV|^2 |\EE'|)$ & v-sync. & $\GG \to \GG'$ \\
subgraph monomorphism & \sumprod\ & $\OO(|\VV|^2\; |\VV'| + |\EE|\;|\EE'|)$ & v-sync. & $\GG \to \GG'$ \\
subgraph supermorphism & \sumprod\ & $\OO((|\VV^2| - |\EE|)\;|\EE'|\; + |\VV'|\;|\VV|^2)$ & v-sync. & $\GG \to \GG'$ \\
homomorphism & \sumprod\ & $\OO(|\EE|\;|\EE'|)$ & v-sync. & $\GG \to \GG'$ \\
graph alignment & \maxsum\ & $\OO(|\VV|^2\;|\VV'| \log(|\VV'|))$ & v-sync. & $\GG \to \GG'$ with general costs\\
max common sub-graph &  &  &  & a variation of graph alignment\\
quadratic assignment &  &  &  & a variation of graph alignment\\
\end{tabu}
}
};
\draw [rounded corners=.5em] (table.north west) rectangle (table.south east);
\end{tikzpicture}
\end{table}

\section*{Future work}
\addcontentsline{toc}{section}{Future work}
Due to breadth of the models and problems that we covered in this thesis, 
our investigation lacks the deserved depth in many cases. This is particularly 
pronounced in the experiments. Moreover, we encountered many new
questions and possibilities while preparing this thesis. Here, we enumerate some of the topics that 
demand a more in depth investigation in the future work
\begin{easylist}
& Many of the problems that we discussed also have efficient LP relaxations that some times
come with approximation guarantees. A comprehensive experimental comparison of message passing
and LP relaxations for these problems, both in terms of speed and accuracy is highly desirable.
& Our algebraic approach to inference suggests that all message passing procedures discussed here, including survey propagation, are also
applicable to the domain of complex numbers. Extensions to this domain not only may allow new applications (\eg using Fourier coefficients as factors)
but may also produce better solutions to many problems that we have studied here (\eg in solving CSPs).
& Our study of using graph homomorphism and its application to finding symmetries is a work in progress. In particular its relation to other methods
such as stochastic block models and spectral techniques needs further investigation.
& Although some preliminary results on Ising model suggested that using sum-product reductions for min-max inference performs much better
than direct min-max message passing, an extensive comparison of these two approaches to min-max inference is missing in our analysis.
& In \refSection{sec:csp_opt}, we noted that several optimization counterparts to CSPs allow  
using binary-search in addition to a direct optimization approach. While using binary search is more expensive for these problems,
we do not know which approach will performs better in practice.
\end{easylist}

%% file: proofs.tex
\begin{algorithm}
\SetKwInOut{Input}{input}\SetKwInOut{Output}{output}
\SetKwFunction{concomp}{ConnectedComponents}
\DontPrintSemicolon
 \Input{Graph $\GG = (\VV,\EE)$ with normalized (weighted)adjacency $\Dn$, maximum iterations $T_{\max}$, damping $\lambda$, threshold $\epsilon_{\max}$.}
 \Output{A clustering $\CC = \{\CC_1,\ldots,\CC_K\}$ of nodes.}
construct the null model\;
$\ph_{i:j} \leftarrow 0 \; \forall (i,j) \in \EE \cup \EE^{\nulll}$\;
 \While(\tcp{the augmentation loop}){$\truemath$}{
  $\epsilon \leftarrow 0$, $T \leftarrow 0$\;
 \While(\tcp{BP loop}){$\epsilon < \epsilon_{\max}$ \hbox{and} $T < T_{\max}$}{
$\epsilon \leftarrow 0$\;
\For{  ${(i,j)} \in \EE \cup \EE^{\nulll}$}{
  $\ph'_{{i:j}} \leftarrow \ph_{{i:j}}$\;
    $\ph_{{i:j}} \leftarrow \;(\Dn_{i,j} - \Dn^{\nulll}_{i,j})$\;
    
   \For(\tcp{update beliefs}){  $\II \in \nb {i:j}$}{
   calculate $\msg{\II}{i:j}$ using \refEq{eq:mIi_clique}\;
   $\ph_{i:j} \leftarrow \ph_{i:j} + \msg{\II}{{i:j}}$\;
   }
   $\epsilon \leftarrow \max \{ \epsilon, \vert \ph_{{i:j}} - \ph'_{{i:j}} \vert\}$\;
   \For(\tcp{update msgs.}){  $\II \in \nb {i:j}$}{
     $\msgt{{i:j}}{\II} \leftarrow \ph_{{i:j}} - \msg{\II}{{i:j}}$\;
     $\msg{{i:j}}{\II} \leftarrow \lambda \msgt{{i:j}}{\II} + (1-\lambda) \msg{{i:j}}{\II}$
   }
 }
 $T \leftarrow T + 1$
}

\For{  $i \in \VV$}{
  \For{  $(i,j), (i,k) \in \EE \cup \EE^{\nulll}$}{
    \lIf{$\ph_{{i:j}} > 0$ \textbf{and} $\ph_{{i:k}} > 0$ \textbf{and} $\ph_{{i:k}} \leq 0$}{
      add the corresponding clique factor to the factor-graph\;
}
}
}
\lIf{no factor was added}{\textbf{break} out of the loop}
\lElse{$\msg{i:j}{\II} \leftarrow 0\; \forall \II, i:j \in \II$}
}
$\CC \leftarrow$ \concomp{$(\VV, \{(i,j) \in \EE \cup \EE^{\nulll}  \mid \ph_{i:j} > 0 \})$}\;

\caption{Message Passing for Modularity Maximization.}\label{alg:clustering}
\end{algorithm}

\begin{algorithm}
\SetKwInOut{Input}{input}\SetKwInOut{Output}{output}
\SetKwFunction{concomp}{ConnectedComponents}
\DontPrintSemicolon
 \Input{Graph $\GG = (\VV,\EE)$, weighted (symmetric) adjacency matrix $\Dn$, maximum iterations $T_{\max}$, damping $\lambda$, threshold $\epsilon_{\max}$.}
 \Output{A subset $\TT \subset \EE$ of the edges in the tour.}
\DontPrintSemicolon
construct the initial factor-graph\;
initialize the messages for degree constraints $\msg{i:j}{\EE(i, \cdot)} \leftarrow 0 \forall i \in \VV, j \in \EE(i, \cdot)$\;
initialize $\ph_{i:j} \leftarrow \Dn_{i,j}\;\; \forall (i,j) \in \EE$\;
 \While(\tcp{the augmentation loop}){$\truemath$}{
  $\epsilon \leftarrow 0$, $T \leftarrow 0$\;
 \While(\tcp{BP loop}){$\epsilon < \epsilon_{\max}$ \textbf{and} $T < T_{\max}$}{
$\epsilon \leftarrow 0$\;
\For(\tcp{including \ $\ff_{\EE(\SS, \cdot)}$, $\ff_{\EE(i, \cdot)}$ (updates all the outgoing messages from this factor)}){ \textbf{each} $\ff_\II$}{
  find three lowest values in  $\{\msg{i:j}{\II} \mid i:j \in \nb \II\}$\;
  \For{\textbf{each} $i:j \in \II$}{
   calculate $\msgt{\II}{i:j}$ using \refEqs{eq:mIi_degree}{eq:mIi_subtour}\;
   $\epsilon_{\II \to i:j} \leftarrow \msgt{\II}{i:j} - \msg{\II}{i:j}$\;
   $\msg{\II}{i:j} \leftarrow \msg{\II}{i:j} + \lambda \epsilon_{\II \to i:j}$\;
   $\ph_{i:j} \leftarrow \ph_{i:j} + \epsilon_{\II \to i:j}$\;
   $\epsilon \leftarrow \max ( \epsilon, \vert \epsilon_{\II \to e} \vert )$
   }
 }
 $T \leftarrow T + 1$
}
$\TT \leftarrow \{ (i,j) \in \EE \mid \ph_{i:j} > 0 \}$\tcp*{respecting degree constraints.}
$\CC \leftarrow$ \concomp{$(\VV, \TT)$}\;
\lIf{$\vert \CC \vert = 1$}{
  return $\TT$
}\lElse{
  augment the factor-graph with $\ff_{\EE(\SS, \cdot)} \; \forall \SS \in \CC$\;
  initialize $\msg{\EE(\SS, \cdot)}{i:j} \leftarrow 0\; \forall \SS \in \CC, i:j \in \EE(\SS, \cdot)$\;
}
}
\caption{Message Passing for TSP}\label{alg:tsp}
\end{algorithm}